\documentclass[10pt,journal,compsoc]{IEEEtran}

\ifCLASSOPTIONcompsoc
\usepackage[nocompress]{cite}
\else
\usepackage{cite}
\fi

\ifCLASSINFOpdf
\usepackage[pdftex]{graphicx}
\DeclareGraphicsExtensions{.pdf,.jpeg,.png}
\DeclareGraphicsExtensions{.eps}
\usepackage{epstopdf}
\else
\fi

\ifCLASSOPTIONcompsoc
\usepackage[caption=false,font=footnotesize,labelfont=sf,textfont=sf]{subfig}
\else
\usepackage[caption=false,font=footnotesize]{subfig}
\fi
\usepackage{booktabs}
\usepackage{amssymb}

\usepackage{mathrsfs}
\usepackage{epsfig}
\usepackage{graphicx}
\usepackage{amsmath}
\usepackage{cite}
\usepackage{enumerate}
\usepackage{cases}
\usepackage{multirow}
\usepackage{verbatim}
\usepackage{amssymb}
\usepackage{CJK}
\usepackage{algorithmicx}

\usepackage{color}
\usepackage{bm}
\usepackage{booktabs}
\usepackage{colortbl}
\usepackage{float}
\usepackage{ragged2e}  
\usepackage{csquotes}
\usepackage[breaklinks=true,bookmarks=false]{hyperref}
\usepackage{diagbox}
\usepackage{threeparttable}
\usepackage{xspace}
\usepackage{array}
\usepackage[table]{xcolor}
\usepackage{overpic}
\usepackage{textcomp}
\usepackage{contour}
\usepackage[table]{xcolor}
\usepackage{threeparttable}
\usepackage{tablefootnote}

\DeclareMathOperator*{\argmin}{argmin}
\newcommand{\tabincell}[2]{\begin{tabular}{@{}#1@{}}#2\end{tabular}}

\usepackage{silence}
\makeatletter
\newcommand{\thickhline}{%
	\noalign {\ifnum 0=`}\fi \hrule height 1pt
	\futurelet \reserved@a \@xhline
}
\makeatother

\makeatletter
\DeclareRobustCommand\onedot{\futurelet\@let@token\@onedot}
\def\@onedot{\ifx\@let@token.\else.\null\fi\xspace}

\begin{document}

	\title{Low-Light Image and Video Enhancement \\Using Deep Learning: A Survey}

	\author{Chongyi Li, Chunle Guo, Linghao Han,  Jun Jiang, Ming-Ming Cheng,~\IEEEmembership{Senior Member,~IEEE},  \\Jinwei Gu,~\IEEEmembership{Senior Member,~IEEE},  and Chen Change Loy,~\IEEEmembership{Senior Member,~IEEE}
		
		\thanks{C. Li and C. C. Loy are with the S-Lab, Nanyang Technological University (NTU), Singapore  (e-mail: chongyi.li@ntu.edu.sg and ccloy@ntu.edu.sg).}
		\thanks{C. Guo, L. Han, and M. M. Cheng are with the College of Computer Science, Nankai University, Tianjin, China  (e-mail: guochunle@nankai.edu.cn, lhhan@mail.nankai.edu.cn, and cmm@nankai.edu.cn).}
		\thanks{J. Jiang and J. Gu are with the SenseTime  (e-mail: jiangjun@sensebrain.site and gujinwei@sensebrain.site).}
		\thanks{C. Li and C. Guo contribute equally.}
		\thanks{C. C. Loy is the corresponding author.}
	}
	
	\markboth{IEEE TRANSACTIONS ON PATTERN ANALYSIS AND MACHINE INTELLIGENCE}%
	{Shell \MakeLowercase{\textit{et al.}}: Bare Demo of IEEEtran.cls for Computer Society Journals}

	\IEEEtitleabstractindextext{%
		\justify  

\begin{abstract}
	\label{sec:Abstrat}
	%
	Low-light image enhancement (LLIE) aims at improving the  perception or interpretability of an image captured in an environment with poor illumination. 
	Recent advances in this area are dominated by deep learning-based solutions, where many learning strategies, network structures, loss functions, training data, etc. have been employed.
	In this paper, we provide a comprehensive survey to cover various aspects ranging from algorithm taxonomy to unsolved open issues. 
	To examine the generalization of existing methods,  we propose a low-light image and video dataset, in which the images and videos are taken by different mobile phones' cameras under diverse illumination conditions.
	Besides, for the first time, we provide a unified online platform that covers many popular LLIE methods, of which the results can be produced through a user-friendly web interface. 
	In addition to qualitative and quantitative evaluation of existing methods on publicly available and our proposed datasets, we also validate their performance in face detection in the dark. 
	This survey together with the proposed dataset and online platform could serve as a reference source for future study and promote the development of this research field. 
	The proposed platform and dataset as well as the collected methods, datasets, and evaluation metrics are publicly available and will be regularly updated. Project page: \href{https://www.mmlab-ntu.com/project/lliv\_survey/index.html}{https://www.mmlab-ntu.com/project/lliv\_survey/index.html}. 
\end{abstract}

		}
	
	\maketitle

	\IEEEdisplaynontitleabstractindextext
	
	\IEEEpeerreviewmaketitle


\IEEEraisesectionheading{\section{Introduction}
	\label{sec:Introduction}}

\begin{figure}[!t]
	\begin{center}
		\begin{tabular}{c@{ }c@{ }c@{ }}
			\includegraphics[width=0.3\linewidth,height=0.2\linewidth]{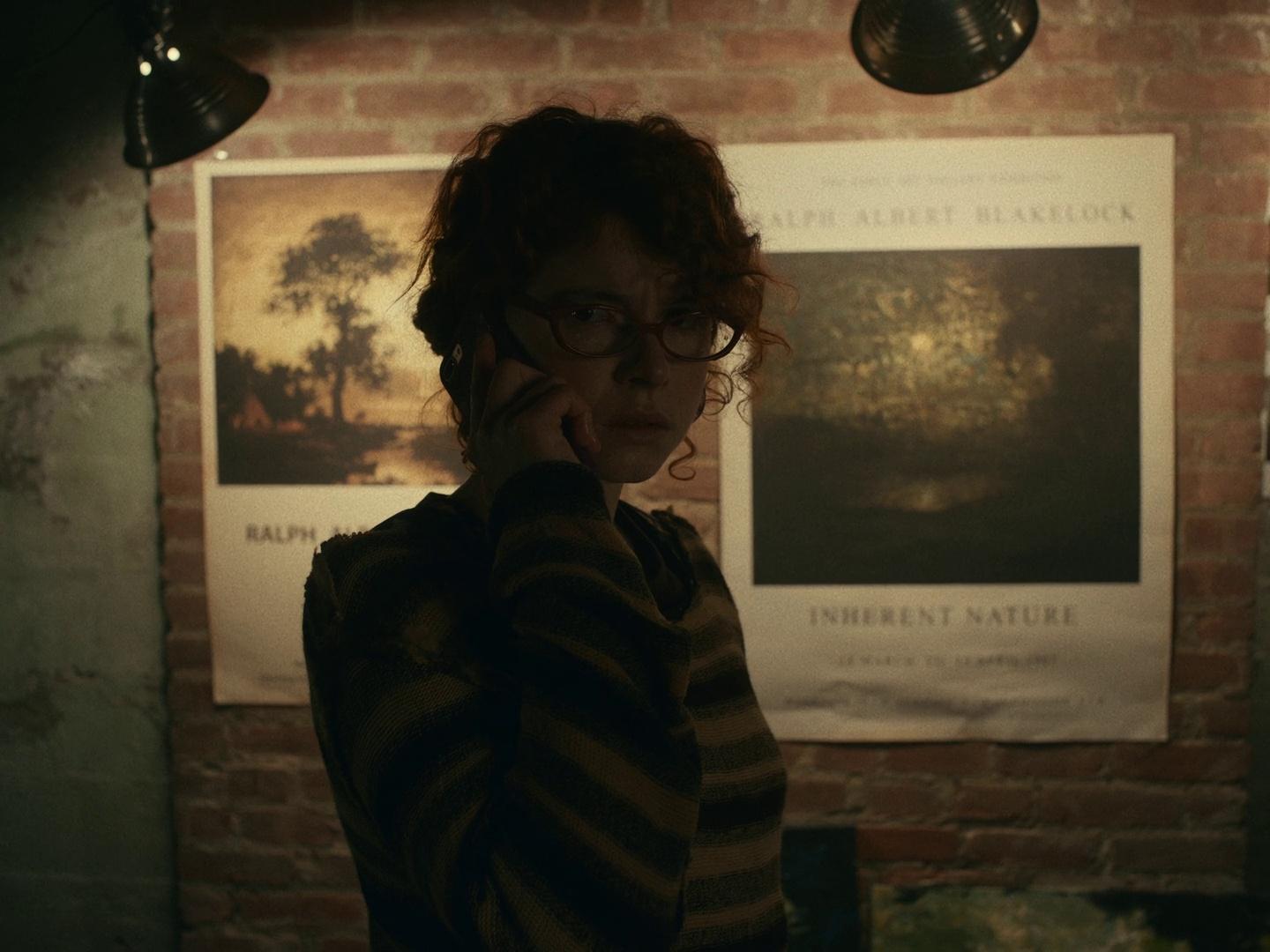}&
			\includegraphics[width=0.3\linewidth,height=0.2\linewidth]{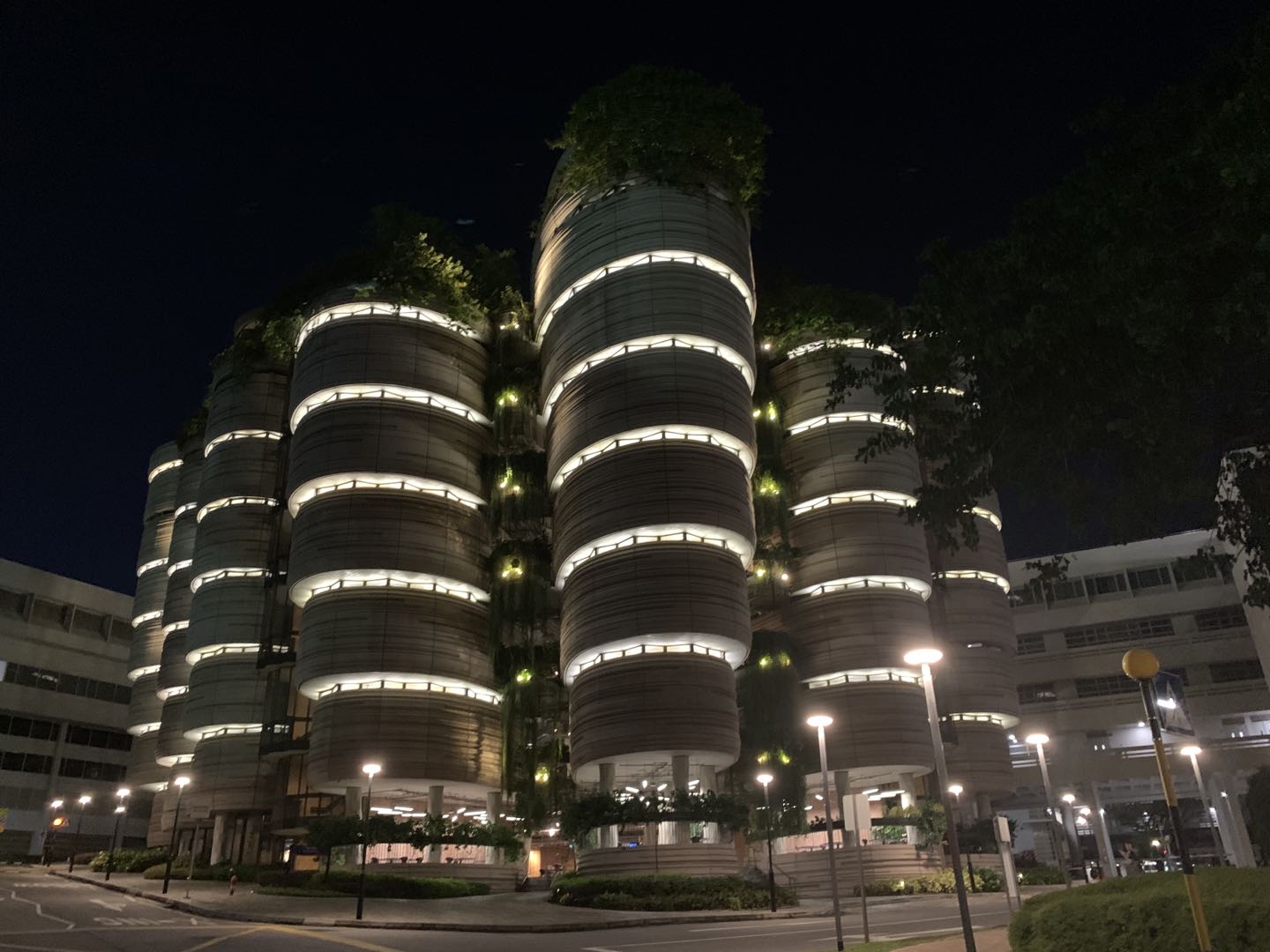}&
			\includegraphics[width=0.3\linewidth,height=0.2\linewidth]{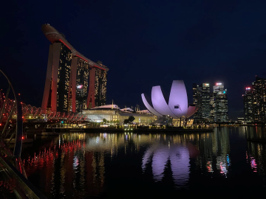}\\
			(a) back lit & (b) uneven light &(c) dim light\\
			\includegraphics[width=0.3\linewidth,height=0.2\linewidth]{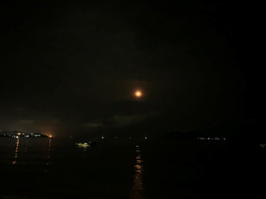}&
			\includegraphics[width=0.3\linewidth,height=0.2\linewidth]{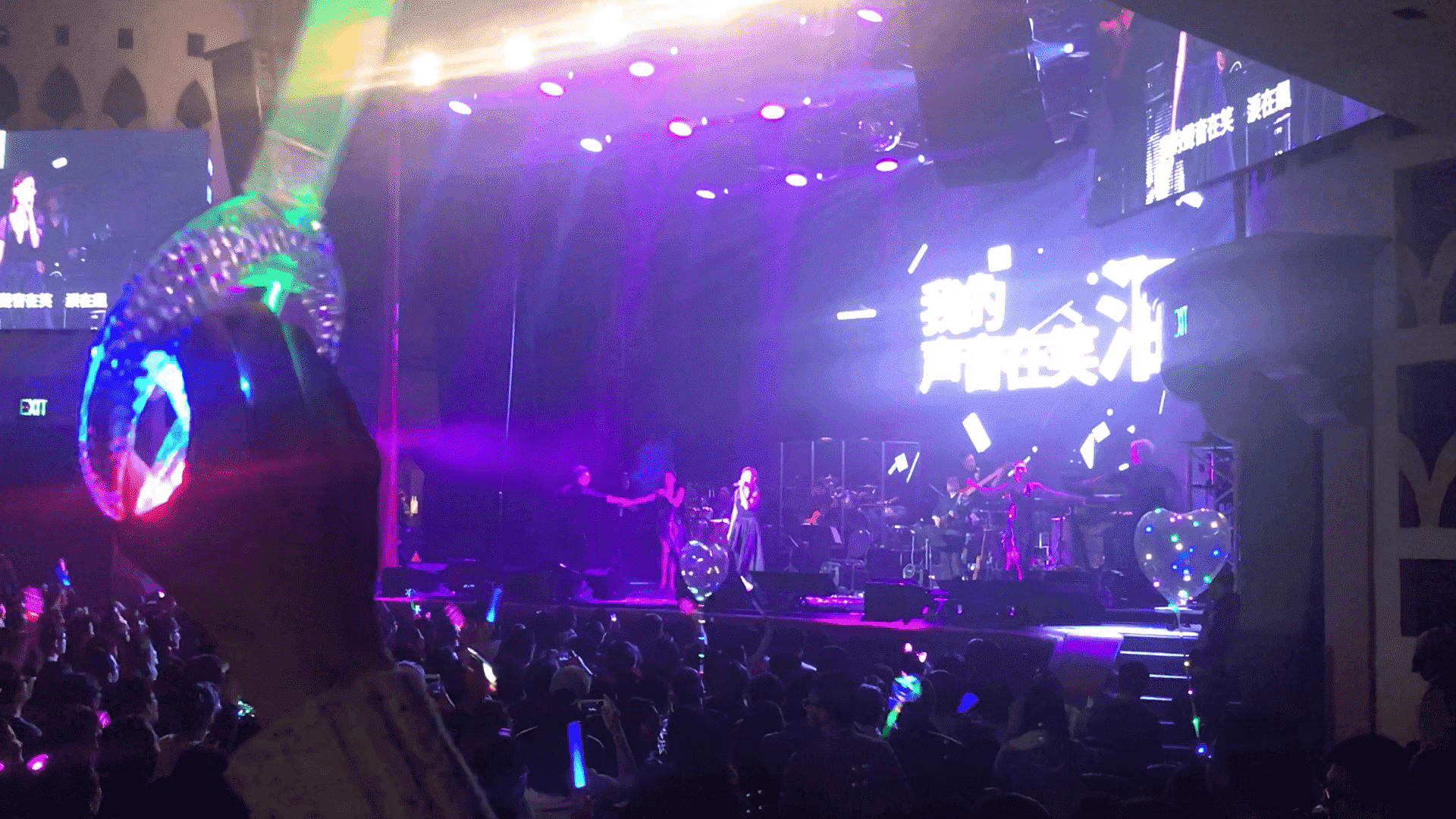}&
			\includegraphics[width=0.3\linewidth,height=0.2\linewidth]{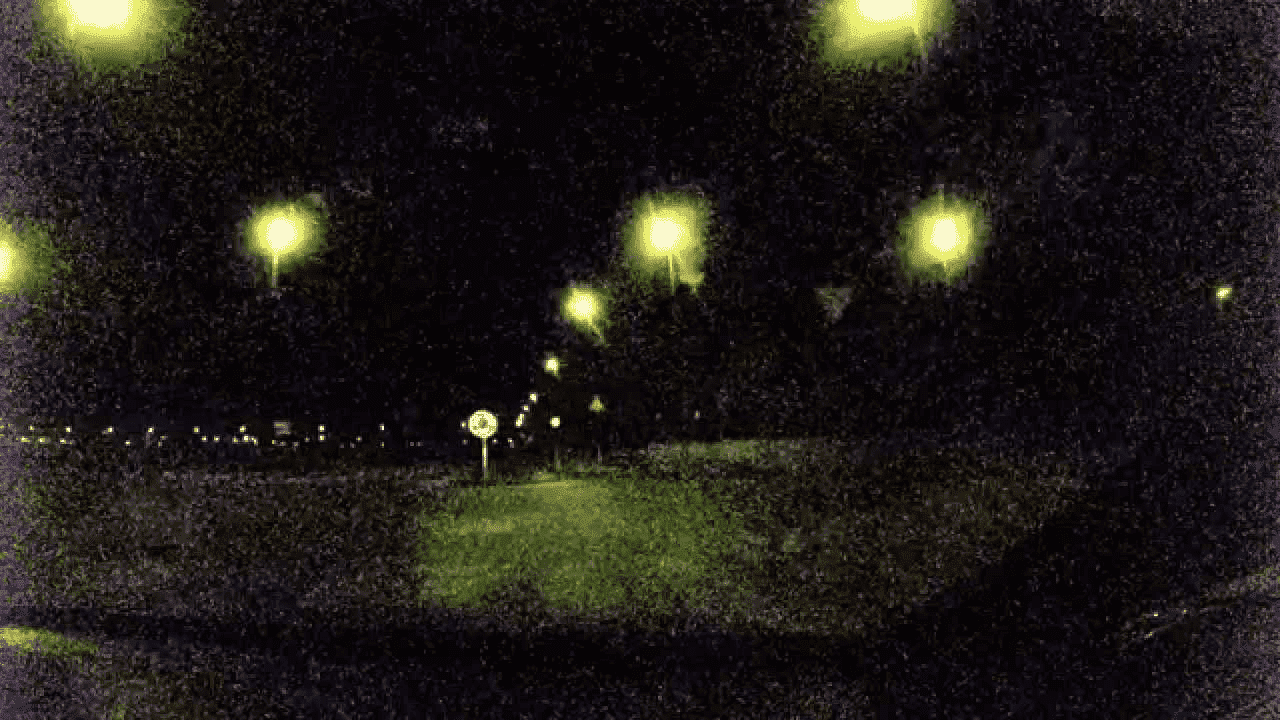}\\
			(d) extremely low &	(e) colored light & (f) boosted noise\\
		\end{tabular}
	\end{center}
	\vspace{-3pt}
	\caption{Examples of images taken under sub-optimal lighting conditions. These images suffer from the buried scene content, reduced contrast, boosted noise, and inaccurate color. }
	\label{fig:example}
	\vspace{-5pt}
\end{figure}

\IEEEPARstart{I}{mages} are often taken under sub-optimal lighting conditions, under the influence of backlit, uneven light, and dim light, due to inevitable environmental and/or technical constraints such as insufficient illumination and limited exposure time. Such images suffer from the compromised aesthetic quality and unsatisfactory transmission of information for high-level tasks such as object tracking, recognition, and detection. Figure \ref{fig:example} shows some examples of the degradations induced by sub-optimal lighting conditions.

Low-light enhancement enjoys a wide range of applications in different areas, including visual surveillance, autonomous driving, and computational photography. In particular, smartphone photography has become ubiquitous and prominent. Limited by the size of the camera aperture, the requirement of real-time processing, and the constraint of memory, taking photographs with a smartphone's camera in a dim environment is especially challenging. There is an exciting research arena of enhancing low-light images and videos in such applications.

\begin{figure*}[thb]
	\centering  \centerline{\includegraphics[width=1\linewidth]{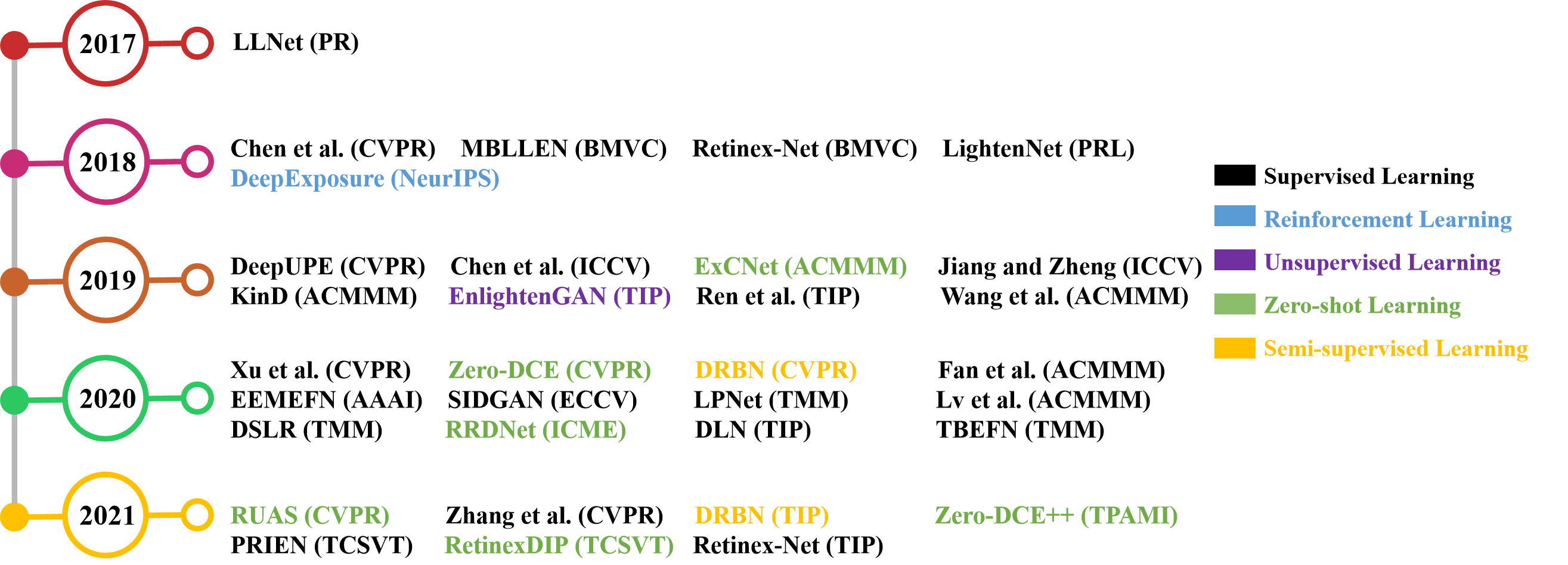}}
	\vspace{-2pt}
	\caption{A concise milestone of deep learning-based low-light image and video enhancement methods.  \textbf{Supervised learning-based methods}: LLNet \cite{LLNet}, Chen et al. \cite{Chen2018}, MBLLEN \cite{LvBMVC2018}, Retinex-Net \cite{ChenBMVC18}, LightenNet \cite{LightenNet}, SCIE \cite{CaiTIP2018}, DeepUPE \cite{DeepUPE}, Chen et al. \cite{ChenICCV19}, Jiang and Zheng \cite{JZICCV19}, Wang et al. \cite{WangACM19}, KinD \cite{ZhangACM19}, Ren et al. \cite{RenTIP2019}, Xu et al. \cite{XuCVPR20}, Fan et al. \cite{FanACM20}, Lv et al. \cite{LvACM20}, EEMEFN~\cite{ZhuAAAI20}, SIDGAN. \cite{TriantafyllidouECCV}, LPNet \cite{LiTMM20}, DLN \cite{WangTIP2020}, TBEFN \cite{TBEFN},  DSLR \cite{DSLR},  Zhang et al. \cite{ZhangCVPR21}, PRIEN \cite{PRIEN}, and Retinex-Net \cite{YangTIP21R}. \textbf{Reinforcement  learning-based method}: DeepExposure \cite{YuNips18}. \textbf{Unsupervised  learning-based method}: EnlightenGAN \cite{EnlightenGAN}. \textbf{Zero-shot learning-based methods}: ExCNet \cite{ZhangACM191}, Zero-DCE \cite{ZeroDCE}, RRDNet \cite{RRDNet}, Zero-DCE++ \cite{ZeroDCE++}, RetinexDIP \cite{RetinexDIP}, and RUAS \cite{RUAS}. \textbf{Semi-supervised learning-based method}: DRBN \cite{YangCVRP20} and DRBN \cite{YangTIP21RR}.}
	\label{fig:milestones}
	\vspace{-4pt}
\end{figure*}

Traditional methods for low-light enhancement include Histogram Equalization-based methods \cite{HaidiTCE,AbdullahTCE} and Retinex model-based methods \cite{Wang2013,Fu2015,LIME,Park2017,Li2018,Gu2019,Ren2020,Hao2020}. The latter received relatively more attention. A typical Retinex model-based approach decomposes a low-light image into a reflection component and an illumination component by priors or regularizations. The estimated reflection component is treated as the enhanced result.
Such methods have some limitations:
\textbf{1)} the ideal assumption that treats the reflection component as the enhanced result does not always hold, especially given various illumination properties, which could lead to unrealistic enhancement such as loss of details and distorted colors, \textbf{2)} the noise is usually ignored in the Retinex model, thus it is remained or amplified in the enhanced results, \textbf{ 3)} finding an effective prior or regularization is challenging. Inaccurate prior or regularization may result in artifacts and color deviations in the enhanced results, and \textbf{4)} the runtime is relatively long because of their complicated optimization process.

Recent years have witnessed the compelling success of deep learning-based LLIE since the first seminal work \cite{LLNet}. Deep learning-based solutions enjoy better accuracy, robustness, and speed over conventional methods, thus attracting increasing attention. A concise milestone of deep learning-based LLIE methods is shown in Figure \ref{fig:milestones}. As shown, 
since 2017, the number of deep learning-based solutions has grown year by year. Learning strategies used in these solutions cover Supervised Learning (SL), Reinforcement Learning (RL), Unsupervised Learning (UL), Zero-Shot Learning (ZSL), and Semi-Supervised Learning (SSL). Note that we only report some representative methods in Figure \ref{fig:milestones}. In fact, there are more than 100 papers on deep learning-based methods from 2017 to 2021. Moreover, although some general photo enhancement methods \cite{WhiteBox,DBLNet,Enhancer,Deng18,Yan16,Chen17,Ni2020TIP,ZengPAMI2020,LiArx2020}  can improve the brightness of images to some extent, we omit them in this survey as they are not designed to handle diverse low-light conditions. We concentrate on deep learning-based solutions that are specially developed for low-light image and video enhancement.

	Despite deep learning has dominated the research of LLIE, an in-depth and comprehensive survey on deep learning-based solutions is lacking. There are two reviews of LLIE \cite{LowlightSurvey,IJCVsurvey21}. Wang et al. \cite{LowlightSurvey} mainly reviews conventional LLIE methods while our work systematically and comprehensively reviews recent advances of deep learning-based LLIE. In comparison to Liu et al. \cite{IJCVsurvey21} that reviews existing LLIE algorithms, measures the machine vision performance of different methods, provides a low-light image dataset serving both low-level and high-level vision enhancement, and develops an enhanced face detector, our survey reviews the low-light image and video enhancement from different aspects and has the following unique characteristics.
	\textbf{1)} Our work mainly focuses on recent advances of deep learning-based low-light image and video enhancement, where we provide in-depth analysis and discussion in various aspects, covering learning strategies, network structures, loss functions, training datasets, test datasets, evaluation metrics, model sizes, inference speed, enhancement performance, etc. Thus, this survey centers on deep learning and its applications in low-light image and video enhancement.
	\textbf{2)} We propose a dataset that contains images and videos captured by different mobile phones' cameras under diverse illumination conditions to evaluate the generalization of existing methods. This new and challenging dataset is a supplement of existing low-light image and video enhancement datasets as such a dataset is lacking in this research area. Besides, we are the first, to the best of our knowledge,  to compare the performance of deep learning-based low-light image enhancement methods on this kind of data. 
	\textbf{3)} We provide an online platform that covers many popular deep learning-based low-light image enhancement methods, where the results can be produced by a user-friendly web interface. With our platform, one without any GPUs can assess the results of different methods for any input images online, which speeds up the development of this research field and helps to create new research.
	We hope that our survey could provide novel insights and inspiration to facilitate the understanding of deep learning-based LLIE, foster research on the raised open issues, and speed up the development of this research field.


\section{Deep Learning-Based LLIE}
\label{sec:Solution}

\subsection{Problem Definition}
We first give a common formulation of the deep learning-based LLIE problem. For a low-light image $I\in \mathbb{R}^{W\times H\times3}$ of width $W$ and height $H$, the  process can be modeled as:
\begin{equation}
	\label{equ_1}
	\widehat{R}=\mathcal{F}(I;\theta),
\end{equation}
where $\widehat{R}\in \mathbb{R}^{W\times H\times3}$ is the enhanced result and $\mathcal{F}$ represents the network with trainable parameters $\theta$. The purpose of deep learning is to find optimal network parameters  $\widehat{\theta}$ that minimizes the error:
\begin{equation}
	\label{equ_2}
	\widehat{\theta}=\argmin_{\theta}\mathcal{L}(\widehat{R},R),
\end{equation}
where $R\in \mathbb{R}^{W\times H\times3}$ is the ground truth, and the loss function $\mathcal{L}(\widehat{R},R)$ drives the optimization of network.
Various loss functions such as supervised loss and unsupervised loss can be used. More details will be presented in Section \ref{sec:Technical}.

\subsection{Learning Strategies}

\begin{figure*}[!t]
		\begin{center}
			\begin{tabular}{c@{ }c@{ }c@{ }c@{ }}
				\includegraphics[width=.23\textwidth]{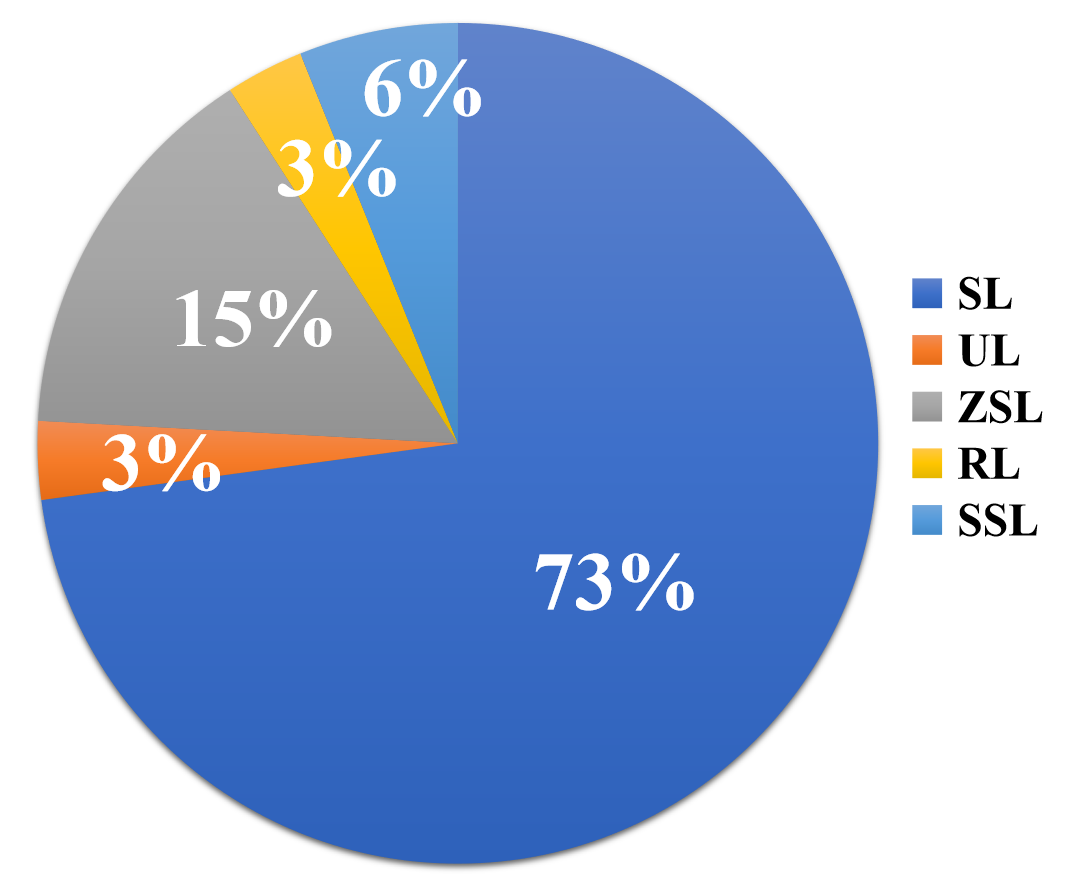}&
				\includegraphics[width=.23\textwidth]{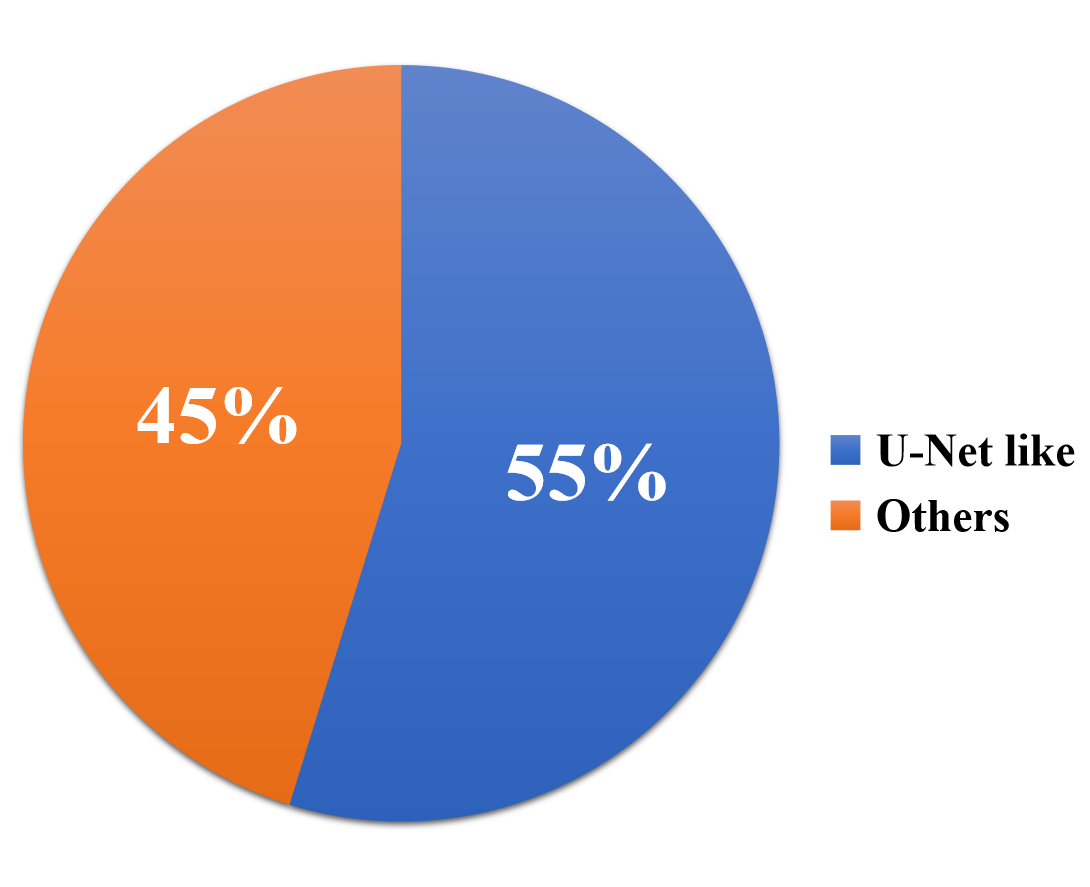}&
				\includegraphics[width=.23\textwidth]{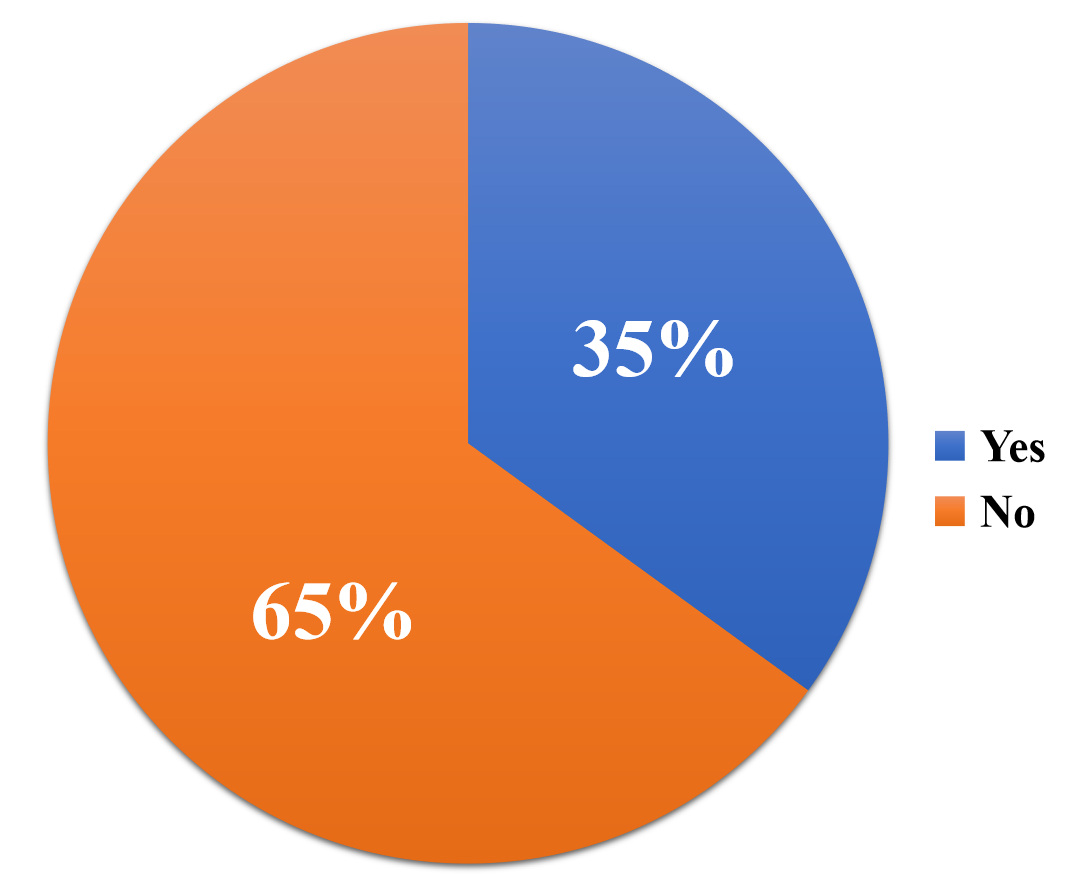}&
				\includegraphics[width=.23\textwidth]{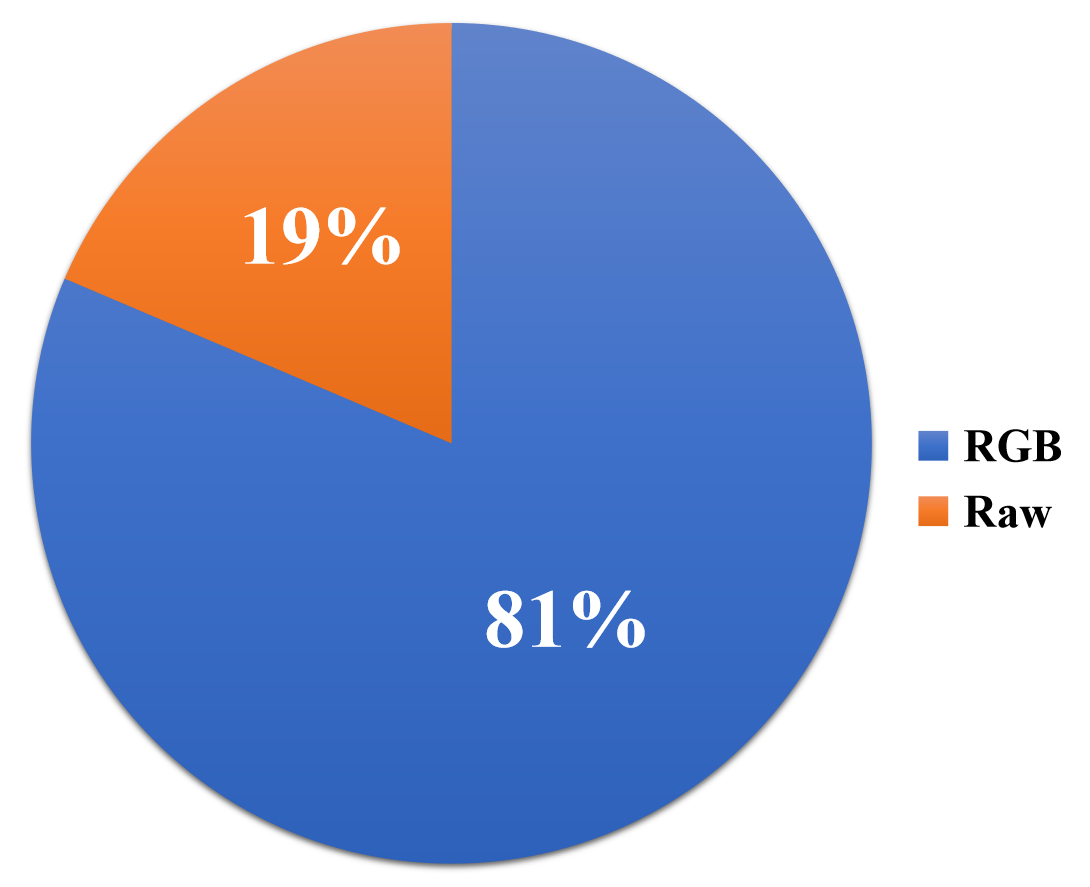}\\
				(a) learning strategy & (b) network structure& (c) Retinex model & (d)  data format \\
				\includegraphics[width=.23\textwidth]{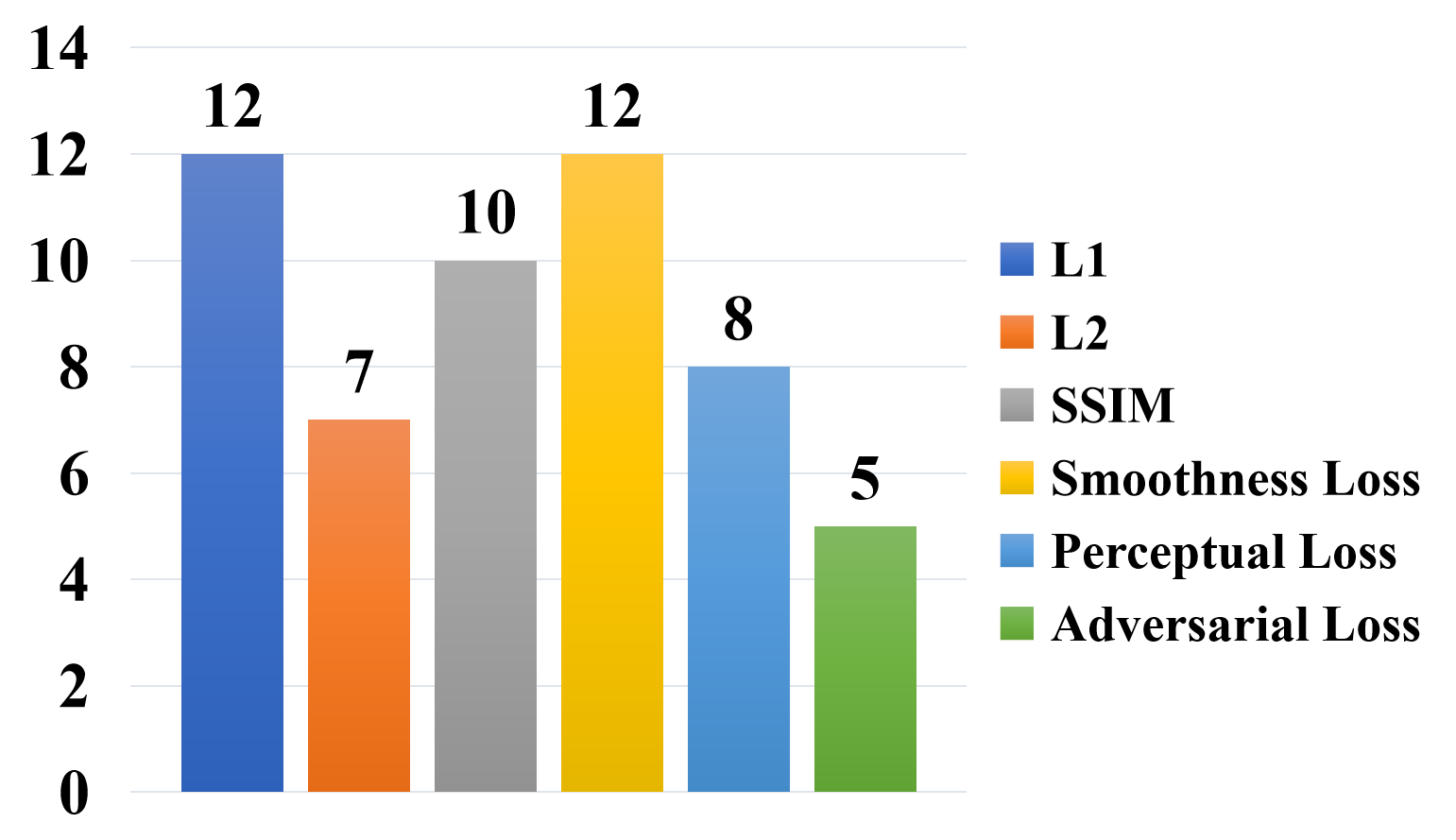}&
				\includegraphics[width=.23\textwidth]{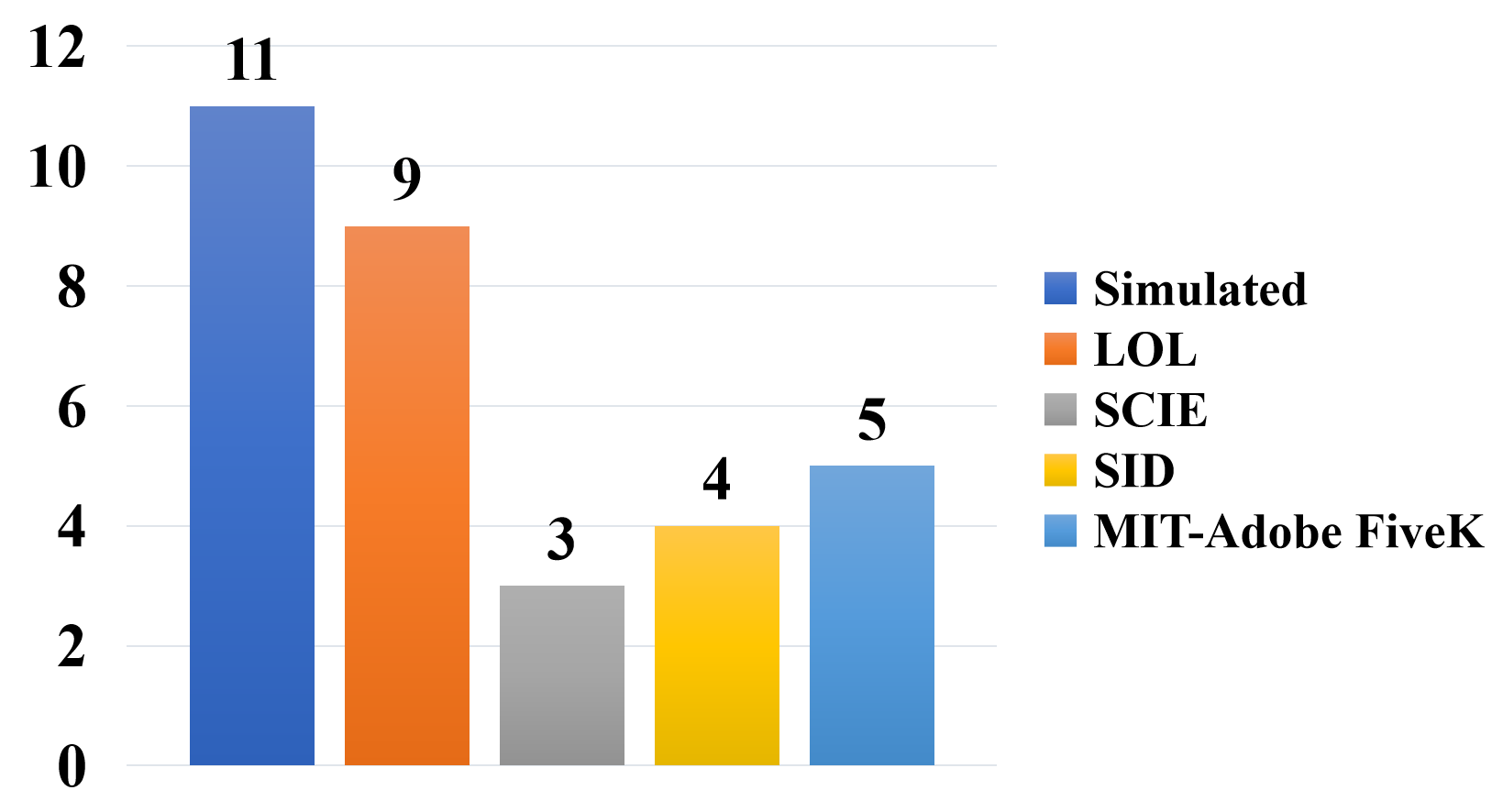}&
				\includegraphics[width=.23\textwidth]{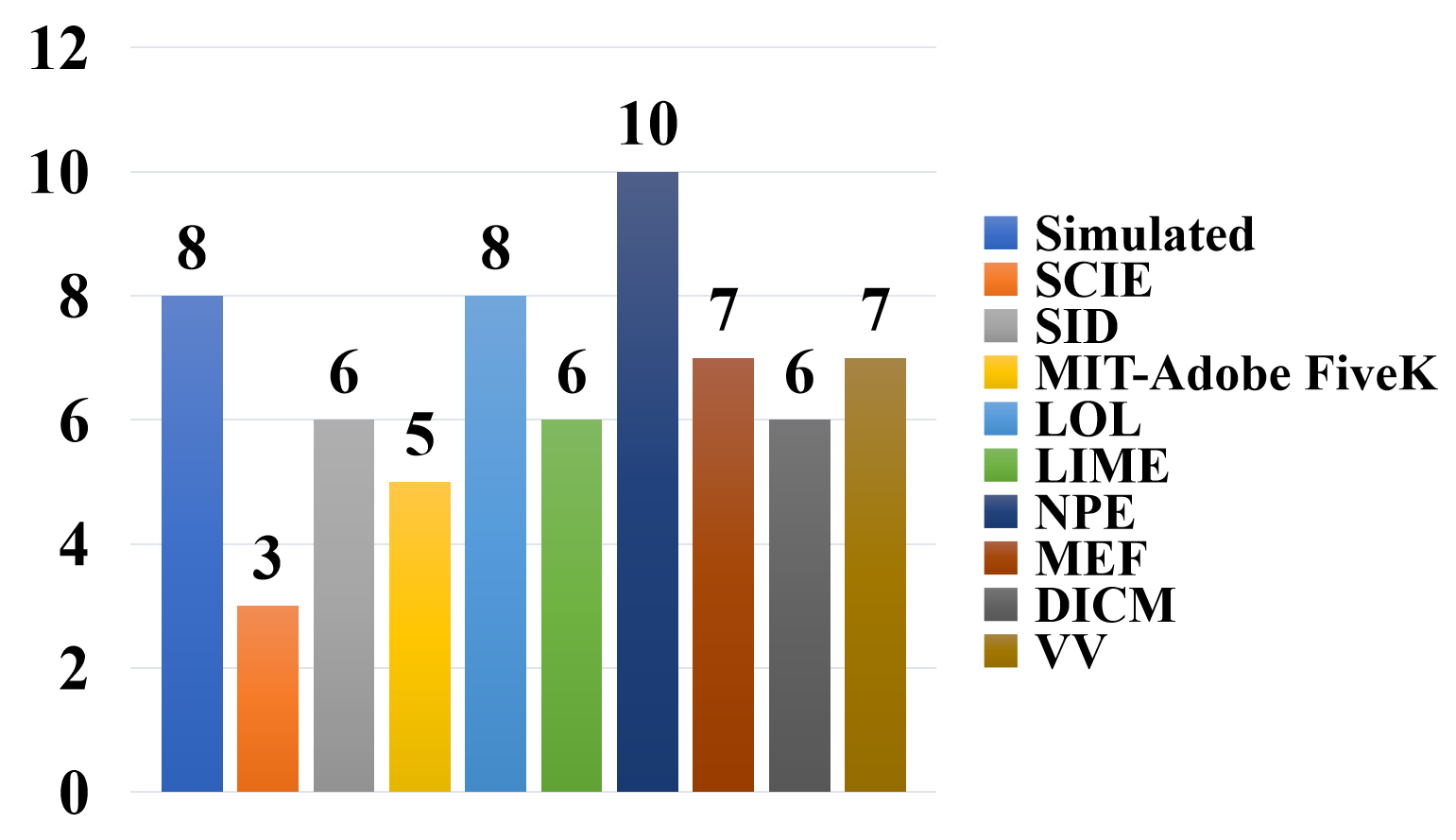}&
				\includegraphics[width=.23\textwidth]{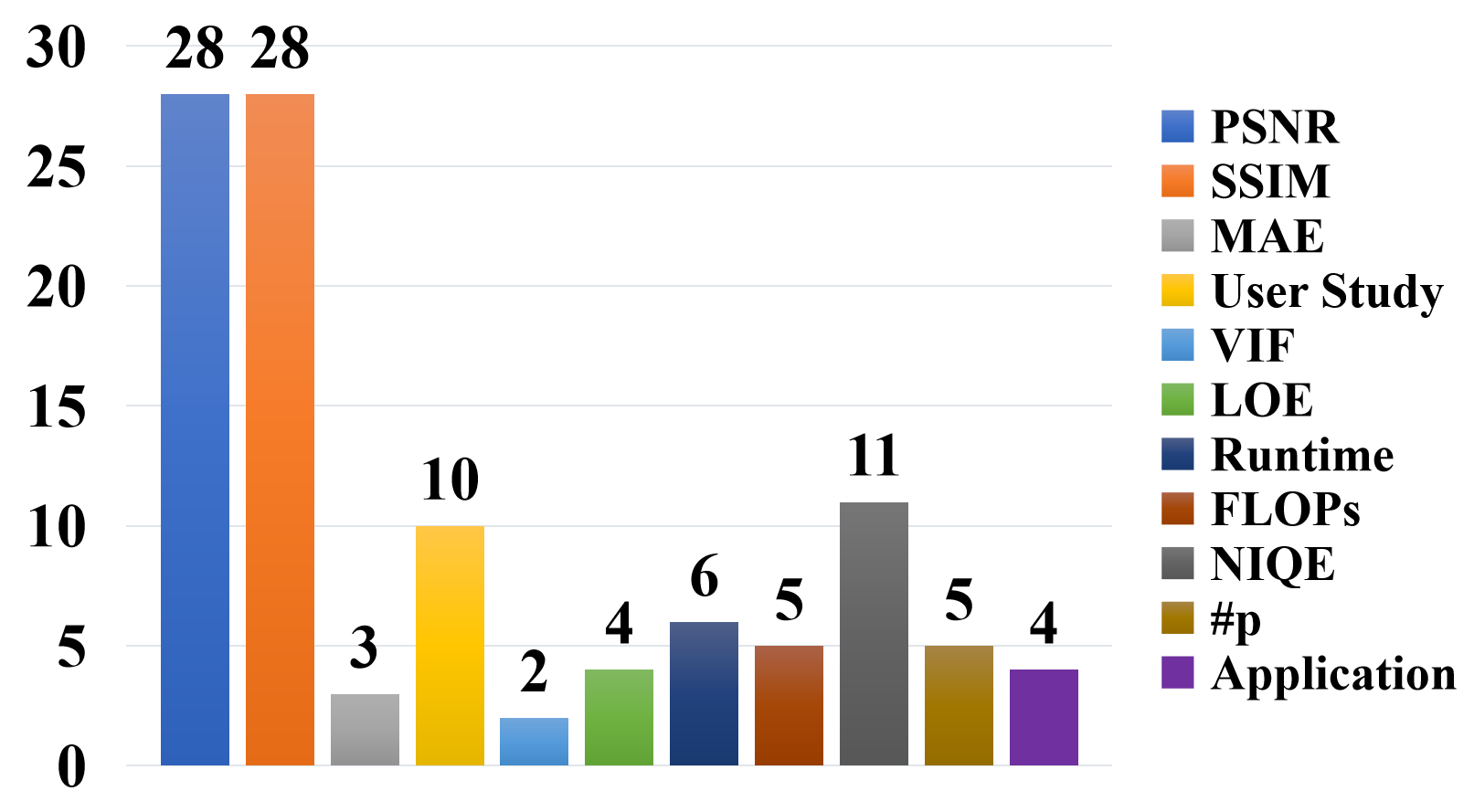}\\
				(e) loss function & (f) training dataset & (g) testing dataset & (h) evaluation metric \\
			\end{tabular}
		\end{center}
		\vspace{-2pt}
		\caption{A statictic analysis of deep learning-based LLIE methods, including learning strategy, network characteristic, Retinex model, data format, loss function, training dataset, testing dataset, and evaluation metric. Best viewed by zooming in.}
		\label{fig:statistic}
		\vspace{-4pt}
	\end{figure*}

According to different learning strategies, we categorize existing LLIE methods into supervised learning, reinforcement learning, unsupervised learning, zero-shot learning, and semi-supervised learning. 
A statistic analysis from different perspectives is presented in Figure \ref{fig:statistic}. In what follows, we review some representative methods of each strategy.

\noindent
\textbf{Supervised Learning.} 
For supervised learning-based LLIE methods,  they are further divided into end-to-end, deep Retinex-based, and realistic data-driven methods.

The first deep learning-based LLIE method LLNet~\cite{LLNet} employs a variant of stacked-sparse
denoising autoencoder \cite{Jain2008} to brighten and denoise low-light images simultaneously.  
This pioneering work inspires the usage of end-to-end networks in LLIE. 
Lv et al. \cite{LvBMVC2018} propose an end-to-end multi-branch enhancement network (MBLLEN). The MBLLEN  improves the performance of LLIE via extracting effective feature representations by a feature extraction module, an enhancement module, and a fusion module. 
The same authors  \cite{LvACM20} propose other three subnetworks including an Illumination-Net, a Fusion-Net, and a Restoration-Net to further improve the performance. 
Ren et al. \cite{RenTIP2019} design a more complex end-to-end network that comprises an encoder-decoder network for image content enhancement and a recurrent neural network for image edge enhancement.  
Similar to Ren et al. \cite{RenTIP2019}, Zhu et al. \cite{ZhuAAAI20} propose a method called EEMEFN. The EEMEFN consists of two stages: multi-exposure fusion and edge enhancement. 
A multi-exposure fusion network, TBEFN~\cite{TBEFN}, is proposed for LLIE. The TBEFN estimates a transfer function in two branches, of which two enhancement results can be obtained. At last, a simple average scheme is employed to fuse these two images and further refine the result via a refinement unit. 
In addition, pyramid network (LPNet) \cite{LiTMM20}, residual network \cite{WangTIP2020}, and Laplacian pyramid \cite{DSLR} (DSLR)
are introduced into LLIE. These methods learn to effectively and efficiently integrate feature representations via commonly used end-to-end network structures for LLIE.
Based on the observation that noise exhibits different levels of contrast in different frequency layers, Xu et al. \cite{Xu2020CVPR} proposed a frequency-based decomposition-and-enhancement network. This network recovers image contents with noise suppression in the low-frequency layer while inferring the details in the high-frequency layer. 
Recently, a progressive-recursive low-light image enhancement network \cite{PRIEN} is proposed, which uses a recursive unit to gradually enhance the input image.
To solve temporal instability when handling low-light videos, Zhang et al. \cite{ZhangCVPR21} propose to learn and infer motion field from a single image then enforce temporal consistency.

In comparison to directly learning an enhanced result in an end-to-end network,  deep Retinex-based methods enjoy better enhancement performance in most cases owing to the physically explicable Retinex theory \cite{Retinex,Jobson1997}. Deep Retinex-based methods usually separately enhance the illuminance component and the reflectance components via specialized subnetworks.
A Retinex-Net \cite{ChenBMVC18} is proposed, which includes a Decom-Net that splits the input image into light-independent reflectance and structure-aware smooth illumination and an Enhance-Net that adjusts the illumination map for low-light enhancement. Recently, the Retinex-Net \cite{ChenBMVC18} is extended by adding new constraints and advanced network designs for better enhancement performance \cite{YangTIP21R}.
To reduce the computational burden, Li et al. \cite{LightenNet} propose a lightweight LightenNet for weakly illuminated image enhancement, which only consists of four layers. 
The LightenNet takes a weakly illuminated image as the input and then estimates its illumination map. Based on the Retinex theory \cite{Retinex,Jobson1997}, the enhanced image is obtained by dividing the input image by the illumination map.
To accurately estimate the illumination map, Wang et al. \cite{Wang2019} extract the global and local features to learn an image-to-illumination mapping by their proposed DeepUPE network.
Zhang et al. \cite{ZhangACM19} separately develop three subnetworks for layer decomposition, reflectance restoration, and illumination adjustment, called KinD. 
Furthermore, the authors alleviate the visual defects left in the results of KinD \cite{ZhangACM19} by a multi-scale illumination attention module. The improved KinD is called KinD++ \cite{GuoIJCV2020}.
To solve the issue that the noise is omitted in the deep Retinex-based methods,
Wang et al. \cite{WangACM19} propose a progressive Retinex network, where an IM-Net estimates the illumination and a NM-Net estimates the noise level. These two subnetworks work in a progressive mechanism until obtaining stable results. 
Fan et al. \cite{FanACM20} integrate semantic segmentation and Retinex model for further improving the enhancement performance in real cases. The core idea is to use semantic prior to guide the enhancement of both the illumination component and the reflectance component.

	Although some methods can achieve decent performance, they show poor generalization capability in real low-light cases due to the usage of synthetic training data. To solve this issue, some works attempt to generate more realistic training data or capture real data.
	Cai et al. \cite{CaiTIP2018}  build a multi-exposure image dataset, where the low-contrast images of different exposure levels have their corresponding high-quality reference images. 
	Each high-quality reference image is obtained by subjectively selecting the best output from 13 results enhanced by different methods. Moreover, a frequency decomposition network is trained on the built dataset and separately enhances the high-frequency layer and the low-frequency layer via a two-stage structure. 
	Chen et al. \cite{Chen2018} collect a real low-light image dataset (SID) and 
	train the U-Net \cite{Unet} to learn a mapping from low-light raw data to the corresponding long-exposure high-quality reference image.
	Further,  Chen et al. \cite{ChenICCV19}  extend the SID dataset to low-light videos (DRV). The DRV contains static videos with the corresponding long-exposure ground truths. To ensure the generalization capability of processing the videos of dynamic scenes, a siamese network is proposed. 
	To enhance the moving objects in the dark, Jiang and Zheng~\cite{JZICCV19} design a co-axis optical system to capture temporally synchronized and spatially aligned low-light and well-lighted video pairs (SMOID). 
	Unlike the DRV video dataset \cite{ChenICCV19},  the SMOID video dataset contains dynamic scenes. To learn the mapping from raw low-light video to well-lighted video, a 3D U-Net-based network is proposed.
	Considering the limitations of previous low-light video datasets such as DRV dataset \cite{ChenICCV19} only containing statistic videos and SMOID dataset \cite{JZICCV19} only having 179 video pairs, 
	Triantafyllidou et al.~\cite{TriantafyllidouECCV} propose a low-light video synthesis pipeline, dubbed SIDGAN. The SIDGAN can produce dynamic video data (raw-to-RGB) by a semi-supervised dual CycleGAN with intermediate domain mapping. 
	To train this pipeline, the real-world videos are collected from Vimeo-90K dataset \cite{Vimeo-90K}. The low-light raw video data and the corresponding long-exposure images are sampled from DRV dataset \cite{ChenICCV19}. 
	With the synthesized training data, this work adopts the same U-Net network as Chen et al. \cite{Chen2018} for low-light video enhancement.

	\noindent
	\textbf{Reinforcement Learning.}
	Without paired training data, Yu et al. \cite{YuNips18}  learn to expose photos with reinforcement adversarial learning, named DeepExposure. Specifically, an input image is first segmented into sub-images according to exposures. For each sub-image, local exposure is learned by the policy network sequentially based on reinforcement learning. The reward evaluation function is approximated by adversarial learning. At last, each local exposure is employed to retouch the input, thus obtaining multiple retouched images under different exposures. The final result is achieved by fusing these images.

	\noindent
	\textbf{Unsupervised Learning.}
	Training a deep model on paired data may result in overfitting and limited generalization capability. To solve this issue, an unsupervised learning method named EnligthenGAN \cite{EnlightenGAN} is proposed.
	The EnlightenGAN adopts an attention-guided U-Net \cite{Unet} as the generator and uses the global-local discriminators to ensure the enhanced results look like realistic normal-light images. 
	In addition to global and local adversarial losses, the global and local self feature preserving losses are proposed to preserve the image content before and after the enhancement. This is a key point for the stable training of such a one-path Generative Adversarial Network (GAN) structure.

	\noindent
	\textbf{Zero-Shot Learning.} 
	The supervised learning, reinforcement learning, and unsupervised learning methods either have limited generalization capability or suffer from unstable training. To remedy these issues, zero-shot learning is proposed to learn the enhancement solely from the testing images. 
	Note that the concept of zero-shot learning in the low-level vision tasks is used to emphasize that the method does not require paired or unpaired training data, which is different from its definition in high-level visual tasks. 
	Zhang et al. \cite{ZhangACM191} propose a zero-shot learning method, called ExCNet, for back-lit image restoration. 
	A network is first used to estimate the S-curve that best fits the input image. Once the S-curve is estimated, the input image is separated into a base layer and a detail layer using the guided filter \cite{He2011}. Then the base layer is adjusted by the estimated S-curve. Finally, the Weber contrast \cite{Whittle1994} is used to fuse the detailed layer and the adjusted base layer. To train the ExCNet, the authors formulate the loss function as a block-based energy minimization problem. 
	Zhu et al. \cite{RRDNet} propose a three-branch CNN, called RRDNet, for underexposed images restoration. The RRDNet decomposes an input image into illumination, reflectance, and noise via iteratively minimizing specially designed loss functions. 
	To drive the zero-shot learning, a combination of Retinex reconstruction loss, texture enhancement loss, and illumination-guided noise estimation loss is proposed. 
	Zhao et al. \cite{RetinexDIP} perform Retinex decomposition via neural networks and then enhance the low-light image based on the Retinex model, called RetinexDIP. Inspired by Deep Image Prior (DIP) \cite{DIP}, RetinexDIP generates the reflectance component and illumination component of an input image by randomly sampled white noise, in which the component characteristics-related losses such as illumination smoothness are used for training.
	Liu et al. \cite{RUAS}  propose a Retinex-inspired unrolling method for LLIE, in which the cooperative architecture search is used to discover lightweight prior architectures of basic blocks and non-reference losses are used to train the network.
	Different from the image reconstruction-based methods \cite{LLNet,LvBMVC2018,RenTIP2019,DSLR,ChenBMVC18,ZhangACM19,GuoIJCV2020}, 
	a deep curve estimation network, Zero-DCE \cite{ZeroDCE}, is proposed. Zero-DCE formulates the light enhancement as a task of image-specific curve estimation, which takes a low-light image as input and produces high-order curves as its output. These curves are used for pixel-wise adjustment on the dynamic range of the input to obtain an enhanced image. Further, an accelerated and lightweight version is proposed, called Zero-DCE++ \cite{ZeroDCE++}. 
	Such curve-based methods do not require any paired or unpaired data during training. They achieve zero-reference learning via a set of non-reference loss functions. Besides, unlike the image reconstruction-based methods that need high computational resources, the image-to-curve mapping only requires lightweight networks, thus achieving a fast inference speed. 
	
	\noindent
	\textbf{Semi-Supervised Learning.}
	To combine the strengths of supervised learning and unsupervised learning, semi-supervised learning has been proposed in recent years.
	Yang et al. \cite{YangCVRP20} propose a semi-supervised deep recursive band network (DRBN). The DRBN first recovers a linear band representation of an enhanced image under supervised learning, and then obtains an improved one by recomposing the given bands via a learnable linear transformation based on unsupervised adversarial learning.  
	The DRBN is extended by introducing  Long Short Term Memory (LSTM) networks and an image quality assessment network pre-trained on an aesthetic visual analysis dataset, which achieves better enhancement performance \cite{YangTIP21RR}.

	Observing Figure \ref{fig:statistic}(a), we can find that supervised learning is the mainstream among deep learning-based LLIE methods, of which the percentage reaches 73\%. This is because supervised learning is relatively easy when paired training data such as LOL \cite{ChenBMVC18}, SID \cite{Chen2018} and diverse low-/normal-light image synthesis approaches are used.  
	However, supervised learning-based methods suffer from some challenges: 
	\textbf{1)} collecting a large-scale paired dataset that covers diverse real-world low-light conditions is difficult, 
	\textbf{2)}  synthetic low-light images do not accurately represent real-world illuminance conditions such as spatially varying lighting and different levels of noise, and 
	\textbf{3)} training a deep model on paired data may result in limited generalization to real-world images of diverse illumination properties. 
	
	Therefore, some methods adopt unsupervised learning, reinforcement learning, semi-supervised learning, and zero-shot learning to bypass the challenges in supervised learning. Although these methods achieve competing performance, they still suffer from some limitations: 
	\textbf{1)} for unsupervised learning/semi-supervised learning methods, how to implement stable training, avoid color deviations, and build the relations of cross-domain information challenges current methods,
	\textbf{2)} for reinforcement learning methods, designing an effective reward mechanism and implementing efficient and stable training are intricate, and 
	\textbf{3)} for zero-shot learning methods, the design of non-reference losses is non-trivial when the color preservation, artifact removal, and gradient back-propagation should be taken into account.

	\begin{table*}
		\rowcolors{1}{gray!20}{white}
		\centering
		\caption{
			{Summary of essential characteristics of representative deep learning-based methods. ``Retinex'' indicates whether the models are Retinex-based or not. ``simulated'' means the testing data are simulated by the same approach as the synthetic training data. ``self-selected'' stands for the real-world images selected by the authors. ``\#P''represents the number of trainable parameters. ``-'' means this item is not available or not indicated in the paper.}
		}
		\vspace{-6pt}
		\label{table:methods}
		\begin{threeparttable}
			\resizebox{1\textwidth}{!}{
				\setlength\tabcolsep{2pt}
				\renewcommand\arraystretch{0.98}
				\begin{tabular}{c|r||c|c|c|c|c|c|c|c|c}
					\hline
					&\textbf{Method}~~~~~~~~~&\textbf{Learning} &\textbf{Network Structure} &\textbf{Loss Function}
					&\textbf{Training Data} & \textbf{Testing Data} &\textbf{Evaluation Metric} & \textbf{Format} &\textbf{Platform} &\textbf{Retinex}\\
					\hline
					\hline
					\multirow{1}{*}{\rotatebox{90}{\textbf{2017}}}
					&LLNet~\cite{LLNet} &SL &SSDA &SRR loss
					&\tabincell{c}{simulated by \\Gamma Correction \& \\Gaussian Noise} &\tabincell{c}{simulated \\ self-selected} &\tabincell{c}{PSNR SSIM} &RGB &Theano &\\
					\hline
					\hline
					\multirow{1}{*}{\rotatebox{90}{\textbf{2018}}}
					&LightenNet~\cite{LightenNet} &SL &four layers &$L_{2}$ loss
					&\tabincell{c}{simulated by \\random illumination values} &\tabincell{c}{simulated\\self-selected} &\tabincell{c}{PSNR MAE\\SSIM \\User Study} &RGB &\tabincell{c}{Caffe\\MATLAB}&\checkmark\\
					&Retinex-Net~\cite{ChenBMVC18} &SL &
					multi-scale network &\tabincell{c}{$L_{1}$ loss smoothness loss \\invariable reflectance loss}
					&\tabincell{c}{LOL\\simulated by \\adjusting histogram} &self-selected &- &RGB &TensorFlow&\checkmark\\
					&MBLLEN~\cite{LvBMVC2018} &SL &multi-branch fusion &\tabincell{c}{SSIM loss region loss\\perceptual loss}
					&\tabincell{c}{simulated by \\Gamma Correction \& \\Poisson Noise} &\tabincell{c}{simulated\\ self-selected} &\tabincell{c}{PSNR SSIM\\AB VIF\\LOE TOMI} &RGB &\tabincell{c}{TensorFlow} & \\
					&SCIE~\cite{CaiTIP2018} &SL  &frequency decomposition &\tabincell{c}{$L_{2}$ loss $L_{1}$ loss SSIM loss}
					&SCIE &SCIE &\tabincell{c}{PSNR FSIM\\Runtime FLOPs} &RGB&\tabincell{c}{Caffe\\MATLAB}&\\
					&Chen et al.~\cite{Chen2018} &SL  &U-Net &$L_{1}$ loss
					&SID &SID &\tabincell{c}{PSNR SSIM} &raw&TensorFlow&\\
					&Deepexposure~\cite{YuNips18} &RL  &\tabincell{c}{policy network\\GAN}  &\tabincell{c}{deterministic policy gradient\\adversarial loss}
					&MIT-Adobe FiveK&MIT-Adobe FiveK &\tabincell{c}{PSNR SSIM} &raw&TensorFlow &\\
					\hline
					\hline
					\multirow{1}{*}{\rotatebox{90}{\textbf{2019}}}
					&Chen et al.~\cite{ChenICCV19} &SL &siamese network &\tabincell{c}{$L_{1}$ loss\\ self-consistency loss}
					&DRV &DRV &\tabincell{c}{PSNR\\ SSIM\\ MAE} &raw &TensorFlow&\\
					&Jiang and Zheng~\cite{JZICCV19}  &SL &3D U-Net &$L_{1}$ loss
					&SMOID &SMOID &\tabincell{c}{PSNR SSIM MSE} &raw &TensorFlow &\\
					&DeepUPE~\cite{Wang2019}  &SL &illumination map &\tabincell{c}{$L_{1}$ loss color loss\\smoothness loss}
					&retouched image pairs&MIT-Adobe FiveK&\tabincell{c}{PSNR SSIM\\User Study} &RGB &TensorFlow &\checkmark\\
					&KinD~\cite{ZhangACM19} &SL &\tabincell{c}{three subnetworks\\U-Net} &\tabincell{c}{reflectance similarity loss\\illumination smoothness loss\\mutual consistency loss\\$L_{1}$ loss $L_{2}$ loss SSIM loss\\texture similarity loss\\illumination adjustment loss}
					&LOL &\tabincell{c}{LOL LIME\\NPE MEF} &\tabincell{c}{PSNR SSIM\\LOE NIQE} &RGB &TensorFlow&\checkmark\\
					&Wang et al.~\cite{WangACM19} &SL &\tabincell{c}{two subnetworks\\pointwise Conv} &$L_{1}$ loss 
					&\tabincell{c}{simulated by \\camera imaging model} &\tabincell{c}{IP100 FNF38\\MPI LOL NPE} &\tabincell{c}{PSNR SSIM\\NIQE} &RGB &Caffe &\checkmark \\
					&Ren et al.~\cite{RenTIP2019} &SL &\tabincell{c}{U-Net like network\\RNN dilated Conv} &\tabincell{c}{$L_{2}$ loss perceptual loss\\adversarial loss} 
					&\tabincell{c}{MIT-Adobe FiveK\\with Gamma correction \\   \& Gaussion noise} &\tabincell{c}{simulated \\self-selected DPED} &\tabincell{c}{PSNR SSIM\\Runtime} &RGB &Caffe&\\
					&EnlightenGAN~\cite{EnlightenGAN} &UL &U-Net like network &\tabincell{c}{adversarial loss\\self feature preserving loss}
					&unpaired real images &\tabincell{c}{NPE LIME\\MEF DICM\\VV BBD-100K\\ExDARK} &\tabincell{c}{User Study NIQE\\Classification} &RGB &PyTorch&\\
					&ExCNet.~\cite{ZhangACM191}  &ZSL &
					\tabincell{c}{fully connected layers} &energy minimization loss &real images
					&$IE_{ps}D$  &\tabincell{c}{User Study\\CDIQA LOD} &RGB &PyTorch&\\
					\hline
					\hline
					\multirow{1}{*}{\rotatebox{90}{\textbf{2020}}}
					&Zero-DCE~\cite{ZeroDCE} &ZSL &U-Net like network &\tabincell{c}{spatial consistency loss\\exposure control loss\\color constancy loss\\illumination smoothness loss}
					&SICE &\tabincell{c}{SICE NPE\\LIME MEF\\DICM VV\\ DARK FACE} &\tabincell{c}{User Study PI\\PNSR SSIM\\MAE Runtime\\Face detection} &RGB &PyTorch &\\
					&DRBN~\cite{YangCVRP20}  &SSL &recursive network &\tabincell{c}{SSIM loss perceptual loss\\adversarial loss}
					&\tabincell{c}{LOL\\images selected by MOS} &LOL &\tabincell{c}{PSNR SSIM\\SSIM-GC} &RGB &PyTorch &\\
					&Lv et al.~\cite{LvACM20}&SL &U-Net like network &\tabincell{c}{Huber loss\\SSIM loss\\perceptual loss\\illumination smoothness loss}
					&\tabincell{c}{simulated by a\\ retouching module} &\tabincell{c}{LOL SICE \\DeepUPE} &\tabincell{c}{User Study PSNR\\SSIM VIF\\LOE NIQE\\ \#P Runtime\\Face detection} &RGB&\tabincell{c}{TensorFlow}&\checkmark\\
					&Fan et al.~\cite{FanACM20} &SL &\tabincell{c}{four subnetworks\\U-Net like network\\feature modulation} &\tabincell{c}{mutual smoothness loss\\reconstruction loss\\illumination smoothness loss\\cross entropy loss\\ consistency loss\\SSIM loss\\gradient loss\\ratio learning loss}
					&\tabincell{c}{simulated by \\illumination adjustment, \\slight color distortion, \\and noise simulation} &\tabincell{c}{simulated\\self-selected} &\tabincell{c}{PSNR SSIM\\NIQE} &RGB & - &\checkmark\\
					&Xu et al.~\cite{Xu2020CVPR}&SL &\tabincell{c}{frequency decomposition\\U-Net like network} &\tabincell{c}{$L_{2}$ loss\\perceptual loss}
					&SID in RGB&\tabincell{c}{SID in RGB \\self-selected} &\tabincell{c}{PSNR SSIM} &RGB &PyTorch&\\
					&EEMEFN~\cite{ZhuAAAI20}&SL &\tabincell{c}{U-Net like network\\edge detection network} & \tabincell{c}{$L_{1}$ loss\\weighted cross-entropy loss}
					&SID &SID &\tabincell{c}{PSNR SSIM} &raw& \tabincell{c}{TensorFlow\\PaddlePaddle}&\\
					&DLN~\cite{WangTIP2020}&SL &\tabincell{c}{residual learning\\interactive factor\\back projection network} &\tabincell{c}{SSIM loss\\total variation loss}
					&\tabincell{c}{simulated by \\illumination adjustment, \\slight color distortion,\\and noise simulation} &\tabincell{c}{simulated \\LOL} &\tabincell{c}{User Study PSNR\\SSIM NIQE} &RGB &PyTorch&\\
					&LPNet~\cite{LiTMM20}&SL &\tabincell{c}{pyramid network} &\tabincell{c}{$L_{1}$ loss\\perceptual loss\\luminance loss}
					&\tabincell{c}{LOL SID in RGB\\MIT-Adobe FiveK} &\tabincell{c}{LOL SID in RGB\\MIT-Adobe FiveK\\ MEF NPE DICM VV} &\tabincell{c}{PSNR SSIM\\NIQE \#P\\FLOPs Runtime} &RGB &PyTorch&\\
					&SIDGAN~\cite{TriantafyllidouECCV} &SL &U-Net &CycleGAN loss
					&SIDGAN &SIDGAN &\tabincell{c}{PSNR SSIM\\TPSNR TSSIM ATWE}  &raw &\tabincell{c}{TensorFlow}&\\
					&RRDNet \cite{RRDNet}&ZSL&three subnetworks &\tabincell{c}{retinex reconstruction loss\\texture enhancement loss\\noise estimation loss}
					&-&\tabincell{c}{NPE LIME\\MEF DICM} &\tabincell{c}{NIQE CPCQI} &RGB&PyTorch &\checkmark\\
					&TBEFN~\cite{TBEFN}&SL&\tabincell{c}{three stages\\U-Net like network} &\tabincell{c}{SSIM loss\\perceptual loss\\smoothness loss}
					&\tabincell{c}{SCIE \\LOL} &\tabincell{c}{SCIE LOL\\DICM MEF\\NPE VV} &\tabincell{c}{PSNR SSIM\\NIQE Runtime\\ \#P FLOPs} &RGB &TensorFlow&\checkmark\\
					&DSLR~\cite{DSLR}&SL&\tabincell{c}{Laplacian pyramid\\U-Net like network} &\tabincell{c}{$L_{2}$ loss\\Laplacian loss\\color loss}
					&MIT-Adobe FiveK &\tabincell{c}{MIT-Adobe FiveK\\self-selected} &\tabincell{c}{PSNR SSIM\\NIQMC NIQE\\BTMQI CaHDC} &RGB& PyTorch&\\
					\hline
					\hline
					\multirow{1}{*}{\rotatebox{90}{\textbf{2021}}}
					&RUAS~\cite{RUAS}  &ZSL &neural architecture search &\tabincell{c}{cooperative loss\\similar loss\\total variation loss}
					&\tabincell{c}{LOL\\MIT-Adobe FiveK} &\tabincell{c}{LOL\\MIT-Adobe FiveK} &\tabincell{c}{PSNR SSIM\\ Runtime \#P FLOPs} &RGB &PyTorch &\checkmark\\
					&Zhang et al.~\cite{ZhangCVPR21}&SL &U-Net &\tabincell{c}{$L_{1}$ loss\\ consistency loss}
					&\tabincell{c}{simulated by illumination  \\adjustmentand noise simulation} &\tabincell{c}{simulated \\self-selected} &\tabincell{c}{User Study PSNR\\SSIM AB\\MABD WE} &RGB&\tabincell{c}{PyTorch}&\\
					&Zero-DCE++~\cite{ZeroDCE++} &ZSL &U-Net like network &\tabincell{c}{spatial consistency loss\\exposure control loss\\color constancy loss\\illumination smoothness loss}
					&SICE &\tabincell{c}{SICE NPE\\LIME MEF\\DICM VV\\ DARK FACE} &\tabincell{c}{User Study PI\\PNSR SSIM \#P\\MAE Runtime\\Face detection FLOPs} &RGB &PyTorch &\\
					&DRBN~\cite{YangTIP21RR}  &SSL &recursive network &\tabincell{c}{perceptual loss\\detail loss quality loss}
					&LOL &LOL &\tabincell{c}{PSNR SSIM\\SSIM-GC} &RGB &PyTorch &\\
					&Retinex-Net~\cite{YangTIP21R} &SL &
					three subnetworks &\tabincell{c}{$L_{1}$ loss $L_{2}$ loss\\SSIM loss\\total variation loss}
					&\tabincell{c}{LOL\\simulated by adjusting histogram} &\tabincell{c}{LOL simulated\\NPE DICM VV} &\tabincell{c}{PNSR SSIM\\UQI OSS User Study} &RGB &PyTorch&\checkmark\\
					&RetinexDIP\cite{RetinexDIP} &ZSL & encoder-decoder networks &\tabincell{c}{reconstruction loss\\illumination-consistency loss\\reflectnce loss \\illumination smoothness loss}&- &\tabincell{c}{DICM, ExDark\\Fusion LIME\\ NASA NPE VV} &\tabincell{c}{NIQE \\NIQMC CPCQI}  &RGB &PyTorch&\checkmark\\
					&PRIEN~\cite{PRIEN} &SL &recursive network &SSIM loss&\tabincell{c}{MEF LOL\\simulated by adjusting histogram} &\tabincell{c}{LOL LIME\\NPE MEF VV} &\tabincell{c}{PNSR SSIM\\LOE TMQI} &RGB &PyTorch &\\
					\hline
				\end{tabular}
			}
		\end{threeparttable}
	\end{table*}


\section{Technical Review and Discussion}
\label{sec:Technical}
In this section, we first summarize the representative deep learning-based LLIE methods in Table \ref{table:methods}, then analyze and discuss their technical characteristics.

\subsection{Network Structure}
Diverse network structures and designs have been used in the existing models, spanning from the basic U-Net, pyramid network, multi-stage network to frequency decomposition network. 
After analyzing Figure \ref{fig:statistic}(b), it can be observed that the U-Net  and U-Net-like networks are mainly adopted network structures in  LLIE. 
This is because U-Net can effectively integrate multi-scale 
features and employ both low-level and high-level features. Such characteristics are essential for achieving satisfactory low-light enhancement.

Nevertheless, some key issues may be ignored in the current LLIE network structures:
\textbf{1)} after going through several convolutional layers, the gradients of an extremely low light image may vanish during the gradient back-propagation due to its small pixel values.  This would degrade the enhancement performance and affect the convergence of network training, 
\textbf{2)} the skip-connections used in the U-Net-like networks might introduce noise and redundant features into the final results. How to effectively filter out the noise and integrate both low-level and high-level features should be carefully considered, and 
\textbf{3)} although some designs and components are proposed for LLIE, most of them are borrowed or modified from related low-level visual tasks. The characteristics of low-light data should be considered when designing the network structures.

\subsection{Combination of Deep Model and Retinex Theory}
As presented in Figure \ref{fig:statistic}(c), almost 1/3 of methods combine the designs of deep networks with the Retinex theory, e.g., designing different subnetworks to estimate the components of the Retinex model and estimating the illumination map to guide the learning of networks.  Despite such a combination can bridge  deep learning-based and model-based methods, their respective weaknesses may be introduced into the final models: 
\textbf{1)} the ideal assumption that the reflectance is the final enhanced result used in Retinex-based LLIE methods would still affect the final results, and 
\textbf{2)}  the risk of overfitting in deep networks still exists despite the use of Retinex theory. How to cream off the best and filter out the impurities should be carefully considered when researchers combine deep learning with the Retinex theory.

\subsection{Data Format}

As shown in Figure \ref{fig:statistic}(d),  RGB data format dominates most methods as it is commonly found as the final imagery form produced by smartphone cameras, Go-Pro cameras, and drone cameras.  Although raw data are limited to specific sensors such as those based on Bayer patterns, the data cover wider color gamut and higher dynamic range. Hence, deep models trained on raw data usually recover clear details and high contrast, obtain vivid color, reduce the effects of noises and artifacts, and improve the brightness of extremely low-light images. In future research, a smooth transformation from  raw data of different patterns to  RGB format would have the potentials to combine the convenience of RGB data and the advantage of high-quality enhancement of raw data for LLIE.

\subsection{Loss Function}
In Figure\ref{fig:statistic}(e), the commonly adopted loss functions in LLIE models include reconstruction loss ($L_{1}$, $L_{2}$, SSIM), perceptual loss, and smoothness loss. Besides, according to different demands and formulations, color loss, exposure loss, adversarial loss, etc are also adopted. We detail representative loss functions as follows.

\noindent
\textbf{Reconstruction Loss.}
	Different reconstruction losses have their advantages and disadvantages. $L_{2}$ loss tends to penalize larger errors, but is tolerant to small errors. $L_{1}$ loss preserves colors and
	luminance well since an error is weighted equally regardless of the local structure. SSIM loss preserves the structure and texture well. Refer to this research paper \cite{lossTCI17} for detailed analysis.

\noindent
\textbf{Perceptual Loss.}
	Perceptual loss \cite{VGGloss}, particularly the feature reconstruction loss, is proposed to constrain the results similar to the ground truth in the feature space. The loss improves the visual quality of results. It is defined as the Euclidean distance between the feature representations of an enhanced result and those of corresponding ground truth. The feature representations are typically extracted from the VGG network \cite{VGG} pre-trained on ImageNet dataset \cite{ImageNet}.

\noindent
\textbf{Smoothness Loss.}
	To remove noise in the enhanced results or preserve the relationship of neighboring pixels, smoothness loss (TV loss) is often used to constrain the enhanced result or the estimated illumination map.

\noindent
\textbf{Adversarial Loss.}
	To encourage enhanced results to be indistinguishable from reference images,  adversarial learning solves a max-min optimization problem \cite{SRGAN,ESRGAN}.

\noindent
\textbf{Exposure Loss.}
	As one of key non-reference losses, exposure loss measures the exposure levels of enhanced results without paired or unpaired images as reference images.

	The commonly used loss functions in LLIE networks are also employed in image reconstruction networks for image super-resolution \cite{SRCNN}, image denoising \cite{XuICIA2015}, image detraining \cite{FuCVPR17,YangCVPR17,FuIJCV21}, and image deblurring \cite{SunCVPR2015}.
	Different from these versatile losses, the specially designed exposure loss for LLIE inspires the design of non-reference losses. A non-reference loss makes a model enjoying better generalization capability.
	It is an on-going research to consider image characteristics for the design of loss functions.

\subsection{Training Datasets}

Figure \ref{fig:statistic}(f) reports the usage of a variety of paired training datasets for training low-light enhancement networks. These datasets include real-world captured datasets and synthetic datasets. We list them in Table \ref{table:training}. 

\noindent
\textbf{Simulated by Gamma Correction.}
Owing to its nonlinearity and simplicity, 
Gamma correction is used to adjust the luminance or tristimulus values in video or still image systems. It is defined by a power-law expression:
\begin{equation}
	\label{equ_12}
	V_{\text{out}}=AV_{\text{in}}^{\gamma},
\end{equation}
where the input $V_{\text{in}}$ and output $V_{\text{out}}$ are typically in the range of [0,1]. The constant $A$ is set to 1 in the common case. The power $\gamma$ controls the luminance of the output. Intuitively, the input is brightened when  $\gamma<$1 while the input is darkened when $\gamma>$1. The input can be the three RGB channels of an image or the luminance-related channels such as $L$ channel in the CIELab color space and $Y$ channel in the YCbCr color space. After adjusting the luminance-related channel using Gamma correction, the corresponding channels in the color space are adjusted by equal proportion to avoid producing artifacts and color deviations. 

To simulate images taken in real-world low-light scenes,  Gaussian noise, Poisson noise, or realistic noise is added to the Gamma corrected images. The low-light image synthesized using Gamma correction can be expressed as:
\begin{equation}
	\label{equ_13}
	I_{\text{low}}=n(g(I_{\text{in}};\gamma)),
\end{equation}
where $n$ represents the noise model, $g(I_{\text{in}};\gamma)$ represents the Gamma correction function with Gamma value $\gamma$, $I_{\text{in}}$ is a normal-light and high-quality image or luminance-related channel. Although this function produces low-light images of different lighting levels by changing the Gamma value $\gamma$, 
it tends to introduce artifacts and color deviations into the synthetic low-light images due to the nonlinear adjustment.

\begin{table}
	\centering
	\setlength\tabcolsep{5pt}
	\caption{
		{Summary of paired training datasets. `Syn' represents Synthetic.}
	}
	\vspace{-6pt}
	\label{table:training}
	\begin{tabular}{r|c|c|c|c}
		\hline
		\textbf{Name}~~~~~~~~~~~~~~~~~~~~~&\textbf{Number}& 
		\textbf{Format} & \textbf{Real/Syn}&\textbf{Video} \\
		\hline
		Gamma Correction &+$\infty$  &RGB 
		&Syn & \\
		Random Illumination &+$\infty$  &RGB 
		&Syn & \\
		\hline
		LOL \cite{ChenBMVC18} &500  &RGB 
		&Real  &\\
		SCIE \cite{CaiTIP2018} &4,413  &RGB 
		&Real &\\
		VE-LOL-L \cite{IJCVsurvey21}  &2,500  &RGB 
		&Real+Syn  &\\
		MIT-Adobe FiveK \cite{Adobe5K}  &5,000  &raw 
		&Real  &\\
		SID \cite{ChenBMVC18}  &5,094  &raw 
		&Real  &\\
		DRV \cite{ChenICCV19}  &202  &raw 
		&Real  &\checkmark\\
		SMOID \cite{JZICCV19}  &179  &raw 
		&Real  &\checkmark\\
		\hline
	\end{tabular}
	\vspace{-4pt}
\end{table}

\noindent
\textbf{Simulated by Random Illumination.}
According to the Retinex model, an image can be decomposed into a reflectance component and an illumination component. Assuming image content is independent of illumination component and local region in the illumination component have the same intensity, a low-light image can be obtained by 
\begin{equation}
	\label{equ_14}
	I_{\text{low}}=I_{\text{in}}L,
\end{equation}
where $L$ is a random illumination value in the range of [0,1]. Noises can be added to the synthetic image. Such a linear function avoids artifacts, but the strong assumption requires the synthesis to operate only on image patches where local regions have the same brightness. A deep model trained on such image patches may lead to sub-optimal performance due to the negligence of context information.

\noindent
\textbf{LOL.}
LOL \cite{ChenBMVC18} is the first paired low-/normal-light image dataset taken in real scenes. The low-light images are collected by changing the exposure time and ISO. LOL contains 500 pairs of low-/normal-light images of size 400$\times$600 saved in RGB format. 

\noindent
\textbf{SCIE.}
SCIE is a multi-exposure image dataset of low-contrast and good-contrast image pairs. It includes multi-exposure sequences of 589 indoor and outdoor scenes. Each sequence has 3 to 18 low-contrast images of different exposure levels, thus containing 4,413 multi-exposure images in total. The 589 high-quality reference images are obtained by selecting from the results of 13 representative enhancement algorithms. That is many multi-exposure images have the same high-contrast reference image.
The image resolutions are between 3,000$\times$2,000 and 
6,000$\times$4,000. The images in SCIE are saved in RGB format.

\noindent
\textbf{MIT-Adobe FiveK.}
MIT-Adobe FiveK \cite{Adobe5K} was collected for global tone adjustment but has been used in LLIE. This is because the input images have low light and low contrast. MIT-Adobe FiveK contains 5,000 images, each of which is retouched by 5 trained photographers towards visually pleasing renditions, akin to a postcard.  The images are all in raw format. To train the networks that can handle images of RGB format, one needs to use Adobe Lightroom to pre-process the images and save them as RGB format following a dedicated pipeline\footnote{\url{https://github.com/nothinglo/Deep-Photo-Enhancer/issues/38\#issuecomment-449786636}}.
The images are commonly resized to have a long edge of 500 pixels.

\noindent
\textbf{SID.}
SID \cite{Chen2018} contains 5,094 raw short-exposure images, each with a corresponding long-exposure reference image. The number of distinct long-exposure reference images is 424. In other words, multiple short-exposure images correspond to the same long-exposure reference image. 
The images were taken using two cameras: Sony $\alpha$7S II and Fujifilm X-T2 in both indoor and outdoor scenes. Thus, the images have different sensor patterns (Sony camera' Bayer sensor and Fuji camera's APS-C X-Trans sensor). 
The resolution is 4,240$\times$2,832 for Sony and 6,000$\times$4,000 for Fuji. Usually, the long-exposure images are processed by libraw (a raw image processing library) and saved in the RGB color space, and randomly cropped 512$\times$512 patches for training.

\noindent
\textbf{VE-LOL.}  VE-LOL \cite{IJCVsurvey21} consists of two subsets: paired VE-LOL-L that is used for training and evaluating LLIE methods and unpaired VE-LOL-H that is used for evaluating the effect of LLIE methods on face detection. Specifically, VE-LOL-L includes 2,500 paired images. Among them, 1,000 pairs are synthetic, while 1,500 pairs are real. VE-LOL-H includes 10,940 unpaired images, where human faces are manually annotated with bounding boxes.  

\noindent
\textbf{DRV.}
DRV \cite{ChenICCV19} contains 202 static raw videos, each of which has a corresponding long-exposure ground truth. Each video was taken at approximately 16 to 18 frames per second in a continuous shooting mode and is with up to 110 frames. The images were taken by a Sony RX100 VI camera in both indoor and outdoor scenes, thus all in  raw format of Bayer pattern. The resolution is 3,672$\times$5,496.

\noindent
\textbf{SMOID.}
SMOID \cite{JZICCV19} contains 179 pairs of videos taken by a co-axis optical system, each of which has 200 frames. Thus, SMOID includes 35,800 extremely low light raw data of Bayer pattern and their corresponding well-lightened RGB counterparts. SMOID consists of moving vehicles and pedestrians under different illumination conditions.

Some issues challenge the aforementioned paired training datasets: 
\textbf{1)} deep models trained on synthetic data may introduce artifacts and color deviations when processing real-world images and videos due to the gap between synthetic data and real data, 
\textbf{2)} the scale and diversity of real training data are unsatisfactory, thus some methods incorporate synthetic data to augment the training data. This may lead to sub-optimal enhancement, and 
\textbf{3)} the input images and corresponding ground truths may exist misalignment due to the effects of motion, hardware, and environment. This would affect the performance of deep networks trained using pixel-wise loss functions.

\subsection{Testing Datasets}
In addition to the testing subsets in the paired datasets \cite{ChenBMVC18,CaiTIP2018,Adobe5K,Chen2018,ChenICCV19,JZICCV19,IJCVsurvey21}, there are several testing data collected from related works or commonly used for experimental comparisons. 
Besides, some datasets such as face detection in the dark \cite{Yuan2019} and detection and recognition in low-light images \cite{ExDark} are employed  to test the effects of LLIE on high-level visual tasks. We summarize the commonly used testing datasets in Table \ref{table:testing} and introduce the representative testing datasets as follows. 

\begin{table}
	\centering
	\caption{
		{Summary of testing datasets.}
	}
	\vspace{-6pt}
	\label{table:testing}
	\begin{tabular}{r|c|c|c|c}
		\hline
		\textbf{Name}~~~~~~~~~&\textbf{Number}& 
		\textbf{Format} & \textbf{Application}&\textbf{Video} \\
		\hline
		LIME \cite{LIME} &10 &RGB & & \\
		NPE \cite{Wang2013} &84  &RGB & & \\
		MEF \cite{LeeTIP2012} &17  &RGB  &  &\\
		DICM \cite{LeeTIP2013}  &64  &RGB & &\\
		VV\tablefootnote{\url{https://sites.google.com/site/vonikakis/datasets}} &24  &RGB &  &\\
		\hline
		BBD-100K \cite{BDD100K}   &10,000  &RGB &\checkmark  &\checkmark\\
		ExDARK \cite{ExDark} &7,363  &RGB  &\checkmark  &\\
		DARK FACE \cite{Yuan2019}  &6,000  &RGB  &\checkmark  &\\
		VE-LOL-H \cite{IJCVsurvey21}  &10,940  &RGB &\checkmark  &\\
		\hline
	\end{tabular}
	\vspace{-4pt}
\end{table}

\noindent
\textbf{BBD-100K.}
BBD-100K \cite{BDD100K} is the largest driving video dataset with 10,000 videos taken over 1,100-hour driving experience across many different times in the day, weather conditions, and driving scenarios, and 10 tasks annotations. The videos taken at nighttime in BBD-100K are used to validate the effects of LLIE on high-level visual tasks and the enhancement performance in real scenarios.

\noindent
\textbf{ExDARK.}
ExDARK~\cite{ExDark} dataset is built for object detection and recognition in low-light images. ExDARK dataset contains 7,363 low-light images from extremely low-light environments to twilight with 12 object classes annotated with image class labels and local object bounding boxes.

\noindent
\textbf{DARK FACE.}
DARK FACE \cite{Yuan2019} dataset contains 6,000 low-light images captured during the nighttime, each of which is labeled with bounding boxes of the human face.

From Figure \ref{table:testing}(g) and Table \ref{table:methods}, we can observe that one prefers using the self-collected testing data in the experiments. The main reasons lie into three-fold: 
\textbf{1)} besides the test partition of paired datasets, there is no acknowledged benchmark for evaluations, \textbf{2)}
the commonly used test sets suffer from some shortcomings such as small scale (some test sets contain 10 images only), repeated content and illumination properties, and unknown experimental settings, and\textbf{ 3)} some of the commonly used testing data are not originally collected for evaluating LLIE. In general, current testing datasets may lead to bias and unfair comparisons.

\subsection{Evaluation Metrics}
Besides human perception-based subjective evaluations, image quality assessment (IQA) metrics, including both full-reference and non-reference IQA metrics, are able to evaluate image quality objectively. In addition, user study,  number of trainable parameters, FLOPs, runtime, and applications also reflect the performance of LLIE models, as shown in Fig. \ref{fig:statistic}(h). 
We will detail them as follows.

\noindent
\textbf{PSNR and MSE.}
PSNR and MSE are widely used IQA metrics. They are always non-negative, and values closer to infinite (PSNR) and  zero (MSE) are better.
Nevertheless, the pixel-wise PSNR and MSE may provide an inaccurate indication of the visual perception of image quality since they neglect the relation of neighboring pixels.

\noindent
\textbf{MAE.}
MAE represents the mean absolute error, serving as a measure of errors between paired observations. 
The smaller the MAE value is, the better similarity is.

\noindent
\textbf{SSIM.}
SSIM is used to measure the similarity between two images. It is a perception-based model that considers image degradation as perceived change in structural information. 
The value 1 is only reachable in the case of two identical sets of data, indicating perfect structural similarity.

\noindent
\textbf{LOE.}
LOE represents the lightness order error that reflects the naturalness of an enhanced image.
For LOE, the smaller the LOE value is, the better the lightness order is preserved.

\noindent
\textbf{Application.} 
Besides improving the visual quality, one of the purposes of image enhancement is to serve high-level visual tasks. Thus, the effects of LLIE on high-level visual applications are commonly examined to validate the performance of different methods.

The current evaluation approaches used in LLIE need to be improved in several aspects: 
\textbf{1)} although the PSNR, MSE, MAE, and SSIM are classic and popular metrics, they are still far from capturing real visual perception of human,
\textbf{2)} some metrics are not originally designed for low-light images. They are used for assessing the fidelity of image information and contrast. Using these metrics may reflect the image quality, but they are far from the real purpose of low-light enhancement,
\textbf{3)} metrics especially designed for low-light images are lacking, except for the LOE metric. Moreover, there is no metric for evaluating low-light video enhancement, and
\textbf{4)} a metric that can balance both the human vision and the machine perception is expected.


\section{Benchmarking and Empirical Analysis}
\label{sec:evaluation}

This section provides empirical analysis and highlights some key challenges in deep learning-based LLIE. To facilitate the analysis, we propose a  low-light image and video dataset to examine the performance of different solutions. We also develop the first online platform, where the results of LLIE models can be produced via a user-friendly web interface. In this section, we conduct extensive evaluations on several benchmarks and our proposed dataset. 

In the experiments, we compare 13 representative RGB format-based methods, including eight supervised learning-based methods (LLNet \cite{LLNet}, LightenNet \cite{LightenNet},  Retinex-Net \cite{ChenBMVC18}, MBLLEN \cite{LvBMVC2018},  KinD \cite{ZhangACM19},  KinD++ \cite{GuoIJCV2020}, TBEFN \cite{TBEFN}, DSLR \cite{DSLR}), one unsupervised learning-based method (EnlightenGAN \cite{EnlightenGAN}), one semi-supervised learning-based method (DRBN \cite{YangCVRP20}), and three zero-shot learning-based methods (ExCNet \cite{ZhangACM191}, Zero-DCE \cite{ZeroDCE}, RRDNet \cite{RRDNet}). Besides, we also compare two raw format-based methods, including SID \cite{Chen18} and EEMEFN \cite{ZhuAAAI20}. Note that RGB format-based methods dominate LLIE. Moreover, most raw format-based methods do not release their code. Thus, we choose two representative methods to provide empirical analysis and insights. For all compared methods, we use the publicly available code to produce their results for fair comparisons.

\begin{table}
	\centering
	\caption{{Summary of LLIV-Phone dataset.  LLIV-Phone dataset contains 120 videos (45,148 images) taken by 18 different mobile phones' cameras. ``\#Video'' and  ``\#Image'' represent the number of videos and images, respectively.}}
	\vspace{-6pt}
	\label{table:LLIVPhone}
	\begin{tabular}{r|c|c|c}
		\hline
		\textbf{Phone's Brand}&\textbf{\#Video}& 
		\textbf{\#Image} &\textbf{Resolution}\\
		\hline
		iPhone 6s &4 &1,029 &1920$\times$1080\\
		iPhone 7 &13 &6,081 &1920$\times$1080\\
		iPhone7 Plus &2 &900 &1920$\times$1080\\
		iPhone8 Plus &1 &489 &1280$\times$720\\
		iPhone 11 &7 &2,200 &1920$\times$1080\\
		iPhone 11 Pro &17 &7,739 &1920$\times$1080\\
		iPhone XS &11 &2,470 &1920$\times$1080\\
		iPhone XR &16 &4,997 &1920$\times$1080\\
		iPhone SE &1 &455 &1920$\times$1080\\
		Xiaomi Mi 9 &2 &1,145 &1920$\times$1080\\
		Xiaomi Mi Mix 3 &6 &2,972 &1920$\times$1080\\
		Pixel 3&4 &1,311 &1920$\times$1080\\
		Pixel 4&3 &1,923 &1920$\times$1080\\
		Oppo R17 &6 &2,126 &1920$\times$1080\\
		Vivo Nex &12 &4,097 &1280$\times$720\\
		LG M322&2 &761 &1920$\times$1080\\
		OnePlus 5T &1 &293 &1920$\times$1080\\
		Huawei Mate 20 Pro & 12&4,160 &1920$\times$1080\\
		\hline
	\end{tabular}
	\vspace{-4pt}
\end{table}

\subsection{A New Low-Light Image and Video Dataset}
\begin{figure}[!t]
	\centering  \centerline{\includegraphics[width=1\linewidth]{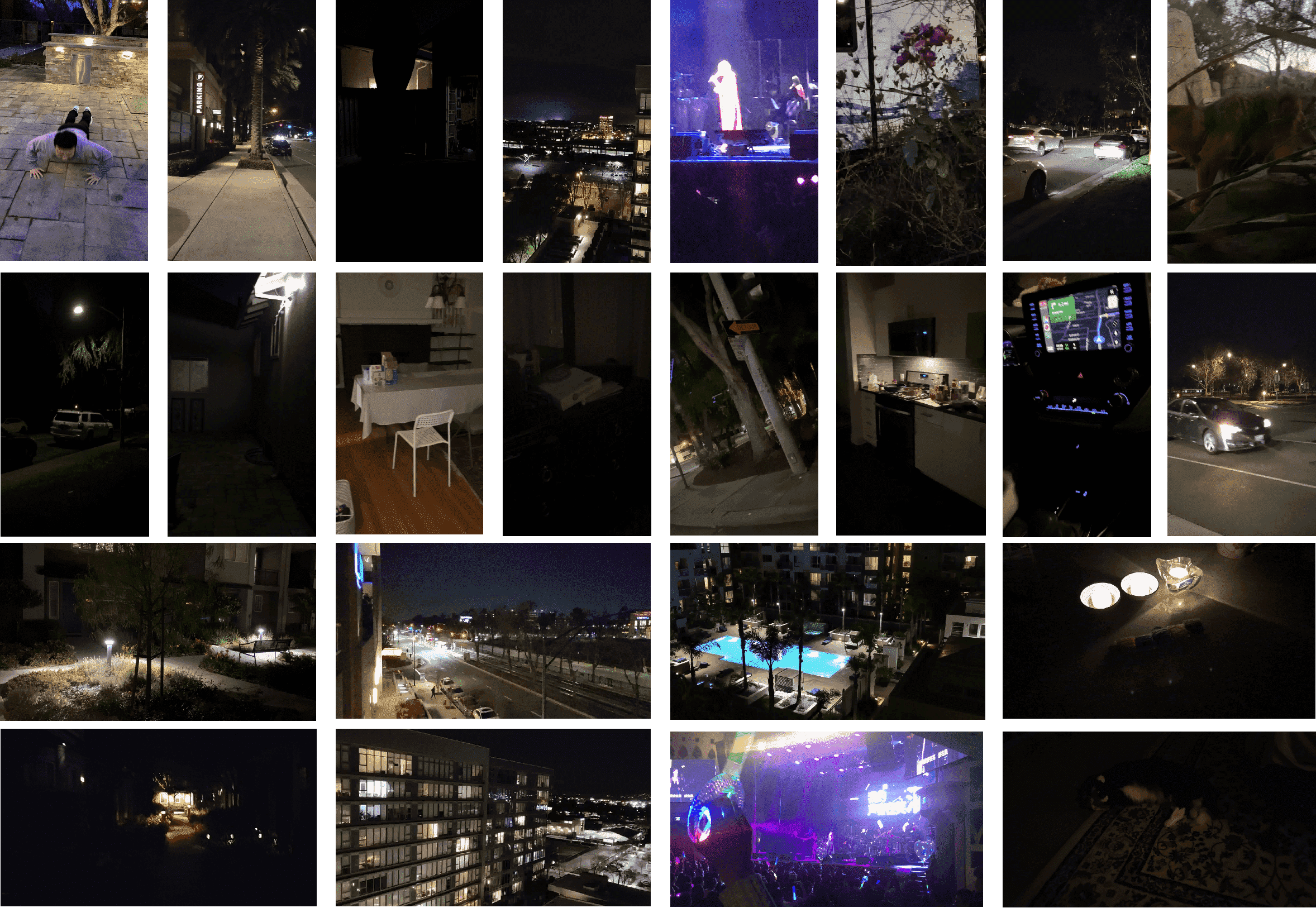}}
	\vspace{-2pt}
	\caption{Several images sampled from the proposed LLIV-Phone dataset. The  images and videos are taken by different devices under diverse lighting conditions and scenes.}
	\label{fig:dataset_samples}
	\vspace{-4pt}
\end{figure}

We propose a  Low-Light Image and Video dataset, called LLIV-Phone, to comprehensively and thoroughly validate the performance of LLIE methods. LLIV-Phone is the largest and most challenging real-world testing dataset of its kind. In particular, the dataset contains 120 videos (45,148 images) taken by 18 different mobile phones' cameras including iPhone 6s, iPhone 7, iPhone7 Plus, iPhone8 Plus, iPhone 11, iPhone 11 Pro, iPhone XS, iPhone XR, iPhone SE, Xiaomi Mi 9, Xiaomi Mi Mix 3, Pixel 3, Pixel 4,  Oppo R17, Vivo Nex, LG M322, OnePlus 5T, Huawei Mate 20 Pro under diverse illumination conditions (e.g., weak lighting, underexposure, moonlight, twilight, dark, extremely dark, back-lit, non-uniform light, and colored light.) in both indoor and outdoor scenes.  A summary of the LLIV-Phone dataset is provided in Table \ref{table:LLIVPhone}. 
We present several samples of LLIV-Phone dataset in Figure \ref{fig:dataset_samples}. 
The LLIV-Phone dataset is available at the project page.

This challenging dataset is collected in real scenes and contains diverse low-light images and videos. Consequently, it is suitable for evaluating the generalization capability of different low-light image and video enhancement models. Notably, the dataset can be used as the training dataset for unsupervised learning and the reference dataset for synthesis methods to generate realistic low-light data.

\begin{figure*} [!t]
	\begin{center}
		\begin{tabular}{c@{ }c@{ }c@{ }c@{ }c@{ }c@{ }c}
			\includegraphics[width=0.18\linewidth]{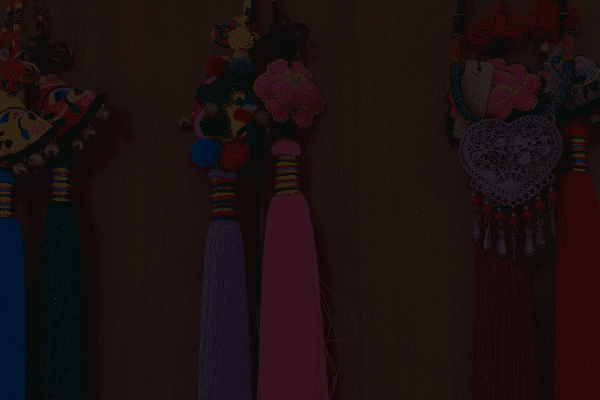}&
			\includegraphics[width=0.18\linewidth]{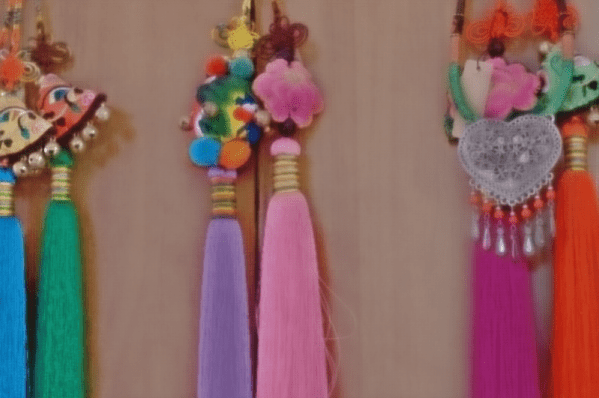}&
			\includegraphics[width=0.18\linewidth]{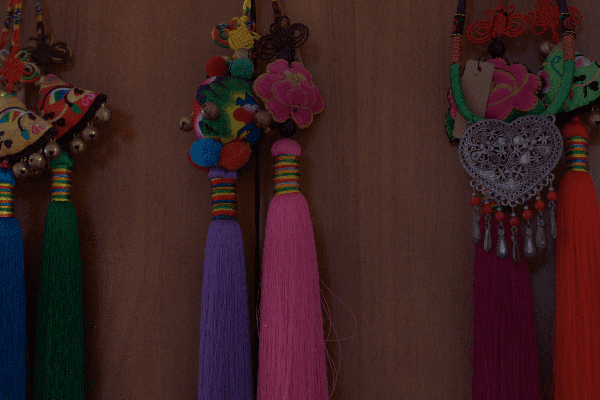}&
			\includegraphics[width=0.18\linewidth]{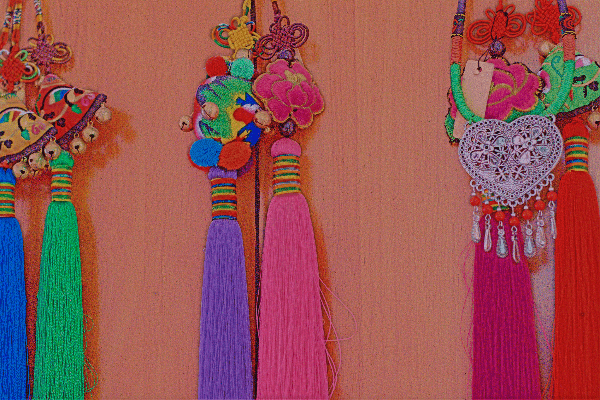}&
			\includegraphics[width=0.18\linewidth]{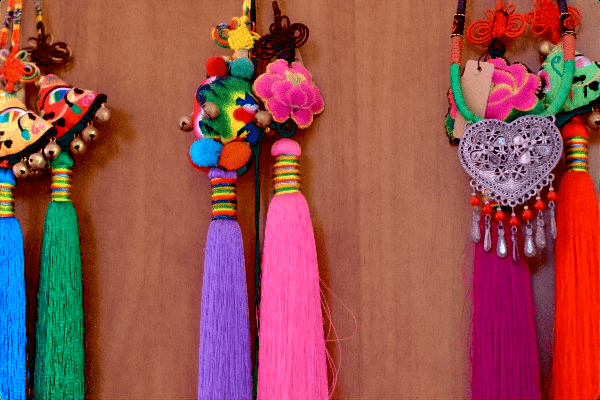}\\
			(a) input  & (b) LLNet \cite{LLNet} & (c) LightenNet \cite{LightenNet} &  (d)  Retinex-Net \cite{ChenBMVC18}  & (e) 	MBLLEN \cite{LvBMVC2018}  \\
			\includegraphics[width=0.18\linewidth]{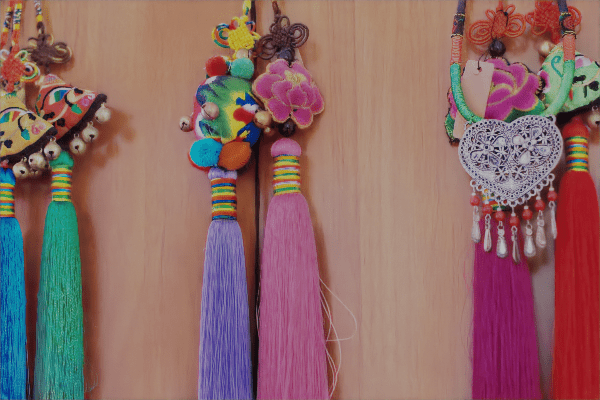}&
			\includegraphics[width=0.18\linewidth]{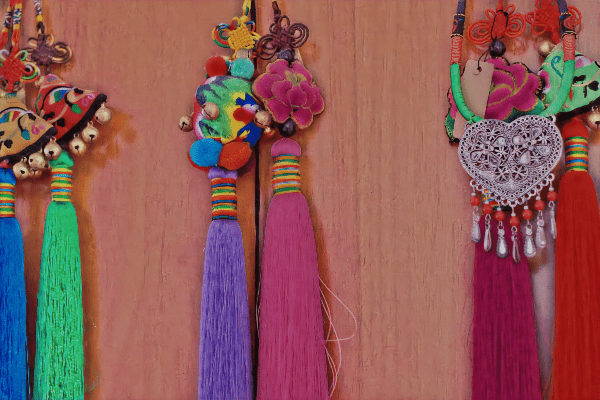}&
			\includegraphics[width=0.18\linewidth]{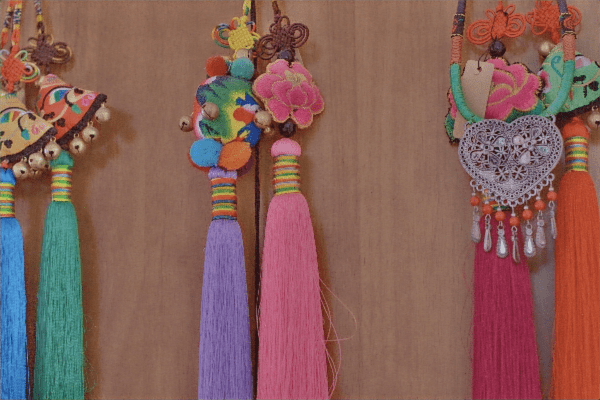}&
			\includegraphics[width=0.18\linewidth]{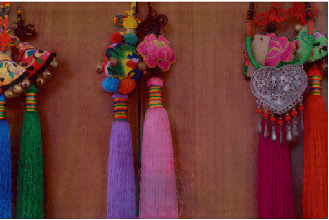}&
			\includegraphics[width=0.18\linewidth]{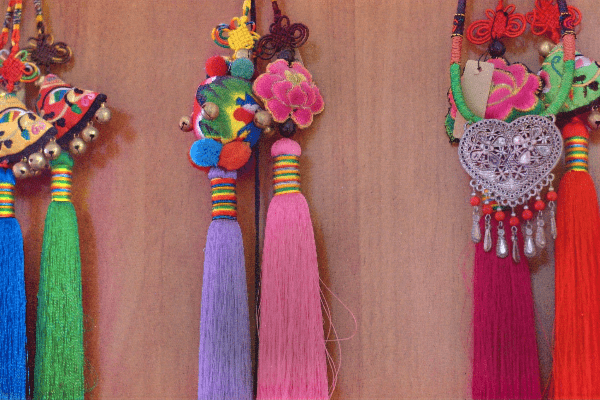}\\
			(f) KinD \cite{ZhangACM19}  & (g) KinD++ \cite{GuoIJCV2020} & (h) TBEFN \cite{TBEFN} &  (i)  	DSLR \cite{DSLR}  & (j) EnlightenGAN \cite{EnlightenGAN}  \\
			\includegraphics[width=0.18\linewidth]{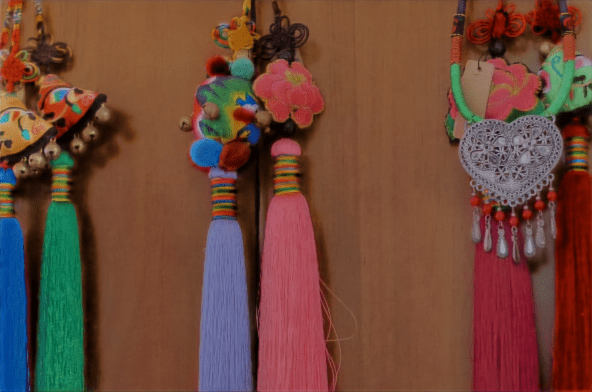}&
			\includegraphics[width=0.18\linewidth]{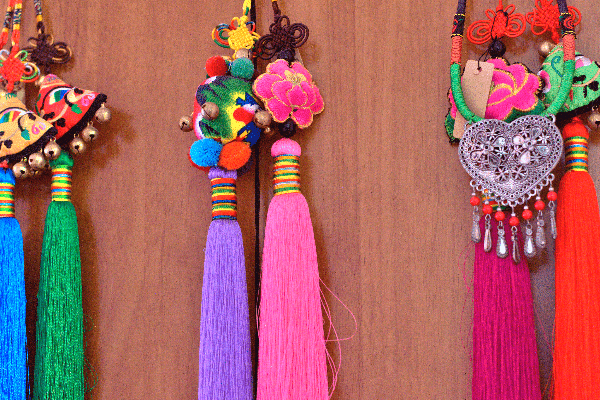}&
			\includegraphics[width=0.18\linewidth]{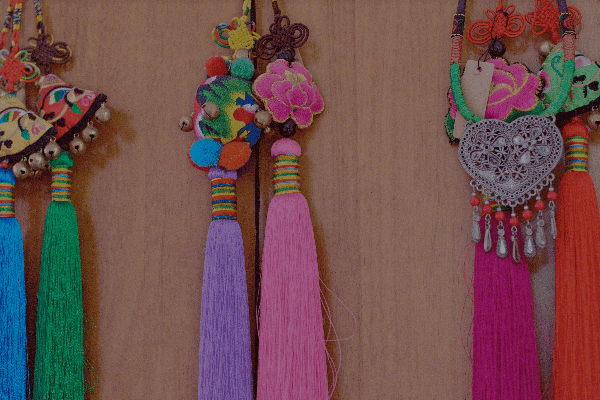}&
			\includegraphics[width=0.18\linewidth]{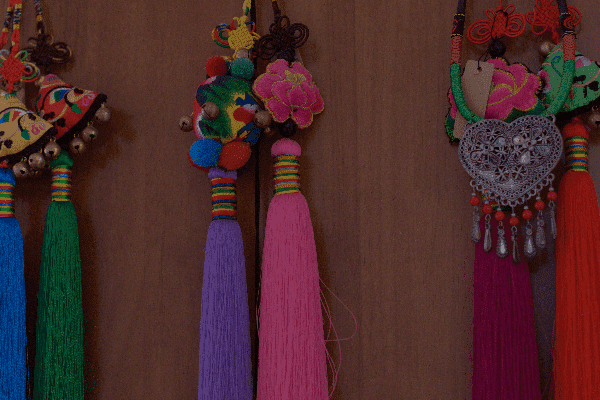}&
			\includegraphics[width=0.18\linewidth]{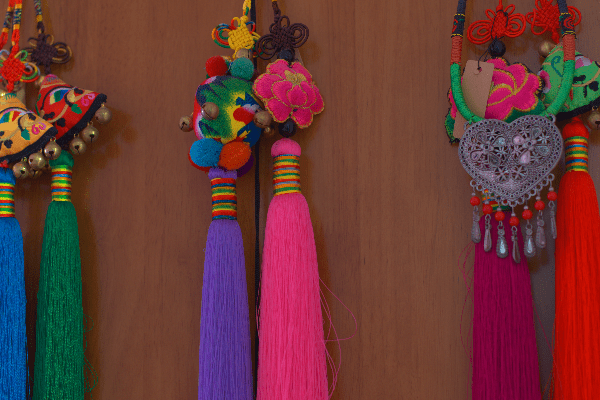}\\
			(k) DRBN \cite{YangCVRP20}  & (l)	ExCNet \cite{ZhangACM191} & (m) Zero-DCE \cite{ZeroDCE} &  (n) 	RRDNet \cite{RRDNet}  & (o) GT  \\
		\end{tabular}
	\end{center}
	\vspace{-2pt}
	\caption{Visual results of different methods on a low-light image sampled from LOL-test dataset.}
	\label{fig:LOL}
	\vspace{-4pt}
\end{figure*}

\begin{figure*}[!t]
	\begin{center}
		\begin{tabular}{c@{ }c@{ }c@{ }c@{ }c@{ }c@{ }c}
			\includegraphics[width=0.18\linewidth]{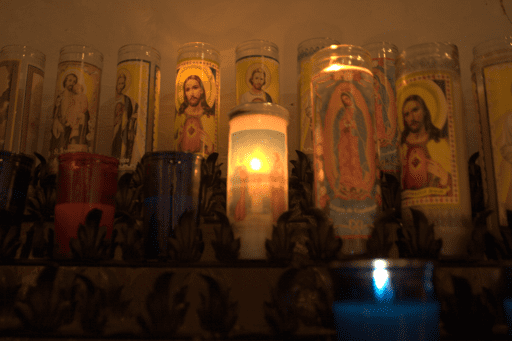}&
			\includegraphics[width=0.18\linewidth]{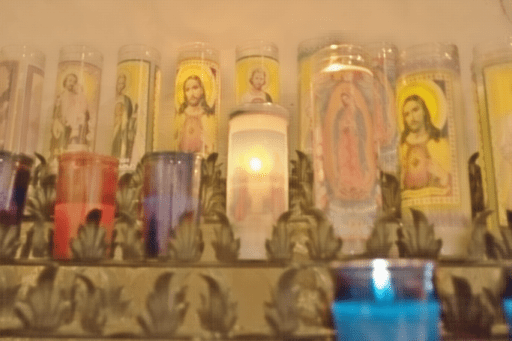}&
			\includegraphics[width=0.18\linewidth]{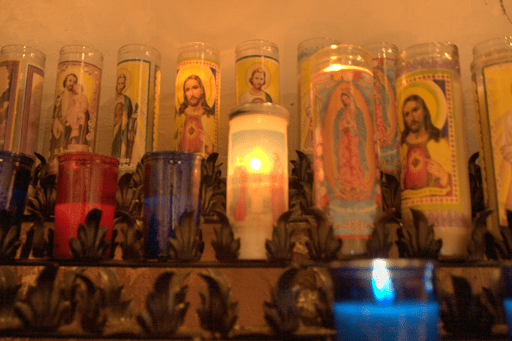}&
			\includegraphics[width=0.18\linewidth]{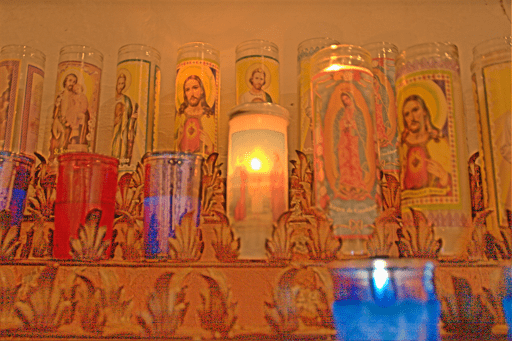}&
			\includegraphics[width=0.18\linewidth]{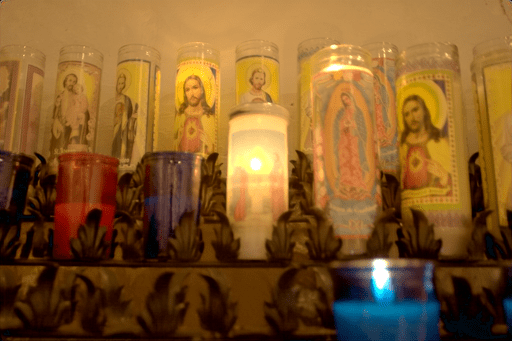}\\
			(a) input  & (b) LLNet \cite{LLNet} & (c) LightenNet \cite{LightenNet} &  (d)  Retinex-Net \cite{ChenBMVC18}  & (e) 	MBLLEN \cite{LvBMVC2018}  \\
			\includegraphics[width=0.18\linewidth]{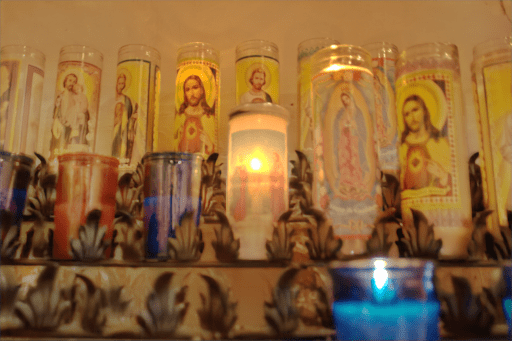}&
			\includegraphics[width=0.18\linewidth]{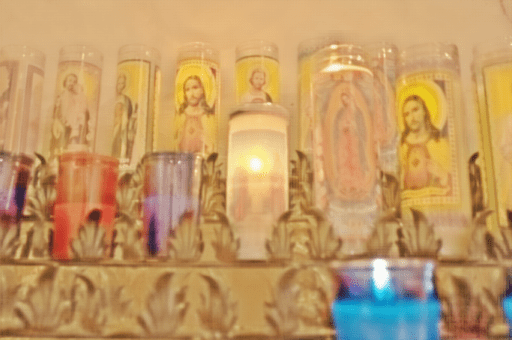}&
			\includegraphics[width=0.18\linewidth]{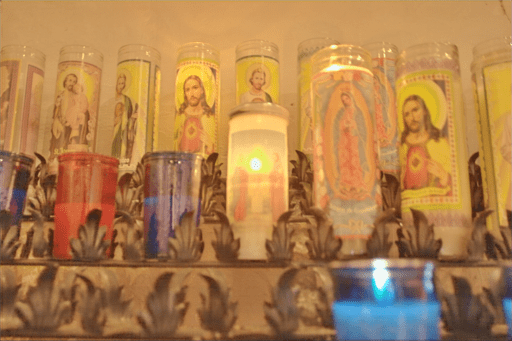}&
			\includegraphics[width=0.18\linewidth]{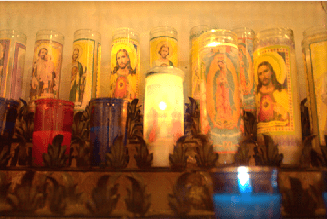}&
			\includegraphics[width=0.18\linewidth]{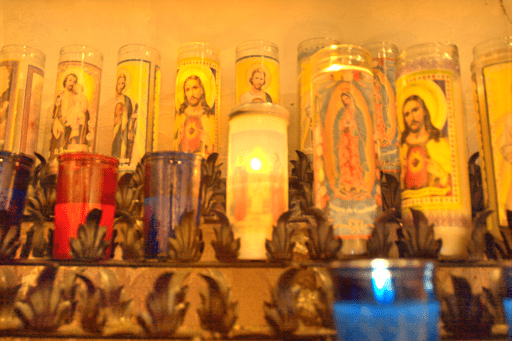}\\
			(f) KinD \cite{ZhangACM19}  & (g) KinD++ \cite{GuoIJCV2020} & (h) TBEFN \cite{TBEFN} &  (i)  	DSLR \cite{DSLR}  & (j) EnlightenGAN \cite{EnlightenGAN}  \\
			\includegraphics[width=0.18\linewidth]{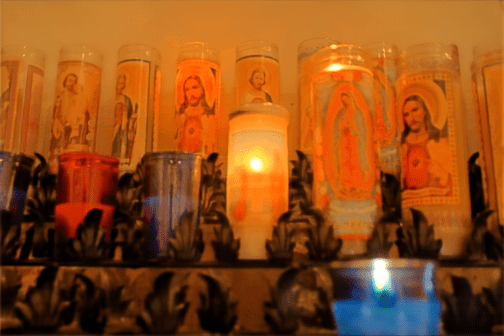}&
			\includegraphics[width=0.18\linewidth]{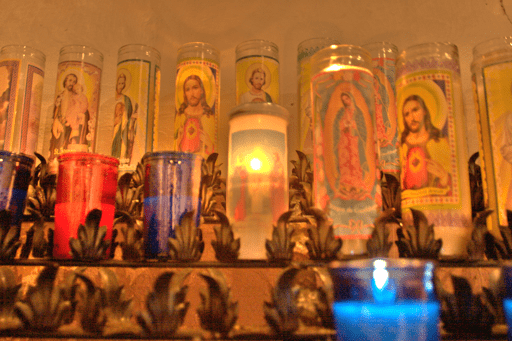}&
			\includegraphics[width=0.18\linewidth]{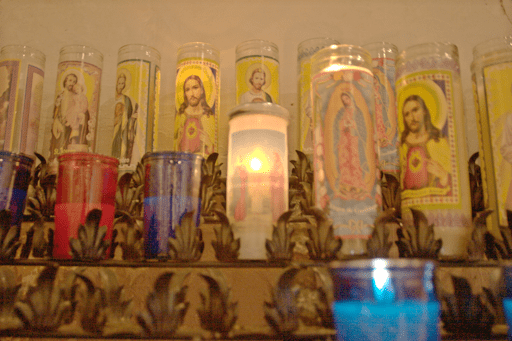}&
			\includegraphics[width=0.18\linewidth]{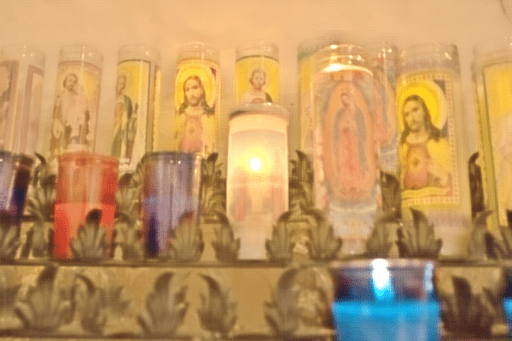}&
			\includegraphics[width=0.18\linewidth]{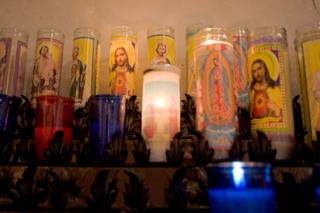}\\
			(k) DRBN \cite{YangCVRP20}  & (l)	ExCNet \cite{ZhangACM191} & (m) Zero-DCE \cite{ZeroDCE} &  (n) 	RRDNet \cite{RRDNet}  & (o) GT  \\
		\end{tabular}
	\end{center}
	\vspace{-2pt}
	\caption{Visual results of different methods on a low-light image sampled from MIT-Adobe FiveK-test dataset.}
	\label{fig:5K}
	\vspace{-4pt}
\end{figure*}

\subsection{Online Evaluation Platform}
Different deep models may be implemented in different platforms such as Caffe, Theano, TensorFlow, and PyTorch. As a result, different algorithms demand different configurations, GPU versions, and hardware specifications. Such requirements are prohibitive to many researchers, especially for beginners who are new to this area and may not even have GPU resources. To resolve these problems, we develop an LLIE online platform, called LLIE-Platform, which is available at \url{http://mc.nankai.edu.cn/ll/}.

To the date of this submission, the LLIE-Platform covers 14 popular deep learning-based LLIE methods including LLNet \cite{LLNet}, LightenNet \cite{LightenNet}, Retinex-Net \cite{ChenBMVC18}, EnlightenGAN \cite{EnlightenGAN}, MBLLEN \cite{LvBMVC2018}, KinD \cite{ZhangACM19}, KinD++ \cite{GuoIJCV2020}, TBEFN \cite{TBEFN}, DSLR \cite{DSLR}, DRBN \cite{YangCVRP20}, ExCNet \cite{ZhangACM191}, Zero-DCE \cite{ZeroDCE}, Zero-DCE++ \cite{ZeroDCE++}, and  RRDNet \cite{RRDNet},  where the results of any input can be produced through a user-friendly web interface.  We will regularly offer new methods on this platform.
We wish that this LLIE-Platform could serve the
growing research community by providing users a flexible interface
to run existing deep learning-based LLIE methods and develop their own new LLIE methods.

\newcommand{\addImg}[1]{\includegraphics[width=0.14\linewidth,height=0.2\linewidth]{figures/#1}}

\begin{figure*}[!t]
	\centering
	\setlength\tabcolsep{0.5pt}
	\begin{tabular}{ccccccccc}
		\addImg{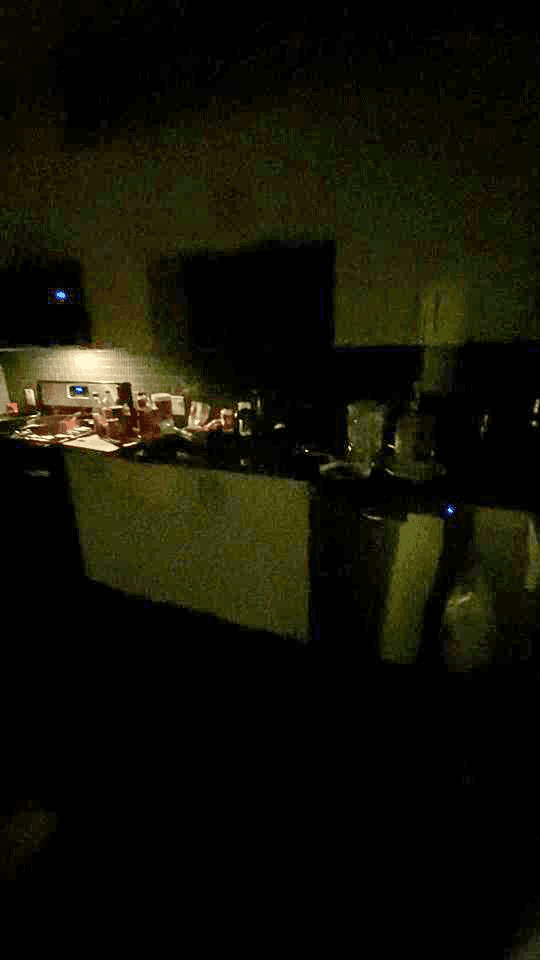} &
		\addImg{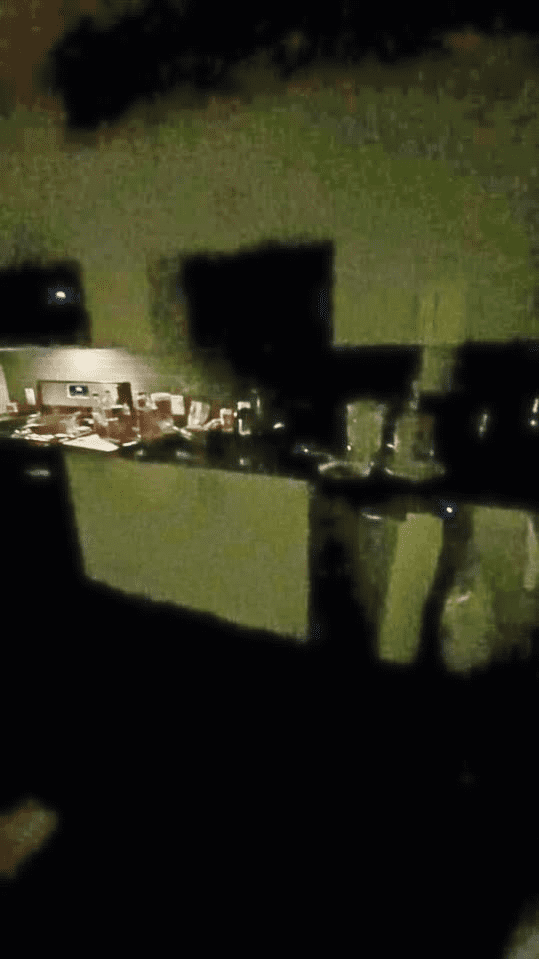}&
		\addImg{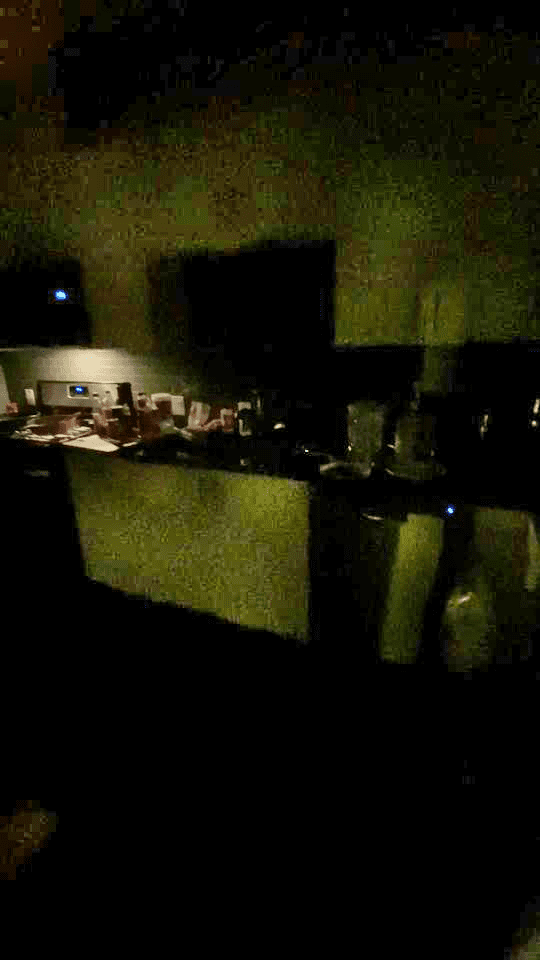}&
		\addImg{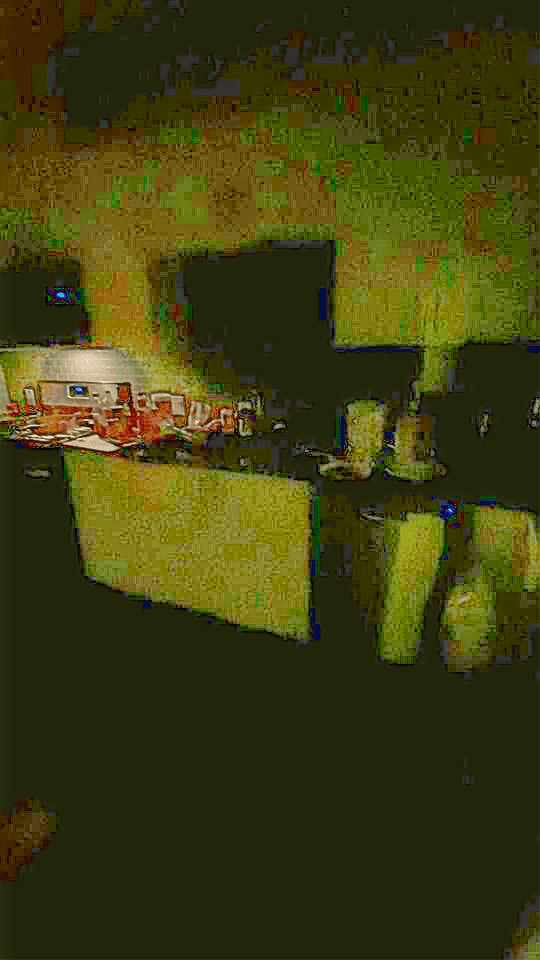}&
		\addImg{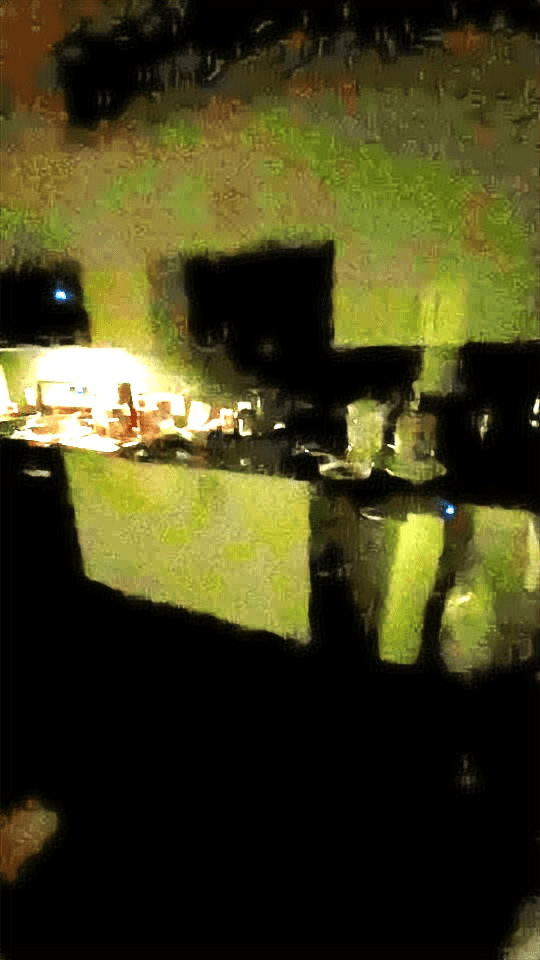}&
		\addImg{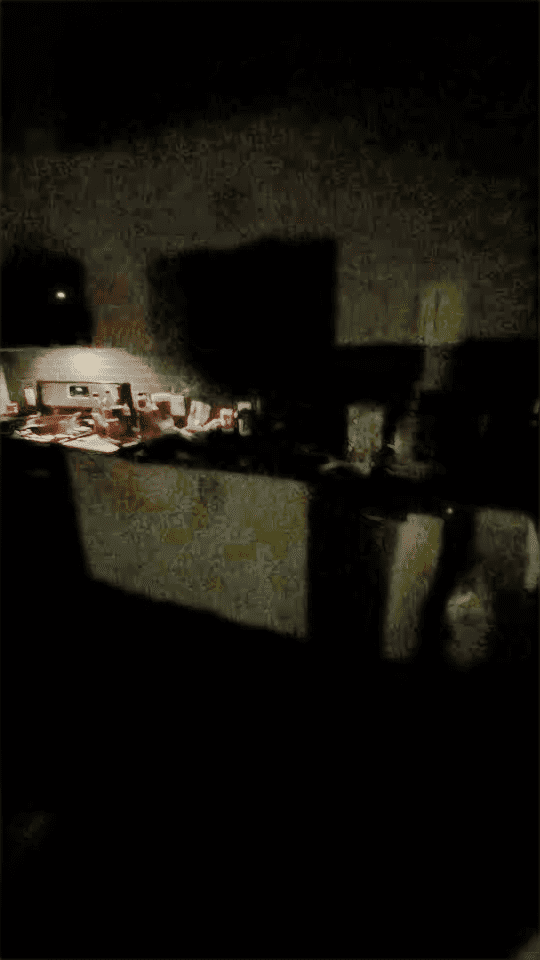}&
		\addImg{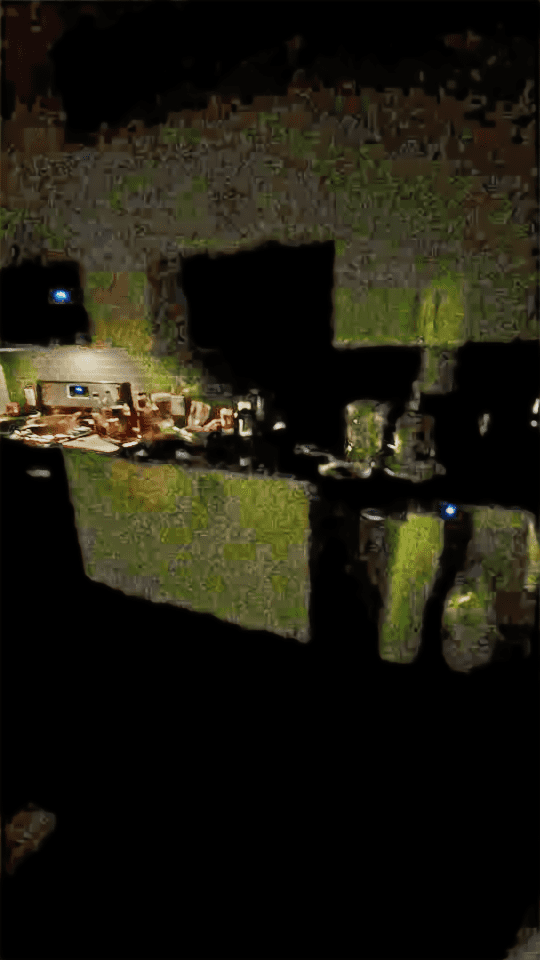}&\\
			(a)   & (b)  & (c)  &  (d)   & (e) & (f)   & (g)   \\
		\addImg{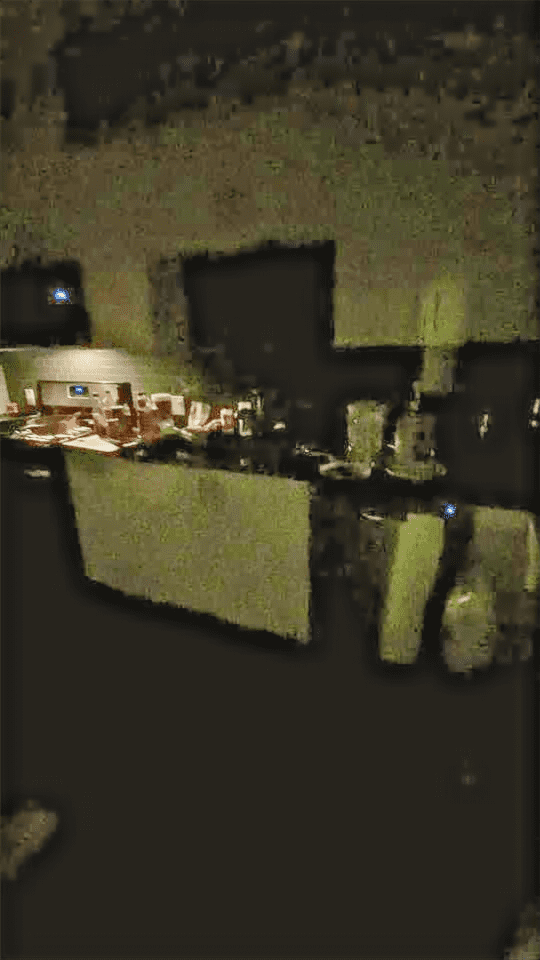}&
		\addImg{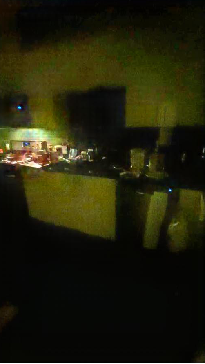}&
		\addImg{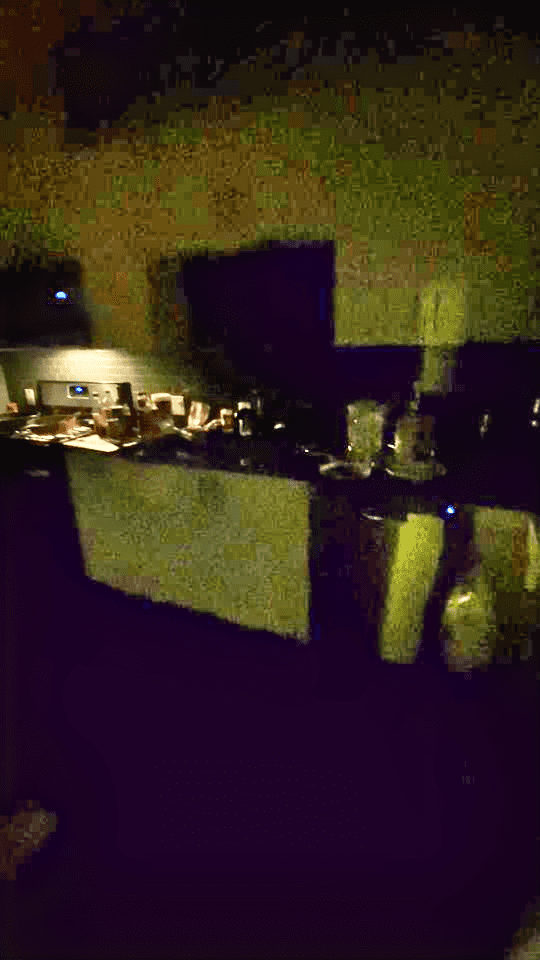}&
		\addImg{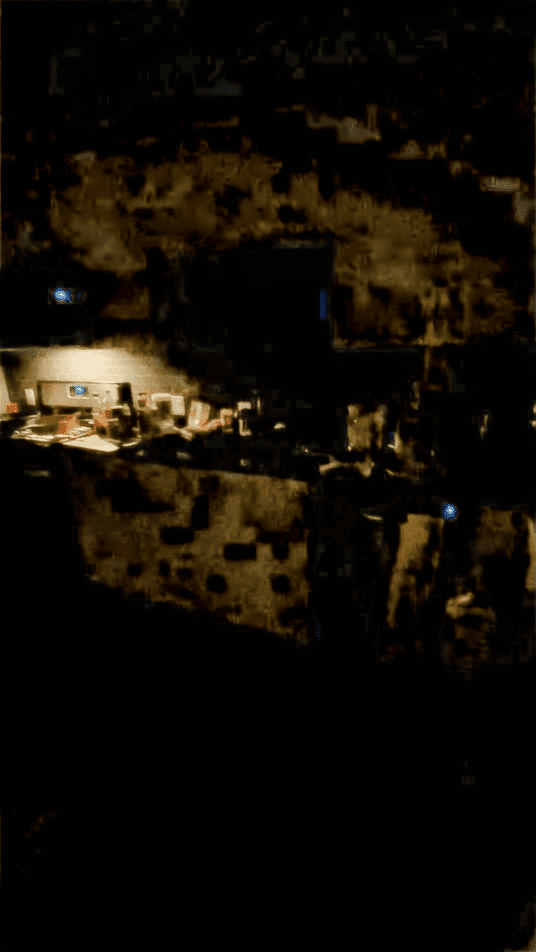}&
		\addImg{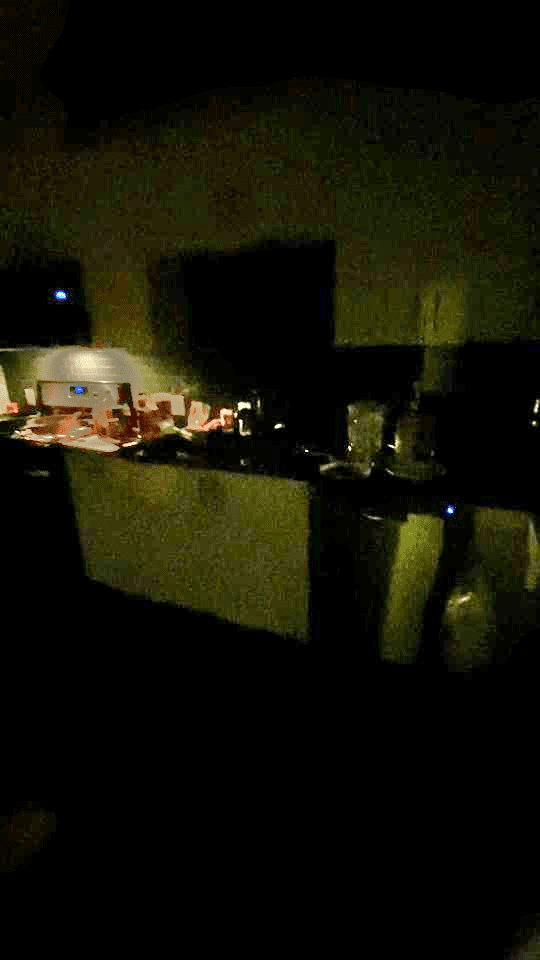}&
		\addImg{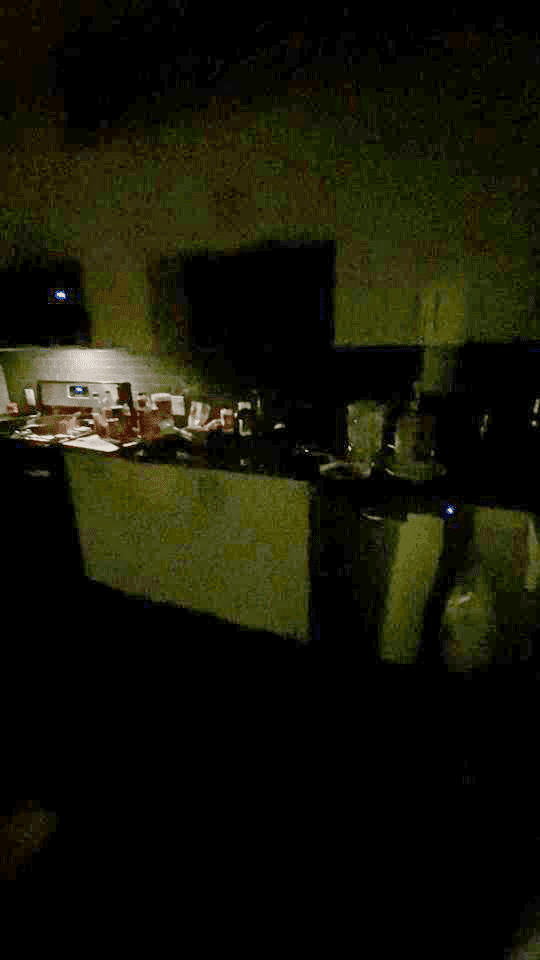}&
		\addImg{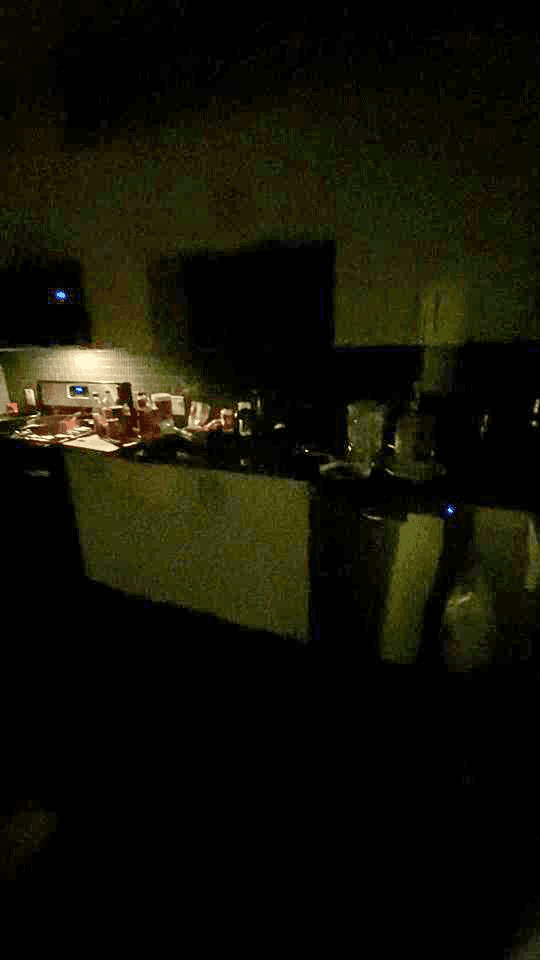}\\
			(h) &  (i)  	  & (j)  & 	(k)  & (l) & (m)  &  (n) 	 \\
	\end{tabular}
	\vspace{-12pt}
	\caption{Visual results of different methods on a low-light image sampled from  LLIV-Phone-imgT dataset. (a) input. (b) LLNet \cite{LLNet}. (c) LightenNet \cite{LightenNet}. (d)  Retinex-Net \cite{ChenBMVC18}. (e) 	MBLLEN \cite{LvBMVC2018}. (f) KinD \cite{ZhangACM19}. (g) KinD++ \cite{GuoIJCV2020}. (h) TBEFN \cite{TBEFN}. (i)  	DSLR \cite{DSLR}. (j) EnlightenGAN \cite{EnlightenGAN}.	(k) DRBN \cite{YangCVRP20}. (l)	ExCNet \cite{ZhangACM191}. (m) Zero-DCE \cite{ZeroDCE}. (n) 	RRDNet \cite{RRDNet}.}
	\label{fig:Phone1}
	\vspace{-4pt}
\end{figure*}

\begin{figure*}[!t]
	\setlength\tabcolsep{0.5pt}
	\centering
	\begin{tabular}{ccccccccc}
		\addImg{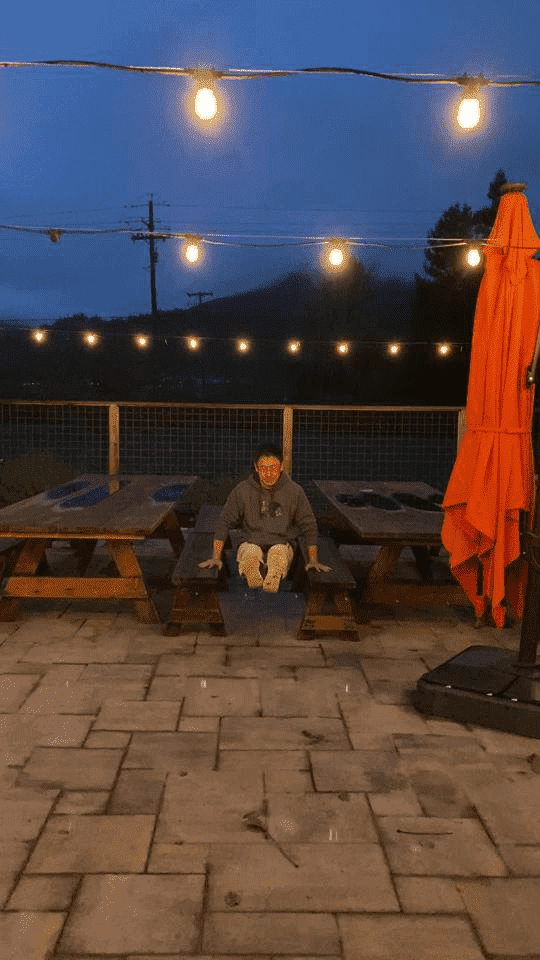}&
		\addImg{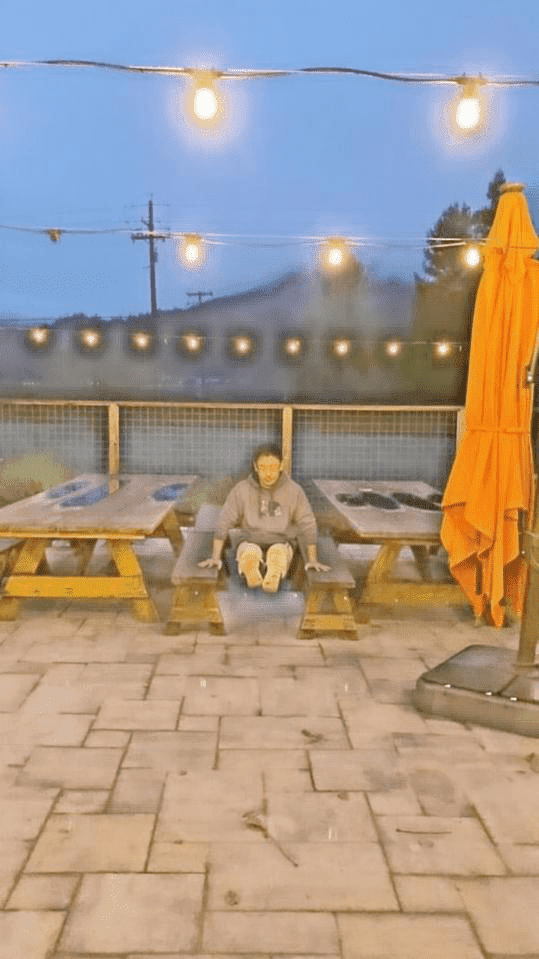}&
		\addImg{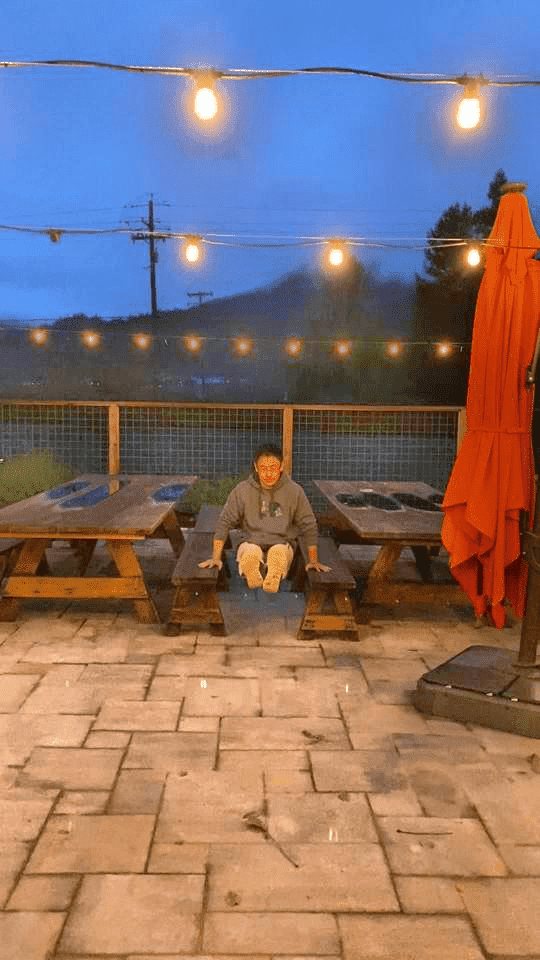}&
		\addImg{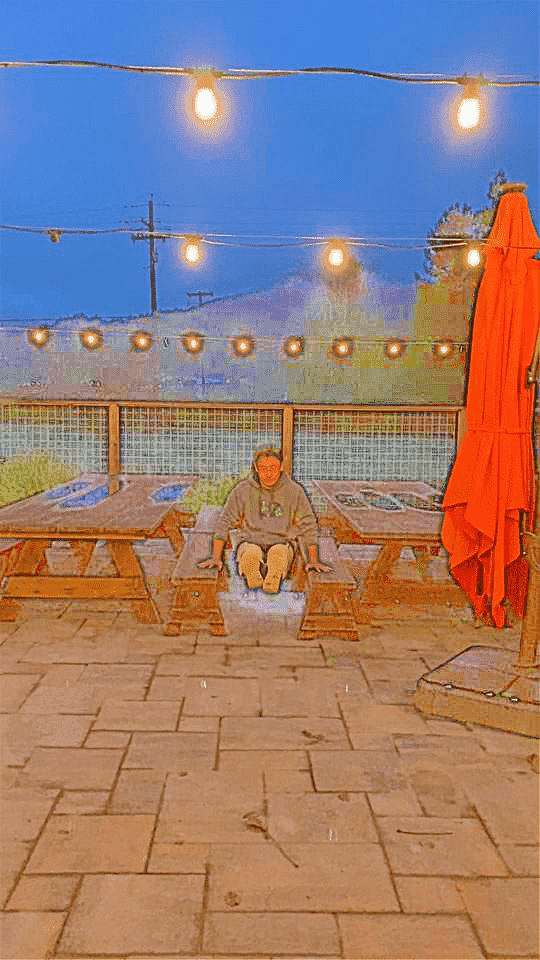}&
		\addImg{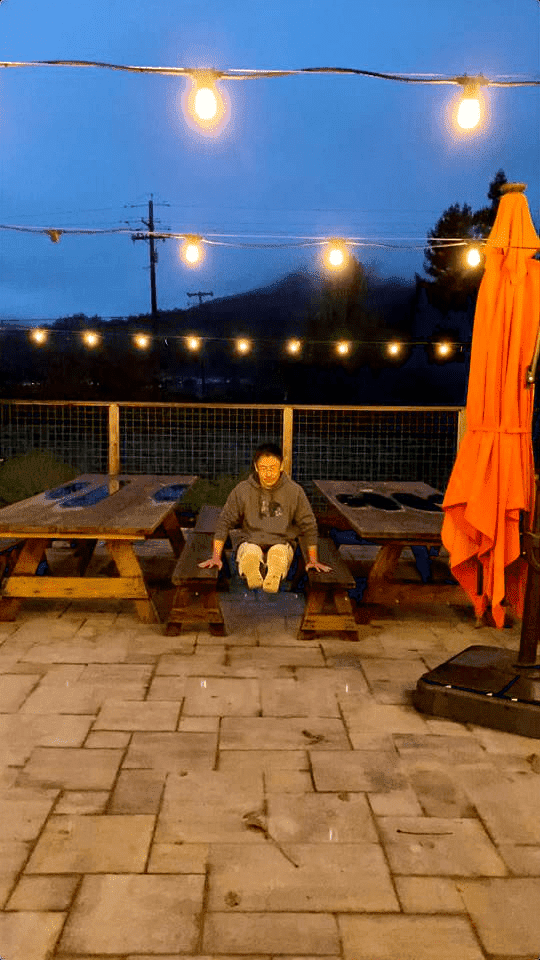}&
		\addImg{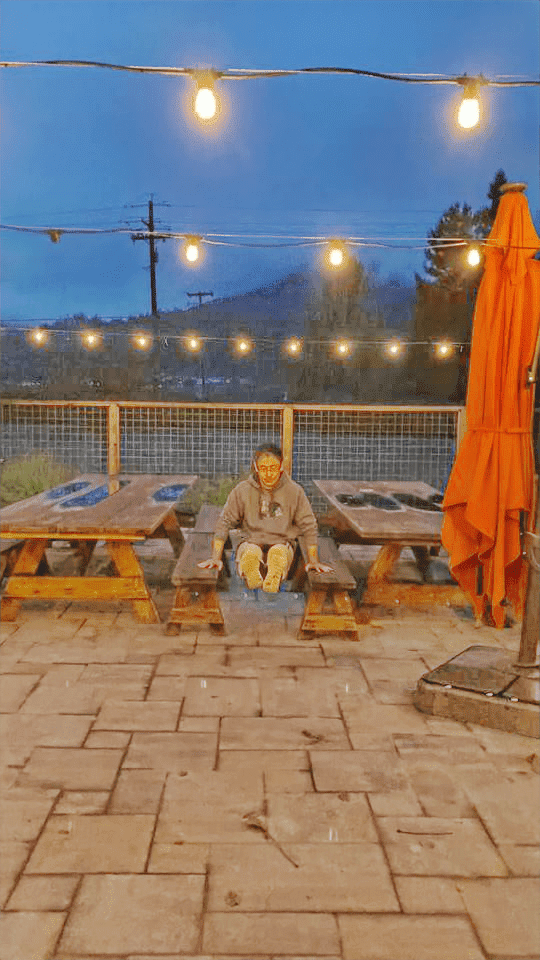}&
		\addImg{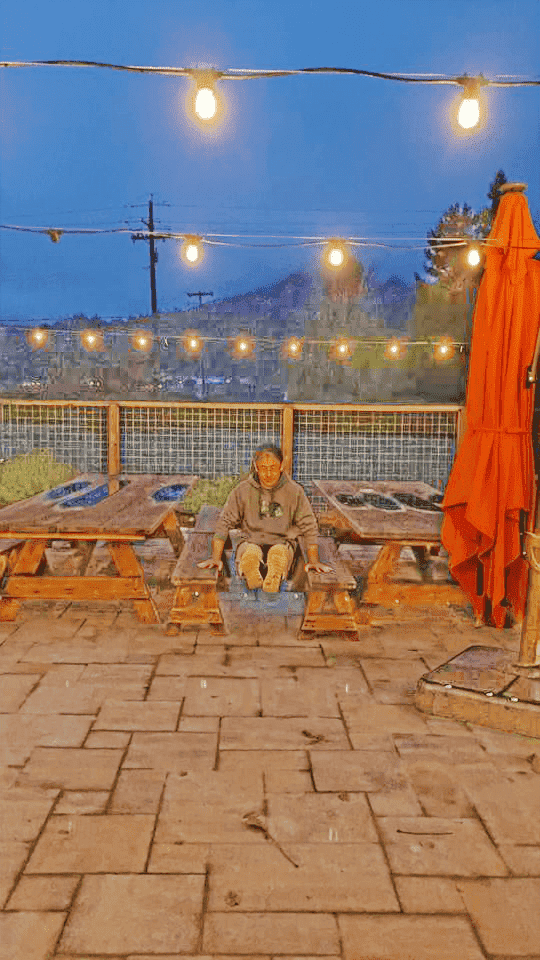}&\\
			(a)   & (b) & (c) &  (d)   & (e) 	 & (f)  & (g)   \\
		\addImg{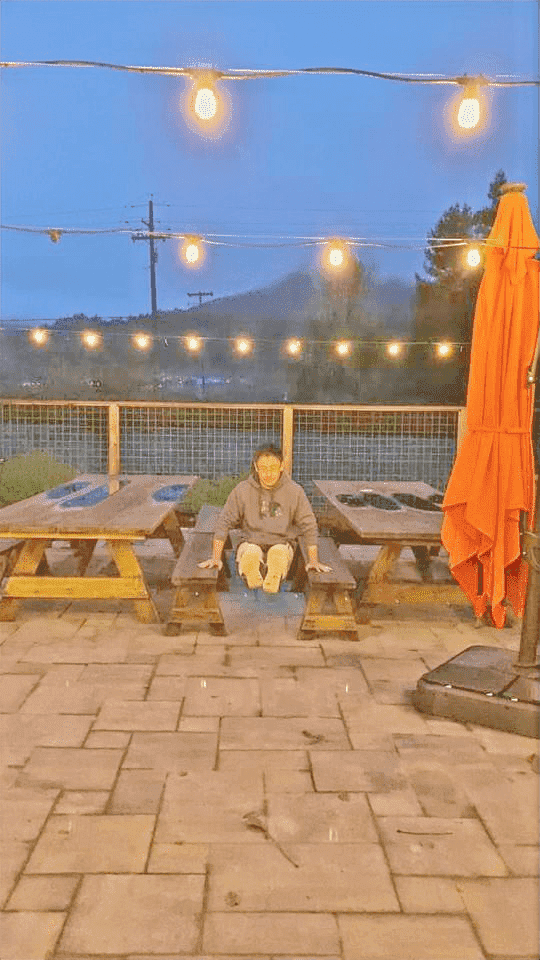}&
		\addImg{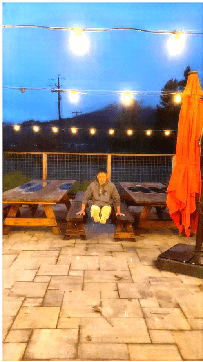}&
		\addImg{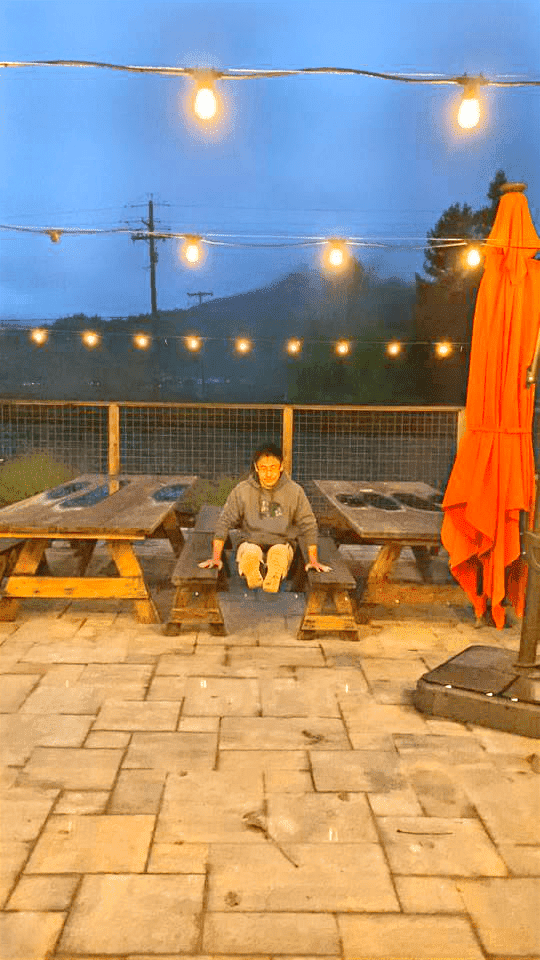}&
		\addImg{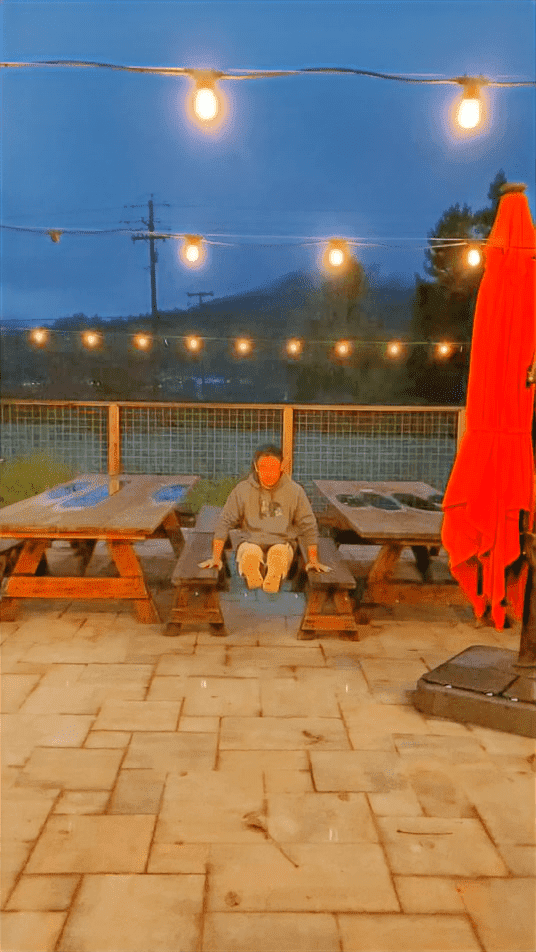}&
		\addImg{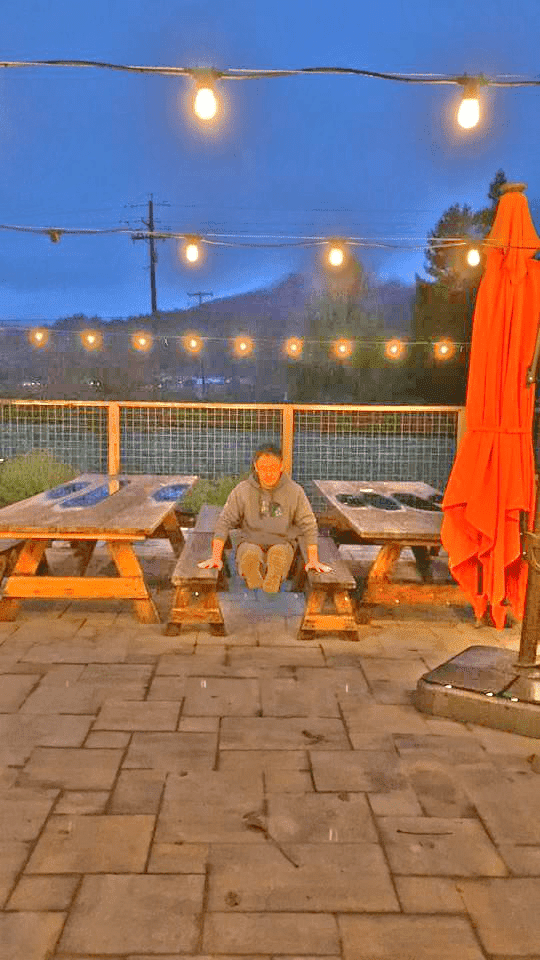}&
		\addImg{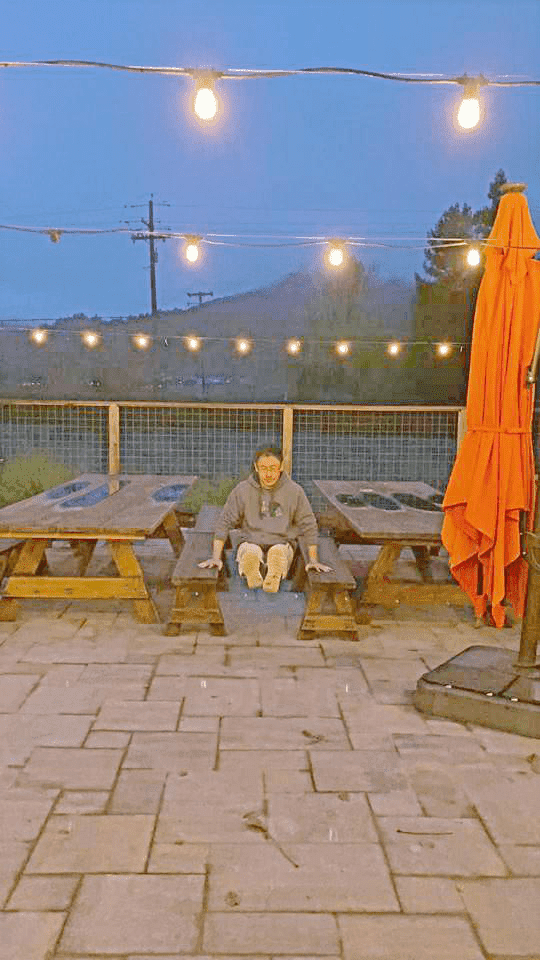}&
		\addImg{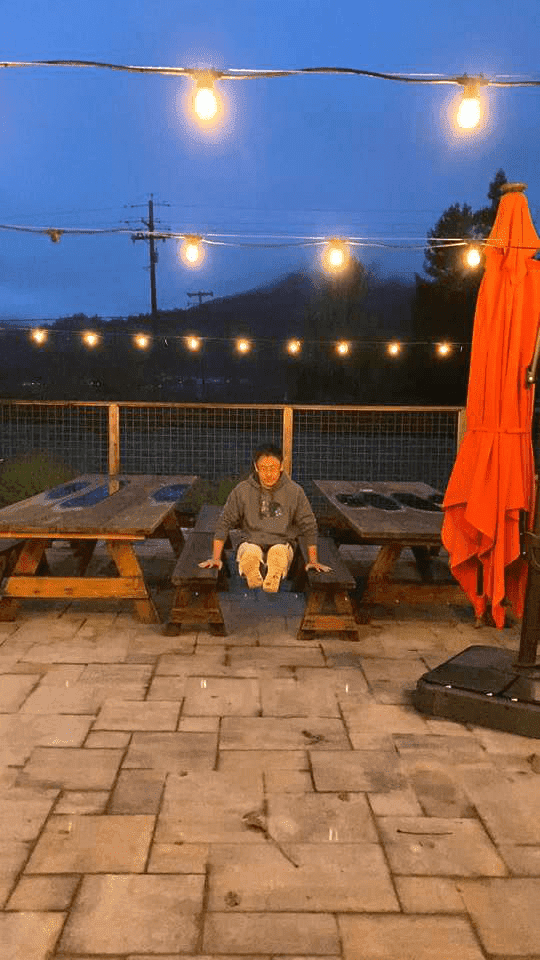}\\
			(h)  &  (i)  	  & (j) & 	(k)  & (l)	 & (m)  &  (n) 	 \\
	\end{tabular}
	\vspace{-12pt}
	\caption{Visual results of different methods on a low-light image sampled from  LLIV-Phone-imgT dataset. (a) input. (b) LLNet \cite{LLNet}. (c) LightenNet \cite{LightenNet}. (d)  Retinex-Net \cite{ChenBMVC18}. (e) 	MBLLEN \cite{LvBMVC2018}. (f) KinD \cite{ZhangACM19}. (g) KinD++ \cite{GuoIJCV2020}. (h) TBEFN \cite{TBEFN}. (i)  	DSLR \cite{DSLR}. (j) EnlightenGAN \cite{EnlightenGAN}.	(k) DRBN \cite{YangCVRP20}. (l)	ExCNet \cite{ZhangACM191}. (m) Zero-DCE \cite{ZeroDCE}. (n) 	RRDNet \cite{RRDNet}.}
	\label{fig:Phone2}
	\vspace{-4pt}
\end{figure*}

\begin{figure*}[!t]
	\begin{center}
		\begin{tabular}{c@{ }c@{ }c@{ }c@{ }c@{ }}
			\rotatebox{90}{~~~~~~~~~Bayer~~~~~~~~~~~~APS-C X-Trans}~&
			\includegraphics[width=0.21\linewidth]{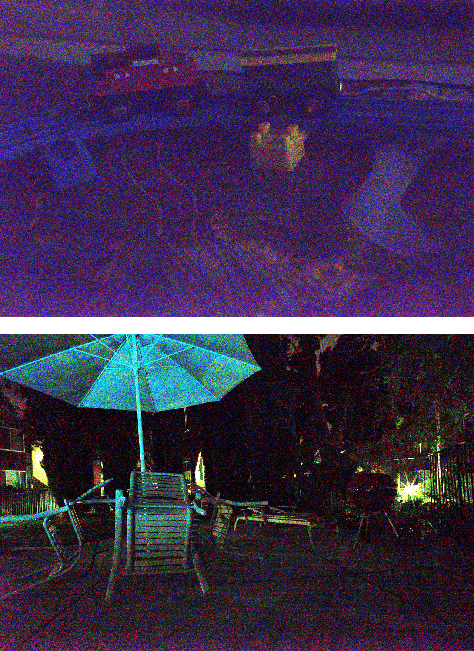}&~~
			\includegraphics[width=0.21\linewidth]{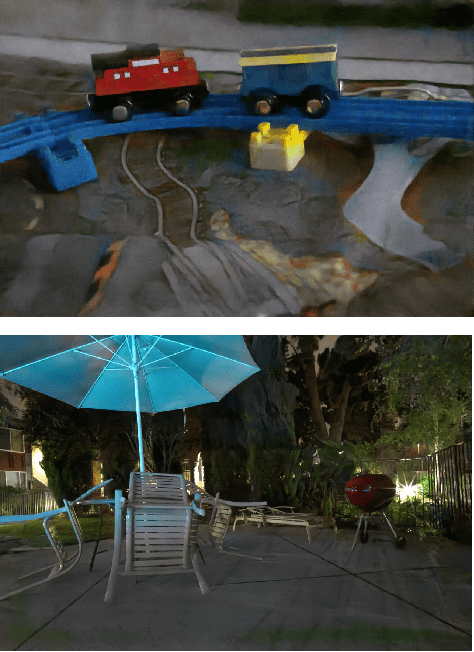}&~~
			\includegraphics[width=0.21\linewidth]{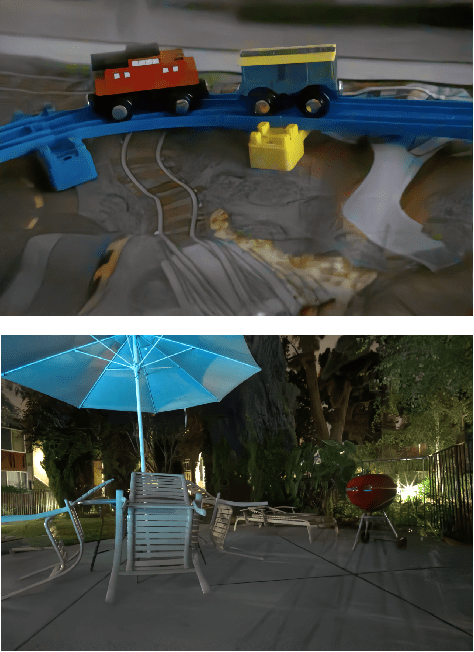}&~~
			\includegraphics[width=0.21\linewidth]{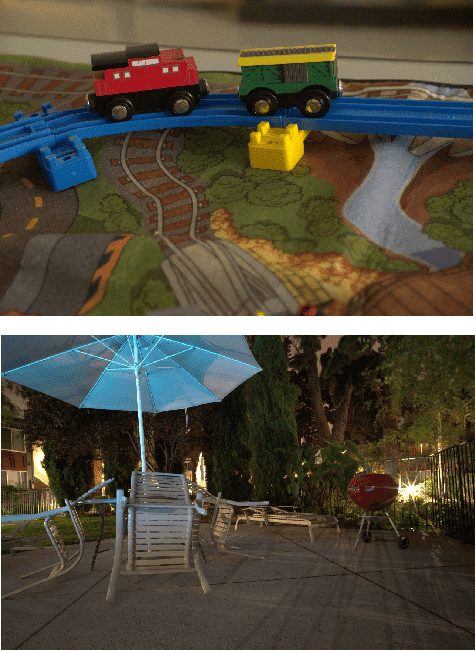}\\
			&~	(a) inputs  &~~ (b) SID \cite{Chen18} &~~ (c) EEMEFN \cite{ZhuAAAI20}&~~  (d) GT   \\
		\end{tabular}
	\end{center}
	\vspace{-2pt}
	\caption{Visual results of different methods on two raw low-light images sampled from  SID-test-Bayer and SID-test-X-Trans test datasets. The inputs are amplified for visualization.}
	\label{fig:raw}
	\vspace{-4pt}
\end{figure*}

\subsection{Benchmarking Results}
To qualitatively and quantitatively evaluate different methods, in addition to the proposed LLIV-Phone dataset, we also adopt the commonly used LOL \cite{ChenBMVC18} and MIT-Adobe FiveK \cite{Adobe5K} datasets for RGB format-based methods, and SID \cite{Chen18} dataset for raw format-based methods. More visual results can be found in the supplementary material. The comparative results on the real low-light videos taken by different mobile phones' cameras can be found at YouTube \url{https://www.youtube.com/watch?v=Elo9TkrG5Oo&t=6s}.

We select five images on average from each video of the LLIV-Phone dataset, forming an image testing dataset with a total of 600 images (denoted as LLIV-Phone-imgT). Furthermore, we randomly select one video from the videos of each phone's brand of LLIV-Phone dataset, forming a video testing dataset with a total of 18 videos (denoted as LLIV-Phone-vidT). We half the resolutions of the frames in both LLIV-Phone-imgT and  LLIV-Phone-vidT because some deep learning-based methods cannot process the full resolution of test images and videos. 
For the LOL dataset, we adopt the original test set including 15 low-light images captured in real scenes for testing, denoted as LOL-test. For the MIT-Adobe FiveK dataset, we follow the protocol in Chen et al. \cite{Enhancer} to decode the images into PNG format and resize them to have a long edge of 512 pixels using Lightroom. We adopt the same testing dataset as Chen et al.  \cite{Enhancer}, MIT-Adobe FiveK-test, including 500 images with the retouching results by expert C as the corresponding ground truths. For the SID dataset, we use the default test set  used in EEMEFN \cite{ZhuAAAI20} for fair comparisons, denoted as SID-test (SID-test-Bayer and SID-test-X-Trans), which is a partial test set of SID \cite{Chen18}. The SID-test-Bayer includes 93 images of the Bayer pattern while the SID-test-X-Trans includes 94 images of the APS-C X-Trans pattern.

\begin{table*}[t]
	\caption{Quantitative comparisons on LOL-test and MIT-Adobe FiveK-test test datasets in terms of MSE ($\times10^3$), PSNR (in dB), SSIM \cite{SSIM}, and LPIPS \cite{LPIPS}. The best result is in {\color{red}{red}} whereas the second and third best results are in {\color{blue}{blue}} and {\color{purple}{purple}} under each case, respectively.}
	\vspace{-6pt}
	\begin{center}
		\begin{tabular}{c|c|c|c|c|c||c|c|c|c}
			\hline
			\multirow{2}{*}{\textbf{Learning}} &\multirow{2}{*}{\textbf{Method}} & \multicolumn{4}{c||}{\textbf{LOL-test}} & \multicolumn{4}{c}{\textbf{MIT-Adobe FiveK-test}} \\
			\cline{3-10}
			\cline{3-10}
			& &	\textbf{MSE}$\downarrow$ & \textbf{PSNR} $\uparrow$ & \textbf{SSIM}$\uparrow$ & \textbf{LPIPS}$\downarrow$ & \textbf{MSE}$\downarrow$ & \textbf{PSNR} $\uparrow$ & \textbf{SSIM}$\uparrow$ & \textbf{LPIPS}$\downarrow$ \\
			\hline
			&input  & 12.613 & 7.773 &0.181 & 0.560& {\color{blue}{1.670}}& {\color{blue}{17.824}} & {\color{purple}{0.779}} &{\color{blue}{0.148}} \\
			\hline
			&LLNet \cite{LLNet} & {\color{red}{1.290}} &  {\color{red}{17.959}} &0.713 &0.360&4.465 & 12.177 &0.645 &0.292\\
			&LightenNet \cite{LightenNet} & 7.614 & 10.301 & 0.402 & 0.394 & 4.127 & 13.579 & 0.744 &{\color{purple}{0.166}}\\
			&Retinex-Net \cite{ChenBMVC18} & 1.651 & 16.774& 0.462 & 0.474 & 4.406 & 12.310 & 0.671 &0.239\\
			SL&MBLLEN \cite{LvBMVC2018}  &1.444&{\color{blue}{17.902}}&0.715& 0.247&  {\color{red}{1.296}} &  {\color{red}{19.781}} & {\color{red}{0.825}}&{\color{red}{0.108}} \\
			&KinD \cite{ZhangACM19} & {\color{purple}{1.431}} & 17.648& {\color{blue}{0.779}}& {\color{red}{0.175}} & 2.675 & 14.535 & 0.741 &0.177 \\
			&KinD++ \cite{GuoIJCV2020} & {\color{blue}{1.298}}&{\color{purple}{17.752}}& {\color{purple}{0.760}}& {\color{blue}{0.198}} & 7.582 & 9.732 &0.568 & 0.336 \\
			&TBEFN \cite{TBEFN} &1.764&17.351&  {\color{red}{0.786}}& {\color{purple}{0.210}} & 3.865 & 12.769 &0.704 &0.178 \\
			&DSLR \cite{DSLR} &3.536&15.050& 0.597&0.337 &{\color{purple}{1.925}} & {\color{purple}{16.632}} &{\color{blue}{0.782}} & 0.167 \\
			\hline
			UL	& EnlightenGAN \cite{EnlightenGAN} &1.998&17.483& 0.677&0.322 &3.628&13.260&0.745& 0.170 \\
			\hline
			SSL	&DRBN \cite{YangCVRP20} &2.359&15.125& 0.472&0.316 &3.314& 13.355&0.378 & 0.281 \\
			\hline
			&	ExCNet \cite{ZhangACM191} &2.292&15.783& 0.515&0.373 &2.927& 13.978&0.710 &0.187 \\			
			ZSL	&Zero-DCE \cite{ZeroDCE} &3.282&14.861& 0.589&0.335 &3.476&13.199&0.709 &0.203\\				
			&	RRDNet \cite{RRDNet} &6.313&11.392&0.468&0.361 &7.057&10.135&0.620 &0.303\\							
			\hline
		\end{tabular}
	\end{center}
	\label{Table:Quan_1}
	\vspace{-4pt}
\end{table*}

\begin{table*}[t]
	\caption{Quantitative comparisons on SID-test test dataset in terms of MSE ($\times10^3$), PSNR (in dB), SSIM \cite{SSIM}, and LPIPS \cite{LPIPS}. The best result is in {\color{red}{red}} under each case. To compute the quantitative scores of input raw data, we use the corresponding camera ISP pipelines provided by Chen et al. \cite{Chen18} to transfer raw data to RGB format.}
	\vspace{-6pt}
	\begin{center}
		\begin{tabular}{c|c|c|c|c|c||c|c|c|c}
			\hline
			\multirow{2}{*}{\textbf{Learning}} &\multirow{2}{*}{\textbf{Method}} & \multicolumn{4}{c||}{\textbf{SID-test--Bayer}} & \multicolumn{4}{c}{\textbf{SID-test--X-Trans}} \\
			\cline{3-10}
			\cline{3-10}
			& &	\textbf{MSE}$\downarrow$ & \textbf{PSNR} $\uparrow$ & \textbf{SSIM}$\uparrow$ & \textbf{LPIPS}$\downarrow$ & \textbf{MSE}$\downarrow$ & \textbf{PSNR} $\uparrow$ & \textbf{SSIM}$\uparrow$ & \textbf{LPIPS}$\downarrow$ \\
			\hline
			&input  &5.378 &11.840  &0.063 &0.711 &4.803 &11.880  &0.075 &0.796 \\
			\hline
			SL&SID \cite{Chen18} &0.140 &28.614   &0.757 &0.465 &0.235  &26.663 &0.680&0.586\\
			&EEMEFN \cite{ZhuAAAI20} &{\color{red}{0.126}} &{\color{red}{29.212}}   &{\color{red}{0.768}} &{\color{red}{0.448}}&{\color{red}{0.191}} &{\color{red}{27.423}}  &{\color{red}{0.695}} &{\color{red}{0.546}}\\				
			\hline
		\end{tabular}
	\end{center}
	\label{Table:Quan_11}
	\vspace{-4pt}
\end{table*}

\noindent
\textbf{Qualitative Comparison.}
We first present the results of different methods on the images sampled from LOL-test and MIT-Adobe FiveK-test datasets in Figures \ref{fig:LOL} and \ref{fig:5K}. 

As shown in Figure \ref{fig:LOL}, all methods improve the brightness and contrast of the input image. However, none of them successfully recovers the accurate color of the input image when the results are compared with the ground truth. In particular, LLNet \cite{LLNet} produces blurring result.  LightenNet \cite{LightenNet} and RRDNet \cite{RRDNet} produce under-exposed results while MBLLEN \cite{LvBMVC2018} and ExCNet \cite{ZhangACM191} over-expose the image. KinD \cite{ZhangACM19}, KinD++  \cite{GuoIJCV2020}, TBEFN \cite{TBEFN}, DSLR \cite{DSLR}, EnlightenGAN \cite{EnlightenGAN}, and DRBN \cite{YangCVRP20} introduce obvious artifacts.
In Figure \ref{fig:5K}, LLNet \cite{LightenNet}, KinD++ \cite{GuoIJCV2020}, TBEFN \cite{TBEFN}, and RRDNet \cite{RRDNet} produce over-exposed results. Retinex-Net \cite{ChenBMVC18}, KinD++ \cite{GuoIJCV2020}, and RRDNet \cite{RRDNet} yield artifacts and blurring in the results. 

We found that the ground truths of MIT-Adobe FiveK dataset still contain some dark regions. This is because the dataset is originally designed for global image retouching, where restoring low light regions is not the main priority in this task. 
We also observed that the input images in LOL dataset and MIT-Adobe FiveK dataset are relatively clean from noise, which is different from real low-light scenes.  
Although some methods \cite{Wang2019, LiTMM20, DSLR} take the MIT-Adobe FiveK dataset as the training or testing dataset, we argue that this dataset is not appropriate for the task of LLIE due to its mismatched/unsatisfactory ground truth for LLIE.

To examine the generalization capability of different methods, we conduct comparisons on the images sampled from our  LLIV-Phone-imgT dataset. The visual results of different methods are shown in Figures \ref{fig:Phone1} and \ref{fig:Phone2}.
As presented in Figure \ref{fig:Phone1}, all methods cannot effectively improve the brightness and remove the noise of the input low-light image. Moreover, Retinex-Net \cite{ChenBMVC18},  MBLLEN \cite{LvBMVC2018}, and DRBN \cite{YangCVRP20} produce obvious artifacts.
In Figure \ref{fig:Phone2}, all methods enhance the brightness of this input image. However, only 
MBLLEN \cite{LvBMVC2018} and RRDNet \cite{RRDNet} obtain visually pleasing enhancement without color deviation, artifacts, and over-/under-exposure.  Notably, for regions with a light source, none of the methods can brighten the image without amplifying the noise around these regions. 
Taking light sources into account for LLIE would be an interesting direction to explore.
The results suggest the difficulty of enhancing the images of the LLIV-Phone-imgT dataset. 
Real low-light images fail most existing LLIE methods due to the limited generalization capability of these methods. The potential reasons are the use of synthetic training data, small-scaled training data, or unrealistic assumptions such as the local illumination consistency and treating the reflectance component as the final result in the Retinex model in these methods.

We further present the visual comparisons of raw format-based methods in Figure \ref{fig:raw}. As shown, the input raw data have obvious noises. Both SID \cite{Chen2018} and EEMEFN \cite{ZhuAAAI20} can effectively remove the effects of noises. In comparison to the simple U-Net structure used in SID \cite{Chen2018}, the more complex structure of EEMEFN \cite{ZhuAAAI20} obtains better brightness recovery. However, their results are far from the corresponding GT, especially for the input of APS-C X-Trans pattern.

\noindent
\textbf{Quantitative Comparison.}
For test sets with ground truth i.e., LOL-test, MIT-Adobe FiveK-test, and SID-test, we adopt MSE, PSNR,  SSIM \cite{SSIM}, and LPIPS \cite{LPIPS} metrics to quantitatively compare different methods. LPIPS \cite{LPIPS} is a deep learning-based image quality assessment metric that measures the perceptual similarity between a result and its corresponding ground truth by deep visual representations. For LPIPS, we employ the AlexNet-based model to compute the perceptual similarity. A lower LPIPS
value suggests a result that is closer to the corresponding ground truth in terms of perceptual similarity. 
In Table \ref{Table:Quan_1} and Table \ref{Table:Quan_11}, we show the quantitative results of RGB format-based methods and raw format-based methods, respectively.

As presented in Table \ref{Table:Quan_1}, the quantitative scores of supervised learning-based methods are better than those of unsupervised learning-based, semi-supervised learning-based, and zero-shot learning-based methods on LOL-test and MIT-Adobe FiveK-test datasets. Among them, LLNet \cite{LLNet} obtains the best MSE and PSNR values on the LOL-test dataset; however, its performance drops on the MIT-Adobe FiveK-test dataset. 
This may be caused by the bias of LLNet \cite{LLNet} towards the LOL dataset since it was trained using the LOL training dataset. 
For the LOL-test dataset, TBEFN \cite{TBEFN} obtains the highest SSIM value while KinD \cite{ZhangACM19} achieves the lowest LPIPS value. There is no winner across these four evaluation metrics on the LOL-test dataset despite the fact that some methods were trained on the LOL training dataset. 
For the MIT-Adobe FiveK-test dataset, MBLLEN \cite{LvBMVC2018} outperforms all compared methods under the four evaluation metrics in spite of being trained on synthetic training data. Nevertheless, MBLLEN \cite{LvBMVC2018} still cannot obtain the best performance on both two test datasets.

As presented in Table \ref{Table:Quan_11}, both SID \cite{Chen18} and EEMEFN \cite{ZhuAAAI20} improve the quality of input raw data. Compared with the quantitative scores of SID \cite{Chen18}, EEMEFN \cite{ZhuAAAI20} achieves consistently better performance across different raw data patterns and evaluation metrics.

For LLIV-Phone-imgT test set, we use the non-reference IQA metrics, i.e., NIQE \cite{NIQE},  perceptual index (PI) \cite{BlauCVPR2018,MaCVIU2017,NIQE}, LOE \cite{Wang2013}, and SPAQ \cite{SPAQ} to quantitatively compare different methods. In terms of LOE, the smaller the LOE value is, the better the lightness order is preserved. For NIQE, the smaller the NIQE value is,  the better the visual quality is. A lower PI value indicates better perceptual quality. SPAQ is devised for the perceptual quality assessment of smartphone photography. A larger SPAQ value suggests better perceptual quality of smartphone photography. The quantitative results are provided in Table \ref{Table:Quan_2}.

\begin{table}[ht]
	\caption{Quantitative comparisons on LLIV-Phone-imgT dataset in terms of NIQE \cite{NIQE}, LOE \cite{Wang2013}, PI  \cite{BlauCVPR2018,MaCVIU2017,NIQE}, and SPAQ \cite{SPAQ}. The best result is in {\color{red}{red}} whereas the second and third best results are in {\color{blue}{blue}} and {\color{purple}{purple}} under each case, respectively.}
	\begin{center}
		\setlength{\tabcolsep}{1.7mm}{
			\begin{tabular}{c|c|c|c|c|c}
				\hline
				\multirow{2}{*}{\textbf{Learning}}&\multirow{2}{*}{\textbf{Method}} & \multicolumn{4}{c}{\textbf{LoLi-Phone-imgT}}\\
				\cline{3-6}
				\cline{3-6}
				&	& \textbf{NIQE}$\downarrow$ & \textbf{LOE} $\downarrow$ & \textbf{PI}$\downarrow$ & \textbf{SPAQ}$\uparrow$ \\
				\hline
				&	input  &6.99 &{\color{red}{0.00}} &5.86 &44.45 \\
				\hline
				&	LLNet \cite{LLNet} &5.86 &{\color{blue}{5.86}} &5.66 &40.56 \\
				&	LightenNet \cite{LightenNet}&5.34 &952.33 &4.58 &45.74 \\
				&	
				Retinex-Net \cite{ChenBMVC18} &   5.01 &790.21 &{\color{red}{3.48}} &{\color{red}{50.95}} \\
				SL&MBLLEN \cite{LvBMVC2018}  &5.08 &220.63 & 4.27 &42.50 \\
				&	KinD \cite{ZhangACM19}&4.97 &405.88 &4.37 &44.79 \\
				&	KinD++ \cite{GuoIJCV2020} &{\color{red}{4.73}}
				&681.97 &{\color{blue}{3.99}} &{\color{blue}{46.89}} \\
				&	TBEFN \cite{TBEFN} &4.81 &552.91 & 4.30 &44.14 \\
				&	DSLR \cite{DSLR}&{\color{blue}{4.77}} &447.98 &4.31 & 41.08\\
				\hline
				UL &EnlightenGAN \cite{EnlightenGAN} &{\color{purple}{4.79}} &821.87 &{\color{purple}{4.19}} &45.48 \\
				\hline
				SSL&	DRBN \cite{YangCVRP20}&5.80 &885.75 &5.54 &42.74 \\	
				\hline
				&	ExCNet \cite{ZhangACM191} &5.55 &723.56 &4.38 &46.74 \\			
				ZSL &	Zero-DCE \cite{ZeroDCE} &5.82 &307.09 & 4.76 &{\color{purple}{46.85}} \\ 			
				&	RRDNet \cite{RRDNet}&5.97 &{\color{purple}{142.89}} &4.84 &45.31 \\ 							
				\hline
		\end{tabular}}
	\end{center}
	\label{Table:Quan_2}
	\vspace{-4pt}
\end{table}

\begin{table}[t]
	\caption{Quantitative comparisons on LLIV-Phone-vidT dataset in terms of average luminance variance (ALV)  score. The best result is in {\color{red}{red}} whereas the second and third best results are in {\color{blue}{blue}} and {\color{purple}{purple}}.}
	\begin{center}
		\begin{tabular}{c|c|c}
			\hline
			\multirow{2}{*}{\textbf{Learning}}&\multirow{2}{*}{\textbf{Method}} & \multicolumn{1}{c}{\textbf{LoLi-Phone-vidT}}\\
			\cline{3-3}
			\cline{3-3}
			&	& \textbf{ALV}$\downarrow$\\
			\hline
			&	input  &185.60   \\
			\hline
			&	LLNet \cite{LLNet} &{\color{blue}{85.72}}  \\
			&	LightenNet \cite{LightenNet}&643.93\\
			&	
			Retinex-Net \cite{ChenBMVC18} &94.05   \\
			SL&MBLLEN \cite{LvBMVC2018}  &113.18  \\
			&	KinD \cite{ZhangACM19}&98.05  \\
			&	KinD++ \cite{GuoIJCV2020} &115.21\\
			&	TBEFN \cite{TBEFN} &{\color{red}{58.69}} \\
			&	DSLR \cite{DSLR}&175.35  \\
			\hline
			UL &EnlightenGAN \cite{EnlightenGAN} &{\color{purple}{90.69}}  \\
			\hline
			SSL&	DRBN \cite{YangCVRP20}&115.04  \\	
			\hline
			&	ExCNet \cite{ZhangACM191} &1375.29   \\			
			ZSL &	Zero-DCE \cite{ZeroDCE} &117.22  \\ 			
			&	RRDNet \cite{RRDNet}&147.11  \\ 							
			\hline
		\end{tabular}
	\end{center}
	\label{Table:Quan_33}
	\vspace{-4pt}
\end{table}

Observing Table \ref{Table:Quan_2}, we can find that the performance of Retinex-Net \cite{ChenBMVC18}, KinD++ \cite{GuoIJCV2020}, and EnlightenGAN \cite{EnlightenGAN} is relatively better than the other methods. Retinex-Net \cite{ChenBMVC18} achieves the best PI and SPAQ scores. The scores suggest the good perceptual quality of the results enhanced by Retinex-Net \cite{ChenBMVC18}. However, from Figure~\ref{fig:Phone1}(d) and Figure~\ref{fig:Phone2}(d), the results of Retinex-Net \cite{ChenBMVC18} evidently suffer from artifacts and color deviations.
Moreover, KinD++ \cite{GuoIJCV2020} attains the lowest NIQE score while the original input achieves the lowest LOE score. For the de-facto standard LOE metric, we question if the lightness order can effectively reflect the enhancement performance. Overall, the non-reference IQA metrics experience biases on the evaluations of the quality of enhanced low-light images in some cases.

To prepare videos in the LLIV-vidT testing set,  we first discard videos without obvious objects in consecutive frames. A total of 10 videos are chosen. For each video, we select one object that appears in all frames. We then use a tracker \cite{DanelljanCVPR17} to track the object in consecutive frames of the input video and ensure the same object appears in the bounding boxes. We discard the frames with inaccurate object tracking. The coordinates of the bounding box in each frame are collected. We employ these coordinates to crop the corresponding regions in the results enhanced by different methods and compute the average luminance variance (ALV) scores of the object in the consecutive frames as: $ALV=\frac{1}{N}\sum\limits_{i=1}^{N}(L_{i}-L_{\text{avg}})^2$, where $N$ is the number of frames of a video, $L_{i}$ represents the average luminance value of the region of bounding box in the $i$th frame, and $L_{\text{avg}}$ denotes the average luminance value of all bounding box regions in the video. A lower ALV value suggests better temporal coherence of the enhanced video. 
The ALV values of different methods averaged over the 10 videos of the LLIV-vidT testing set are shown in Table \ref{Table:Quan_33}. 
The ALV values of different methods on each video can be found in the supplementary material. Besides, we follow Jiang and Zheng \cite{JZICCV19} to plot their luminance curves in the supplementary material.

As shown in Table \ref{Table:Quan_33}, 	TBEFN \cite{TBEFN} obtains the best temporal coherence in terms of ALV value whereas LLNet \cite{LLNet} and EnlightenGAN \cite{EnlightenGAN} rank the second and third best, respectively. In contrast, the ALV value of ExCNet \cite{ZhangACM191}, as the worst performer, reaches 1375.29. This is because the performance of the zero-reference learning-based ExCNet \cite{ZhangACM191} is unstable for the enhancement of consecutive frames.  ExCNet \cite{ZhangACM191} can effectively improve the brightness of some frames while it does not work well on other frames.

\begin{table*}[t]
	\caption{Quantitative comparisons of computational complexity in terms of runtime (in second), number of trainable parameters (\#Parameters) (in M), and FLOPs (in G).  The best result is in {\color{red}{red}} whereas the second and third best results are in {\color{blue}{blue}} and {\color{purple}{purple}} under each case, respectively.  `-' indicates the result is not available.}
	\vspace{-6pt}
	\begin{center}
		\begin{tabular}{c|c|c|c|c|c}
			\hline
			\textbf{Learning}&	\textbf{Method}  
			& \textbf{RunTime}$\downarrow$ & \textbf{\#Parameters} $\downarrow$ & \textbf{FLOPs}$\downarrow$ & \textbf{Platform}\\
			\hline
			&	LLNet \cite{LLNet} &36.270 &17.908  &4124.177 & Theano\\
			&	LightenNet \cite{LightenNet} & -&{\color{red}{0.030}}  &{\color{red}{30.540}} &MATLAB\\
			&	Retinex-Net \cite{ChenBMVC18} &0.120 &0.555 &587.470 &TensorFlow\\
			SL&	MBLLEN \cite{LvBMVC2018}  &13.995&{\color{purple}{0.450}}&301.120 &TensorFlow\\
			&	KinD \cite{ZhangACM19} &0.148 &8.160&574.954 &TensorFlow\\
			&	KinD++ \cite{GuoIJCV2020} &1.068&8.275&12238.026 &TensorFlow\\
			&	TBEFN \cite{TBEFN} &{\color{purple}{0.050}}&0.486&108.532  &TensorFlow\\
			&	DSLR \cite{DSLR} &0.074&14.931&{\color{purple}{96.683}}  &PyTorch\\
			\hline
			UL&	EnlightenGAN \cite{EnlightenGAN} &{\color{blue}{0.008}}&8.637& 273.240 &PyTorch\\
			\hline
			SSL	&	DRBN \cite{YangCVRP20} &0.878&0.577&196.359 &PyTorch\\	
			\hline
			&	ExCNet \cite{ZhangACM191} &23.280&8.274&- &PyTorch\\			
			ZSL	&	Zero-DCE \cite{ZeroDCE} &{\color{red}{0.003}}&{\color{blue}{0.079}}&{\color{blue}{84.990}} &PyTorch \\				
			&	RRDNet \cite{RRDNet} &167.260 &0.128&- &PyTorch\\								
			\hline
		\end{tabular}
	\end{center}
	\label{Table:Quan_4}
	\vspace{-4pt}
\end{table*}

\subsection{Computational Complexity}

In Table \ref{Table:Quan_4}, we compare the computational complexity of RGB format-based methods, including runtime, trainable parameters, and FLOPs averaged over 32 images of size 1200$\times$900$\times$3 using an NVIDIA 1080Ti GPU. We omit LightenNet \cite{LightenNet} for fair comparisons because only the CPU version of its code is publicly available.
Besides, we do not report the FLOPs of ExCNet \cite{ZhangACM191} and RRDNet \cite{RRDNet} as the number depends on the input images (different inputs require different numbers of iterations).

As presented in Table \ref{Table:Quan_4}, Zero-DCE \cite{ZeroDCE}  has the shortest runtime because it only estimates several curve parameters via a lightweight network. As a result, its number of trainable parameters and FLOPs are much fewer. Moreover, the number of trainable parameters and FLOPs of LightenNet  \cite{LightenNet} are the least among the compared methods. This is because LightenNet  \cite{LightenNet} estimates the illumination map of input image via a tiny network of four convolutional layers. In contrast, the FLOPs of LLNet \cite{LLNet} and KinD++ \cite{GuoIJCV2020} are extremely large, reaching 4124.177G and 12238.026G, respectively.  The runtime of SSL-based ExCNet \cite{ZhangACM191} and RRDNet \cite{RRDNet} is long due to the time-consuming optimization process.

\begin{figure}[!t]
	\centering  \centerline{\includegraphics[width=1\linewidth]{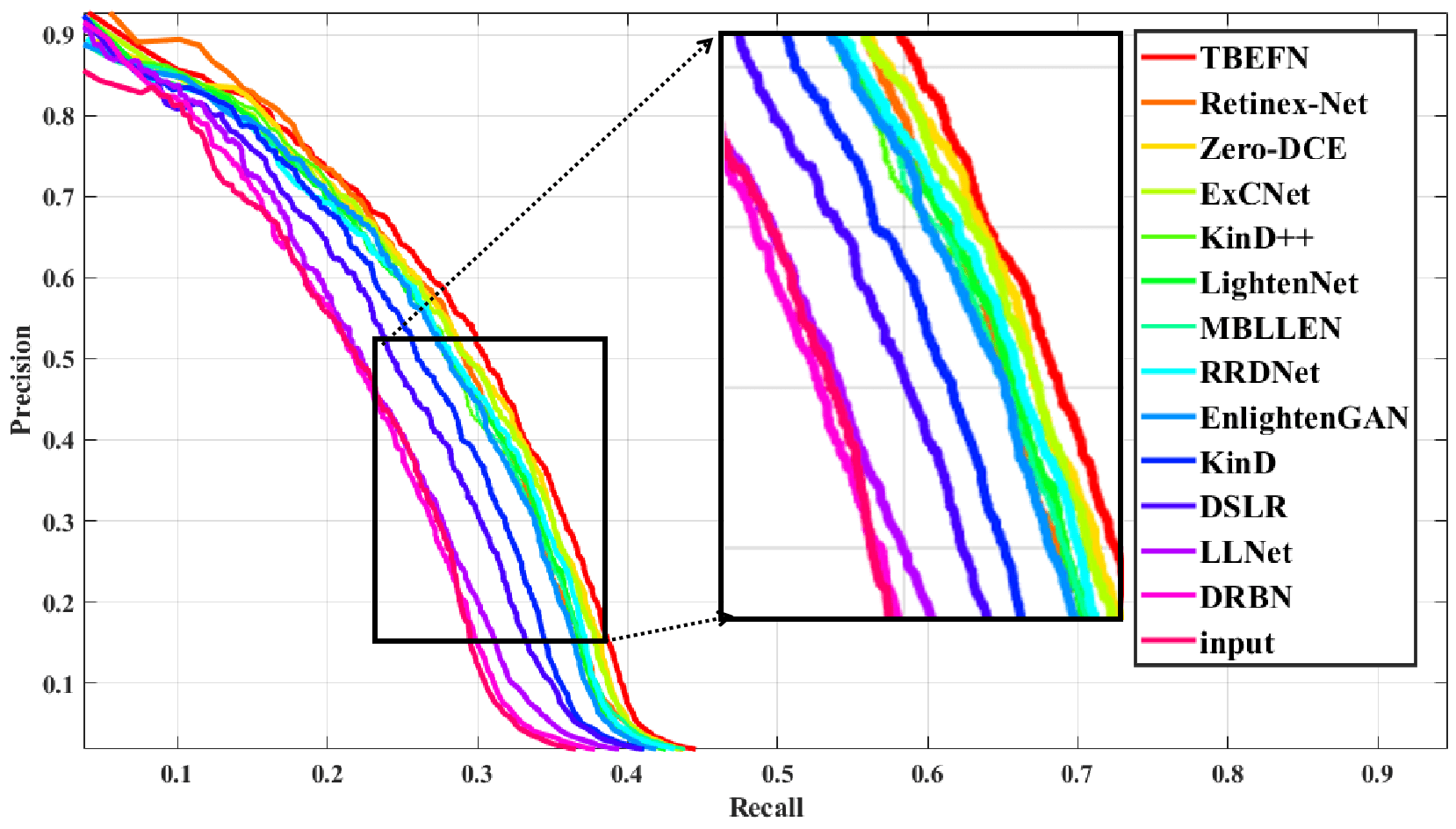}}
	\vspace{-2pt}
	\caption{The P-R curves  of face detection in the dark.}
	\label{fig:face}
	\vspace{-4pt}
\end{figure}

\begin{table}[!t]
	\caption{Quantitative comparisons of AP under different IoU thresholds of face detection in the dark. The best result is in {\color{red}{red}} whereas the second and third best results are in {\color{blue}{blue}} and {\color{purple}{purple}} under each case, respectively.}
	\begin{center}
		\begin{tabular}{c|c|c|c|c}
			\hline
			
			\multirow{2}{*}{\textbf{Learning}}&\multirow{2}{*}{\textbf{Method}}&\multicolumn{3}{c}{\textbf{IoU thresholds}} \\
			\cline{3-5}
			\cline{3-5}
			&	 & \textbf{0.5} & \textbf{0.6}  & \textbf{0.7}\\
			\hline
			&input  &0.195  &0.061   &0.007\\
			\hline
			&	LLNet \cite{LLNet} &0.208  &0.063&0.006\\
			&	LightenNet \cite{LightenNet} &0.249  &0.085&{\color{purple}{0.010}}\\
			&	Retinex-Net \cite{ChenBMVC18} &{\color{blue}{0.261}}  &{\color{red}{0.101}}&{\color{red}{0.013}}\\
			SL&	MBLLEN \cite{LvBMVC2018}  &0.249  &{\color{purple}{0.092}}&{\color{purple}{0.010}}\\
			&	KinD \cite{ZhangACM19} &0.235  &0.081&{\color{purple}{0.010}}\\
			&	KinD++ \cite{GuoIJCV2020} &0.251  &0.090&{\color{blue}{0.011}}\\
			&	TBEFN \cite{TBEFN} &{\color{red}{0.268}}  &{\color{blue}{0.099}}&{\color{blue}{0.011}}\\
			&	DSLR \cite{DSLR} &0.223  &0.067&0.007\\
			\hline
			UL&	EnlightenGAN \cite{EnlightenGAN} &0.246  &0.088&{\color{blue}{0.011}}\\
			\hline
			SSL	&	DRBN \cite{YangCVRP20} &0.199  &0.061&0.007\\	
			\hline
			&	ExCNet \cite{ZhangACM191} &0.256  &{\color{purple}{0.092}}&{\color{purple}{0.010}}\\			
			ZSL	&	Zero-DCE \cite{ZeroDCE}  &{\color{purple}{0.259}}  &{\color{purple}{0.092}}&{\color{blue}{0.011}}\\				
			&	RRDNet \cite{RRDNet} &0.248  &0.083&{\color{purple}{0.010}}\\								
			\hline
		\end{tabular}
	\end{center}
	\label{Table:Quan_8}
	\vspace{-4pt}
\end{table}

\subsection {Application-Based Evaluation}

We investigate the performance of low-light image enhancement methods on face detection in the dark. 
Following the setting presented in Guo et al. \cite{ZeroDCE}, we use the DARK FACE dataset \cite{Yuan2019} that is composed of images with faces taken in the dark. Since the bounding boxes of the test set are not publicly available, we perform the evaluation on 500 images randomly sampled from the training and validation sets.
The  Dual Shot Face Detector (DSFD) \cite{DSFD} trained on WIDER FACE dataset \cite{Widerface} is used as the face detector. We feed the results of different LLIE methods to the DSFD \cite{DSFD} and depict the precision-recall (P-R) curves under 0.5 IoU threshold in Figure \ref{fig:face}. We compare the average precision (AP) under different IoU thresholds using the evaluation tool\footnote{\url{https://github.com/Ir1d/DARKFACE_eval_tools}} provided in DARK FACE dataset \cite{Yuan2019} in Table \ref{Table:Quan_8}. 

As shown in Figure \ref{fig:face}, all the deep learning-based solutions improve the performance of face detection in the dark, suggesting the effectiveness of deep learning-based LLIE solutions for face detection in the dark. As shown in Table \ref{Table:Quan_8}, the AP scores of best performers under different IoU thresholds range from 0.268 to 0.013 and the AP scores of input under different IoU thresholds are very low. The results suggest that there is still room for improvement. It is noteworthy that Retinex-Net \cite{ChenBMVC18}, Zero-DCE \cite{ZeroDCE}, and TBEFN \cite{TBEFN} achieve relatively robust performance on face detection in the dark. We show the visual results of different methods in Figure \ref{fig:face_visual}. Although Retinex-Net \cite{ChenBMVC18} performs better than other methods on the AP score, its visual result contains obvious artifacts and unnatural textures. In general, Zero-DCE \cite{ZeroDCE} obtains a good balance between the AP score and the perceptual quality for face detection in the dark. 	Note that the results of face detection in the dark are related to not only the enhanced results but also the face detector including the detector model and the training data of the detector. Here, we only take the pre-trained DSFD \cite{DSFD} as an example to validate the low-light image enhancement performance of different methods to some extent.

\subsection{Discussion}
From the experimental results, we obtain several interesting observations and insights:

\begin{figure*}[!t]
	\begin{center}
		\begin{tabular}{c@{ }c@{ }c@{ }c@{ }}
			\includegraphics[width=0.24\linewidth]{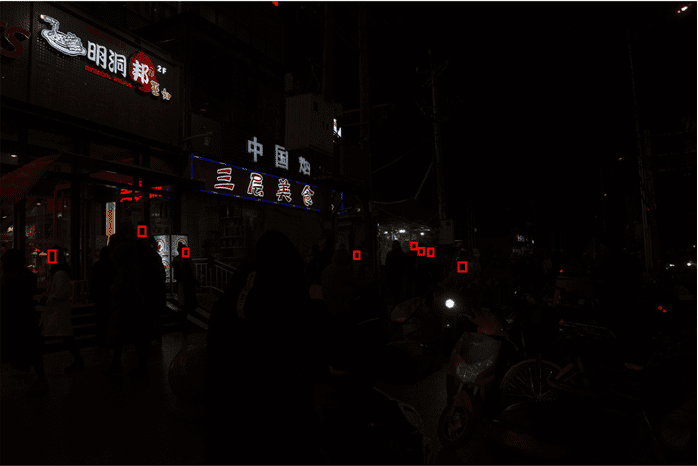}&
			\includegraphics[width=0.24\linewidth]{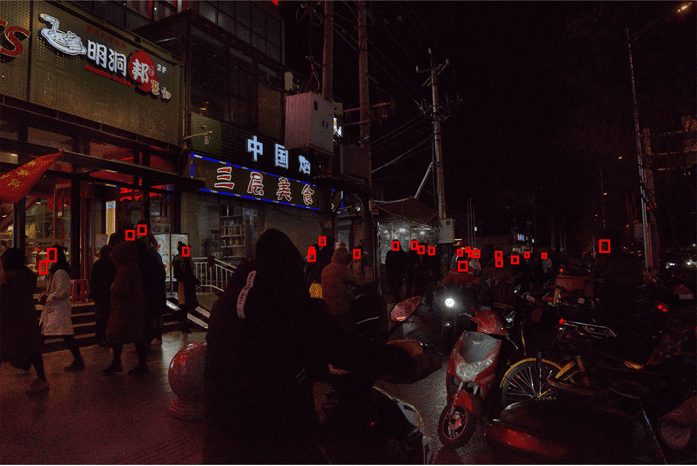}&
			\includegraphics[width=0.24\linewidth]{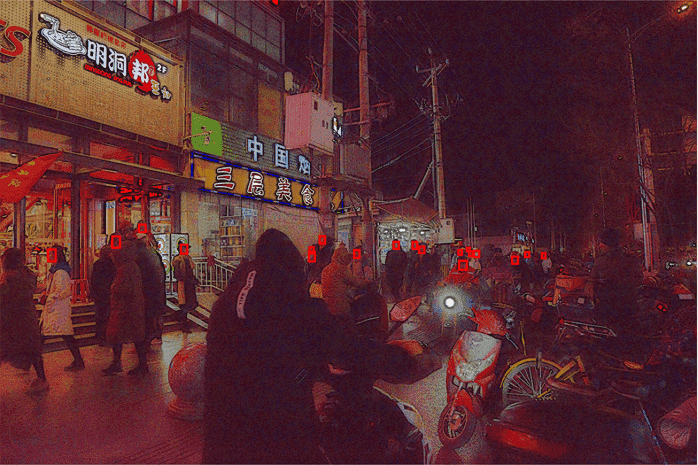}&
			\includegraphics[width=0.24\linewidth]{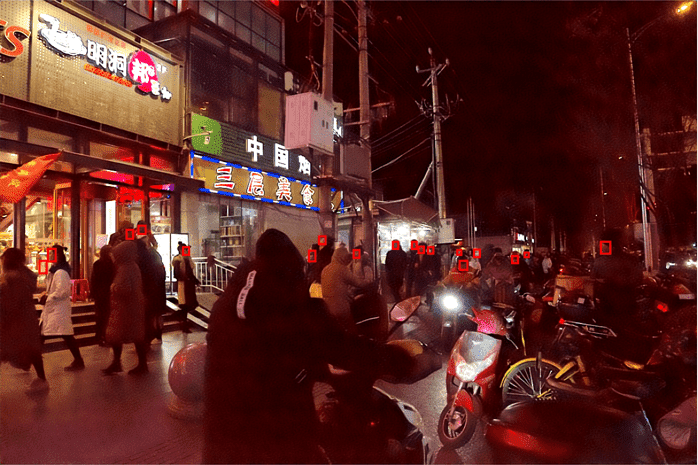}\\
			(a) input  & (b) LightenNet \cite{LightenNet} &  (c) Retinex-Net \cite{ChenBMVC18} & (d) 	MBLLEN \cite{LvBMVC2018} \\
			\includegraphics[width=0.24\linewidth]{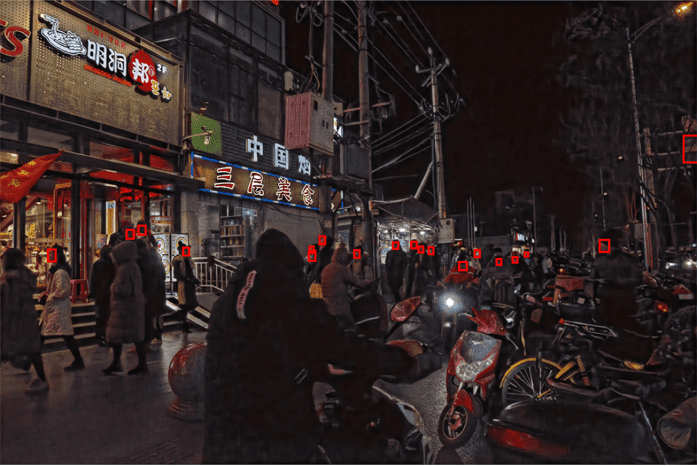}&
			\includegraphics[width=0.24\linewidth]{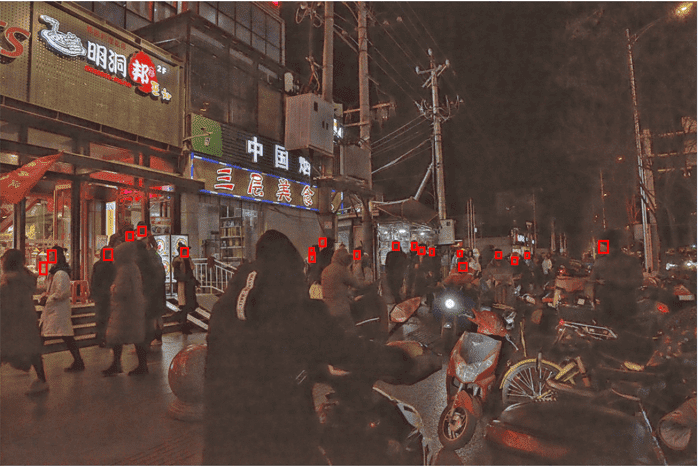}&
			\includegraphics[width=0.24\linewidth]{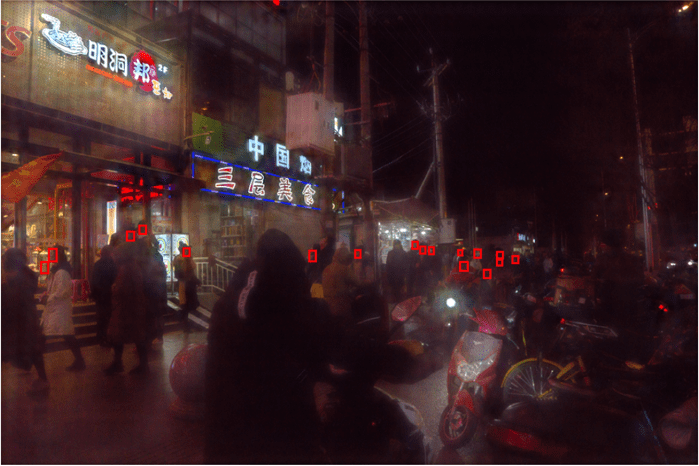}&
			\includegraphics[width=0.24\linewidth]{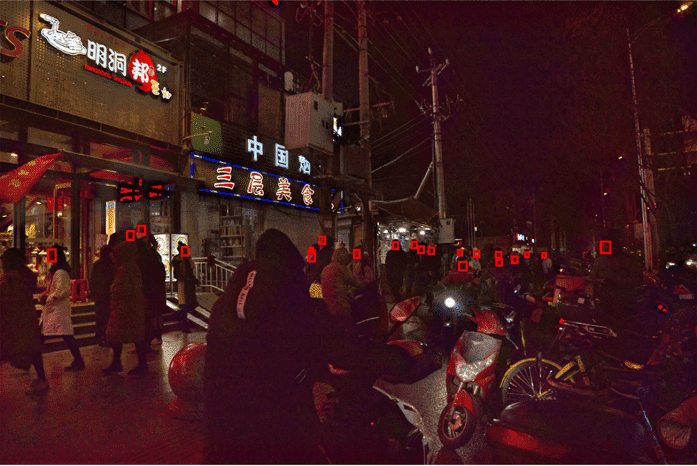}\\
			(e)  KinD++ \cite{GuoIJCV2020}  &	(f) TBEFN \cite{TBEFN} &  (g) 	DSLR \cite{DSLR} 	  & (h) EnlightenGAN \cite{EnlightenGAN} \\
			\includegraphics[width=0.24\linewidth]{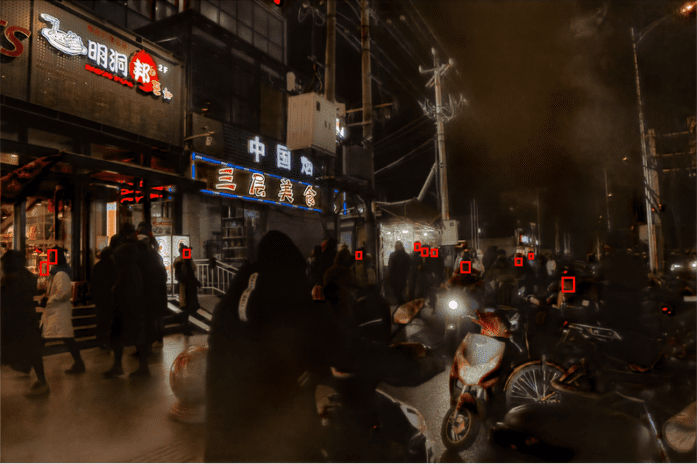}&
			\includegraphics[width=0.24\linewidth]{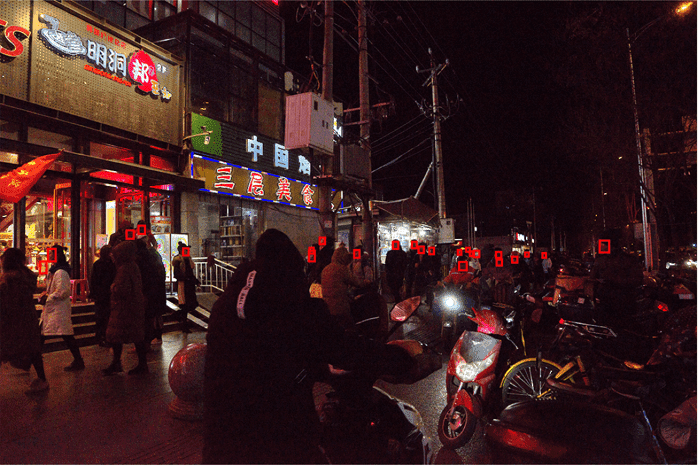}&
			\includegraphics[width=0.24\linewidth]{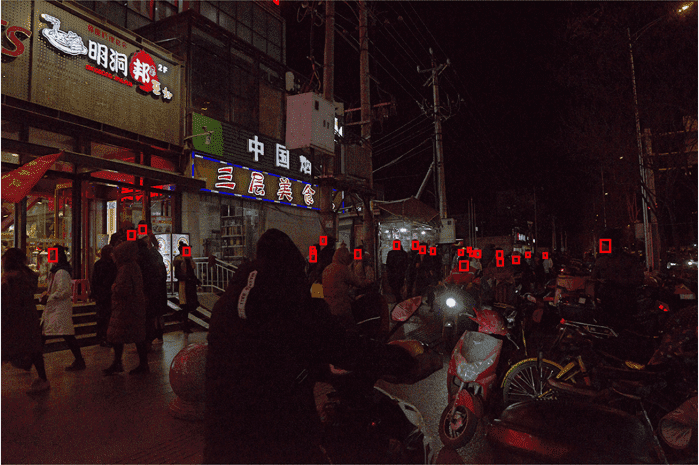}&
			\includegraphics[width=0.24\linewidth]{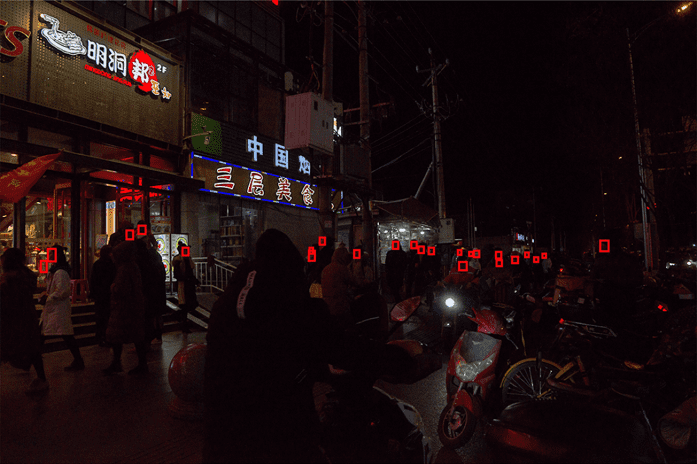}\\
			(i) DRBN \cite{YangCVRP20} & (j)	ExCNet \cite{ZhangACM191} & (k) Zero-DCE \cite{ZeroDCE} &  (l) RRDNet \cite{RRDNet}	 \\
		\end{tabular}
	\end{center}
	\vspace{-2pt}
	\caption{Visual results of different methods on a low-light image sampled from  DARK FACE dataset. Better see with zoom in for the bounding boxes of faces.}
	\label{fig:face_visual}
	\vspace{-4pt}
\end{figure*}

1) The performance of different methods significantly varies  based on the test datasets and evaluation metrics. In terms of the full-reference IQA metrics on commonly used test datasets, MBLLEN \cite{LvBMVC2018}, KinD++ \cite{GuoIJCV2020}, and DSLR \cite{DSLR} are generally better  than other compared methods. For real-world low-light images taken by mobile phones, supervised learning-based Retinex-Net \cite{ChenBMVC18} and KinD++ \cite{GuoIJCV2020}  obtain better scores measured in the non-reference IQA metrics. For real-world low-light videos taken by mobile  phones, TBEFN \cite{TBEFN} preserves the temporal coherence better. When coming to the computational efficiency, LightenNet  \cite{LightenNet} and Zero-DCE \cite{ZeroDCE} are outstanding. From the aspect of face detection in the dark, TBEFN \cite{TBEFN}, Retinex-Net \cite{ChenBMVC18}, and  Zero-DCE \cite{ZeroDCE} rank the first three. No method always wins. Overall, Retinex-Net \cite{ChenBMVC18}, \cite{TBEFN}, Zero-DCE \cite{ZeroDCE}, and DSLR \cite{DSLR} are better choice in most cases. For raw data, EEMEFN \cite{ZhuAAAI20} obtains relatively better qualitative and quantitative performance than SID \cite{Chen18}. However, from the visual results,  EEMEFN \cite{ZhuAAAI20} and \cite{Chen18} cannot recover the color well when compared with the corresponding ground truth.

2) LLIV-Phone dataset fails most methods. The generalization capability of existing methods needs further improvements. It is worth noting that it is inadequate to use only the average luminance variance to evaluate the performance of different methods for low-light video enhancement. More effective and comprehensive assessment metrics would guide the development of low-light video enhancement towards the right track.

3) Regarding learning strategies, supervised learning achieves better performance in most cases, but requires high computational resources and paired training data. In comparison, zero-shot learning is more appealing in practical applications because it does not require paired or unpaired training data. Consequently, zero-shot learning-based methods enjoy better generalization capability. However,  their quantitative performance is inferior to other methods. 

4) There is a gap between visual results and quantitative IQA scores. In other words, a good visual appearance does not always yield a good IQA score. The relationships between human perception and IQA scores are worth more investigation. Pursuing better visual perception or quantitative scores depends on specific applications. For instance, to show the results to observers, more attention should be paid to visual perception. In contrast, accuracy is more important when LLIE methods are applied to face detection in the dark. Thus, more comprehensive comparisons should be performed when comparing different methods.

5) Deep learning-based LLIE methods are beneficial to face detection in the dark. Such results further support the significance of enhancing low-light images and videos. However, in comparison to the high accuracy of face detection in normal-light images, the accuracy of face detection in the dark is extremely low, despite using LLIE  methods.

6) In comparison to RGB format-based LLIE methods, raw format-based LLIE methods usually recover details better, obtain more vivid color, and reduce noises and artifacts more effectively. This is because raw data contain more information such as wider color gamut and higher dynamic range. However, raw format-based LLIE methods are limited to specific sensors and formats such as the Bayer pattern of the Sony camera and the APS-C X-Trans pattern of the Fuji camera. In contrast,  RGB format-based LLIE methods are more convenient and versatile since RGB images are commonly found as the final imagery form produced by mobile devices. However, RGB format-based LLIE methods cannot cope well with cases that exhibit low light and excessive noise.

\section{Open Issues}
\label{sec:Issue}

In this section, we summarize the open issues in low-light image and video enhancement as follows.

\noindent
\textbf{Generalization Capability.} Although existing methods can produce some visually pleasing results, they have limited generalization capability. For example, a method trained on MIT-Adobe FiveK dataset \cite{Adobe5K} cannot effectively enhance the low-light images of LOL dataset \cite{ChenBMVC18}. Albeit synthetic data are used to augment the diversity of training data, the models trained on the combination of real and synthetic data cannot solve this issue well. Improving the generalization capability of LLIE methods is an unsolved open issue.

\noindent
\textbf{Removing Unknown Noises.} Observing the results of existing methods on the low-light images captured by different types of phones' cameras, we can find that these methods cannot remove the noises well and even amplify the noises, especially when the types of noises are unknown. Despite some methods add Gaussian and/or Poisson noises in their training data, the noise types are different from real noises, thus the performance of these methods is unsatisfactory in real scenarios. Removing unknown noises is still unsolved.

\noindent
\textbf{Removing Unknown Artifacts.} One may enhance a low-light image downloaded from the Internet. The image may have gone through a serial of degradations such as JPEG compression or editing. Thus, the image may contain unknown artifacts. Suppressing unknown artifacts still challenges existing low-light image and video enhancement methods.

\noindent
\textbf{Correcting Uneven Illumination.} Images taken in real scenes usually exhibit uneven illumination. For example, an image captured at night has both dark regions and normal-light or over-exposed regions such as the regions of light sources. Existing methods tend to brighten both the dark regions and the light source regions, affecting the visual quality of the enhanced result. It is expected to enhance dark regions but suppress over-exposed regions. However, this open issue is not solved well in existing LLIE methods.

\noindent
\textbf{Distinguishing Semantic Regions.} Existing methods tend to enhance a low-light image without considering the semantic information of its different regions. For example, the black hair of a man in a low-light image is enhanced to be off-white as the black hair is treated as the low-light regions. An ideal enhancement method is expected to only enhance the low-light regions induced by external environments. How to distinguish semantic regions is an open issue. 

\noindent
\textbf{Using Neighbouring Frames.} 
	Despite some methods that have been proposed to enhance low-light videos, they commonly process a video frame-by-frame. How to make full use of the neighboring frames to improve the enhancement performance and speed up the processing speed is an unsolved open issue. For example, the well-lit regions of neighboring frames are used to enhance the current frame. For another example, the estimated parameters for processing neighboring frames can be reused to enhance the current frame for reducing the time of parameter estimation.

\section{Future Research Directions}
\label{sec:Future}

Low-light enhancement is a challenging research topic. As can be observed from the experiments presented in Section~\ref{sec:evaluation} and the unsolved open issues in Section~\ref{sec:Issue}, there is still room for improvement. We suggest potential future research directions as follows.

\noindent
\textbf{Effective Learning Strategies.}
As aforementioned,  current LLIE models mainly adopt supervised learning that requires massive paired training data and may overfit on a specific dataset.  Although some researchers attempted to introduce unsupervised learning into LLIE, the inherent relationships between LLIE and these learning strategies are not clear and their effectiveness in LLIE needs further improvements. Zero-shot learning has shown robust performance for real scenes while not requiring paired training data. The unique advantage suggests zero-shot learning as a potential research direction, especially on the formulation of zero-reference losses, deep priors, and optimization strategies.  

\noindent
\textbf{Specialized Network Structures.}
A network structure can significantly affect enhancement performance. As previously analyzed, most LLIE deep models employ U-Net or U-Net-like structures. Though they have achieved promising performance in some cases, the investigation if such an encoder-decoder network structure is most suitable for the LLIE task is still lacking.
Some network structures require a high memory footprint and long inference time due to their
large parameter space. Such network structures are unacceptable for practical applications. 
Thus, it is worthwhile to investigate a more effective network structure for LLIE, considering the characteristics of low-light images such as non-uniform illumination, small pixel values, noise suppression, and color constancy. 
One can also design more efficient network structures via taking into account the local similarity of low-light images or considering more efficient operations such as depthwise separable convolution layer \cite{MobileNets} and self-calibrated convolution \cite{Calibrate}. Neural architecture search (NAS) technique \cite{NAS,deeplab} may be considered to obtain more effective and efficient LLIE network structures. Adapting the transformer architecture \cite{Transformer1616,TransformerSR} into LLIE may be a potential and interesting research direction.

\noindent
\textbf{Loss Functions.}
Loss functions constrain the relationships between an input image and ground truth and drive the optimization of deep networks. In LLIE, the commonly used loss functions are borrowed from related vision tasks. 
Thus, designing loss functions that are more well-suited for LLIE is desired. Recent studies have shown the possibility of using deep neural networks to approximate human visual perception of image quality \cite{FangCVPR20,NIMA}. These ideas and fundamental theories could be used to guide the designs of loss functions for low-light enhancement networks.

\noindent
\textbf{Realistic Training Data.}
Although there are several training datasets for LLIE, their authenticity, scales, and diversities fall behind real low-light conditions. Thus, as shown in Section \ref{sec:evaluation}, current LLIE deep models cannot achieve satisfactory performance when encountering low-light images captured in real-world scenes.
More efforts are needed to study the collection of large-scale and diverse real-world paired LLIE training datasets or to generate more realistic synthetic data. 

\noindent
\textbf{Standard Test Data.}
Currently, there is no well-accepted LLIE evaluation benchmark. Researchers prefer selecting their test data that may bias to their proposed methods. Despite some researchers leave some paired data as test data, the division of training and test partitions are mostly ad-hoc across the literature.
Consequently, conducting a fair comparison among different methods is often laborious if not impossible. Besides, some test data are either easy to be handled or not originally collected for low-light enhancement. 
It is desired to have a standard low-light image and video test dataset, which includes a large number of test samples with the corresponding ground truths, covering diverse scenes and challenging illumination conditions.  

\noindent
\textbf{Task-Specific Evaluation Metrics.}
The commonly adopted evaluation metrics in LLIE can reflect the image quality to some extent. However, how to measure how good a result is enhanced by an LLIE method still challenges current IQA metrics, especially for non-reference measurements.
The current IQA metrics either focus on human visual perceptual such as subjective quality or emphasize machine perceptual such as the effects on high-level visual tasks.  Therefore, 
more works are expected in this research direction to make efforts on designing more accurate and task-specific evaluation metrics for LLIE.

\noindent
\textbf{Robust Generalization Capability.}
Observing the experimental results on real-world test data, most methods fail due to their limited generalization capability. The poor generalization is caused by several factors such as synthetic training data, small-scaled training data, ineffective network structures, or unrealistic assumptions.
It is important to explore ways to improve the generalization.

\noindent
\textbf{Extension to Low-Light Video Enhancement.}
Unlike the rapid development of video enhancement in other low-level vision tasks such as video deblurring \cite{KimICCV17}, video denoising \cite{EhretCVPR19}, and video super-resolution \cite{BasicVSR}, low-light video enhancement receives less attention.
A direct application of existing LLIE methods to videos often leads to unsatisfactory results and flickering artifacts. 
More efforts are needed to remove visual flickering effectively, exploit the temporal information between neighboring frames, and speed up the enhancement speed. 

\noindent
\textbf{Integrating Semantic Information.}
Semantic information is crucial for low-light enhancement. It guides the networks to distinguish different regions in the process of enhancement. 
A network without access to semantic priors can easily deviate the original color of a region, e.g., turning black hair to gray color after enhancement. Therefore, integrating semantic priors into LLIE  models is a promising research direction. Similar work has been done on image super-resolution \cite{SFTGAN,GLEAN} and face restoration \cite{DFDNet}.

	\ifCLASSOPTIONcompsoc
	\section*{Acknowledgments}
	\else
	\section*{Acknowledgment}
	\fi
This study is supported under the RIE2020 Industry Alignment Fund Industry Collaboration Projects (IAF-ICP) Funding Initiative, as well as cash and in-kind contribution from the industry partner(s). It is also partially supported by the NTU SUG  and NAP grant. Chunle Guo is sponsored by CAAI-Huawei MindSpore Open Fund.
	
	\ifCLASSOPTIONcaptionsoff
	\newpage
	\fi
	
	{
		\bibliographystyle{IEEEtran}
		\bibliography{bibliography}
	}
	\begin{IEEEbiography}[{\includegraphics[width=1in,height=1.25in,clip,keepaspectratio]{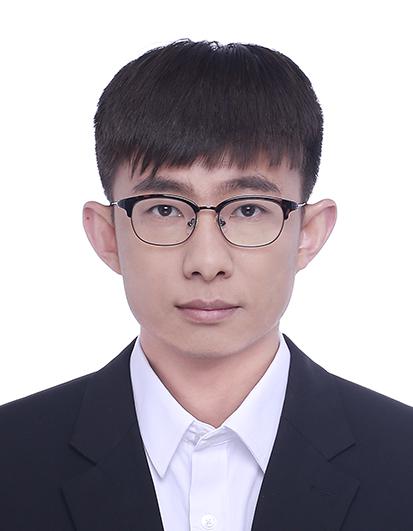}}]{Chongyi Li}  is a Research Assistant Professor with the School of Computer Science and Engineering, Nanyang Technological University, Singapore. He received the Ph.D. degree from Tianjin University, China in  2018. From 2016 to 2017, he was a joint-training Ph.D. Student with Australian National University, Australia.  Prior to joining NTU, he was a postdoctoral fellow with City University of Hong Kong and Nanyang Technological University from 2018 to 2021. His current research focuses on image processing, computer vision, and deep learning, particularly in the domains of image restoration and enhancement. He serves as an associate editor of the Journal of Signal, Image and Video Processing and a lead guest editor of the IEEE Journal of Oceanic Engineering.
	\end{IEEEbiography}
	\begin{IEEEbiography}[{\includegraphics[width=1in,height=1.25in,clip,keepaspectratio]{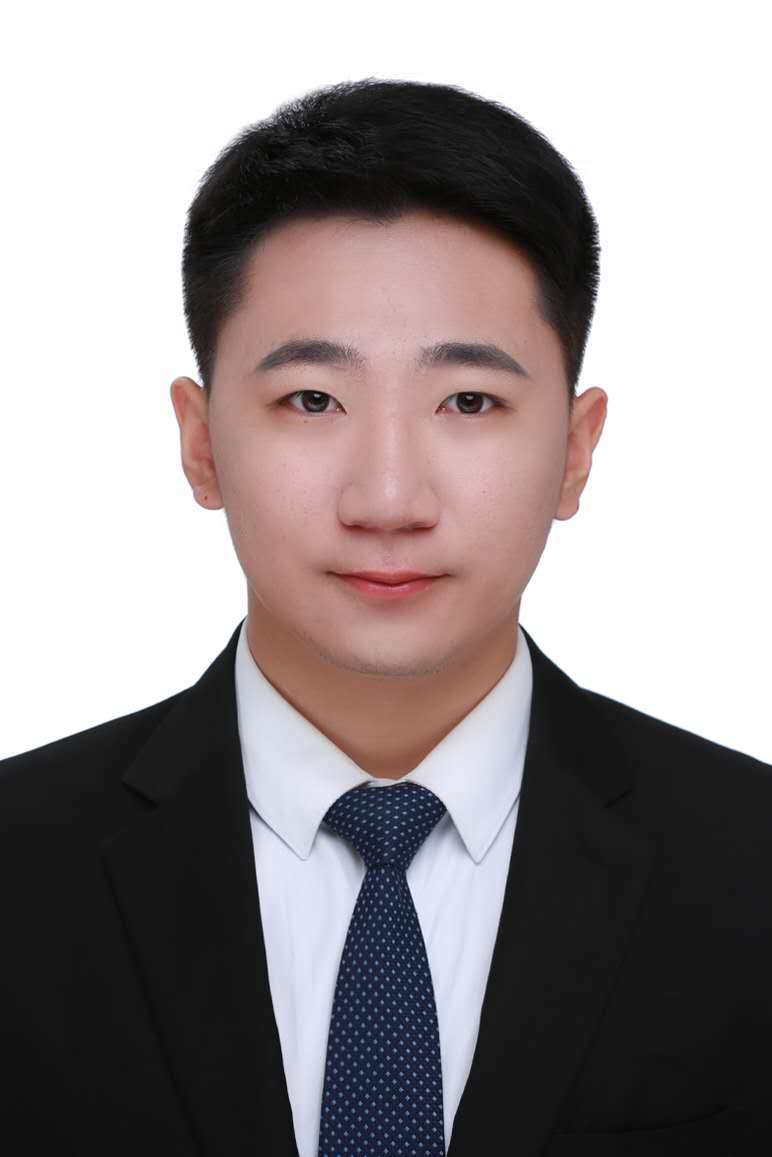}}]{Chunle Guo}  received his PhD degree from Tianjin University in China under the supervision of Prof. Jichang Guo. He conducted the Ph.D. research as a Visiting Student with the School of Electronic Engineering and Computer Science, Queen Mary University of London (QMUL), UK.  He continued his research as a Research Associate with the Department of Computer Science, City University of Hong Kong (CityU), from 2018 to 2019. Now he is a postdoc research fellow working with Prof. Ming-Ming Cheng at Nankai University. His research interests lie in image processing, computer vision, and deep learning.
	\end{IEEEbiography}
	\begin{IEEEbiography}[{\includegraphics[width=1in,height=1.25in,clip,keepaspectratio]{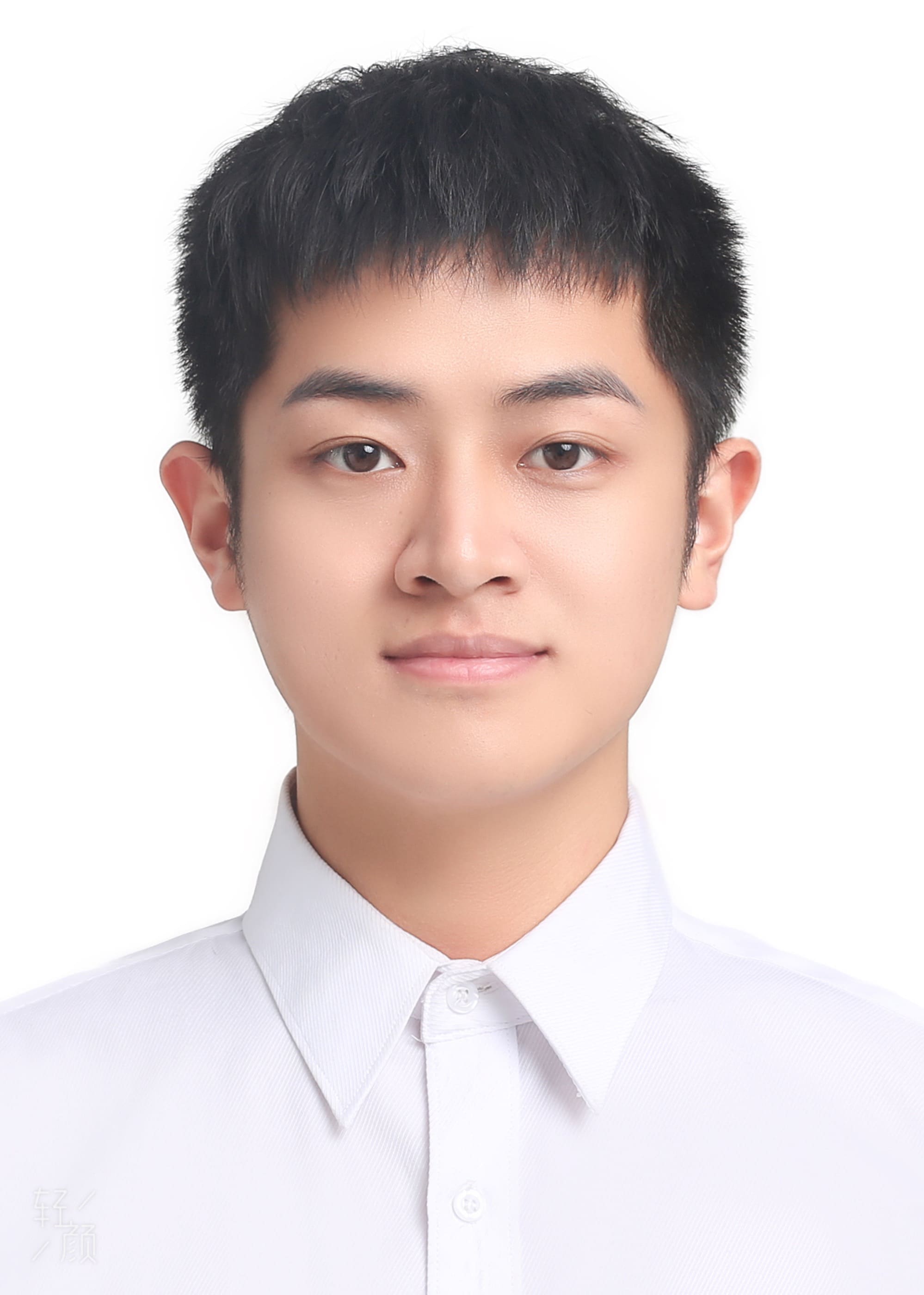}}]{Linhao Han}  is currently a master student at the College of Computer Science, Nankai
		University, under the supervision of Prof. Ming-Ming Cheng. His research interests include
		deep learning and computer vision.
	\end{IEEEbiography}
	\begin{IEEEbiography}[{\includegraphics[width=1in,height=1.25in,clip,keepaspectratio]{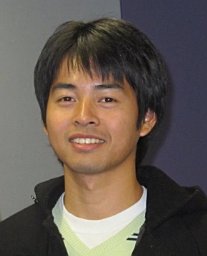}}]{Jun Jiang} received the PhD degree in Color Science from Rochester Institute of Technology in 2013. He is a Senior Researcher in SenseBrain focusing on algorithm development to improve image quality on smartphone cameras. His research interest includes computational photography, low-level computer vision, and deep learning.
	\end{IEEEbiography}
	\begin{IEEEbiography}[{\includegraphics[width=1in,height=1.25in,clip,keepaspectratio]{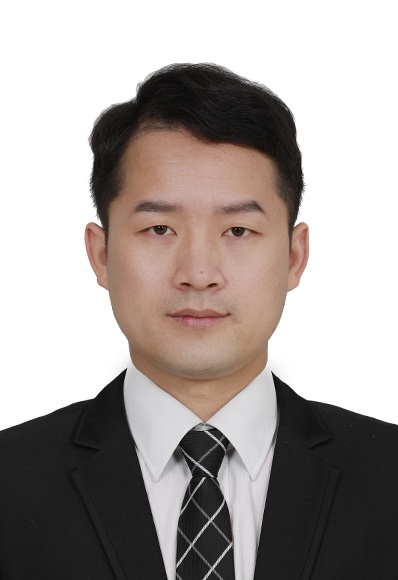}}]{Ming-Ming Cheng} (Senior Member, IEEE) received the Ph.D. degree from Tsinghua University in 2012. Then he did two years research fellowship with Prof. Philip Torr at Oxford. He is currently a Professor at Nankai University and leading the Media Computing Laboratory. His research interests include computer graphics, computer vision, and image processing. He received research awards, including the ACM China Rising Star Award, the IBM Global SUR Award, and the CCF-Intel Young Faculty Researcher Program. He is on the Editorial Board Member of IEEE Transactions on Image Processing (TIP).
	\end{IEEEbiography}
	\begin{IEEEbiography}[{\includegraphics[width=1in,height=1.25in,clip,keepaspectratio]{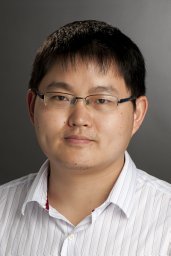}}]{Jinwei Gu} (Senior Member, IEEE) is the R\&D Executive Director of SenseTime USA. His current research focuses on low-level computer vision, computational photography, smart visual sensing and perception, and robotics. He obtained his Ph.D. degree in 2010 from Columbia University, and his B.S and M.S. from Tsinghua University, in 2002 and 2005 respectively. Before joining SenseTime, he was a senior research scientist in NVIDIA Research from 2015 to 2018.  Prior to that, he was an assistant professor in Rochester Institute of Technology from 2010 to 2013, and a senior researcher in the media lab of Futurewei Technologies from 2013 to 2015. He is an associate editor for IEEE Transactions on Computational Imaging and an IEEE senior member since 2018.
	\end{IEEEbiography}
	\begin{IEEEbiography}[{\includegraphics[width=1in,height=1.25in,clip,keepaspectratio]{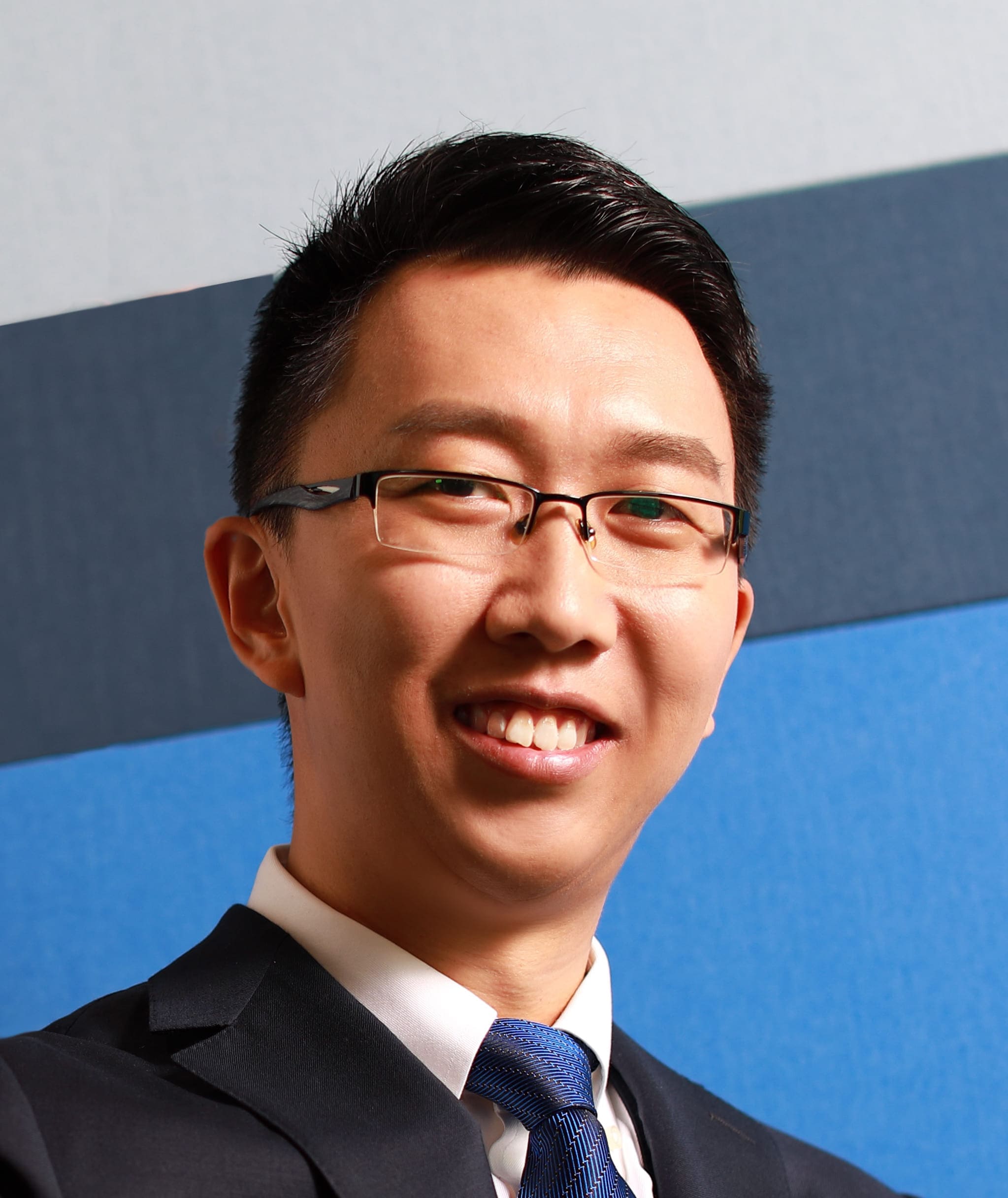}}]{Chen Change Loy} (Senior Member, IEEE) is an Associate Professor with the School of Computer Science and Engineering, Nanyang Technological University, Singapore. He is also an Adjunct Associate Professor at The Chinese University of Hong Kong. He received his Ph.D. (2010) in Computer Science from the Queen Mary University of London. Prior to joining NTU, he served as a Research Assistant Professor at the MMLab of The Chinese University of Hong Kong, from 2013 to 2018. He was a postdoctoral researcher at Queen Mary University of London and Vision Semantics Limited, from 2010 to 2013. He serves as an Associate Editor of the IEEE Transactions on Pattern Analysis and Machine Intelligence and International Journal of Computer Vision. He also serves/served as an Area Chair of major conferences such as ICCV, CVPR, ECCV and AAAI. His research interests include image/video restoration and enhancement, generative tasks, and representation learning. 
		
	\end{IEEEbiography}
	
\end{document}


\title{Low-Light Image and Video Enhancement \\Using Deep Learning: A Survey\\ (Supplementary Material)}

\author{Chongyi Li, Chunle Guo, Linghao Han,  Jun Jiang, Ming-Ming Cheng,~\IEEEmembership{Senior Member,~IEEE},  \\Jinwei Gu,~\IEEEmembership{Senior Member,~IEEE},  and Chen Change Loy,~\IEEEmembership{Senior Member,~IEEE}\\
	~\\
	Project Page: \url{https://www.mmlab-ntu.com/project/lliv_survey/index.html}.

\thanks{C. Li and C. C. Loy are with the S-Lab, Nanyang Technological University (NTU), Singapore  (e-mail: chongyi.li@ntu.edu.sg and ccloy@ntu.edu.sg).}
\thanks{C. Guo, L. Han, and M.-M. Cheng are with the College of Computer Science, Nankai University, Tianjin, China  (e-mail: guochunle@nankai.edu.cn, lhhan@mail.nankai.edu.cn, and cmm@nankai.edu.cn).}
\thanks{J. Jiang and J. Gu are with the SenseTime  (e-mail: jiangjun@sensebrain.site and gujinwei@sensebrain.site).}
\thanks{C. Li and C. Guo contribute equally.}
\thanks{C. C. Loy is the corresponding author.}
}

\markboth{IEEE TRANSACTIONS ON PATTERN ANALYSIS AND MACHINE INTELLIGENCE}%
{Shell \MakeLowercase{\textit{et al.}}: Bare Demo of IEEEtran.cls for Computer Society Journals}

\IEEEtitleabstractindextext{%
\justify  
\begin{itemize}
	\item In this supplementary material, we provide more visual comparisons of the results enhanced by different methods on a variety of input scenes sampled from different testing benchmark datasets. Besides, we follow Jiang and Zheng \cite{JZICCV19} to plot the luminance curves for comparing the temporal coherence of enhanced videos. The smoother the curve is, the better the method for temporal coherence is. We also provide the average luminance variance value (the smaller, the better).
	\item  We also upload the video results of different methods to YouTube at \url{https://www.youtube.com/watch?v=Elo9TkrG5Oo&t=6s}.
	\item One can test the performance of different methods with any inputs on our online platform at \url{http://mc.nankai.edu.cn/ll/}.
	\item We collect low-light image and video enhancement methods, datasets, and evaluation metrics and periodically update the content in at \url{https://github.com/Li-Chongyi/Lighting-the-Darkness-in-the-Deep-Learning-Era-Open}. 
	\item We release our proposed dataset at \url{https://drive.google.com/file/d/1QS4FgT5aTQNYy-eHZ_A89rLoZgx_iysR/view}.
\end{itemize}

~\\
In what follows, we present the visual results of different methods. Specifically, 
~\\

\noindent
Figures \ref{fig:LOL1}, \ref{fig:LOL2}, and \ref{fig:LOL3} show the results enhanced by different deep learning-based low-light image enhancement methods on the low-light images sampled from LOL-test dataset \cite{ChenBMVC18}.
~\\

\noindent
Figures \ref{fig:5K1}, \ref{fig:5K2}, and \ref{fig:5K3} show the results enhanced by different deep learning-based low-light image enhancement methods on the low-light images sampled from MIT-Adobe FiveK-test dataset \cite{Adobe5K}.
~\\

\noindent
Figures \ref{fig:Phone1}, \ref{fig:Phone2}, and \ref{fig:Phone3} show the results enhanced by different deep learning-based low-light image enhancement methods on the low-light images sampled from our proposed LLIV-Phone-imgT dataset.
~\\

\noindent
Figures \ref{fig:Sony} and \ref{fig:Fuji} show the results enhanced by different deep learning-based low-light image enhancement methods on the raw low-light images sampled from SID-test dataset \cite{Chen2018}.
~\\

\noindent
Figures \ref{fig:face_visual1}, \ref{fig:face_visual2}, and \ref{fig:face_visual3} show the results enhanced by different deep learning-based low-light image enhancement methods on the low-light images sampled from  DARK FACE dataset  \cite{Yuan2019} and their face detection resutls.
~\\ 

\noindent
Figures \ref{fig:curve1}, \ref{fig:curve2}, \ref{fig:curve3}, \ref{fig:curve4}, \ref{fig:curve5}, \ref{fig:curve6}, \ref{fig:curve7}, \ref{fig:curve8}, \ref{fig:curve9}, and \ref{fig:curve10} show the luminance curves of the videos enhanced by different methods. These ten low-light videos were taken by different mobile phones' cameras and sampled from our proposed LLIV-Phone-vidT dataset.
~\\

}

\maketitle

\IEEEdisplaynontitleabstractindextext

\IEEEpeerreviewmaketitle

\begin{figure*} [h]
	\begin{center}
		\begin{tabular}{c@{ }c@{ }c@{ }}
			\includegraphics[width=0.3\linewidth]{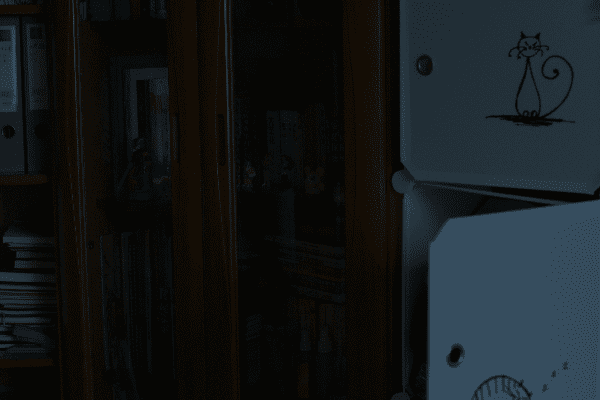}&
			\includegraphics[width=0.3\linewidth]{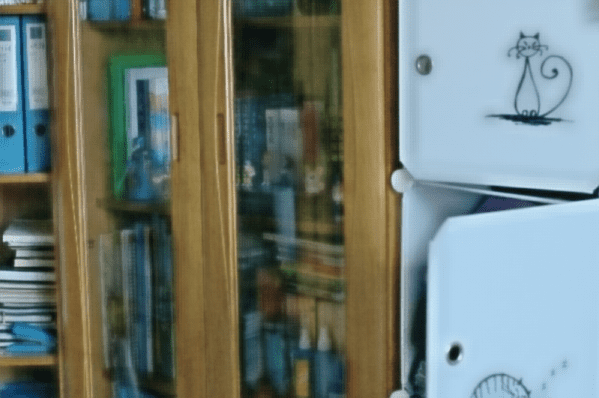}&
			\includegraphics[width=0.3\linewidth]{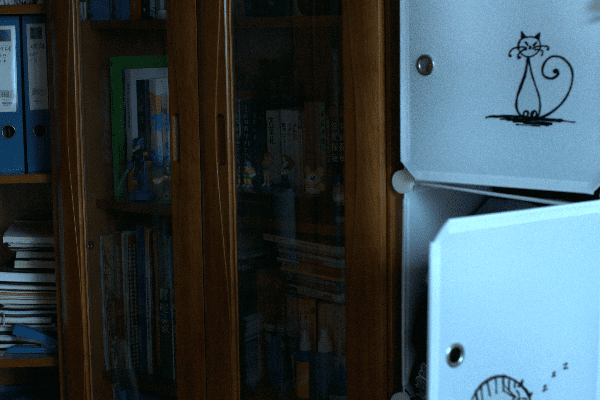}\\
			(a) input  & (b) LLNet \cite{LLNet}  &  (c) LightenNet \cite{LightenNet}\\
			\includegraphics[width=0.3\linewidth]{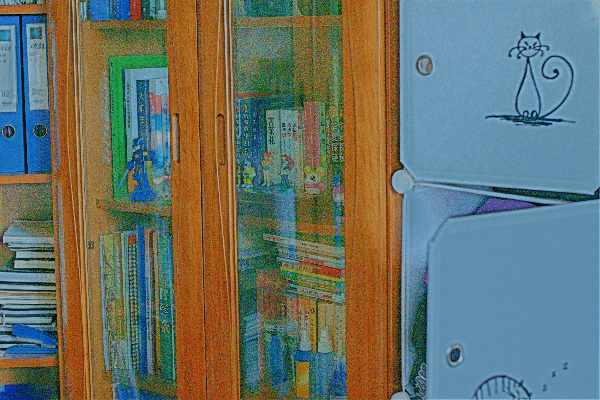}&
			\includegraphics[width=0.3\linewidth]{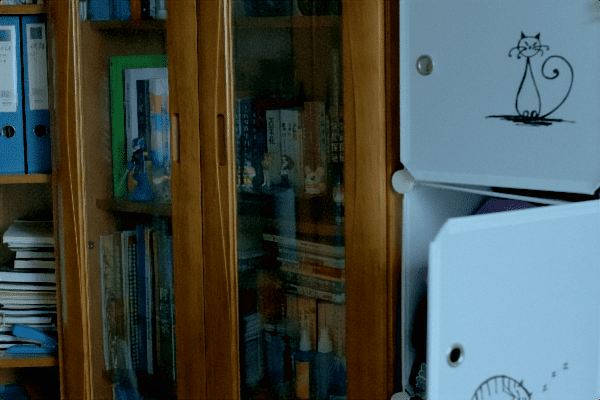}&
			\includegraphics[width=0.3\linewidth]{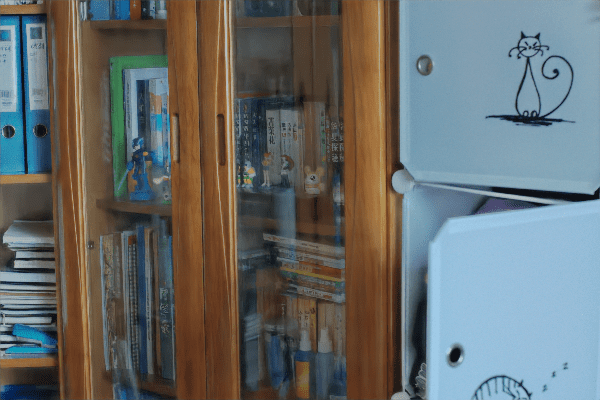}\\
			 (d) Retinex-Net \cite{ChenBMVC18} & (e) MBLLEN \cite{LvBMVC2018} & (f) KinD \cite{ZhangACM19}\\
			\includegraphics[width=0.3\linewidth]{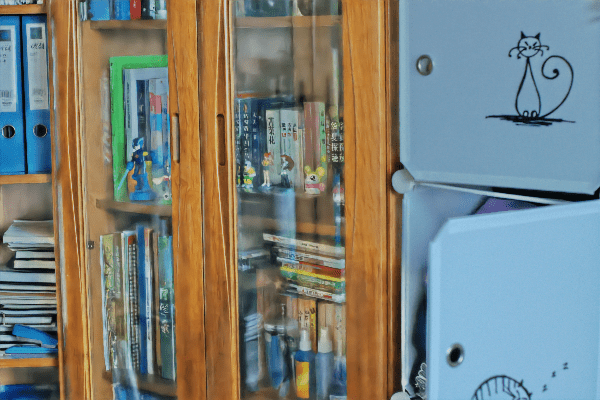}&
			\includegraphics[width=0.3\linewidth]{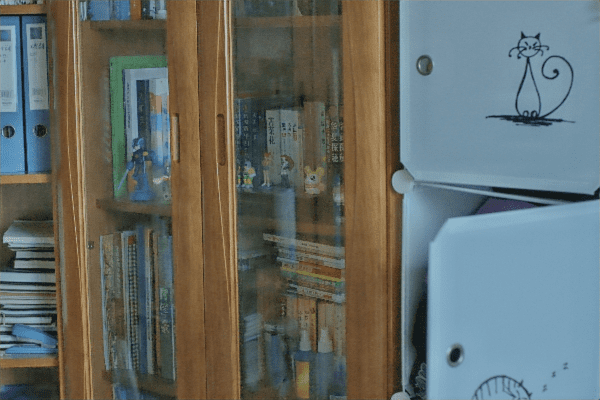}&
			\includegraphics[width=0.3\linewidth]{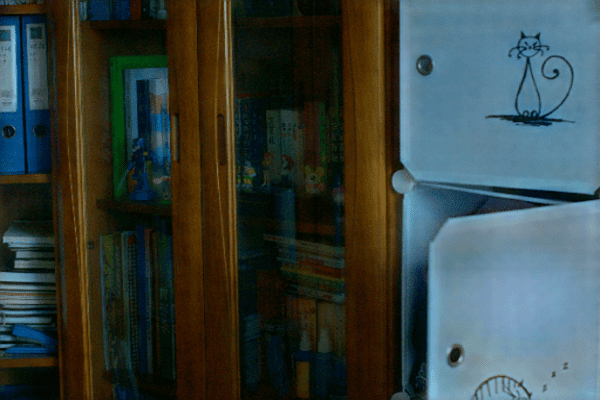}\\
			(g) KinD++ \cite{GuoIJCV2020} & (h) TBEFN \cite{TBEFN} &  (i)  	DSLR \cite{DSLR}\\
			\includegraphics[width=0.3\linewidth]{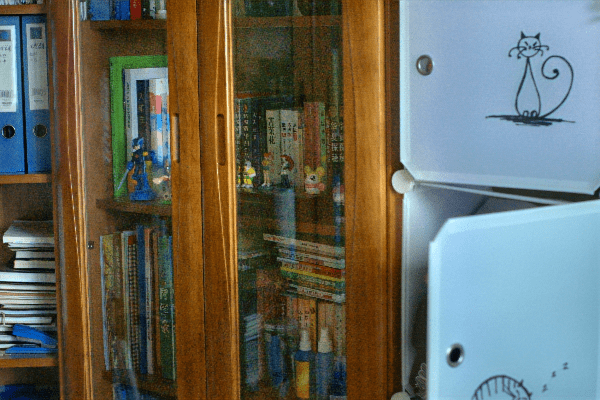}&
			\includegraphics[width=0.3\linewidth]{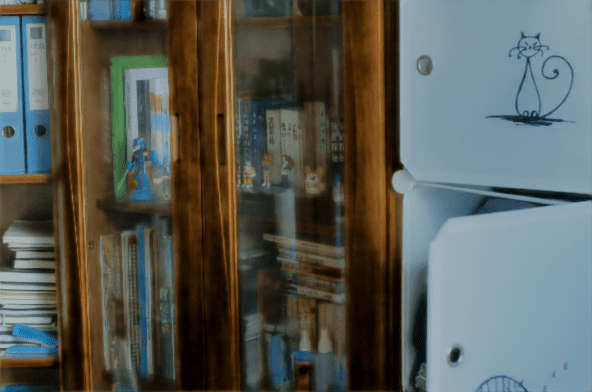}&
			\includegraphics[width=0.3\linewidth]{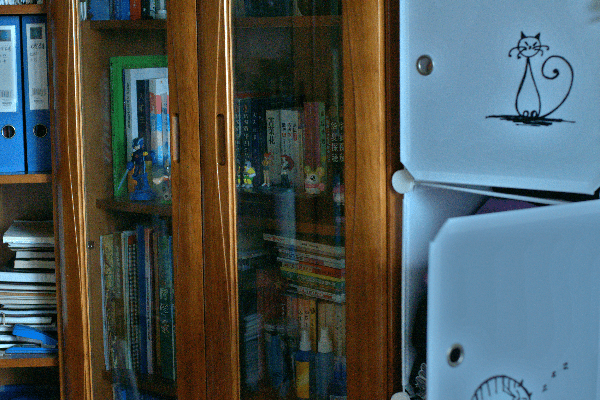}\\
			 (j) EnlightenGAN \cite{EnlightenGAN} & (k) DRBN \cite{YangCVRP20} & (l) ExCNet \cite{ZhangACM191}\\
			\includegraphics[width=0.3\linewidth]{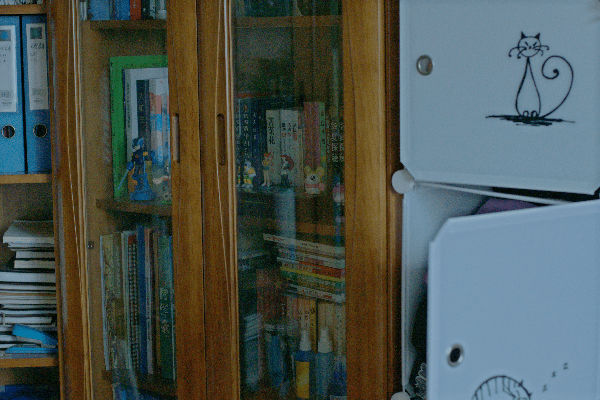}&
			\includegraphics[width=0.3\linewidth]{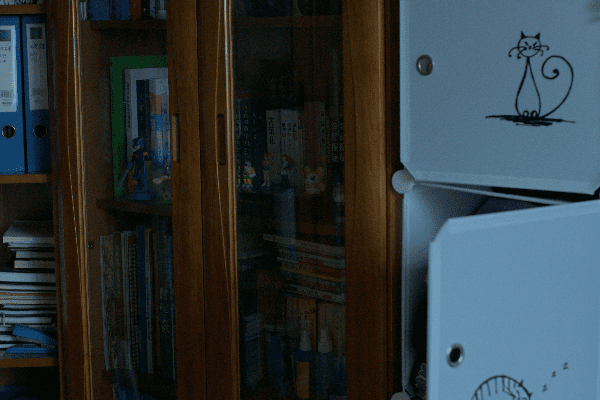}&
			\includegraphics[width=0.3\linewidth]{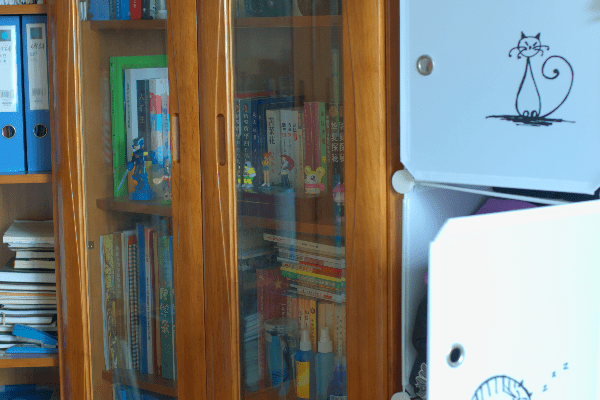}\\
			  (m) Zero-DCE \cite{ZeroDCE} &  (n) 	RRDNet \cite{RRDNet}  & (o) GT  \\
		\end{tabular}
	\end{center}
	\caption{Visual results of different methods on a low-light image sampled from LOL-test dataset \cite{ChenBMVC18}.}
	\label{fig:LOL1}
\end{figure*}

\begin{figure*} [h]
	\begin{center}
		\begin{tabular}{c@{ }c@{ }c@{ }}
			\includegraphics[width=0.3\linewidth]{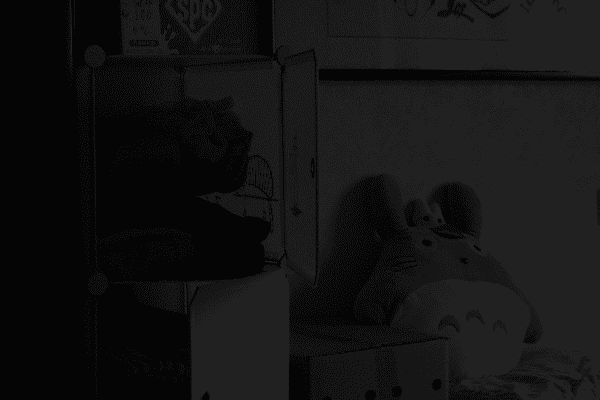}&
			\includegraphics[width=0.3\linewidth]{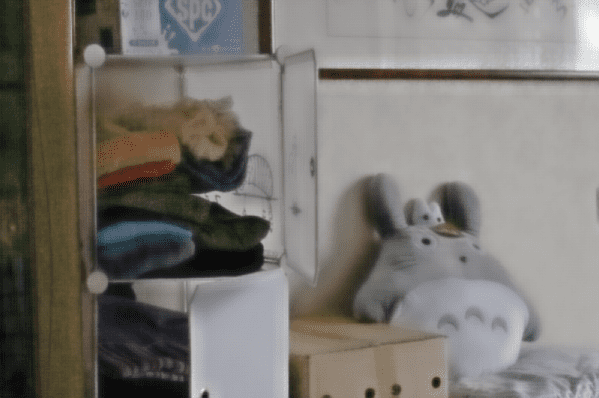}&
			\includegraphics[width=0.3\linewidth]{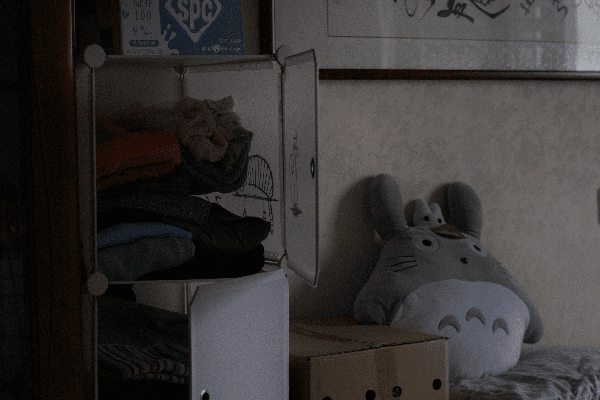}\\
			(a) input  & (b) LLNet \cite{LLNet}  &  (c) LightenNet \cite{LightenNet}\\
			\includegraphics[width=0.3\linewidth]{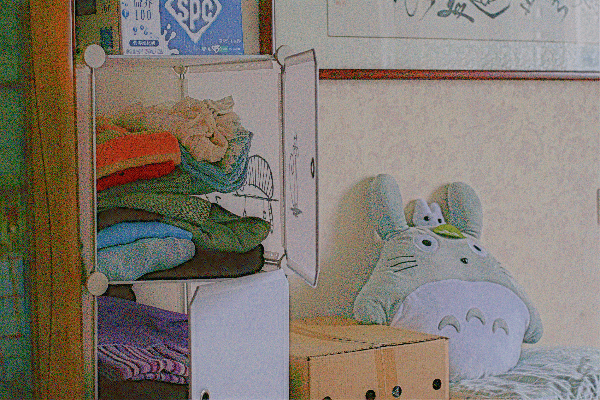}&
			\includegraphics[width=0.3\linewidth]{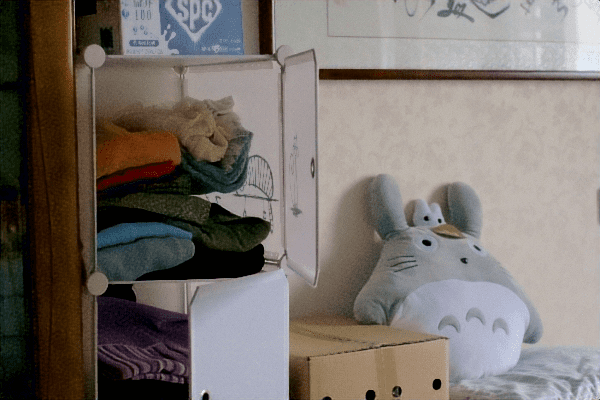}&
			\includegraphics[width=0.3\linewidth]{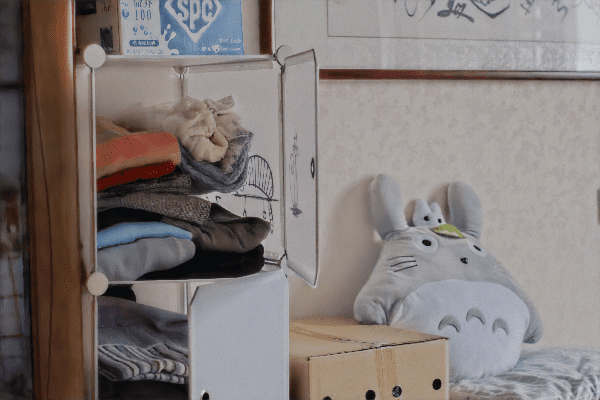}\\
			(d) Retinex-Net \cite{ChenBMVC18} & (e) MBLLEN \cite{LvBMVC2018} & (f) KinD \cite{ZhangACM19}\\
			\includegraphics[width=0.3\linewidth]{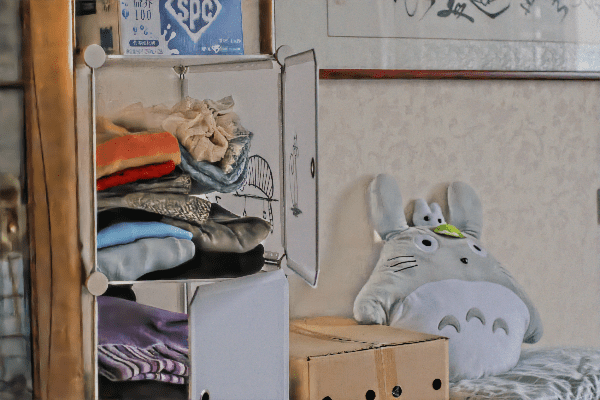}&
			\includegraphics[width=0.3\linewidth]{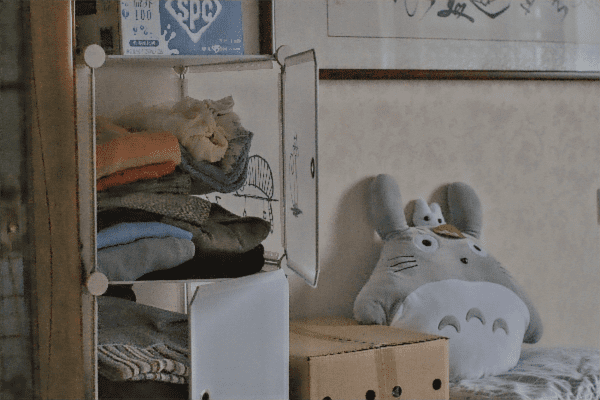}&
			\includegraphics[width=0.3\linewidth]{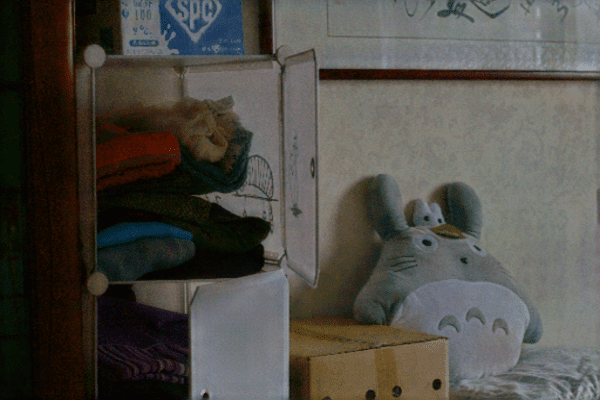}\\
			(g) KinD++ \cite{GuoIJCV2020} & (h) TBEFN \cite{TBEFN} &  (i)  	DSLR \cite{DSLR}\\
			\includegraphics[width=0.3\linewidth]{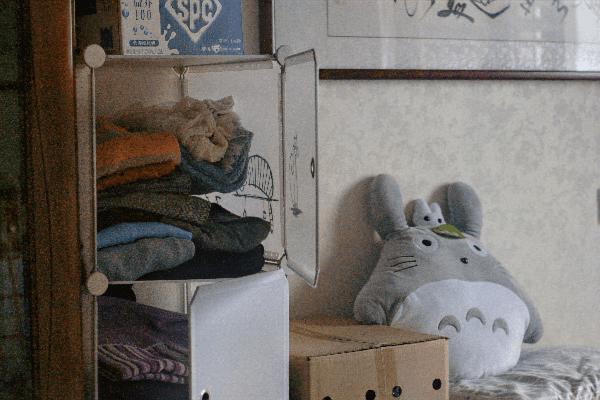}&
			\includegraphics[width=0.3\linewidth]{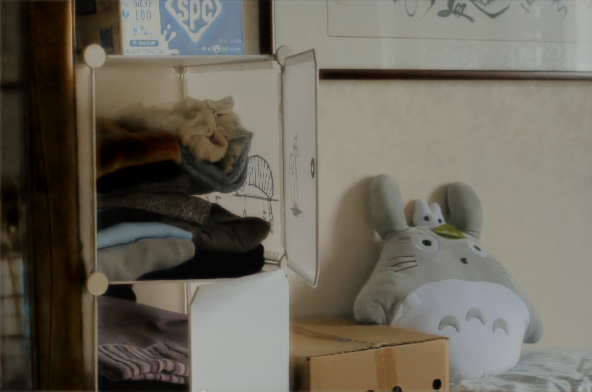}&
			\includegraphics[width=0.3\linewidth]{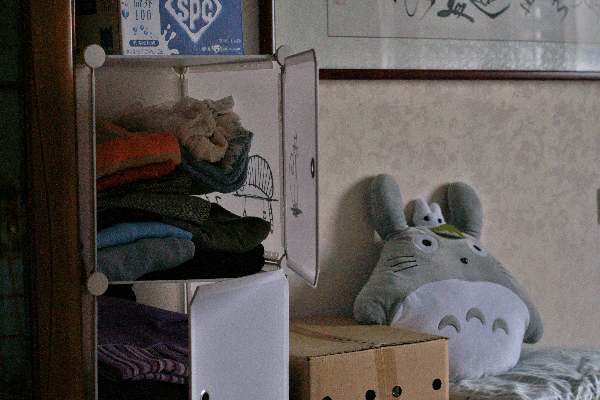}\\
			(j) EnlightenGAN \cite{EnlightenGAN} & (k) DRBN \cite{YangCVRP20} & (l) ExCNet \cite{ZhangACM191}\\
			\includegraphics[width=0.3\linewidth]{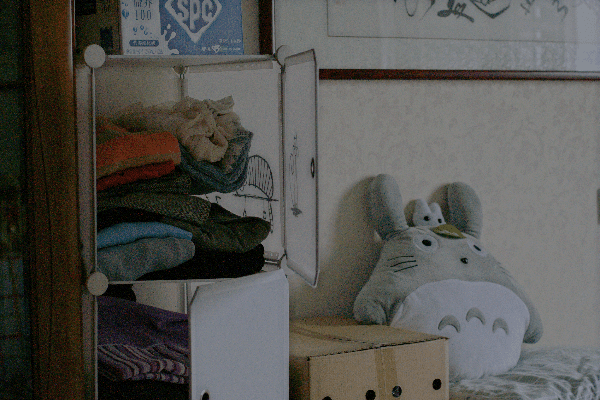}&
			\includegraphics[width=0.3\linewidth]{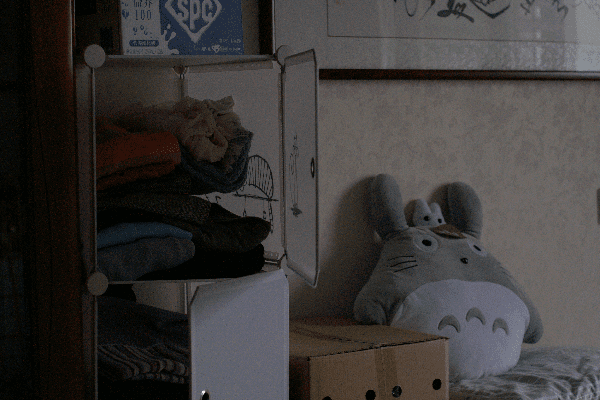}&
			\includegraphics[width=0.3\linewidth]{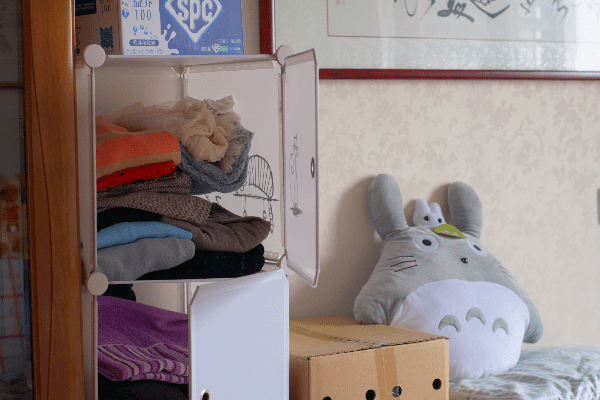}\\
			(m) Zero-DCE \cite{ZeroDCE} &  (n) 	RRDNet \cite{RRDNet}  & (o) GT  \\
		\end{tabular}
	\end{center}
	\caption{Visual results of different methods on a low-light image sampled from LOL-test dataset \cite{ChenBMVC18}.}
	\label{fig:LOL2}
\end{figure*}

\begin{figure*} [h]
	\begin{center}
		\begin{tabular}{c@{ }c@{ }c@{ }}
			\includegraphics[width=0.3\linewidth]{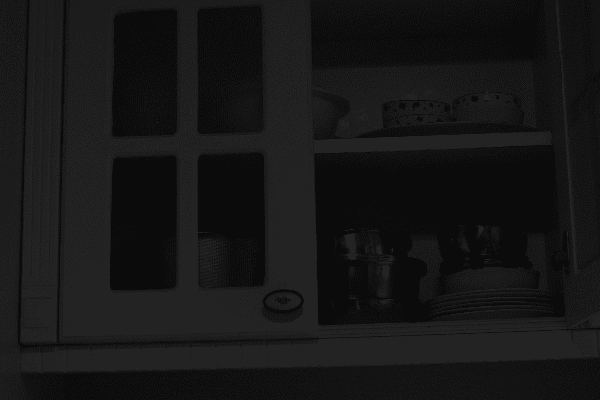}&
			\includegraphics[width=0.3\linewidth]{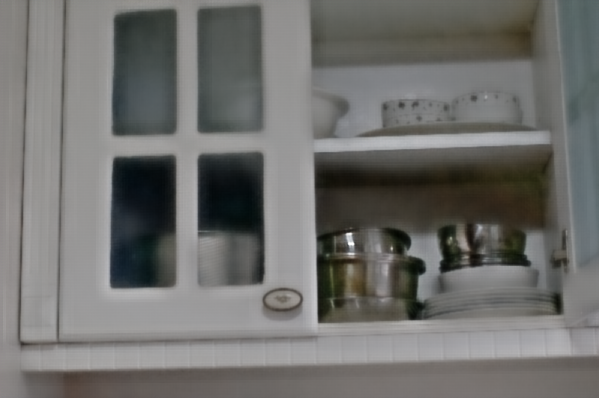}&
			\includegraphics[width=0.3\linewidth]{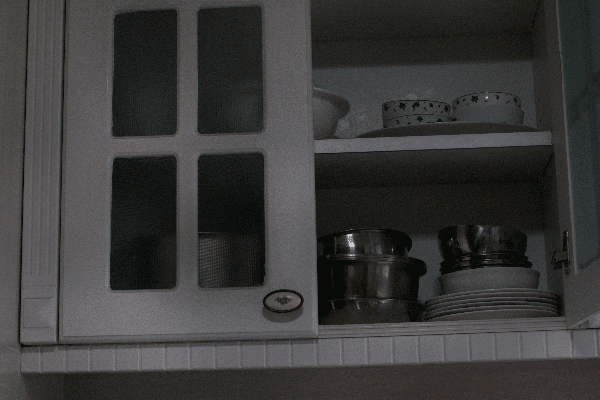}\\
			(a) input  & (b) LLNet \cite{LLNet}  &  (c) LightenNet \cite{LightenNet}\\
			\includegraphics[width=0.3\linewidth]{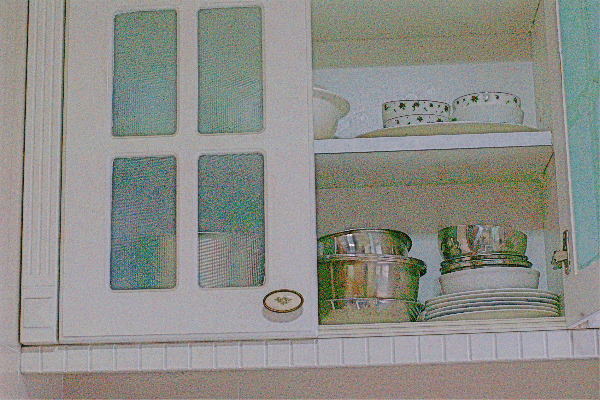}&
			\includegraphics[width=0.3\linewidth]{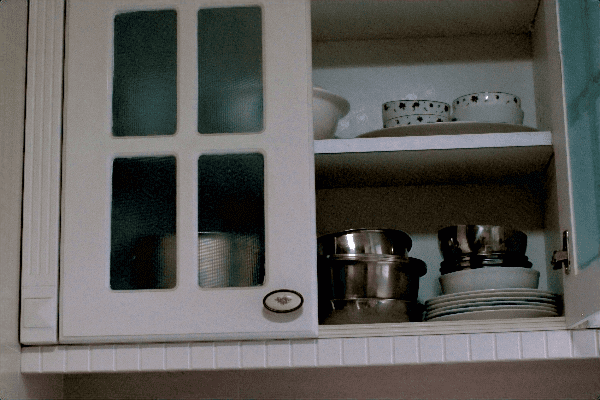}&
			\includegraphics[width=0.3\linewidth]{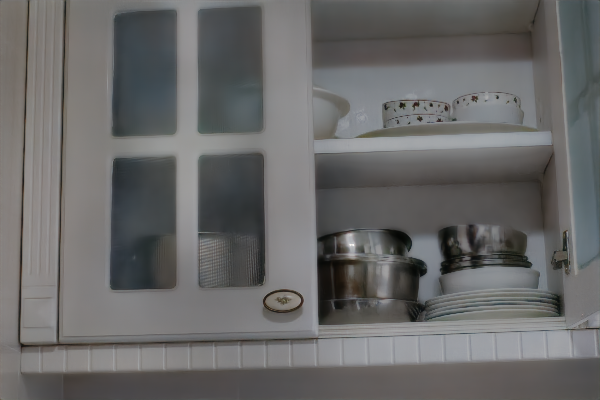}\\
			(d) Retinex-Net \cite{ChenBMVC18} & (e) MBLLEN \cite{LvBMVC2018} & (f) KinD \cite{ZhangACM19}\\
			\includegraphics[width=0.3\linewidth]{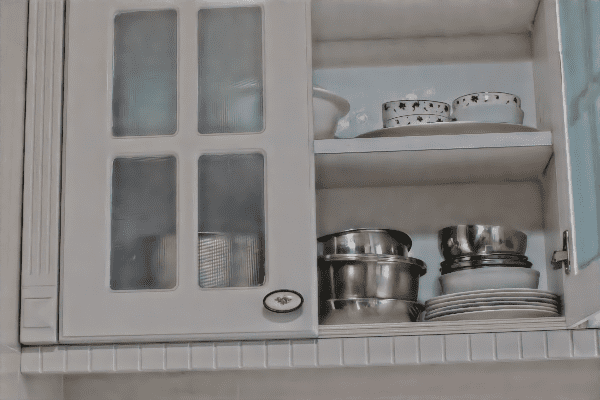}&
			\includegraphics[width=0.3\linewidth]{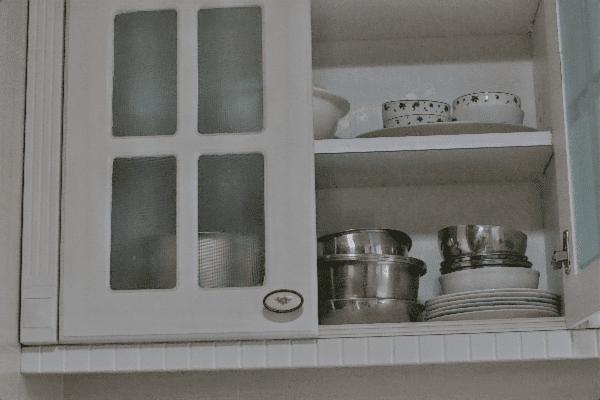}&
			\includegraphics[width=0.3\linewidth]{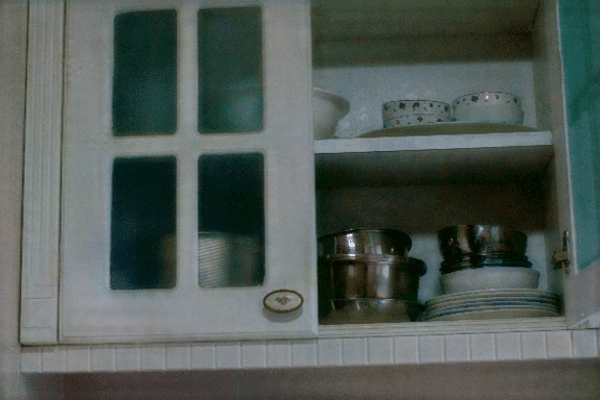}\\
			(g) KinD++ \cite{GuoIJCV2020} & (h) TBEFN \cite{TBEFN} &  (i)  	DSLR \cite{DSLR}\\
			\includegraphics[width=0.3\linewidth]{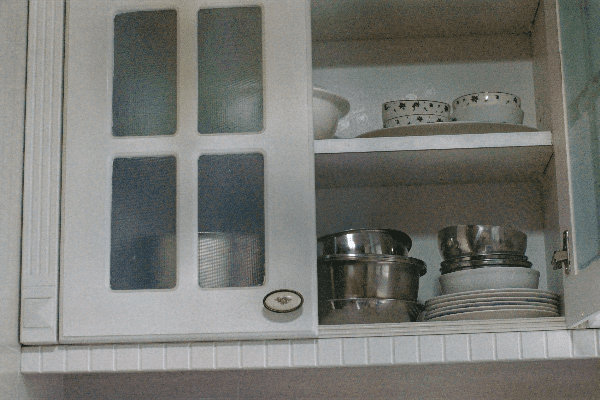}&
			\includegraphics[width=0.3\linewidth]{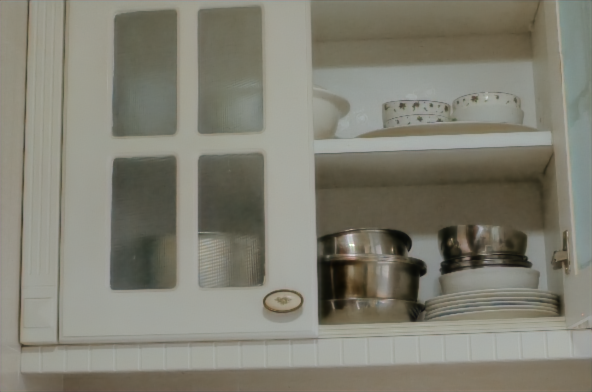}&
			\includegraphics[width=0.3\linewidth]{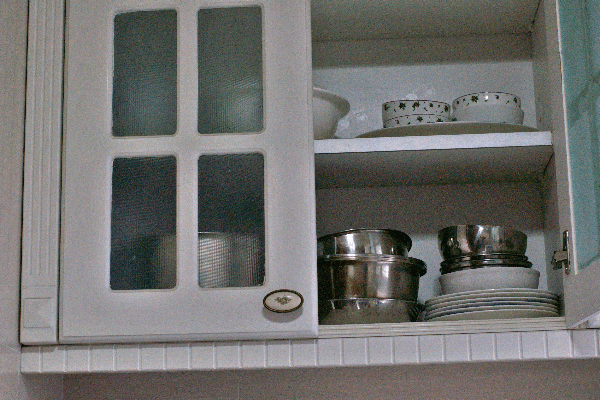}\\
			(j) EnlightenGAN \cite{EnlightenGAN} & (k) DRBN \cite{YangCVRP20} & (l) ExCNet \cite{ZhangACM191}\\
			\includegraphics[width=0.3\linewidth]{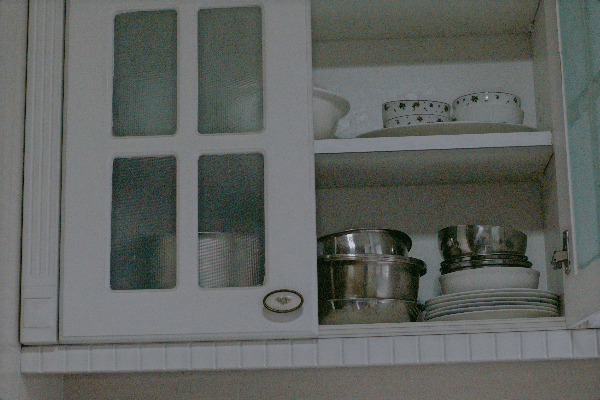}&
			\includegraphics[width=0.3\linewidth]{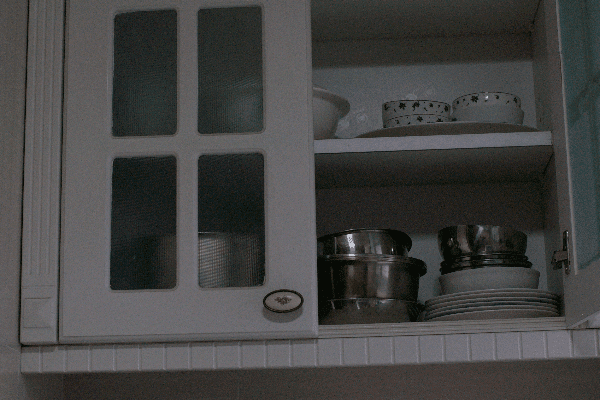}&
			\includegraphics[width=0.3\linewidth]{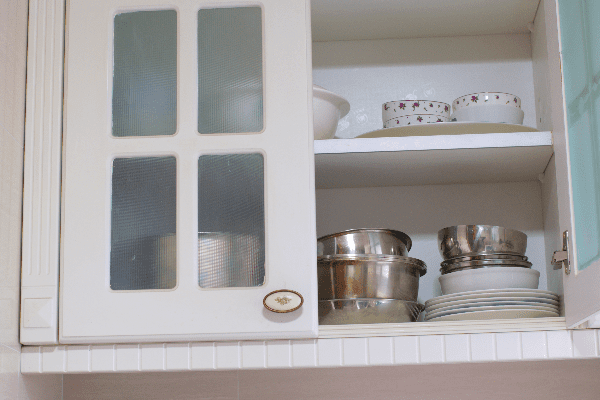}\\
			(m) Zero-DCE \cite{ZeroDCE} &  (n) 	RRDNet \cite{RRDNet}  & (o) GT  \\
		\end{tabular}
	\end{center}
	\caption{Visual results of different methods on a low-light image sampled from LOL-test dataset \cite{ChenBMVC18}.}
	\label{fig:LOL3}
\end{figure*}

\begin{figure*} [h]
	\begin{center}
		\begin{tabular}{c@{ }c@{ }c@{ }}
			\includegraphics[width=0.3\linewidth]{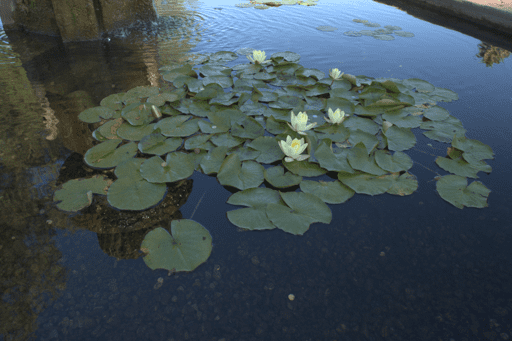}&
			\includegraphics[width=0.3\linewidth]{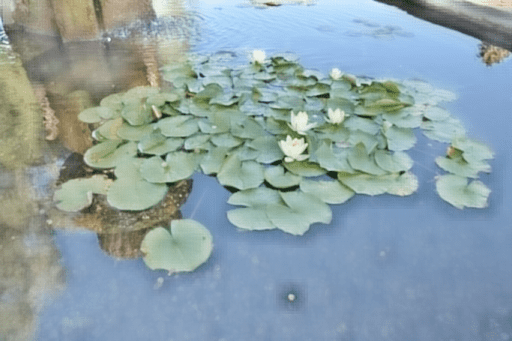}&
			\includegraphics[width=0.3\linewidth]{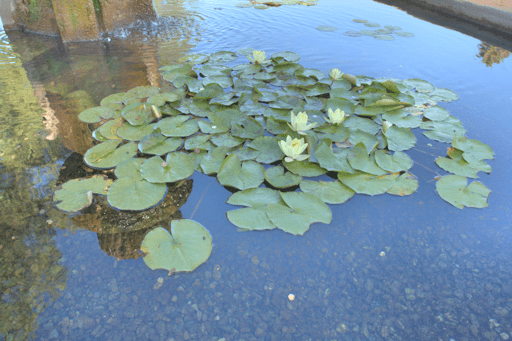}\\
			(a) input  & (b) LLNet \cite{LLNet}  &  (c) LightenNet \cite{LightenNet}\\
			\includegraphics[width=0.3\linewidth]{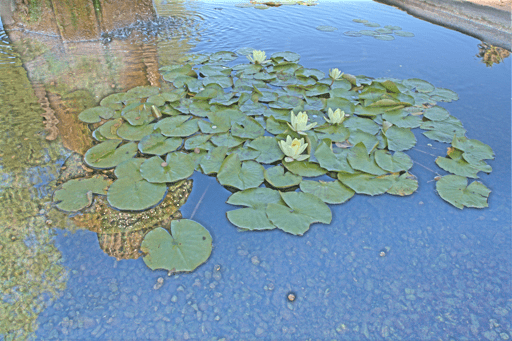}&
			\includegraphics[width=0.3\linewidth]{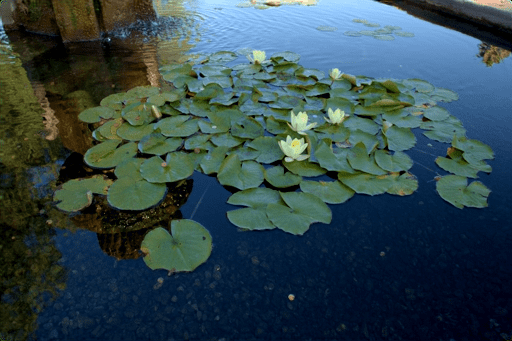}&
			\includegraphics[width=0.3\linewidth]{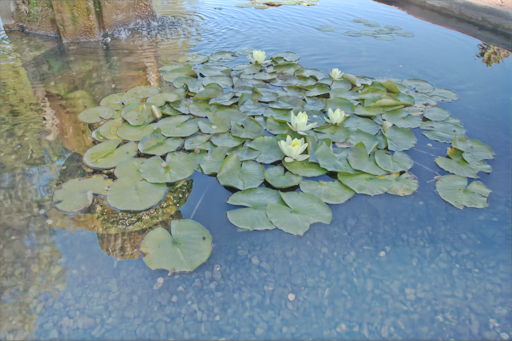}\\
			(d) Retinex-Net \cite{ChenBMVC18} & (e) MBLLEN \cite{LvBMVC2018} & (f) KinD \cite{ZhangACM19}\\
			\includegraphics[width=0.3\linewidth]{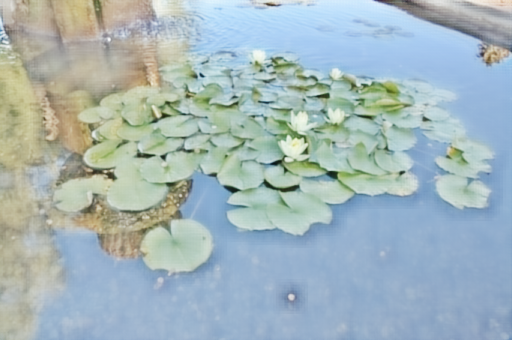}&
			\includegraphics[width=0.3\linewidth]{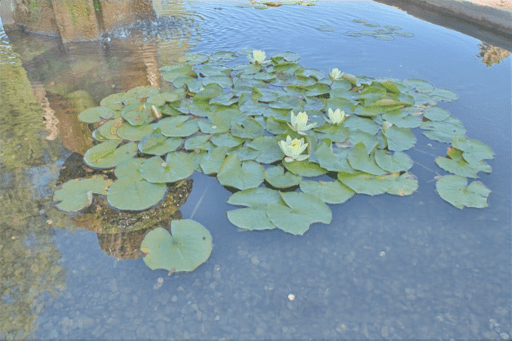}&
			\includegraphics[width=0.3\linewidth]{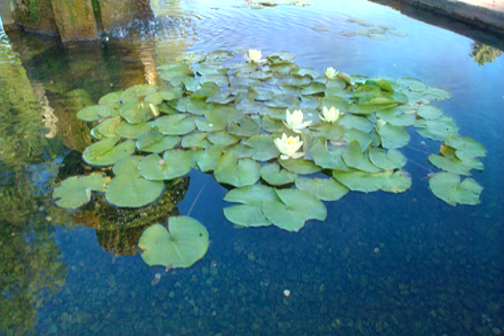}\\
			(g) KinD++ \cite{GuoIJCV2020} & (h) TBEFN \cite{TBEFN} &  (i)  	DSLR \cite{DSLR}\\
			\includegraphics[width=0.3\linewidth]{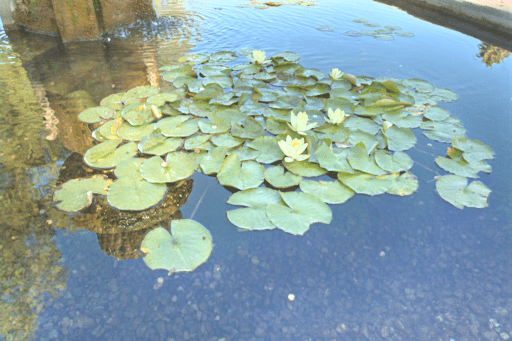}&
			\includegraphics[width=0.3\linewidth]{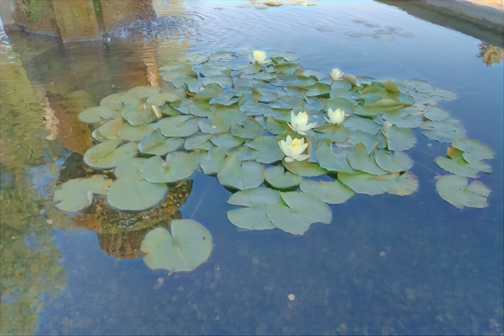}&
			\includegraphics[width=0.3\linewidth]{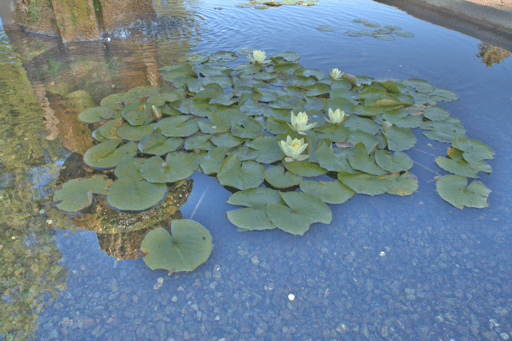}\\
			(j) EnlightenGAN \cite{EnlightenGAN} & (k) DRBN \cite{YangCVRP20} & (l) ExCNet \cite{ZhangACM191}\\
			\includegraphics[width=0.3\linewidth]{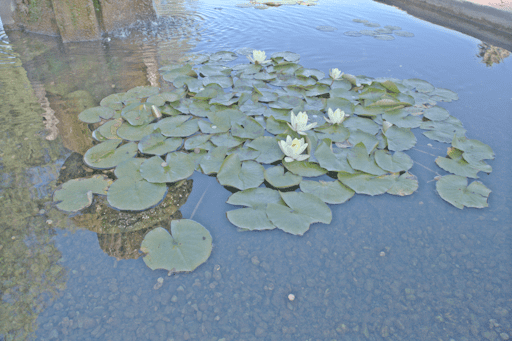}&
			\includegraphics[width=0.3\linewidth]{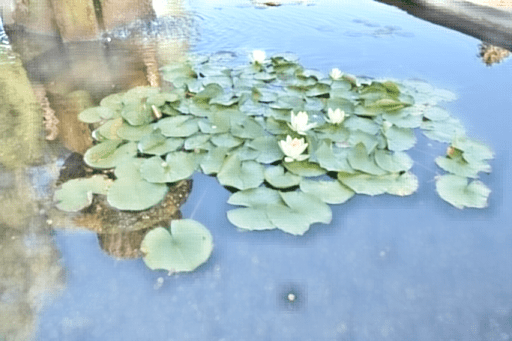}&
			\includegraphics[width=0.3\linewidth]{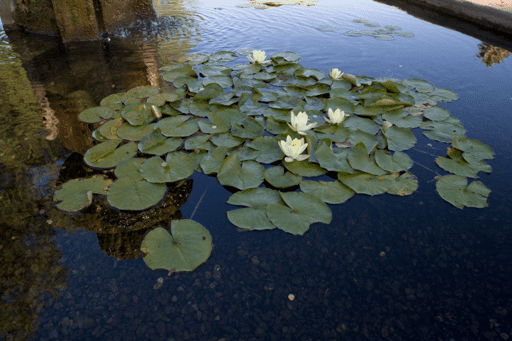}\\
			(m) Zero-DCE \cite{ZeroDCE} &  (n) 	RRDNet \cite{RRDNet}  & (o) GT  \\
		\end{tabular}
	\end{center}
	\caption{Visual results of different methods on a low-light image sampled from MIT-Adobe FiveK-test dataset \cite{Adobe5K}.}
	\label{fig:5K1}
\end{figure*}

\begin{figure*} [h]
	\begin{center}
		\begin{tabular}{c@{ }c@{ }c@{ }}
			\includegraphics[width=0.3\linewidth]{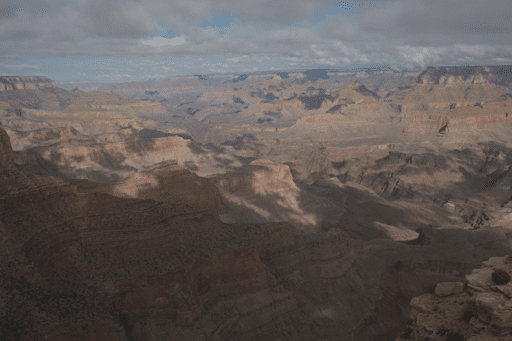}&
			\includegraphics[width=0.3\linewidth]{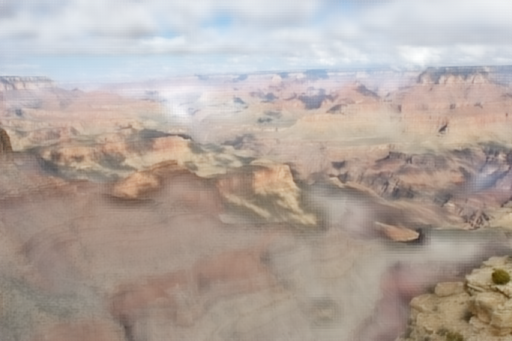}&
			\includegraphics[width=0.3\linewidth]{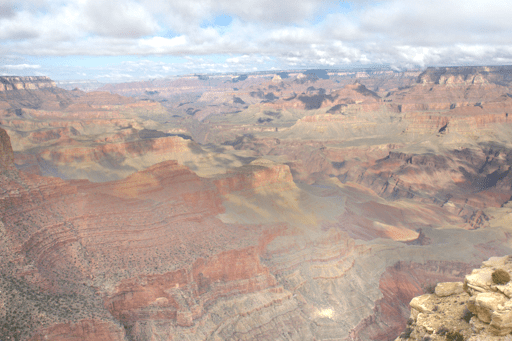}\\
			(a) input  & (b) LLNet \cite{LLNet}  &  (c) LightenNet \cite{LightenNet}\\
			\includegraphics[width=0.3\linewidth]{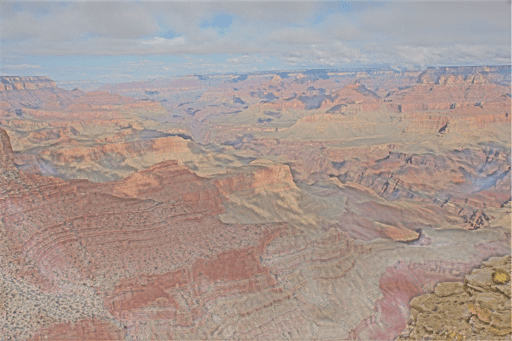}&
			\includegraphics[width=0.3\linewidth]{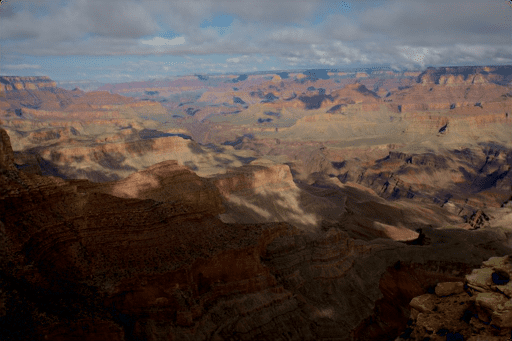}&
			\includegraphics[width=0.3\linewidth]{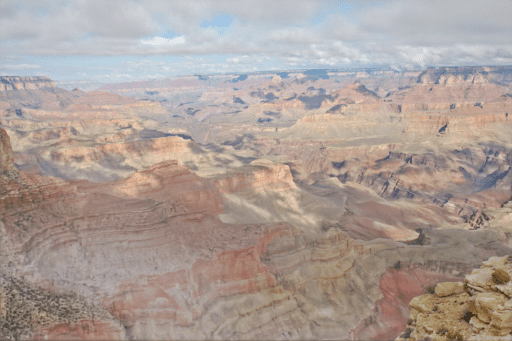}\\
			(d) Retinex-Net \cite{ChenBMVC18} & (e) MBLLEN \cite{LvBMVC2018} & (f) KinD \cite{ZhangACM19}\\
			\includegraphics[width=0.3\linewidth]{supple_figures/5K_KinD_a0018-kme_234.png}&
			\includegraphics[width=0.3\linewidth]{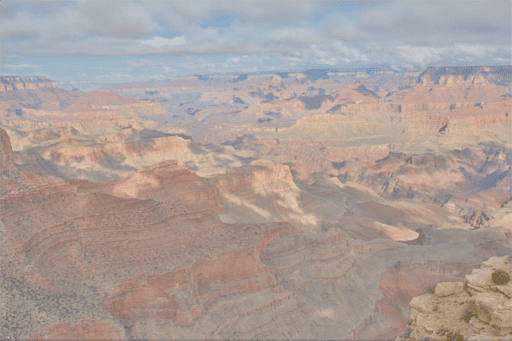}&
			\includegraphics[width=0.3\linewidth]{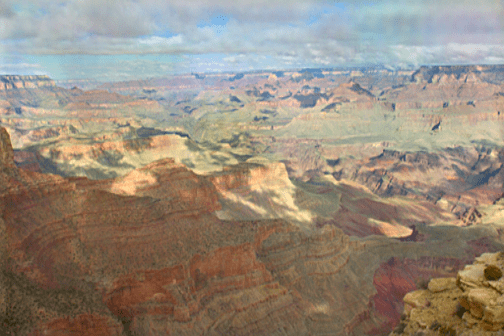}\\
			(g) KinD++ \cite{GuoIJCV2020} & (h) TBEFN \cite{TBEFN} &  (i)  	DSLR \cite{DSLR}\\
			\includegraphics[width=0.3\linewidth]{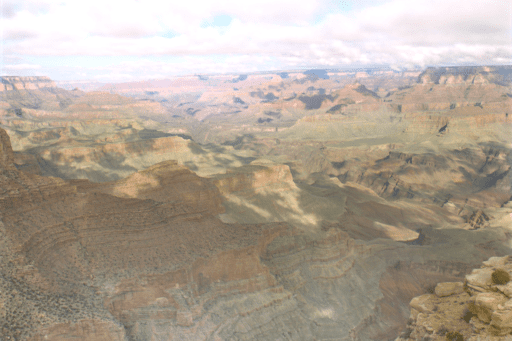}&
			\includegraphics[width=0.3\linewidth]{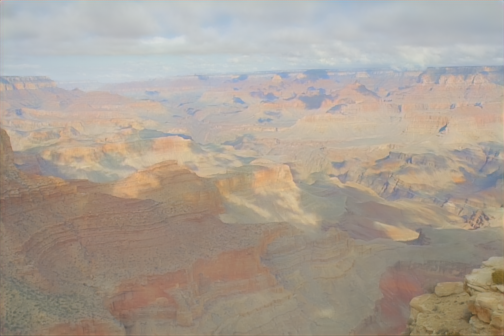}&
			\includegraphics[width=0.3\linewidth]{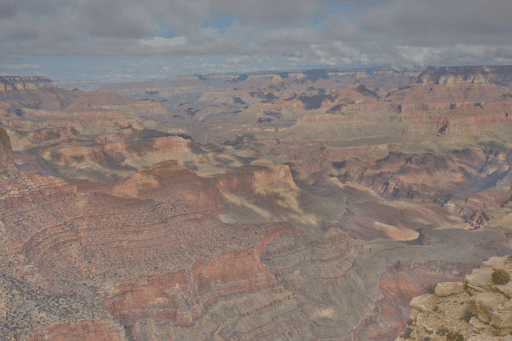}\\
			(j) EnlightenGAN \cite{EnlightenGAN} & (k) DRBN \cite{YangCVRP20} & (l) ExCNet \cite{ZhangACM191}\\
			\includegraphics[width=0.3\linewidth]{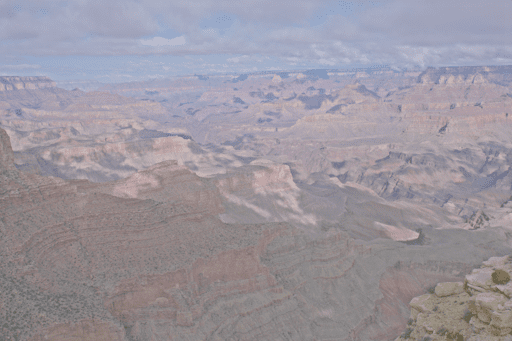}&
			\includegraphics[width=0.3\linewidth]{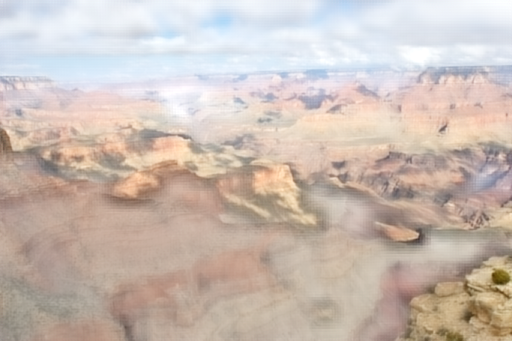}&
			\includegraphics[width=0.3\linewidth]{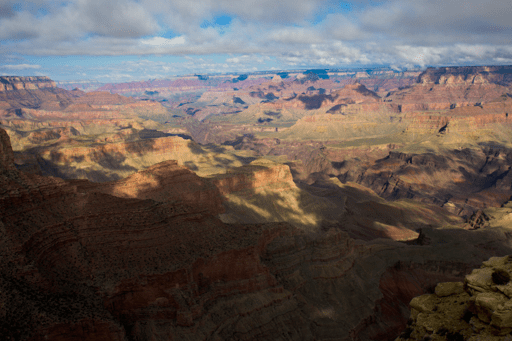}\\
			(m) Zero-DCE \cite{ZeroDCE} &  (n) 	RRDNet \cite{RRDNet}  & (o) GT  \\
		\end{tabular}
	\end{center}
	\caption{Visual results of different methods on a low-light image sampled from MIT-Adobe FiveK-test dataset \cite{Adobe5K}.}
	\label{fig:5K2}
\end{figure*}

\begin{figure*} [h]
	\begin{center}
		\begin{tabular}{c@{ }c@{ }c@{ }}
			\includegraphics[width=0.3\linewidth]{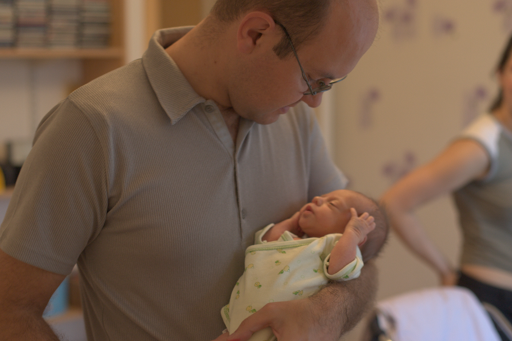}&
			\includegraphics[width=0.3\linewidth]{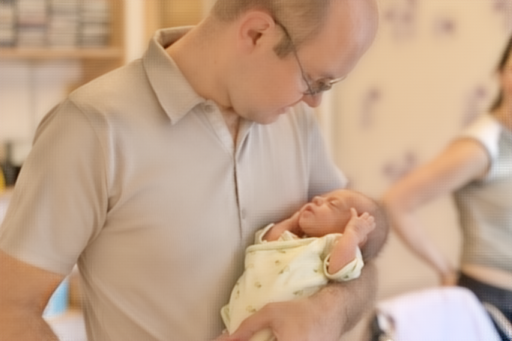}&
			\includegraphics[width=0.3\linewidth]{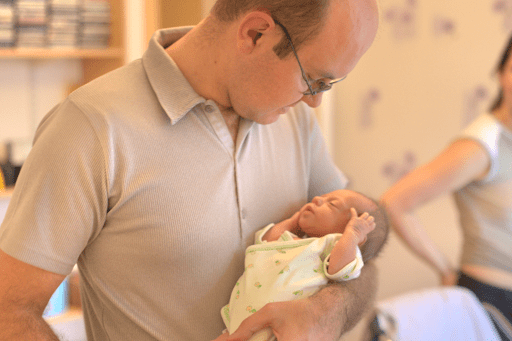}\\
			(a) input  & (b) LLNet \cite{LLNet}  &  (c) LightenNet \cite{LightenNet}\\
			\includegraphics[width=0.3\linewidth]{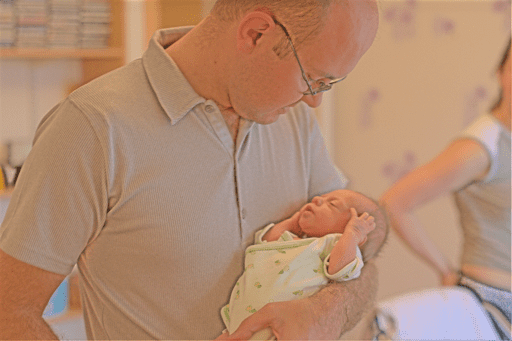}&
			\includegraphics[width=0.3\linewidth]{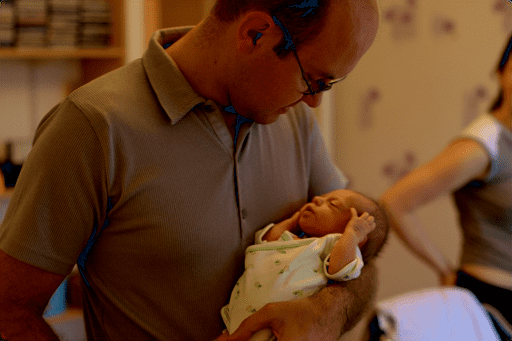}&
			\includegraphics[width=0.3\linewidth]{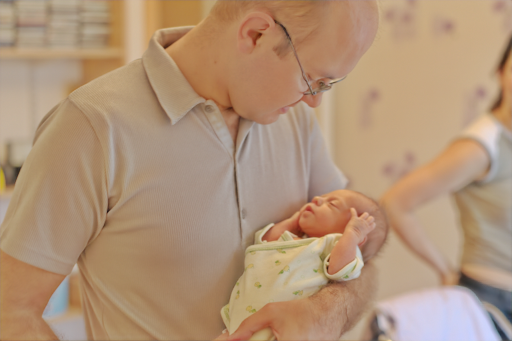}\\
			(d) Retinex-Net \cite{ChenBMVC18} & (e) MBLLEN \cite{LvBMVC2018} & (f) KinD \cite{ZhangACM19}\\
			\includegraphics[width=0.3\linewidth]{supple_figures/5K_KinD_a0055-050729_194412__I2E5282.png}&
			\includegraphics[width=0.3\linewidth]{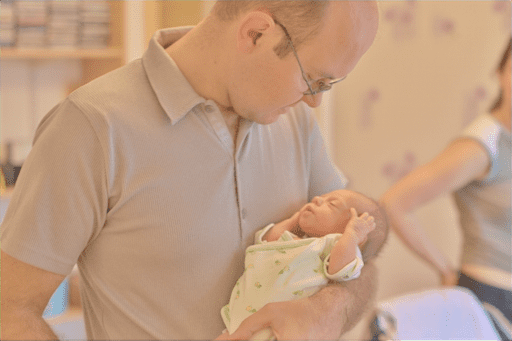}&
			\includegraphics[width=0.3\linewidth]{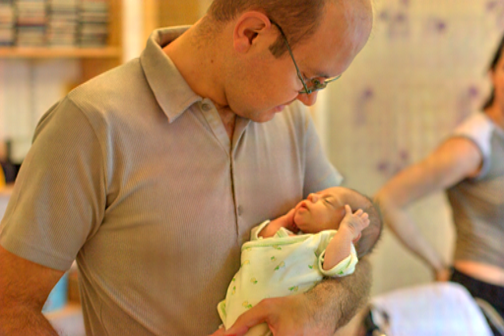}\\
			(g) KinD++ \cite{GuoIJCV2020} & (h) TBEFN \cite{TBEFN} &  (i)  	DSLR \cite{DSLR}\\
			\includegraphics[width=0.3\linewidth]{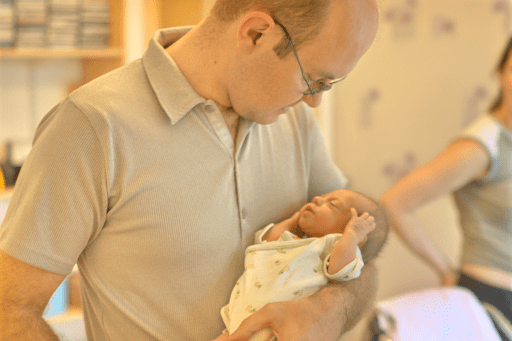}&
			\includegraphics[width=0.3\linewidth]{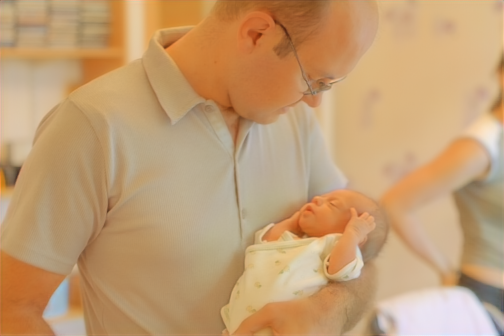}&
			\includegraphics[width=0.3\linewidth]{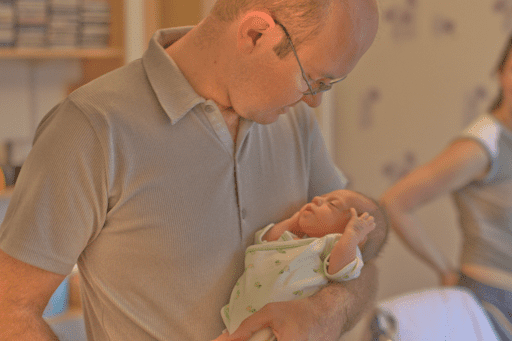}\\
			(j) EnlightenGAN \cite{EnlightenGAN} & (k) DRBN \cite{YangCVRP20} & (l) ExCNet \cite{ZhangACM191}\\
			\includegraphics[width=0.3\linewidth]{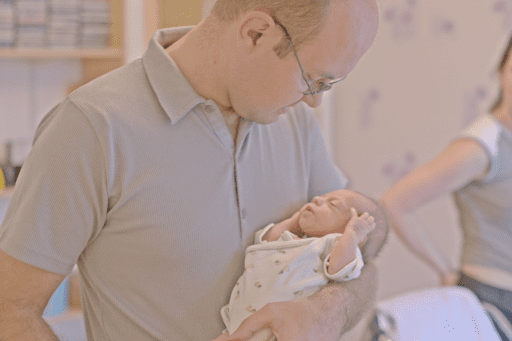}&
			\includegraphics[width=0.3\linewidth]{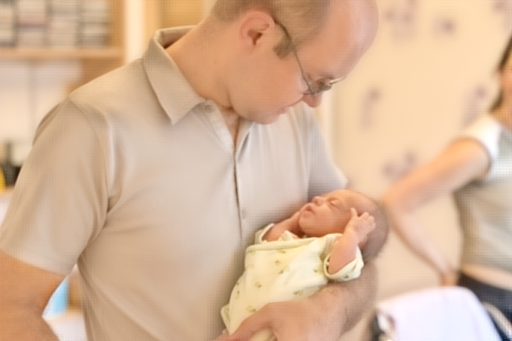}&
			\includegraphics[width=0.3\linewidth]{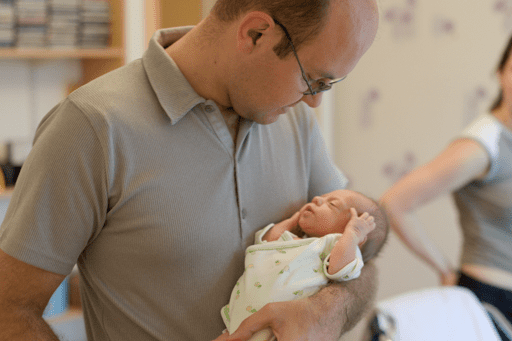}\\
			(m) Zero-DCE \cite{ZeroDCE} &  (n) 	RRDNet \cite{RRDNet}  & (o) GT  \\
		\end{tabular}
	\end{center}
	\caption{Visual results of different methods on a low-light image sampled from MIT-Adobe FiveK-test dataset \cite{Adobe5K}.}
	\label{fig:5K3}
\end{figure*}

\begin{figure*} [h]
	\begin{center}
		\begin{tabular}{c@{ }c@{ }c@{ }}
			\includegraphics[width=0.3\linewidth]{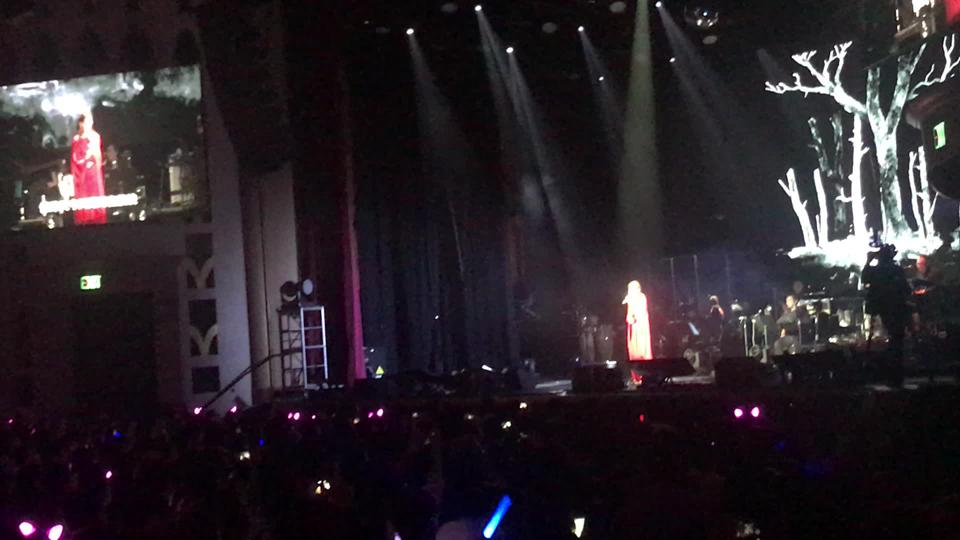}&
			\includegraphics[width=0.3\linewidth]{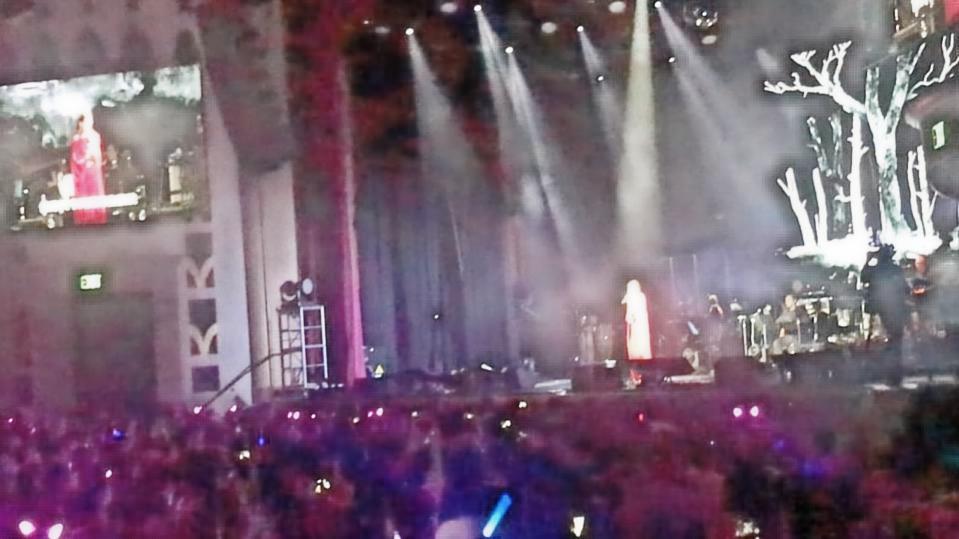}&
			\includegraphics[width=0.3\linewidth]{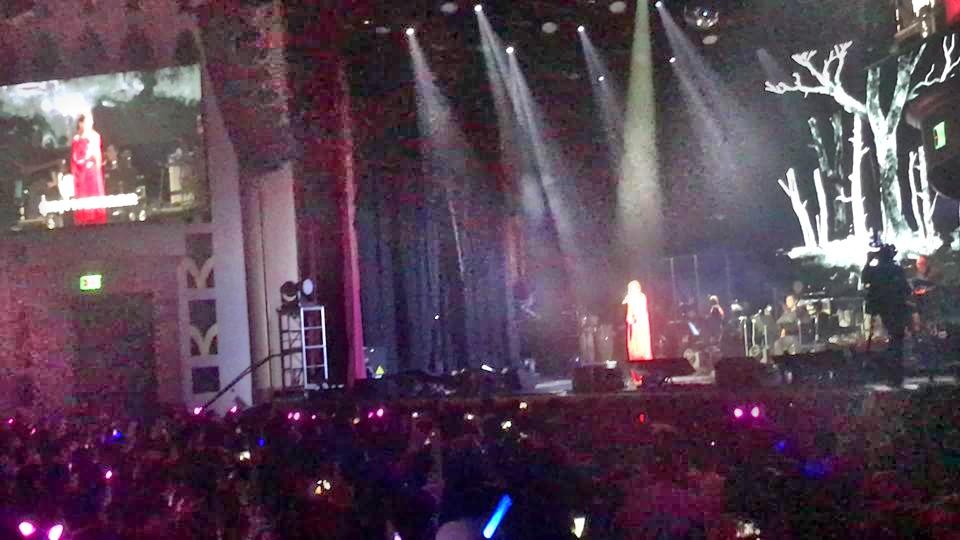}\\
			(a) input  & (b) LLNet \cite{LLNet}  &  (c) LightenNet \cite{LightenNet}\\
			\includegraphics[width=0.3\linewidth]{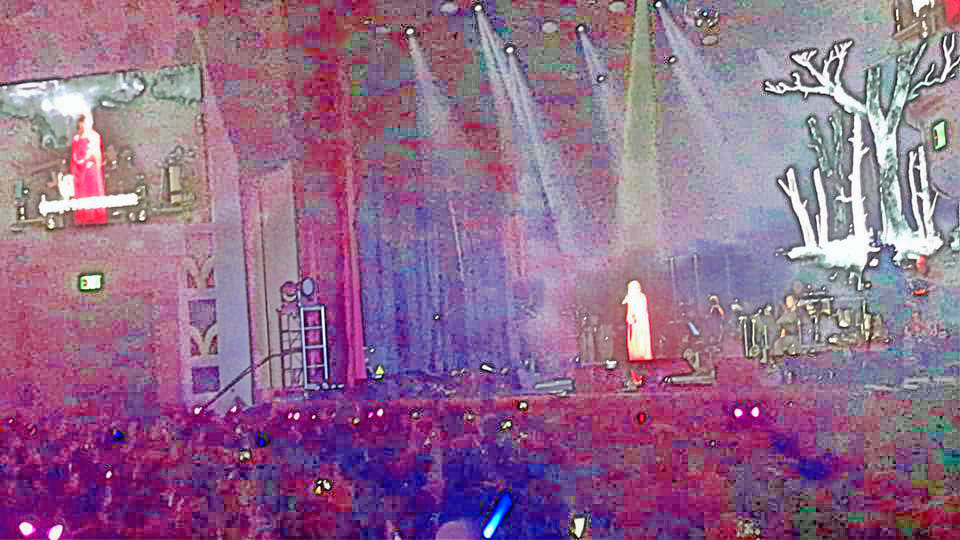}&
			\includegraphics[width=0.3\linewidth]{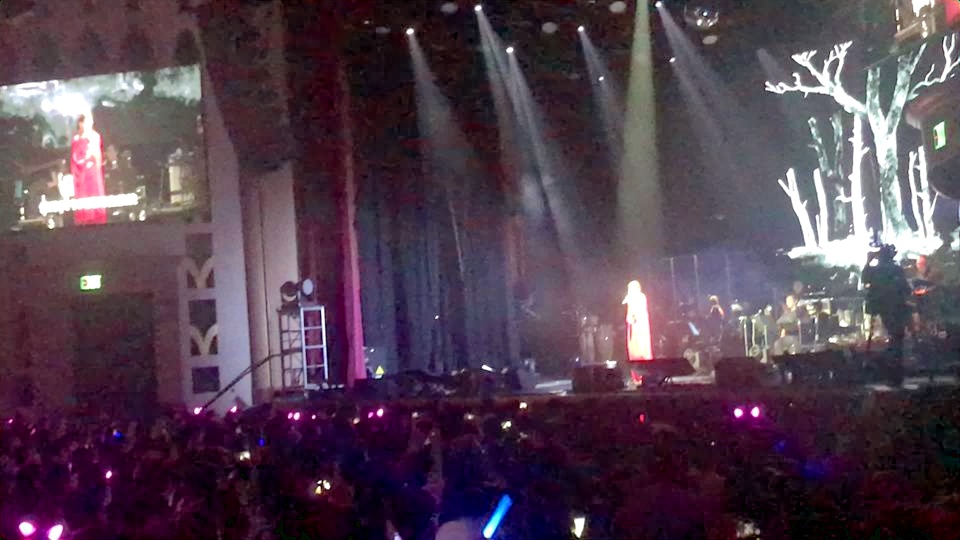}&
			\includegraphics[width=0.3\linewidth]{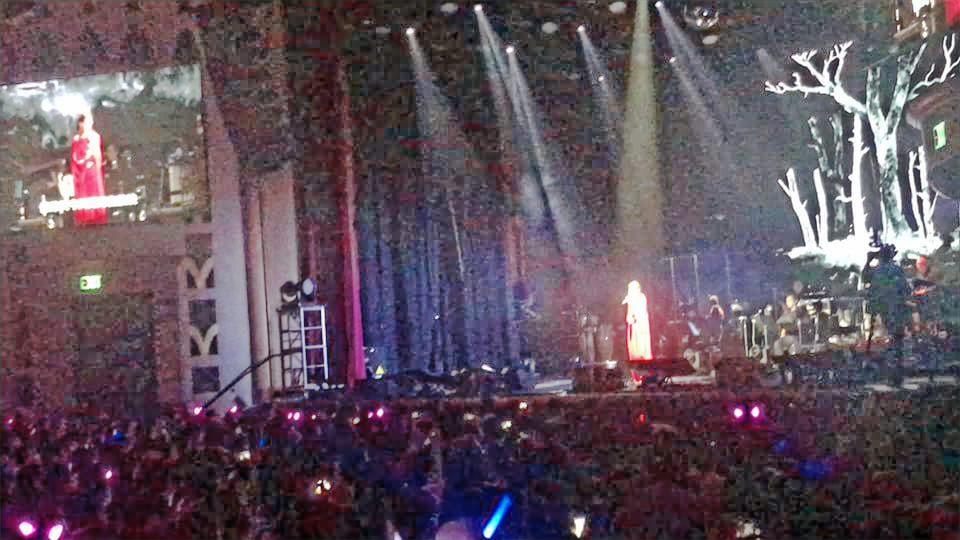}\\
			(d) Retinex-Net \cite{ChenBMVC18} & (e) MBLLEN \cite{LvBMVC2018} & (f) KinD \cite{ZhangACM19}\\
			\includegraphics[width=0.3\linewidth]{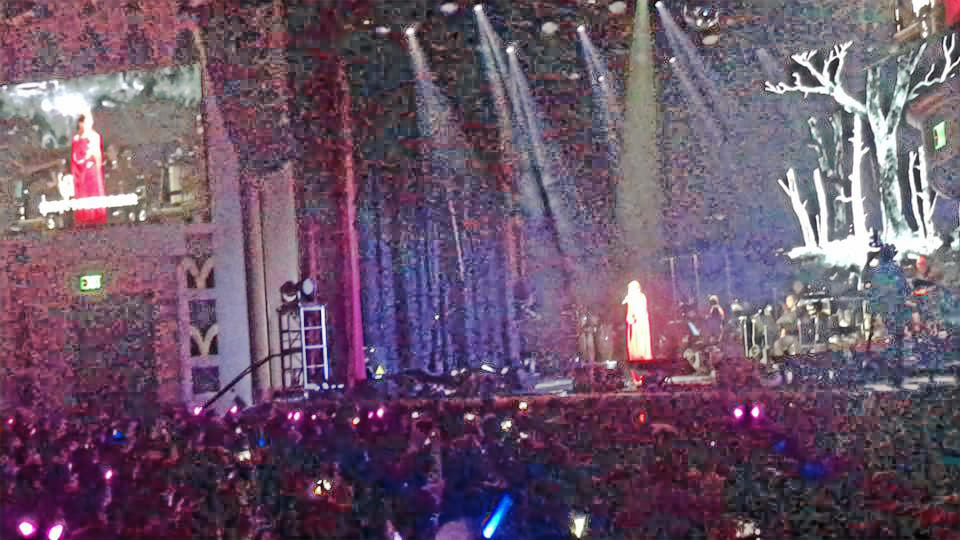}&
			\includegraphics[width=0.3\linewidth]{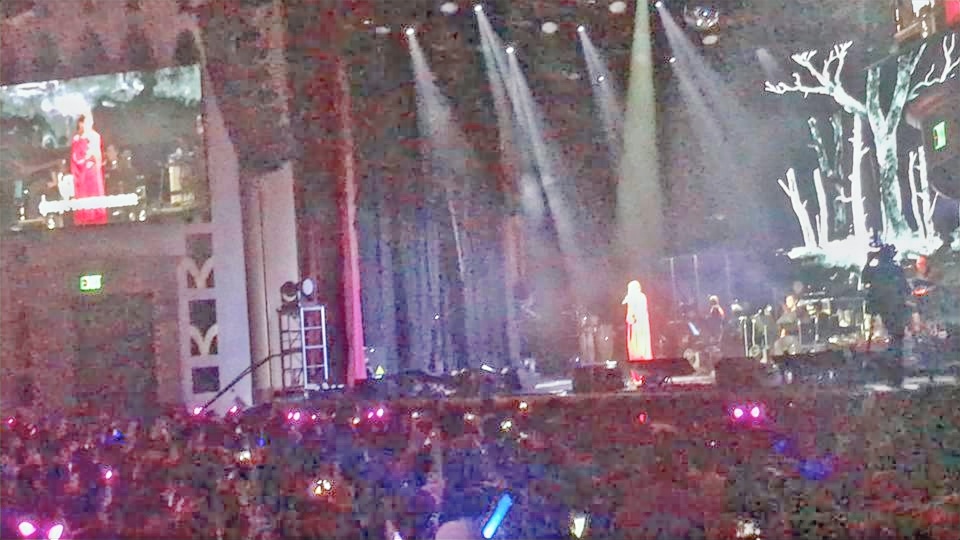}&
			\includegraphics[width=0.3\linewidth]{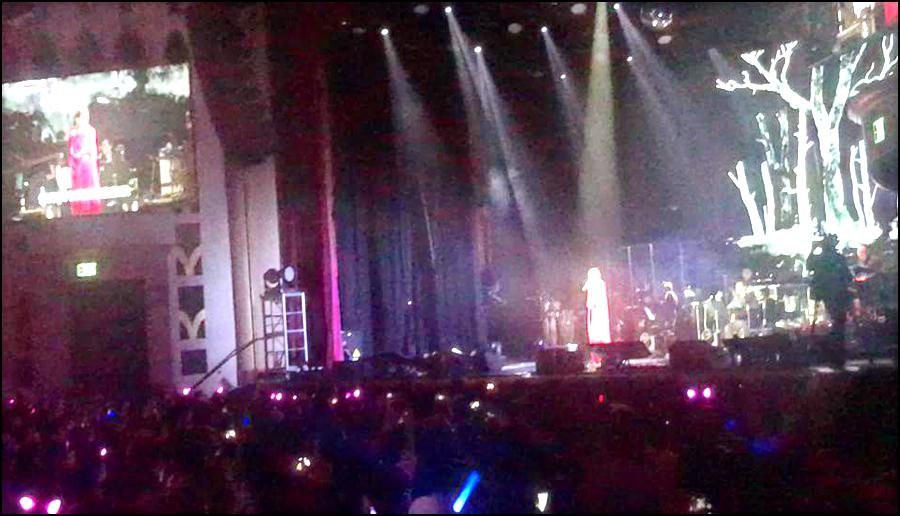}\\
			(g) KinD++ \cite{GuoIJCV2020} & (h) TBEFN \cite{TBEFN} &  (i)  	DSLR \cite{DSLR}\\
			\includegraphics[width=0.3\linewidth]{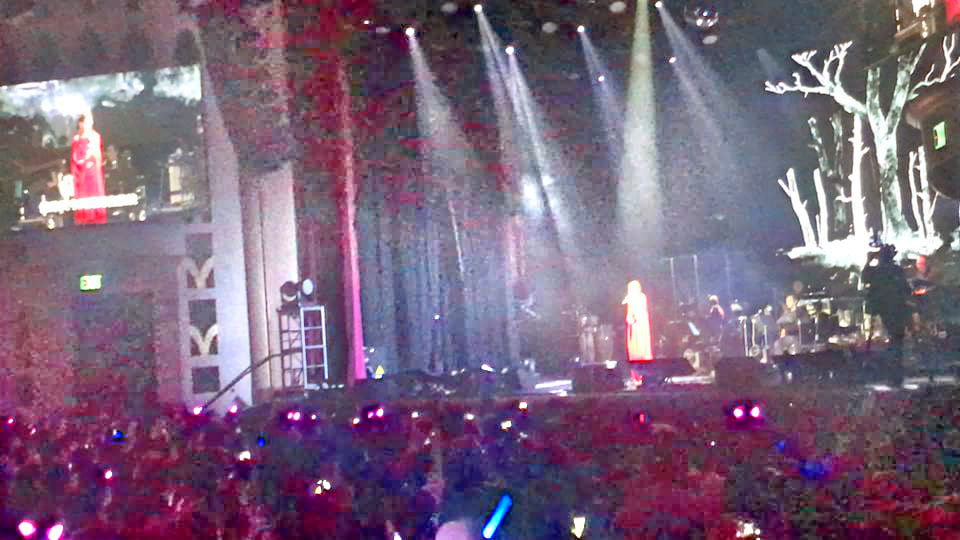}&
			\includegraphics[width=0.3\linewidth]{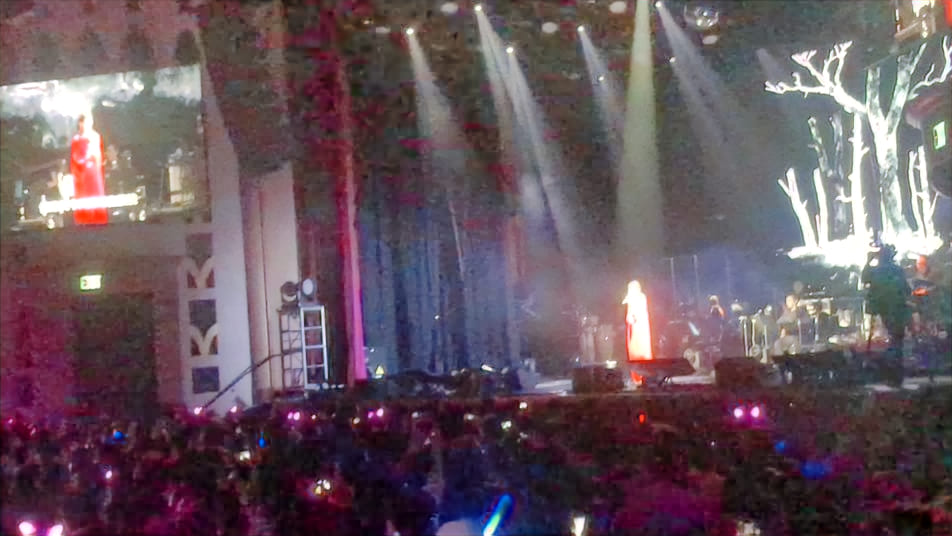}&
			\includegraphics[width=0.3\linewidth]{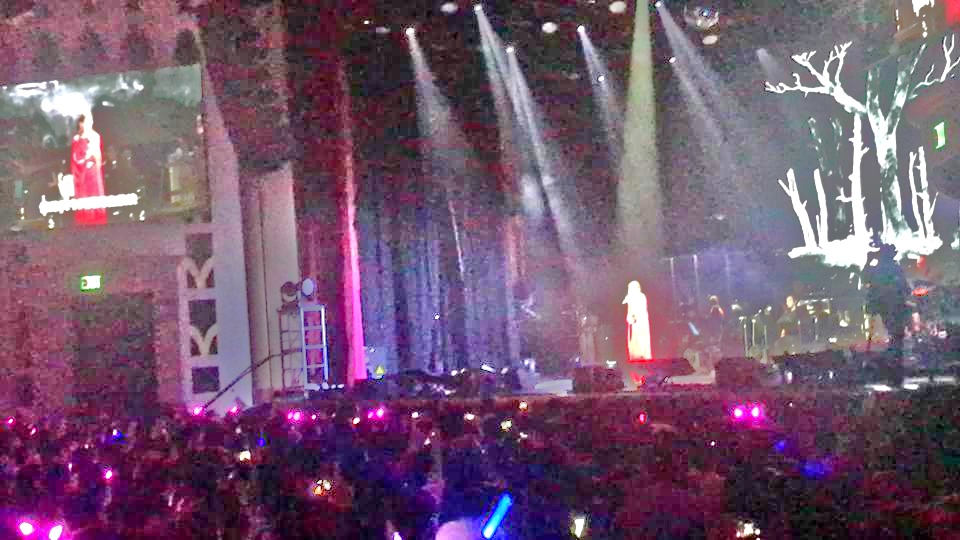}\\
			(j) EnlightenGAN \cite{EnlightenGAN} & (k) DRBN \cite{YangCVRP20} & (l) ExCNet \cite{ZhangACM191}\\
			\includegraphics[width=0.3\linewidth]{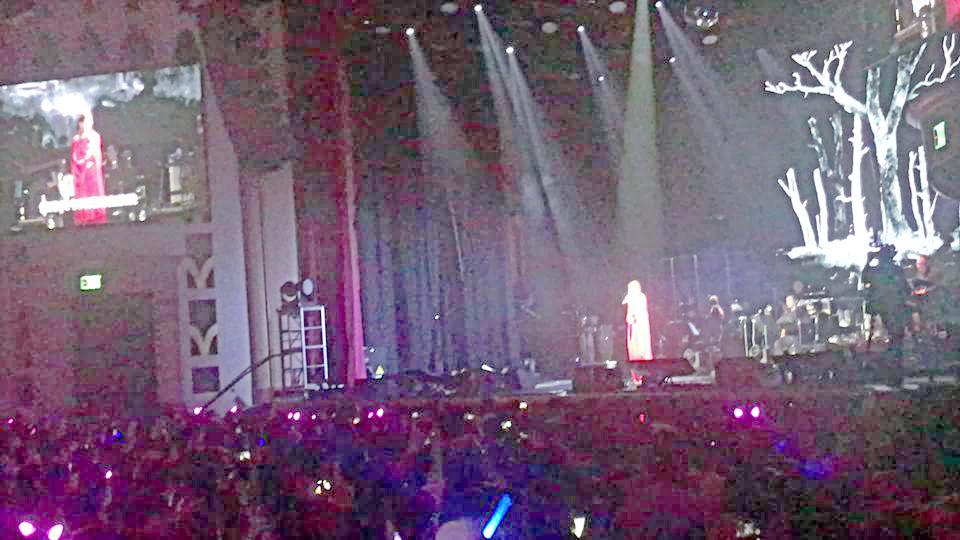}&
			\includegraphics[width=0.3\linewidth]{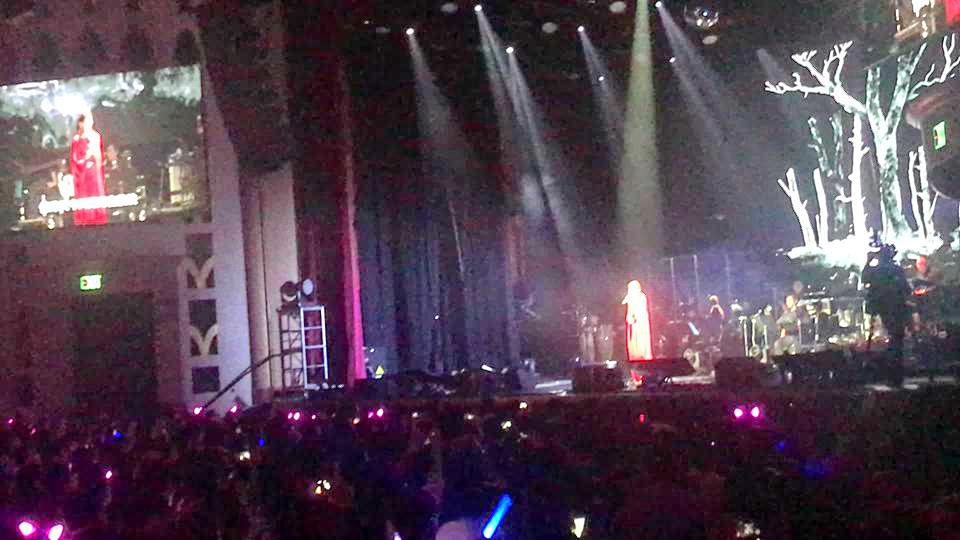}&
              ~\\
			(m) Zero-DCE \cite{ZeroDCE} &  (n) 	RRDNet \cite{RRDNet}  &  \\
		\end{tabular}
	\end{center}
	\caption{Visual results of different methods on a low-light image sampled from LLIV-Phone-imgT dataset.}
	\label{fig:Phone1}
\end{figure*}

\begin{figure*} [h]
	\begin{center}
		\begin{tabular}{c@{ }c@{ }c@{ }}
			\includegraphics[width=0.3\linewidth]{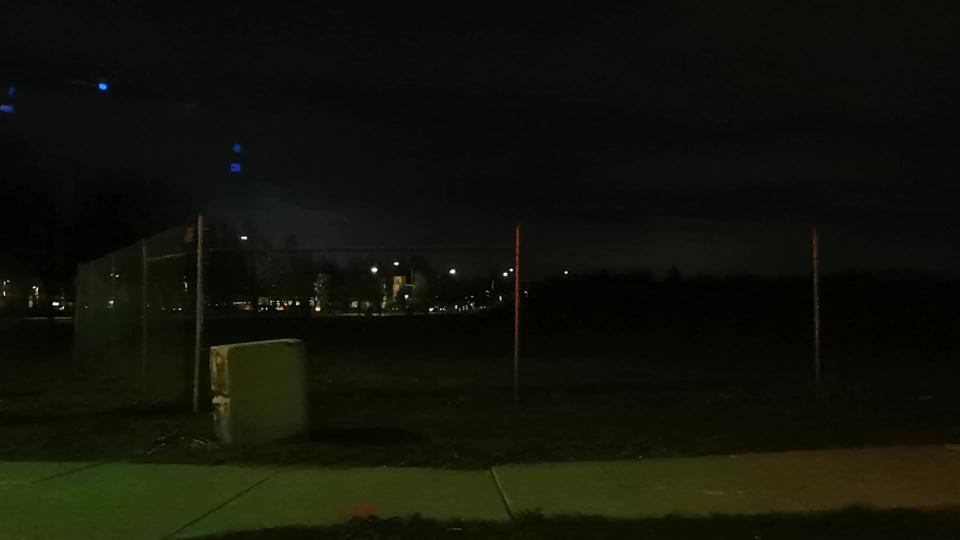}&
			\includegraphics[width=0.3\linewidth]{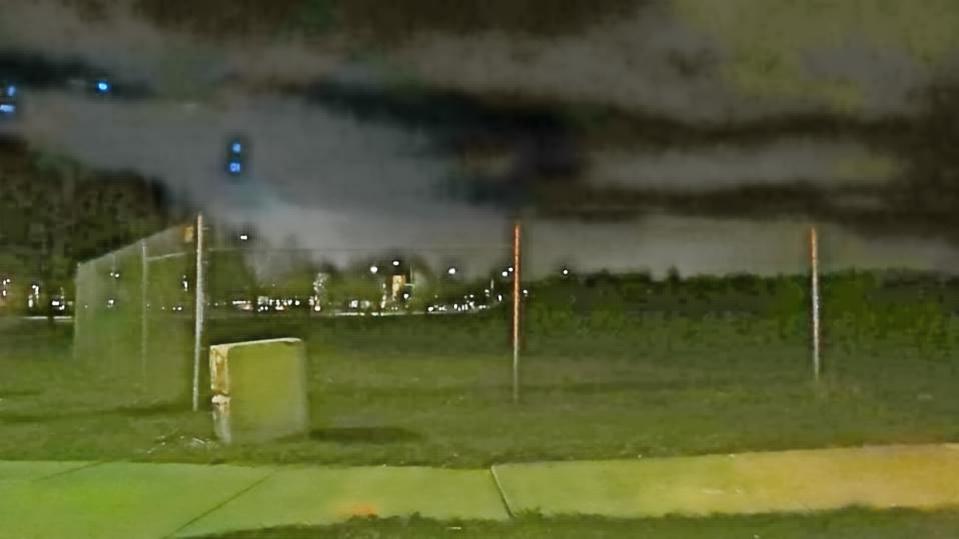}&
			\includegraphics[width=0.3\linewidth]{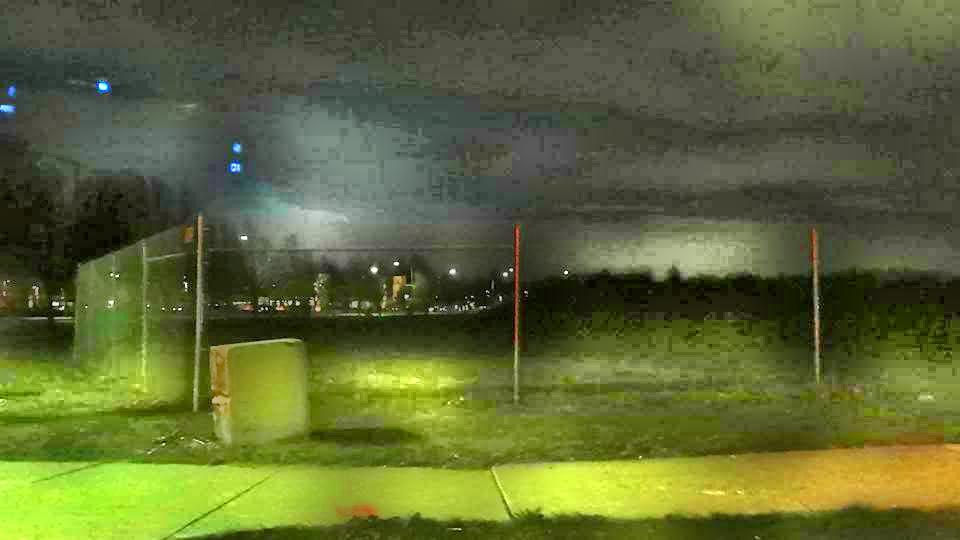}\\
			(a) input  & (b) LLNet \cite{LLNet}  &  (c) LightenNet \cite{LightenNet}\\
			\includegraphics[width=0.3\linewidth]{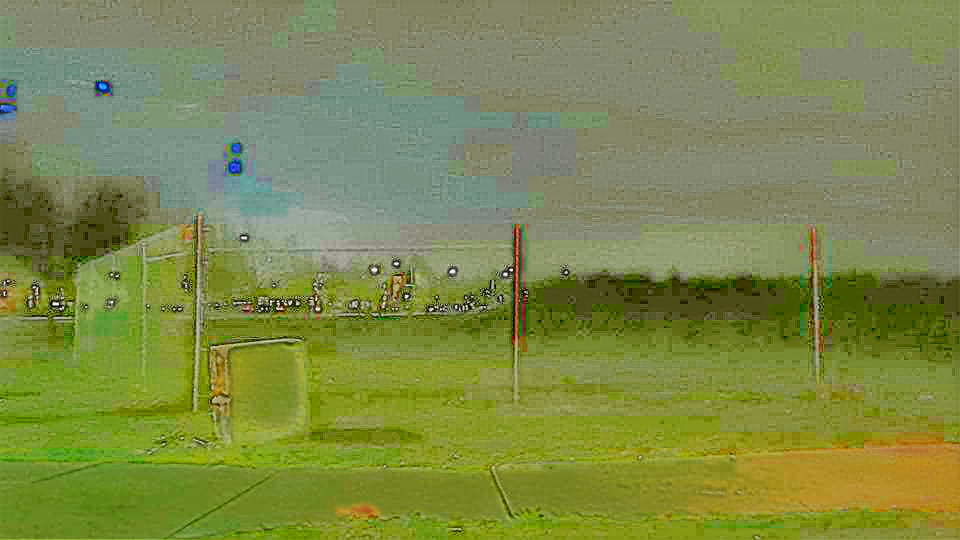}&
			\includegraphics[width=0.3\linewidth]{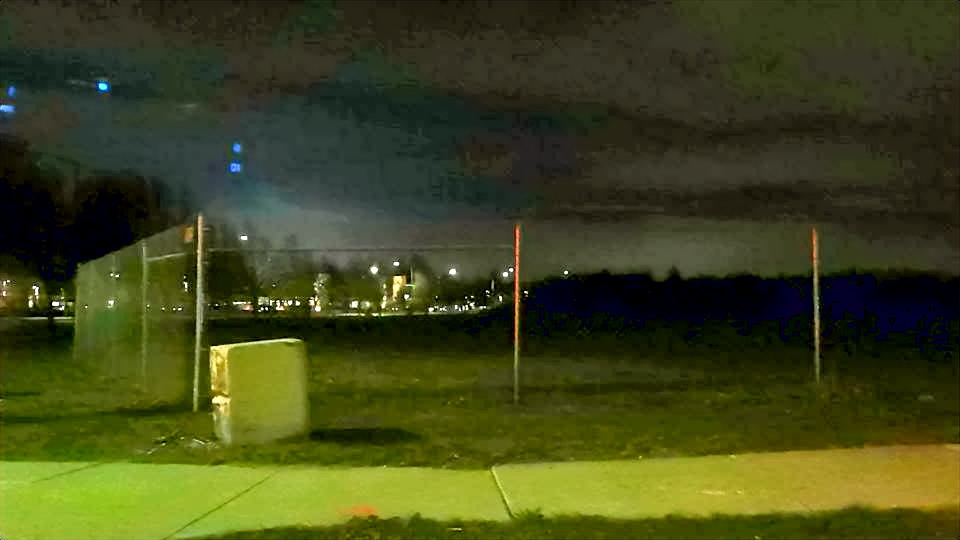}&
			\includegraphics[width=0.3\linewidth]{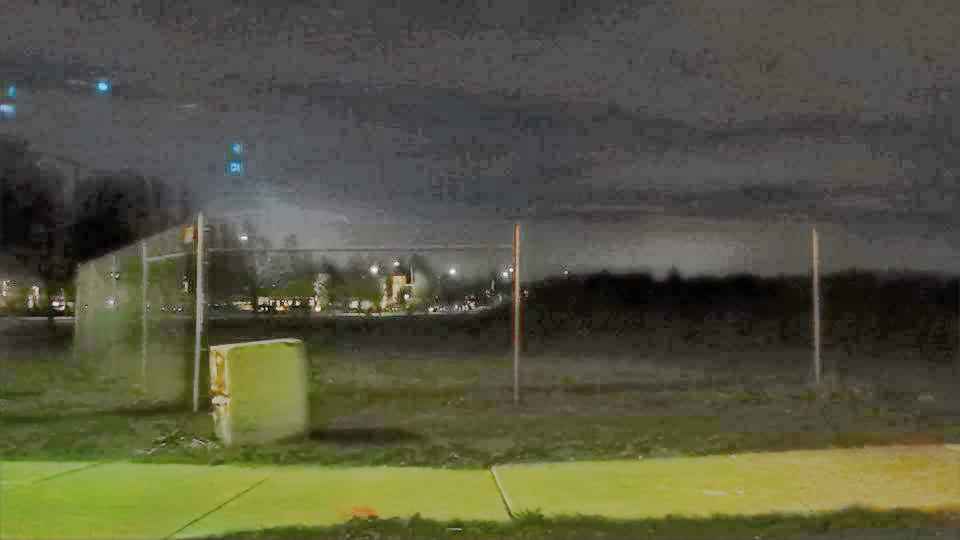}\\
			(d) Retinex-Net \cite{ChenBMVC18} & (e) MBLLEN \cite{LvBMVC2018} & (f) KinD \cite{ZhangACM19}\\
			\includegraphics[width=0.3\linewidth]{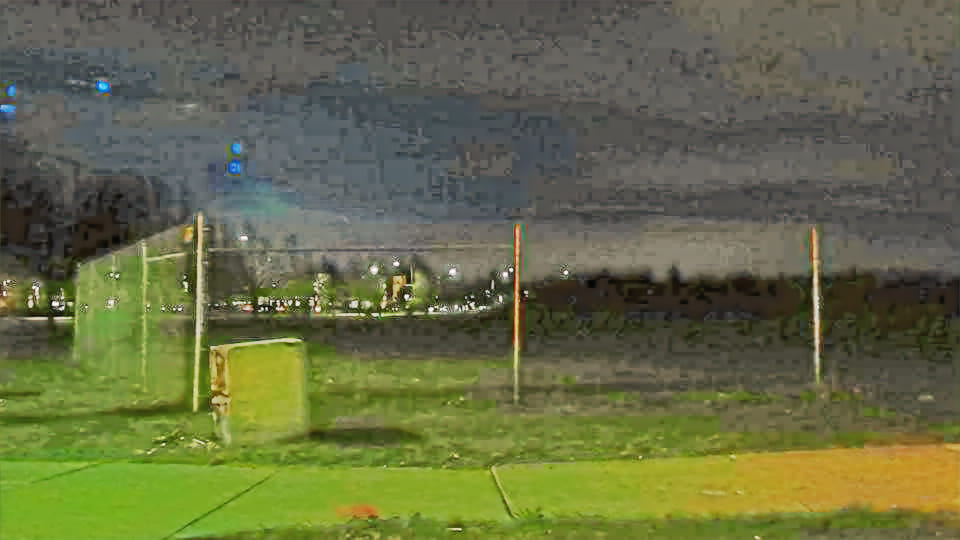}&
			\includegraphics[width=0.3\linewidth]{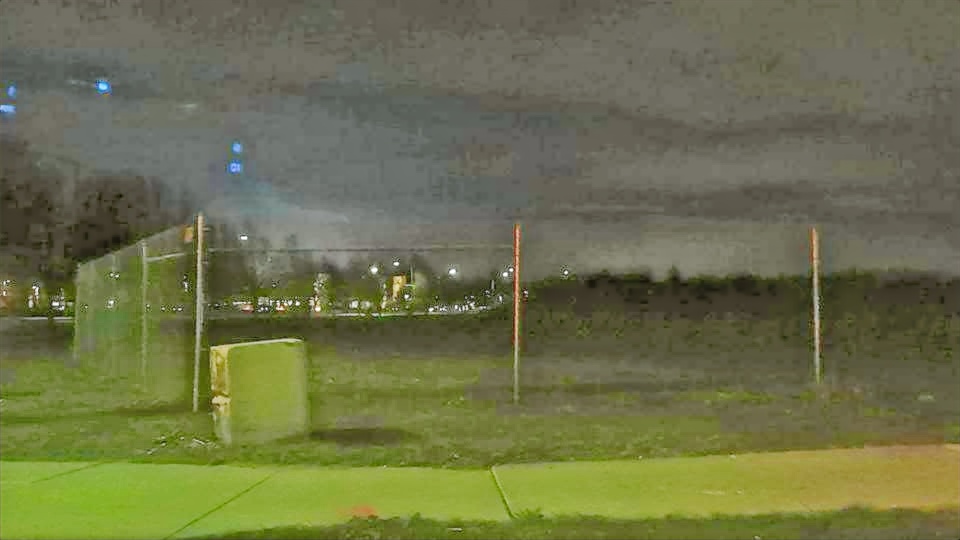}&
			\includegraphics[width=0.3\linewidth]{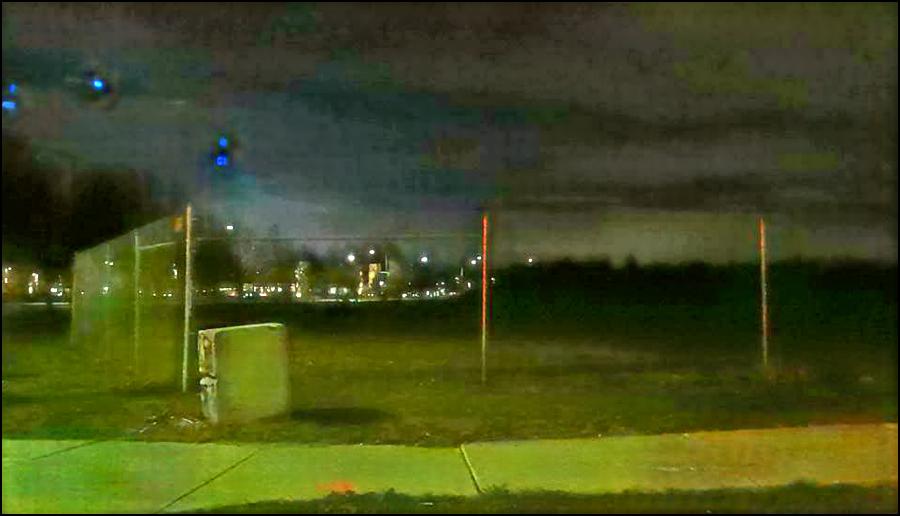}\\
			(g) KinD++ \cite{GuoIJCV2020} & (h) TBEFN \cite{TBEFN} &  (i)  	DSLR \cite{DSLR}\\
			\includegraphics[width=0.3\linewidth]{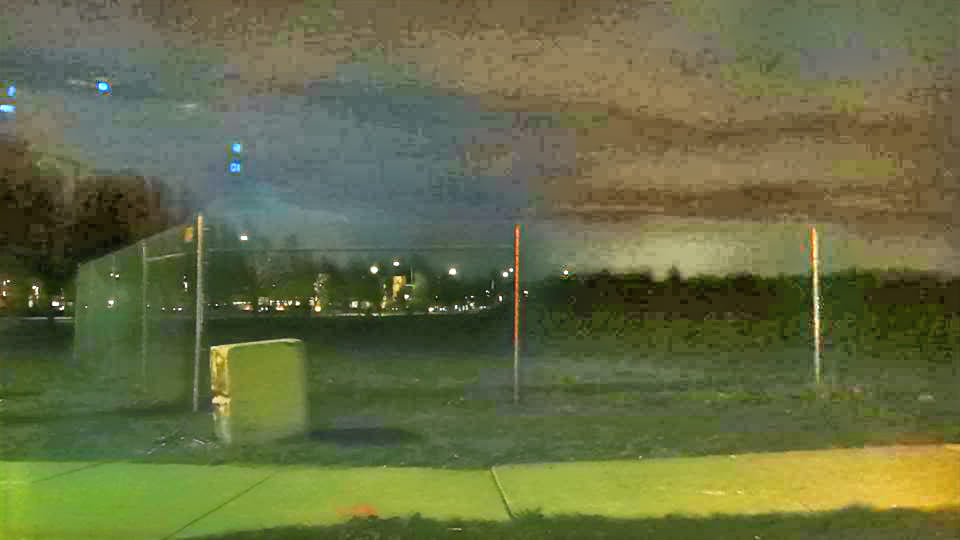}&
			\includegraphics[width=0.3\linewidth]{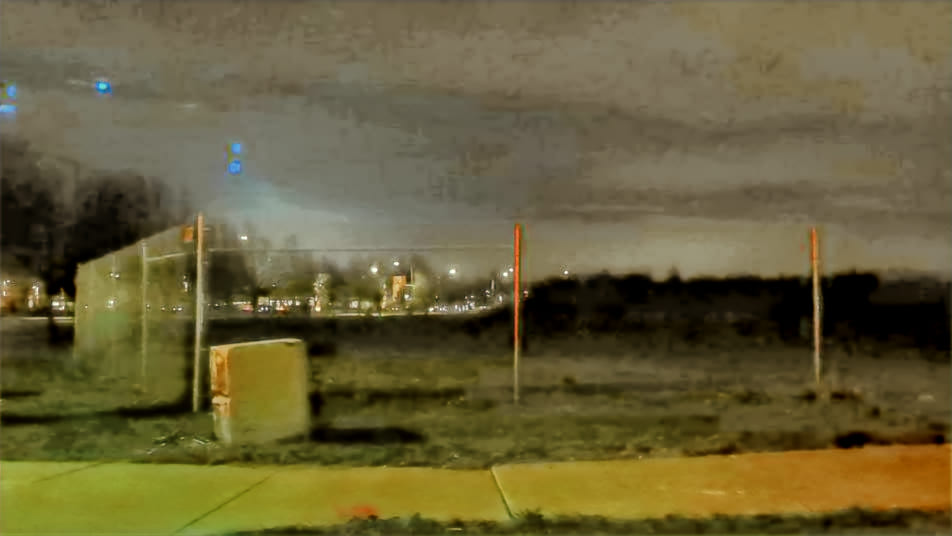}&
			\includegraphics[width=0.3\linewidth]{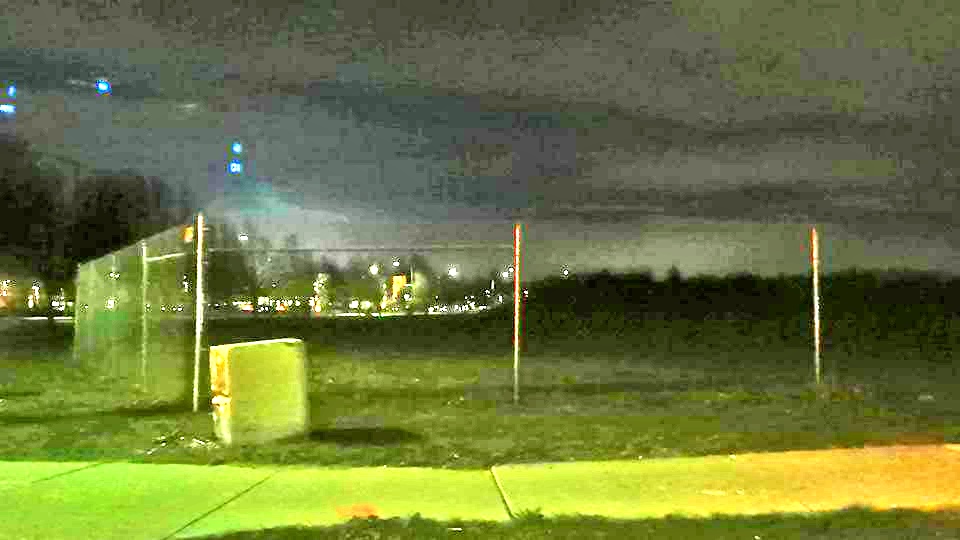}\\
			(j) EnlightenGAN \cite{EnlightenGAN} & (k) DRBN \cite{YangCVRP20} & (l) ExCNet \cite{ZhangACM191}\\
			\includegraphics[width=0.3\linewidth]{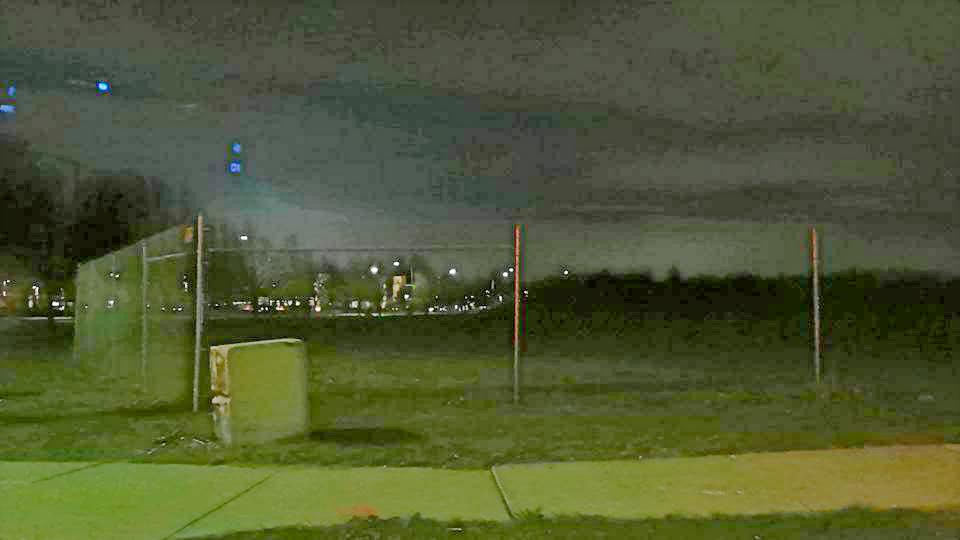}&
			\includegraphics[width=0.3\linewidth]{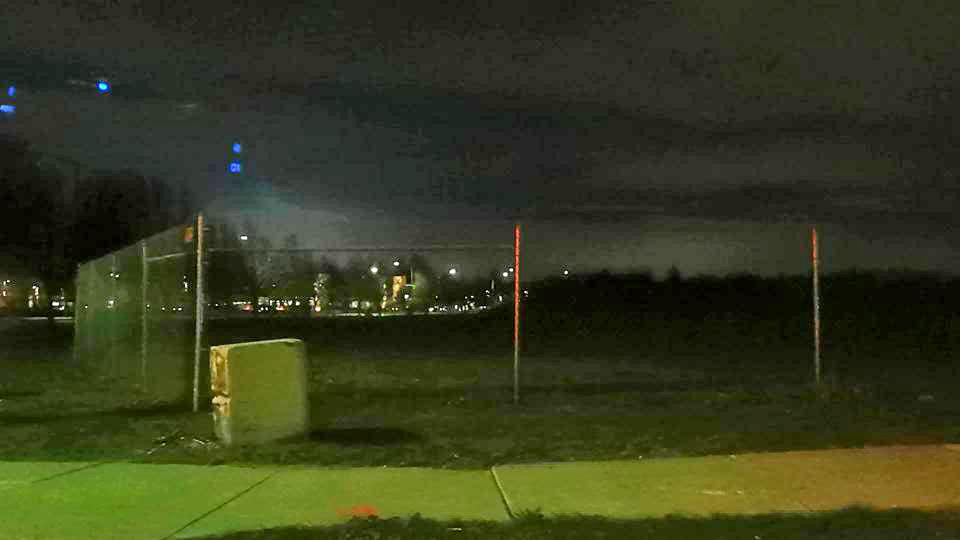}&
			~\\
			(m) Zero-DCE \cite{ZeroDCE} &  (n) 	RRDNet \cite{RRDNet}  &  \\
		\end{tabular}
	\end{center}
	\caption{Visual results of different methods on a low-light image sampled from LLIV-Phone-imgT dataset.}
	\label{fig:Phone2}
\end{figure*}

\begin{figure*} [h]
	\begin{center}
		\begin{tabular}{c@{ }c@{ }c@{ }}
			\includegraphics[width=0.3\linewidth]{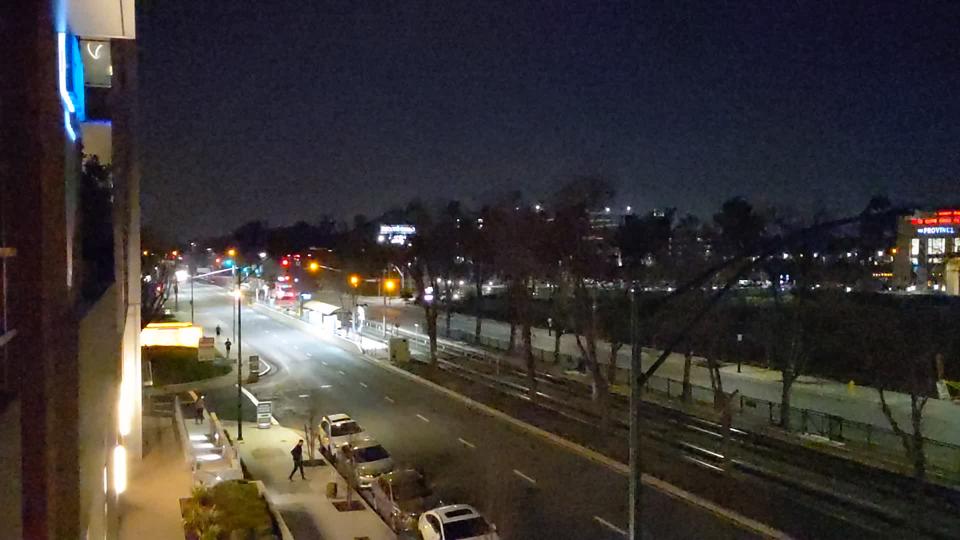}&
			\includegraphics[width=0.3\linewidth]{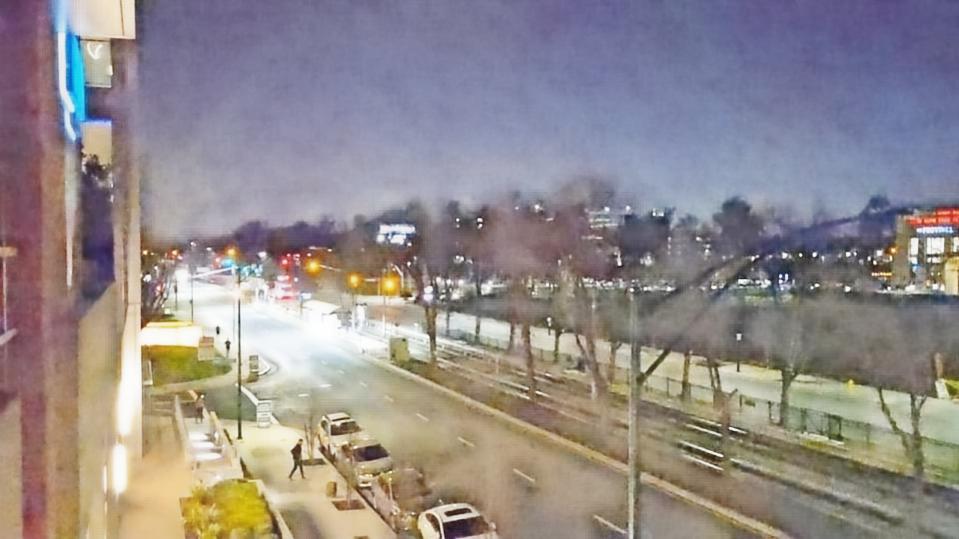}&
			\includegraphics[width=0.3\linewidth]{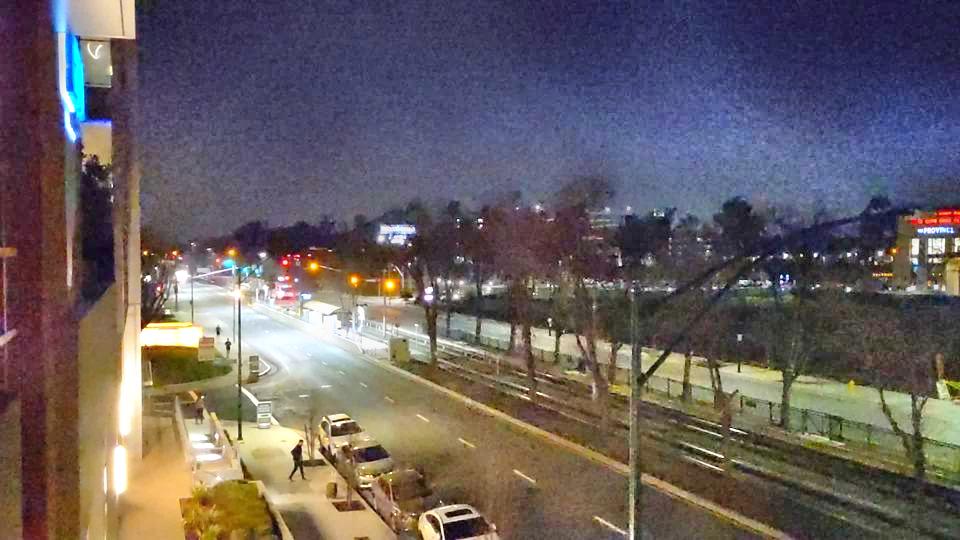}\\
			(a) input  & (b) LLNet \cite{LLNet}  &  (c) LightenNet \cite{LightenNet}\\
			\includegraphics[width=0.3\linewidth]{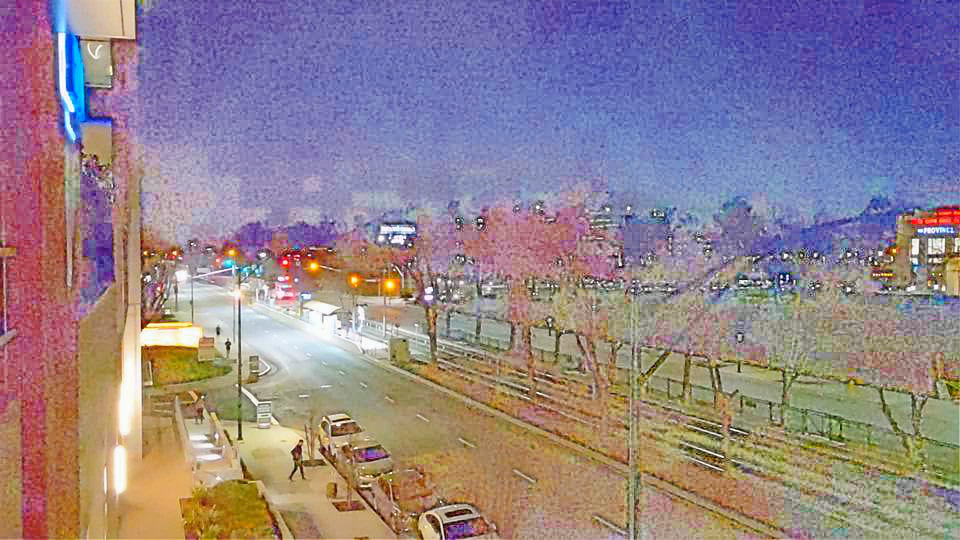}&
			\includegraphics[width=0.3\linewidth]{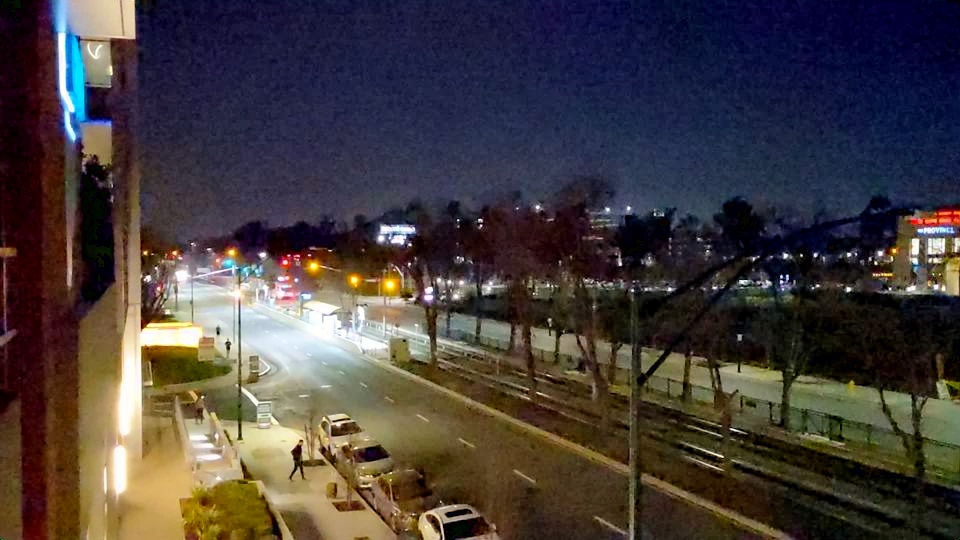}&
			\includegraphics[width=0.3\linewidth]{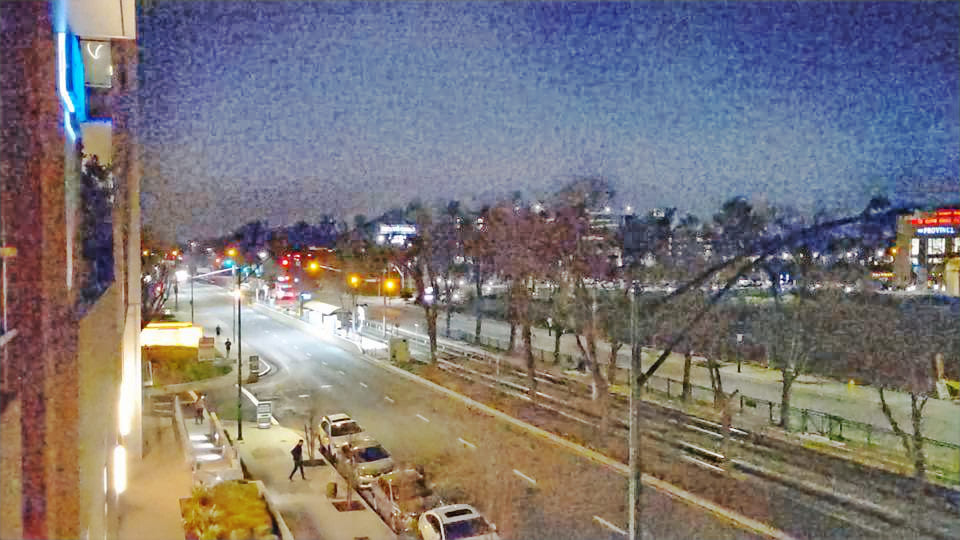}\\
			(d) Retinex-Net \cite{ChenBMVC18} & (e) MBLLEN \cite{LvBMVC2018} & (f) KinD \cite{ZhangACM19}\\
			\includegraphics[width=0.3\linewidth]{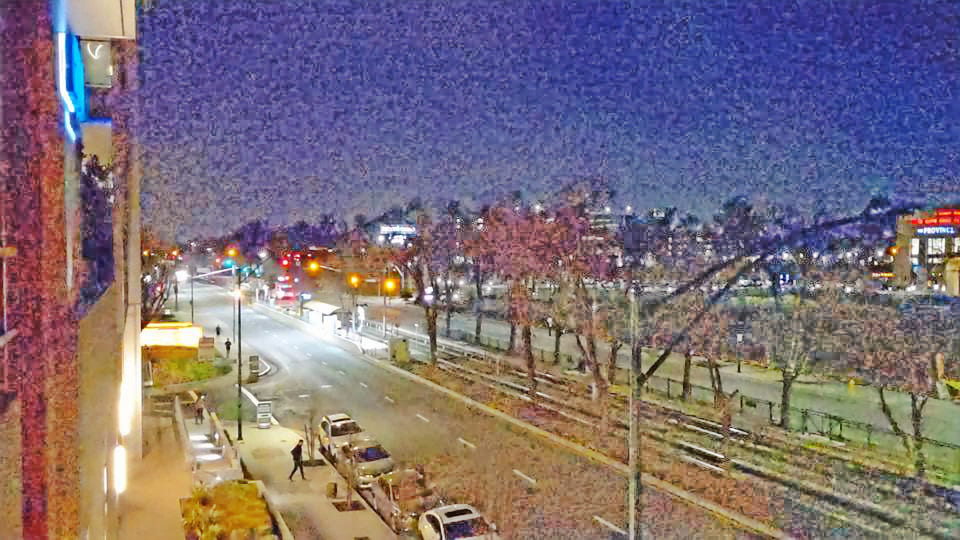}&
			\includegraphics[width=0.3\linewidth]{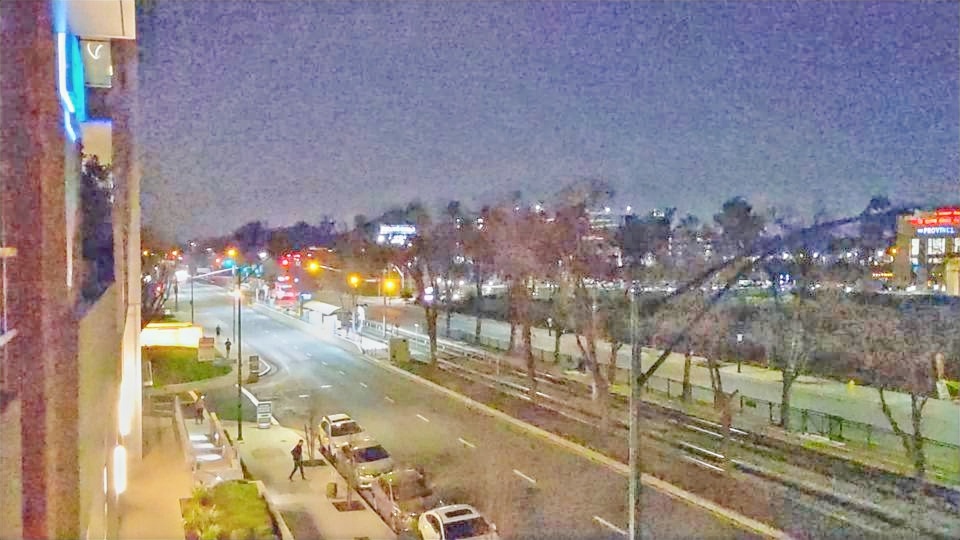}&
			\includegraphics[width=0.3\linewidth]{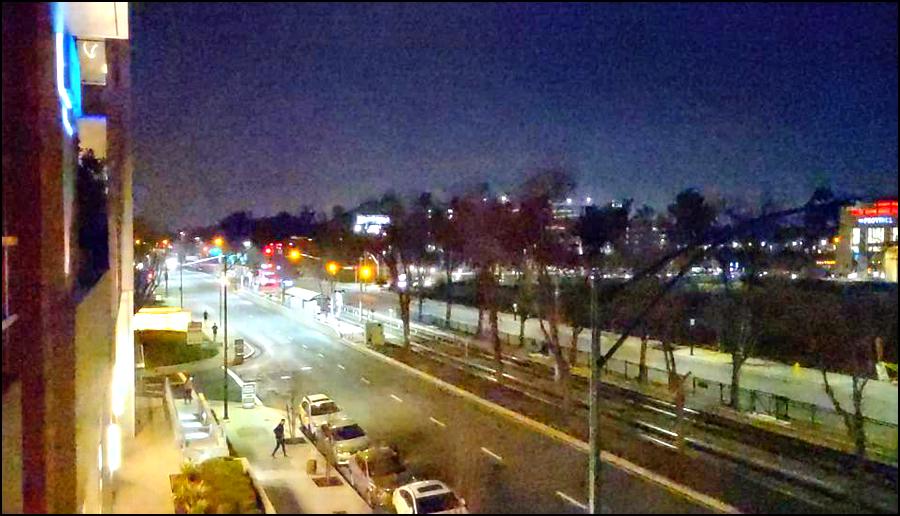}\\
			(g) KinD++ \cite{GuoIJCV2020} & (h) TBEFN \cite{TBEFN} &  (i)  	DSLR \cite{DSLR}\\
			\includegraphics[width=0.3\linewidth]{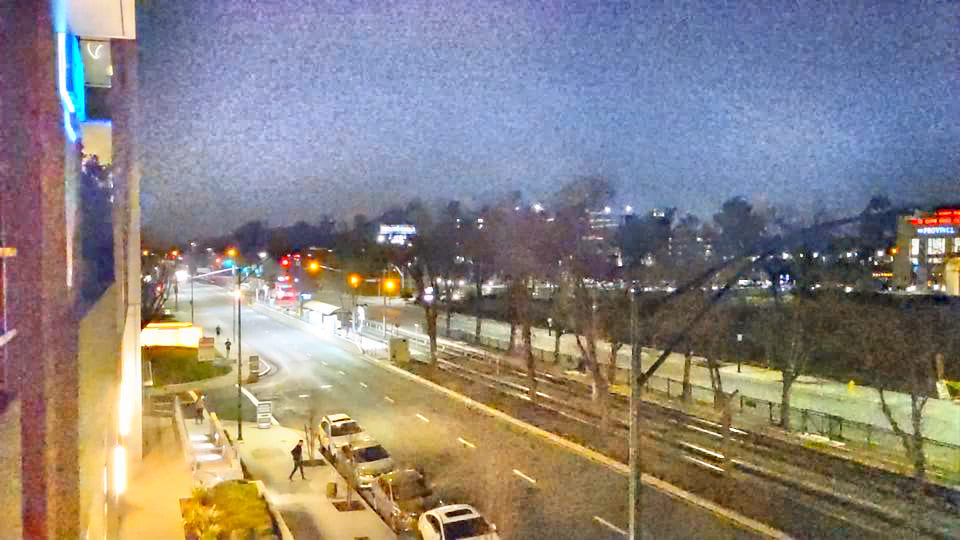}&
			\includegraphics[width=0.3\linewidth]{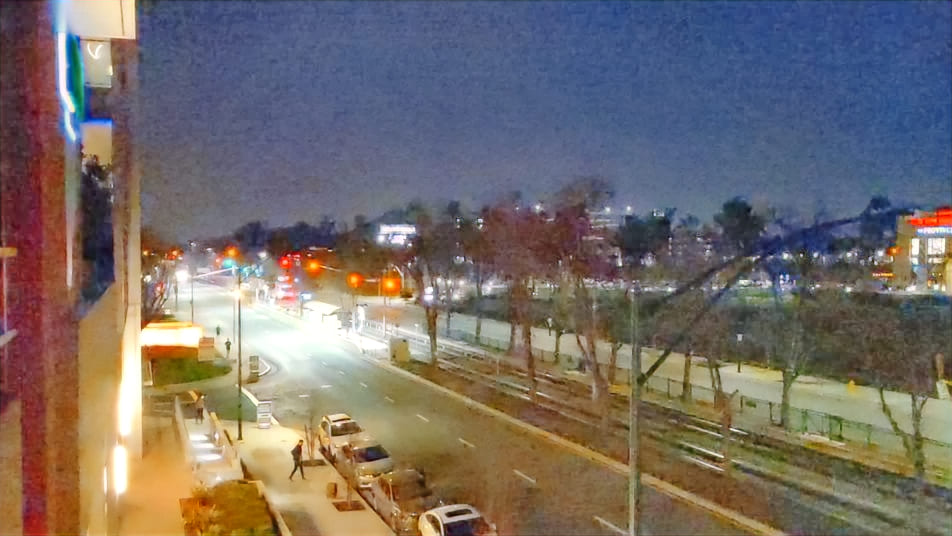}&
			\includegraphics[width=0.3\linewidth]{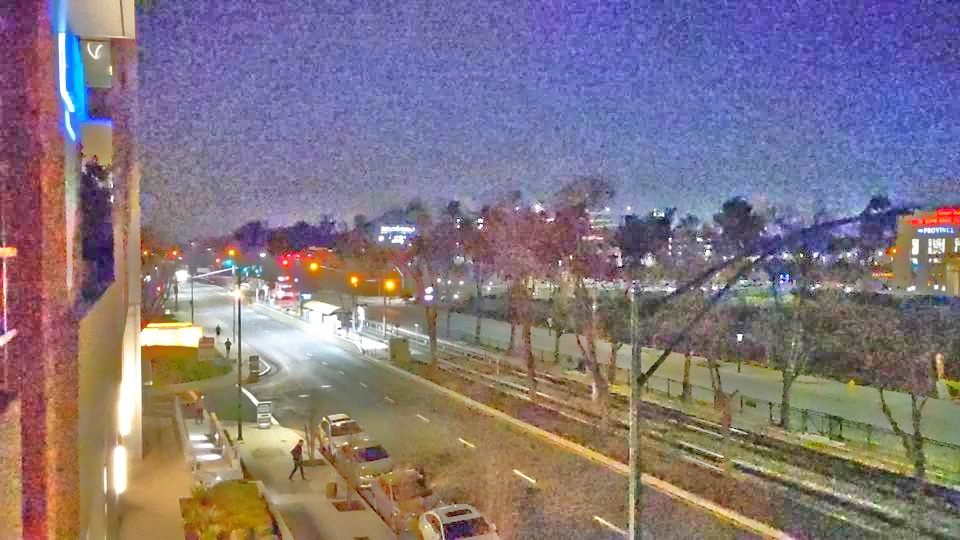}\\
			(j) EnlightenGAN \cite{EnlightenGAN} & (k) DRBN \cite{YangCVRP20} & (l) ExCNet \cite{ZhangACM191}\\
			\includegraphics[width=0.3\linewidth]{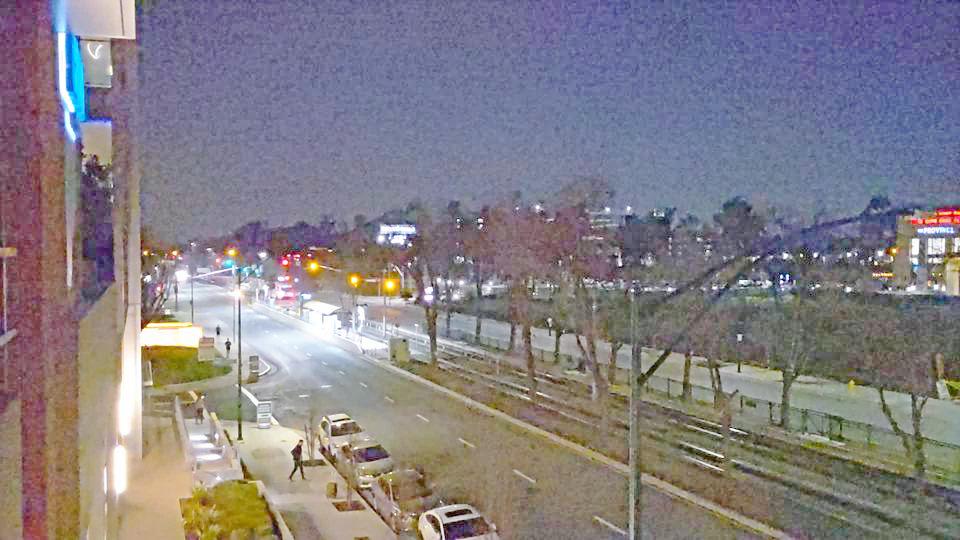}&
			\includegraphics[width=0.3\linewidth]{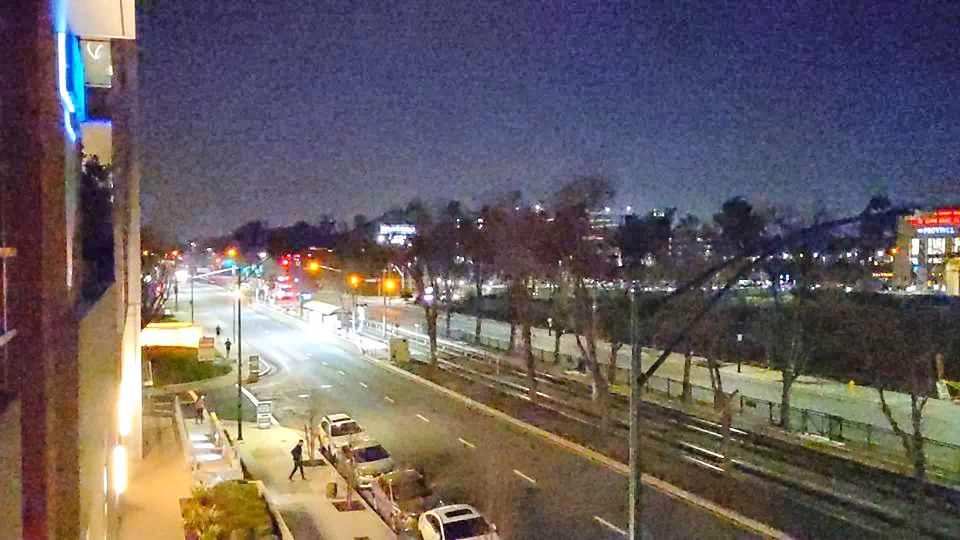}&
			~\\
			(m) Zero-DCE \cite{ZeroDCE} &  (n) 	RRDNet \cite{RRDNet}  &  \\
		\end{tabular}
	\end{center}
	\caption{Visual results of different methods on a low-light image sampled from LLIV-Phone-imgT dataset.}
	\label{fig:Phone3}
\end{figure*}

\begin{figure*}[!t]
	\begin{center}
		\begin{tabular}{c@{ }c@{ }c@{ }c@{ }}
			\includegraphics[width=0.25\linewidth]{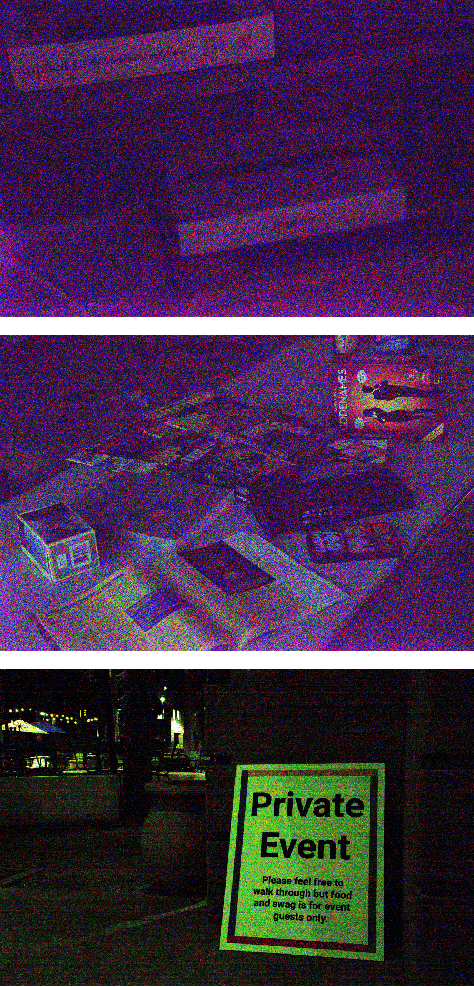}&
			\includegraphics[width=0.25\linewidth]{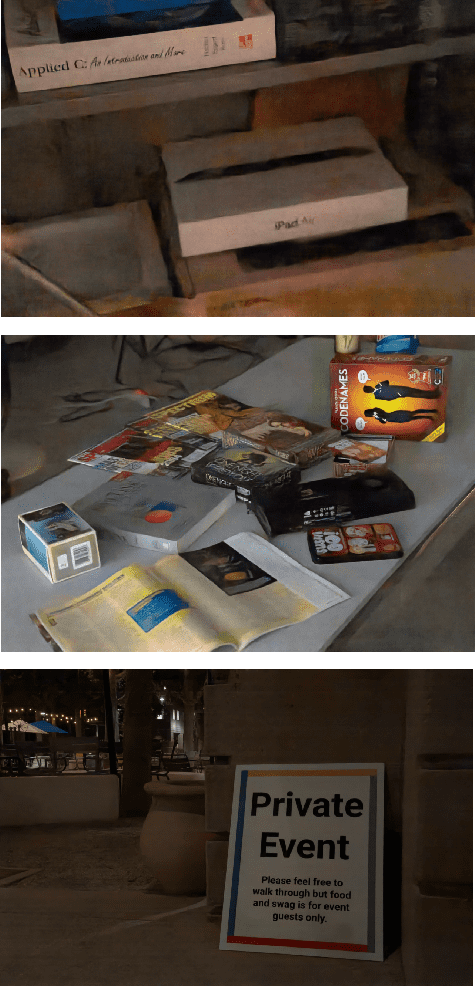}&
			\includegraphics[width=0.25\linewidth]{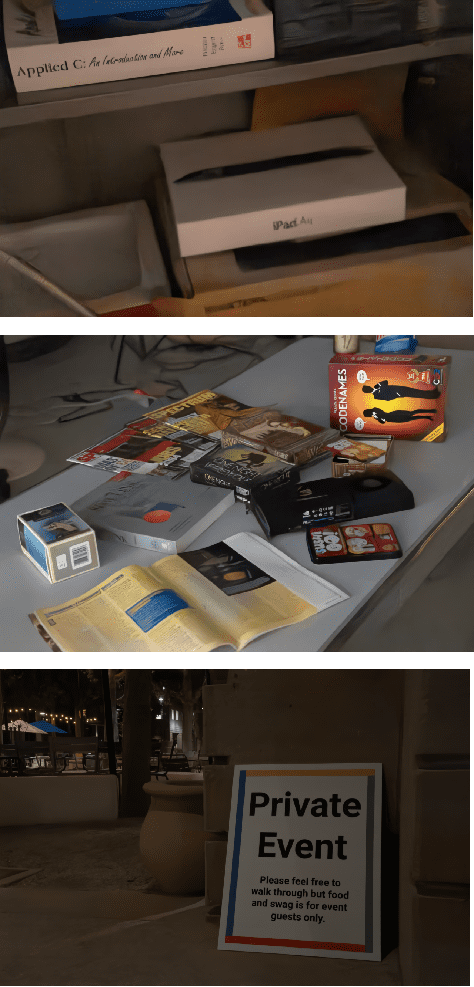}&
			\includegraphics[width=0.25\linewidth]{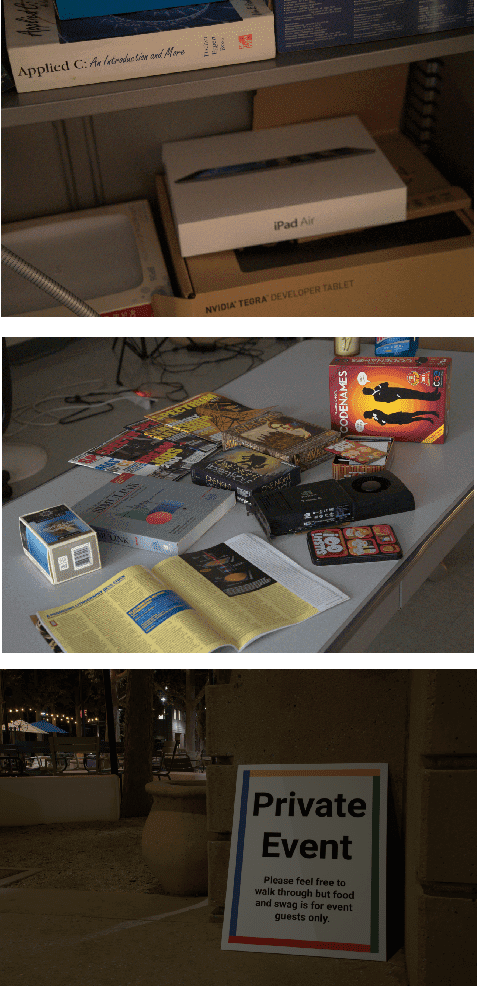}\\
			(a) inputs  & (b) SID \cite{Chen18} & (c) EEMEFN \cite{ZhuAAAI20}&  (d) GT   \\
		\end{tabular}
	\end{center}
	\vspace{-2pt}
	\caption{Visual results of different methods on raw low-light images of Bayer pattern sampled from  SID-test-Bayer test dataset. The inputs are amplified for visualization.}
	\label{fig:Sony}
	\vspace{-4pt}
\end{figure*}

\begin{figure*}[!t]
	\begin{center}
		\begin{tabular}{c@{ }c@{ }c@{ }c@{ }}
			\includegraphics[width=0.25\linewidth]{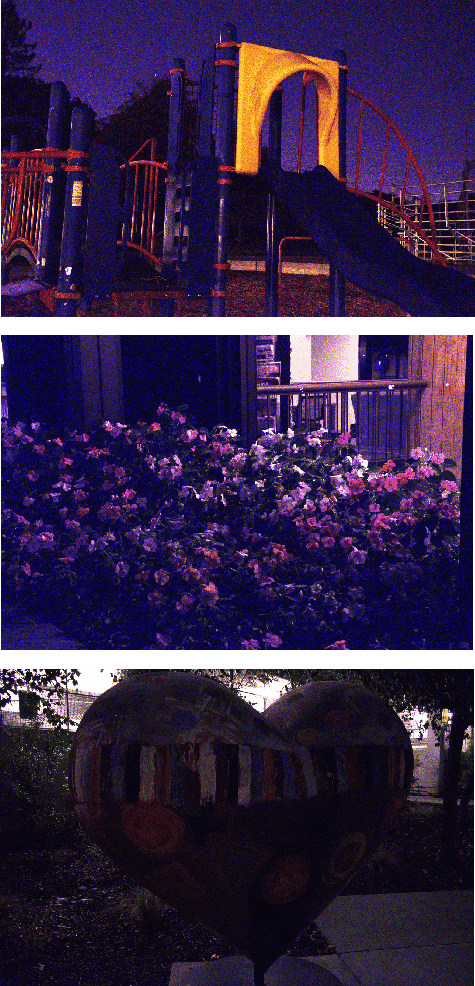}&
			\includegraphics[width=0.25\linewidth]{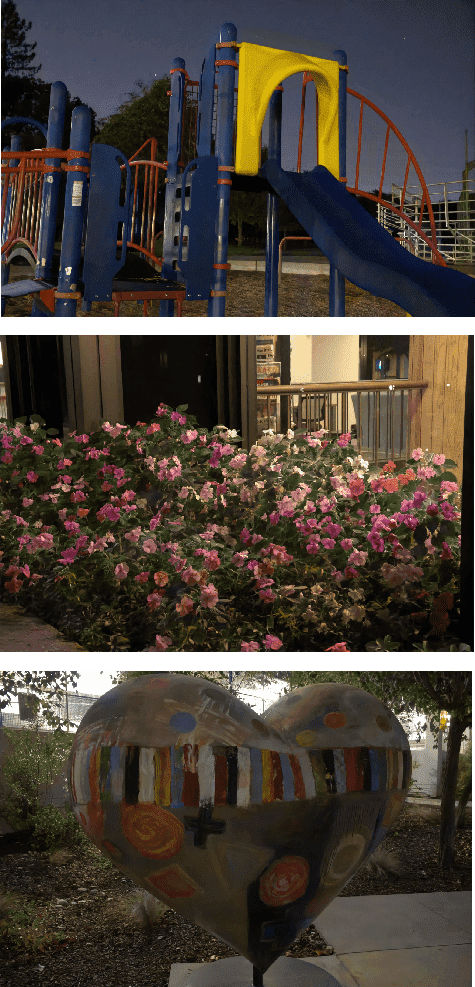}&
			\includegraphics[width=0.25\linewidth]{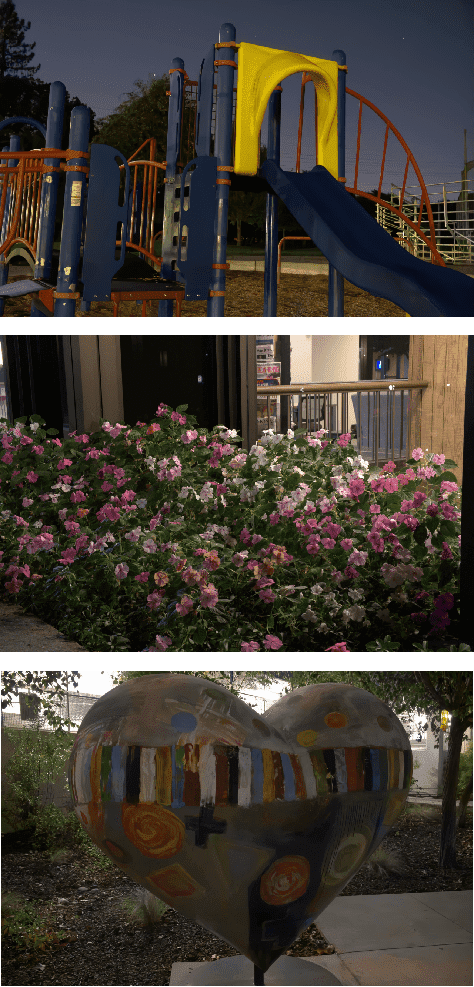}&
			\includegraphics[width=0.25\linewidth]{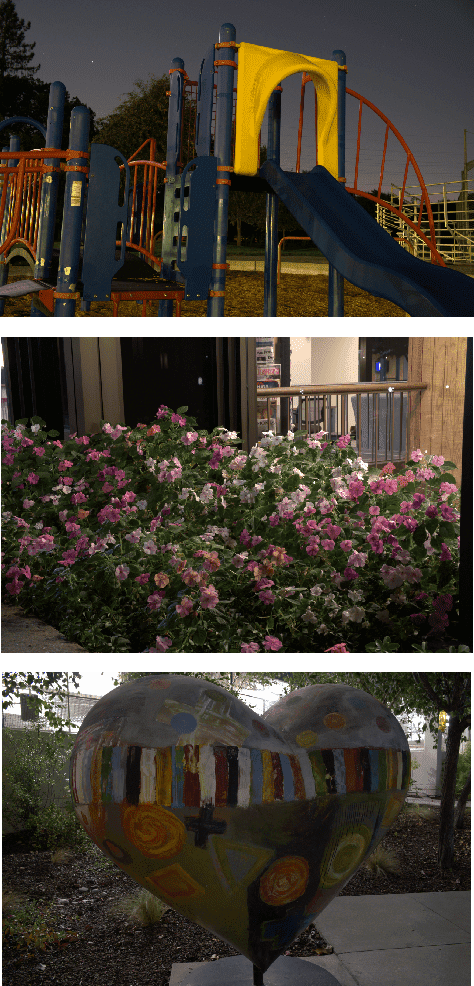}\\
			(a) inputs  & (b) SID \cite{Chen18} & (c) EEMEFN \cite{ZhuAAAI20}&  (d) GT   \\
		\end{tabular}
	\end{center}
	\vspace{-2pt}
	\caption{Visual results of different methods on raw low-light images of APS-C X-Trans pattern sampled from  SID-test-X-Trans test dataset. The inputs are amplified for visualization.}
	\label{fig:Fuji}
	\vspace{-4pt}
\end{figure*}

\begin{figure*} [h]
	\begin{center}
		\begin{tabular}{c@{ }c@{ }c@{ }}
			\includegraphics[width=0.3\linewidth]{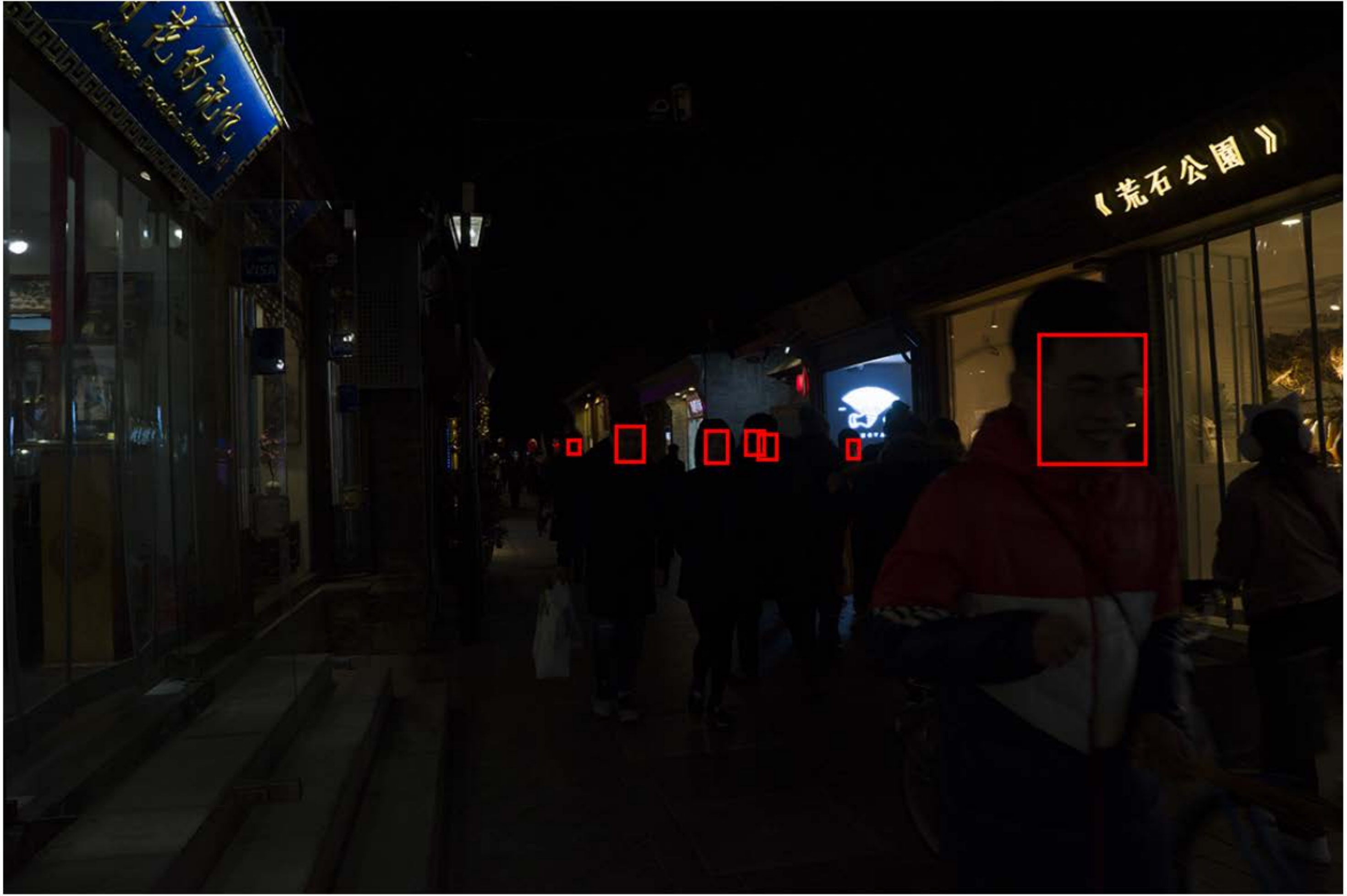}&
			\includegraphics[width=0.3\linewidth]{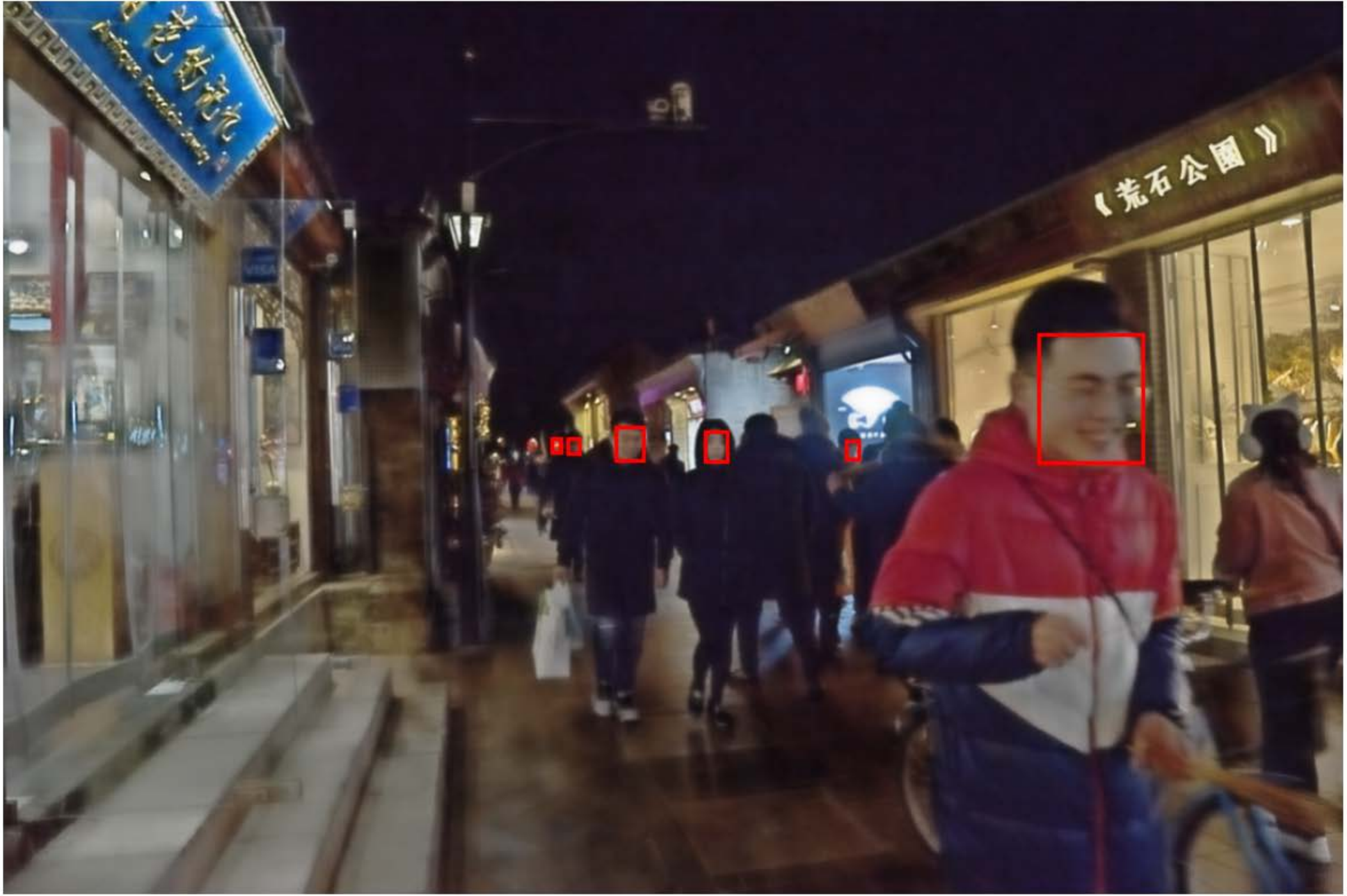}&
			\includegraphics[width=0.3\linewidth]{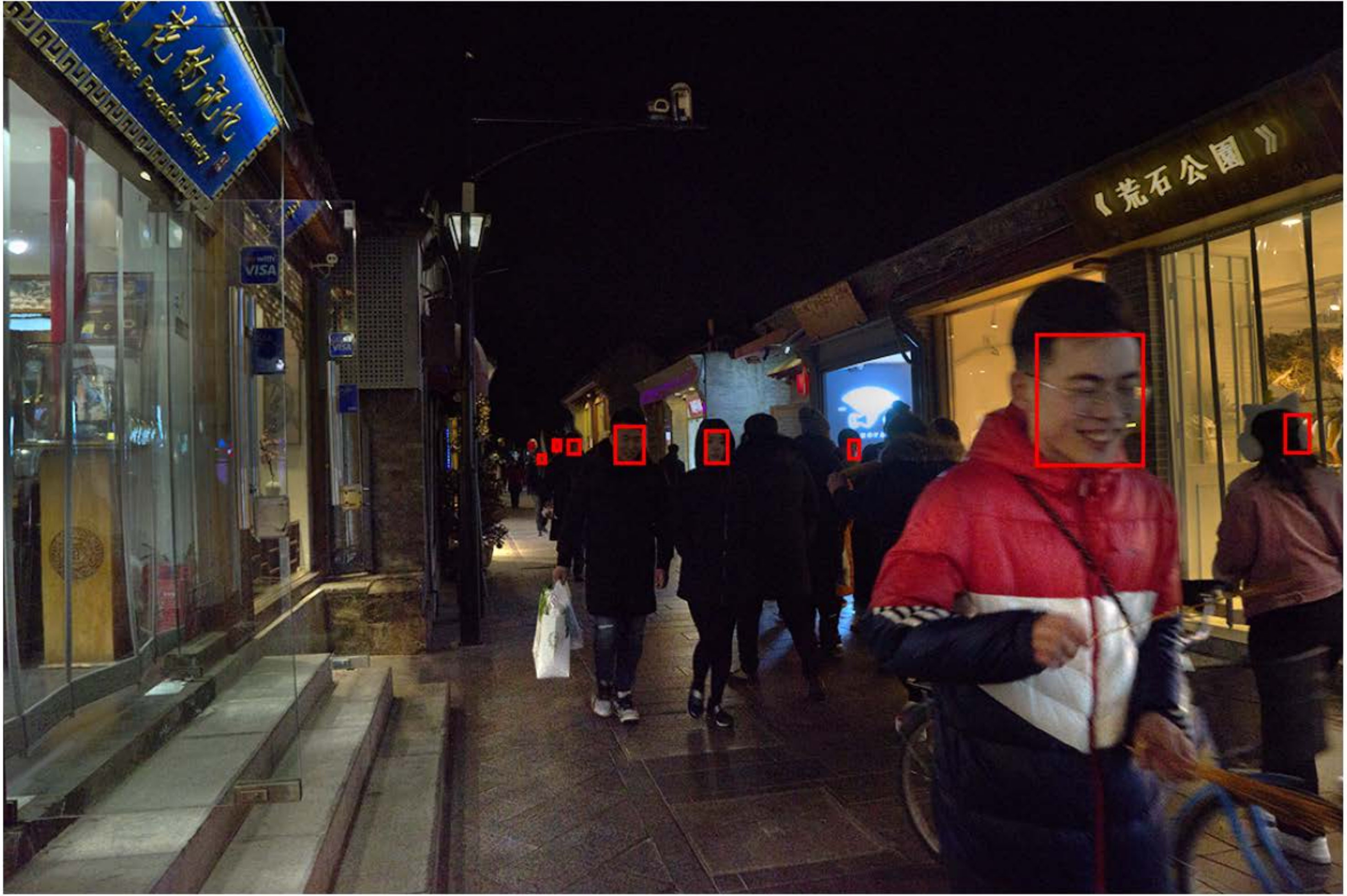}\\
			(a) input  & (b) LLNet \cite{LLNet}  &  (c) LightenNet \cite{LightenNet}\\
			\includegraphics[width=0.3\linewidth]{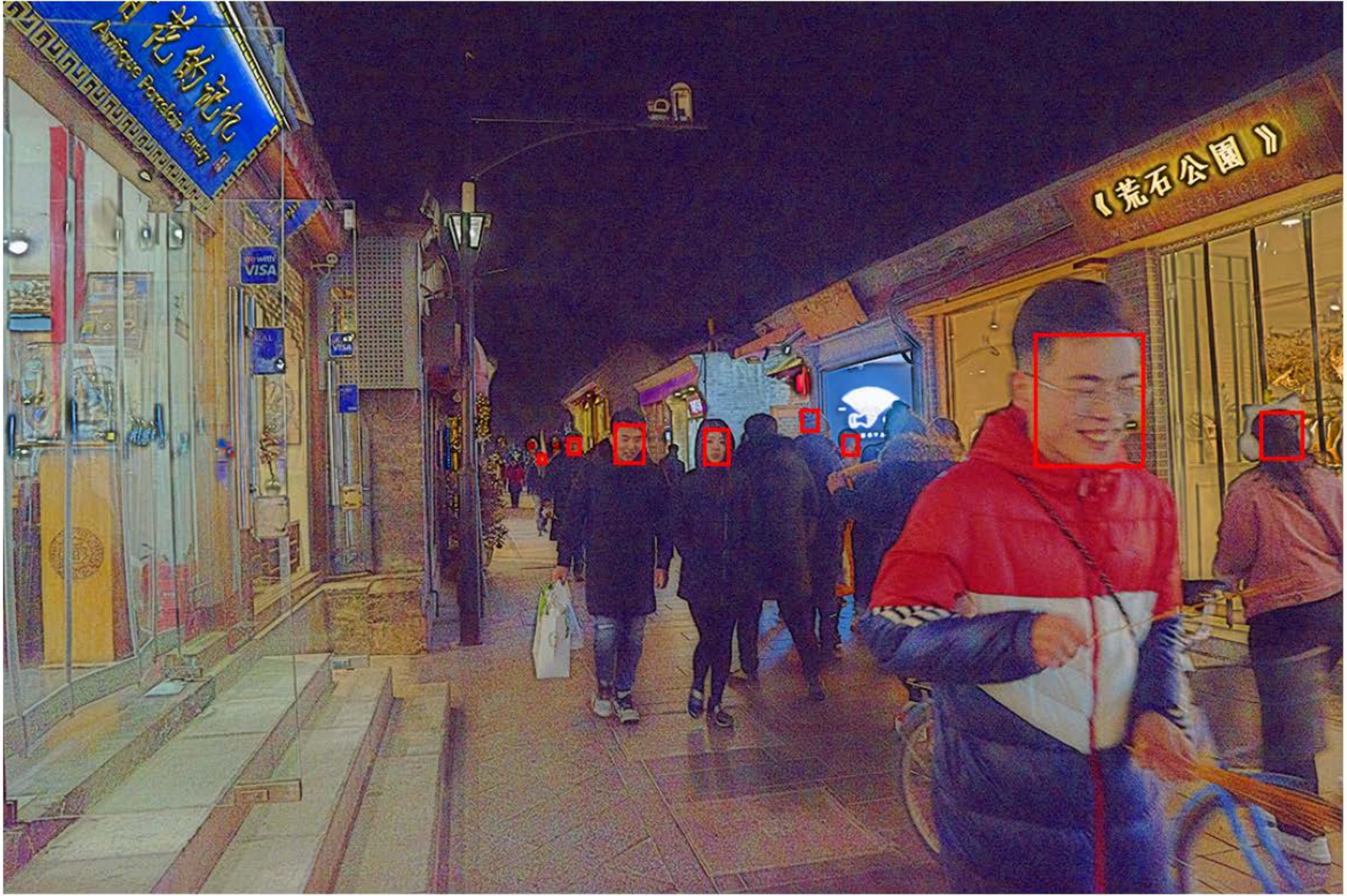}&
			\includegraphics[width=0.3\linewidth]{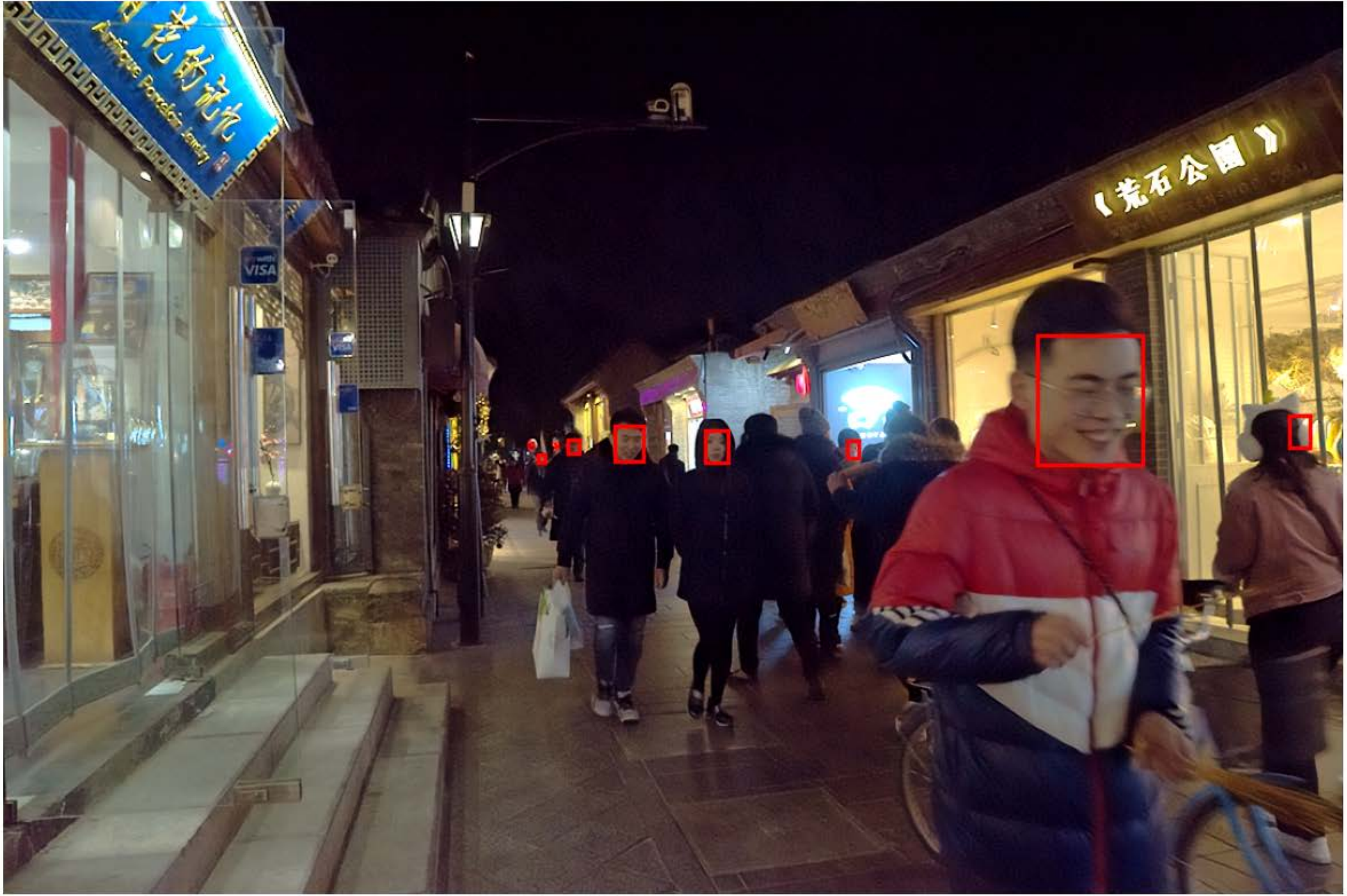}&
			\includegraphics[width=0.3\linewidth]{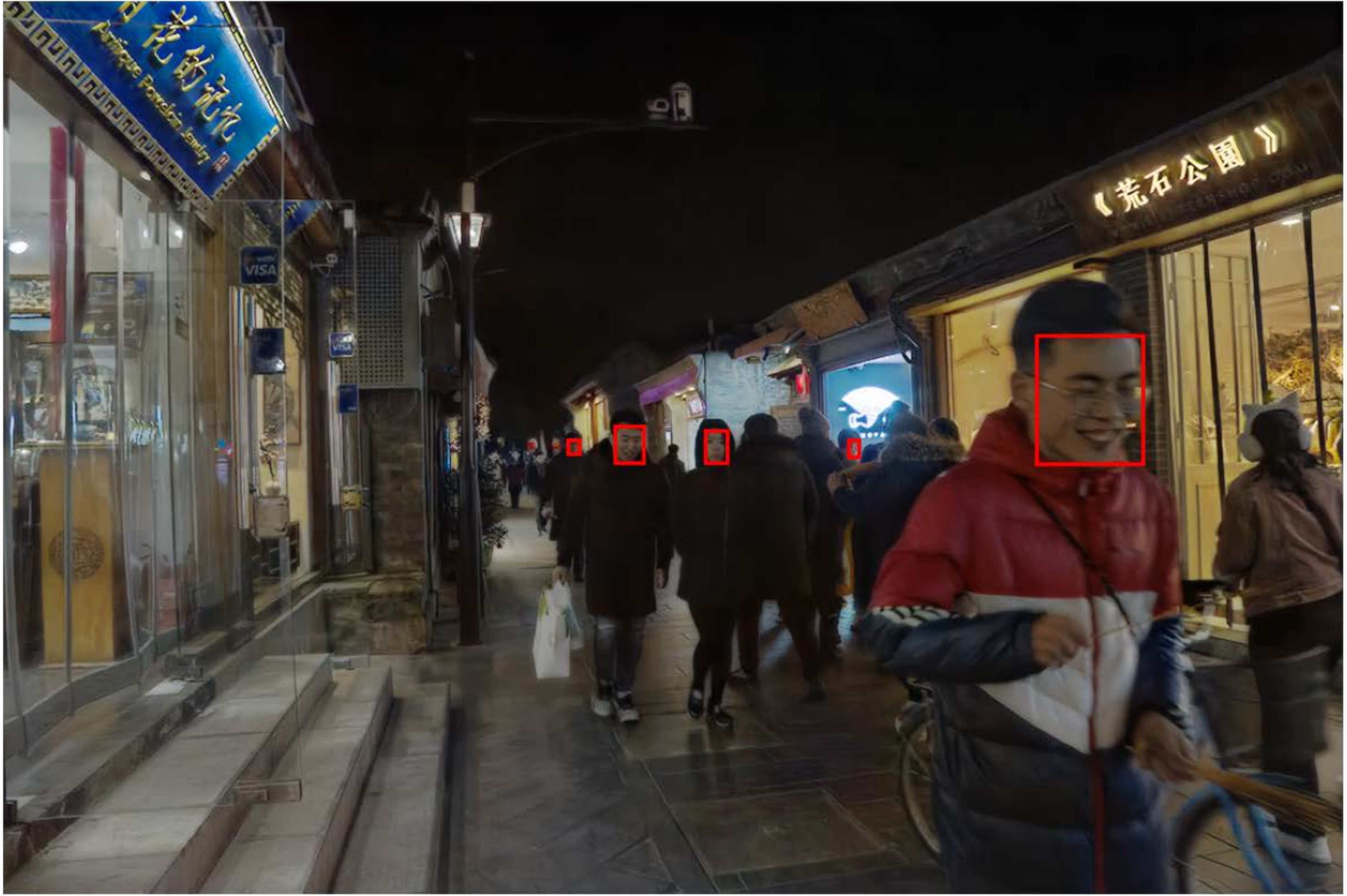}\\
			(d) Retinex-Net \cite{ChenBMVC18} & (e) MBLLEN \cite{LvBMVC2018} & (f) KinD \cite{ZhangACM19}\\
			\includegraphics[width=0.3\linewidth]{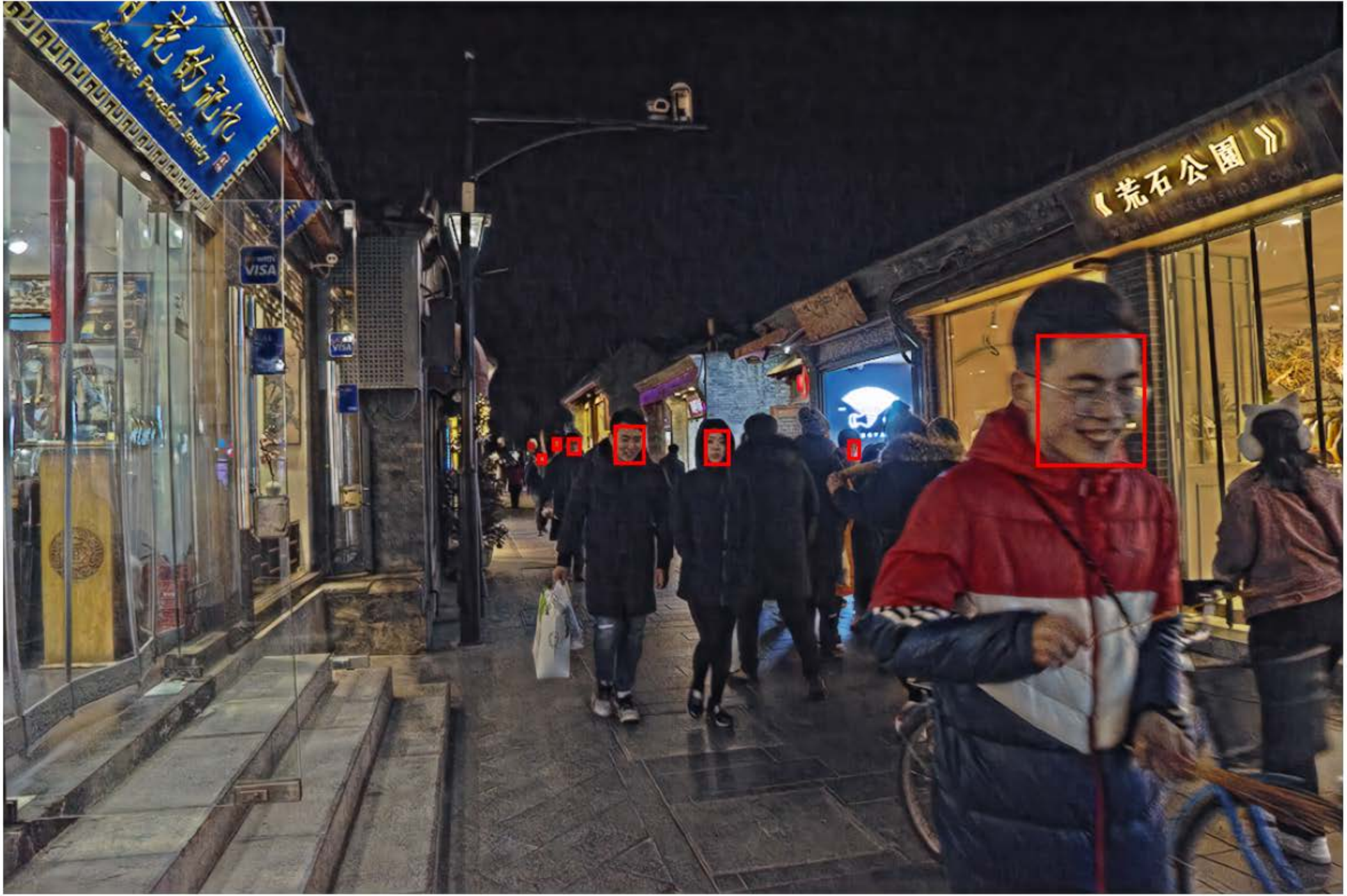}&
			\includegraphics[width=0.3\linewidth]{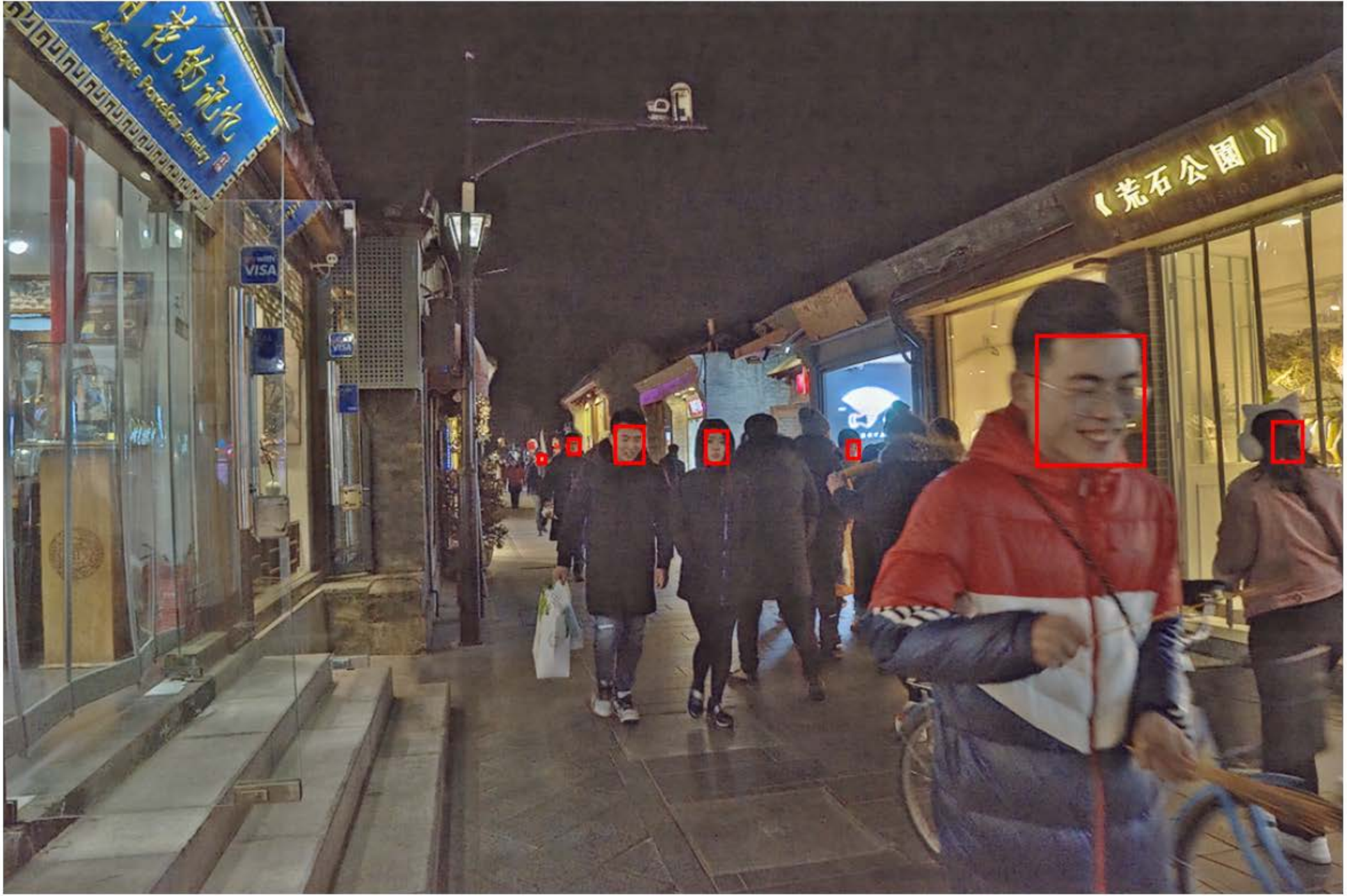}&
			\includegraphics[width=0.3\linewidth]{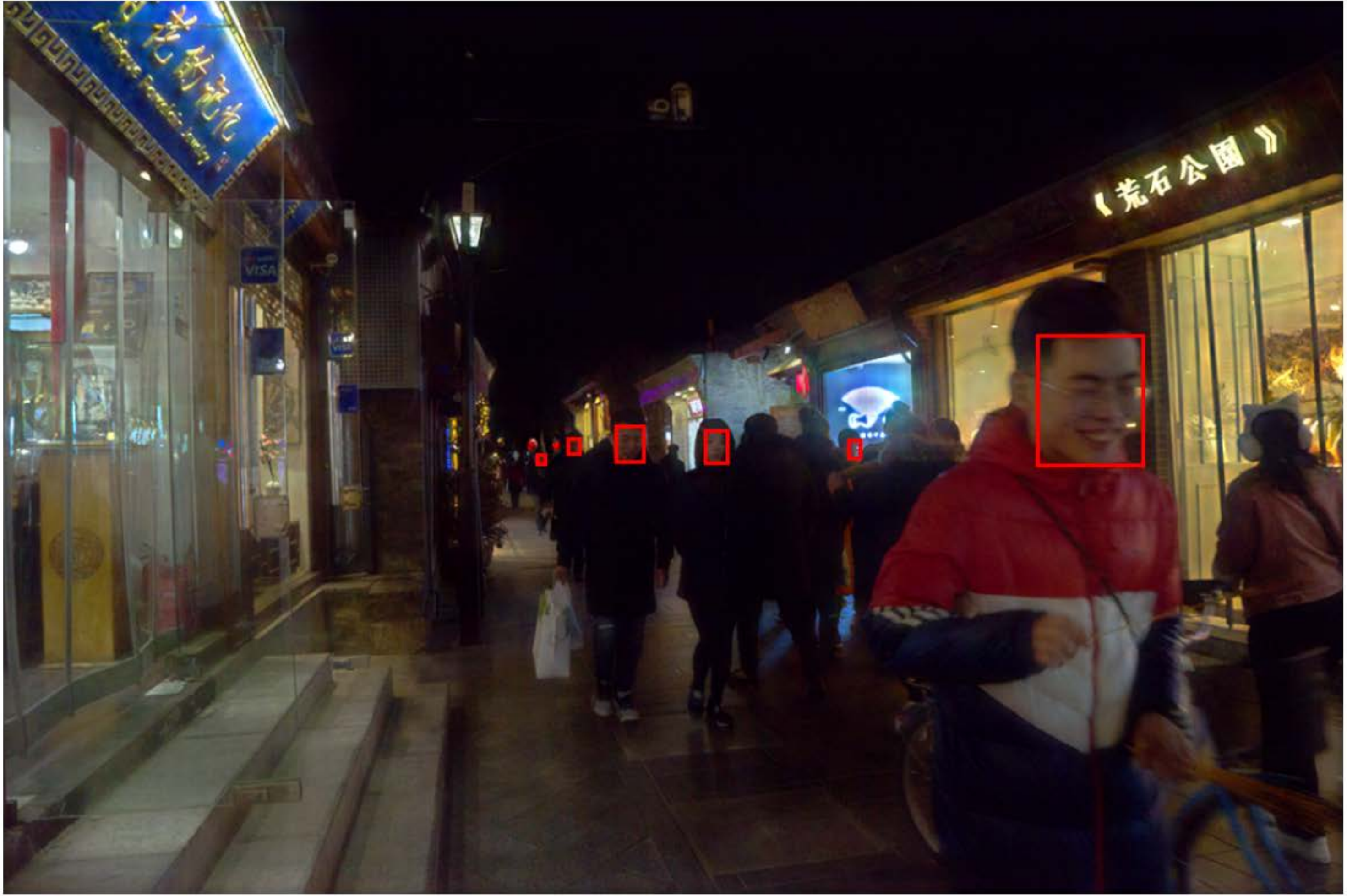}\\
			(g) KinD++ \cite{GuoIJCV2020} & (h) TBEFN \cite{TBEFN} &  (i)  	DSLR \cite{DSLR}\\
			\includegraphics[width=0.3\linewidth]{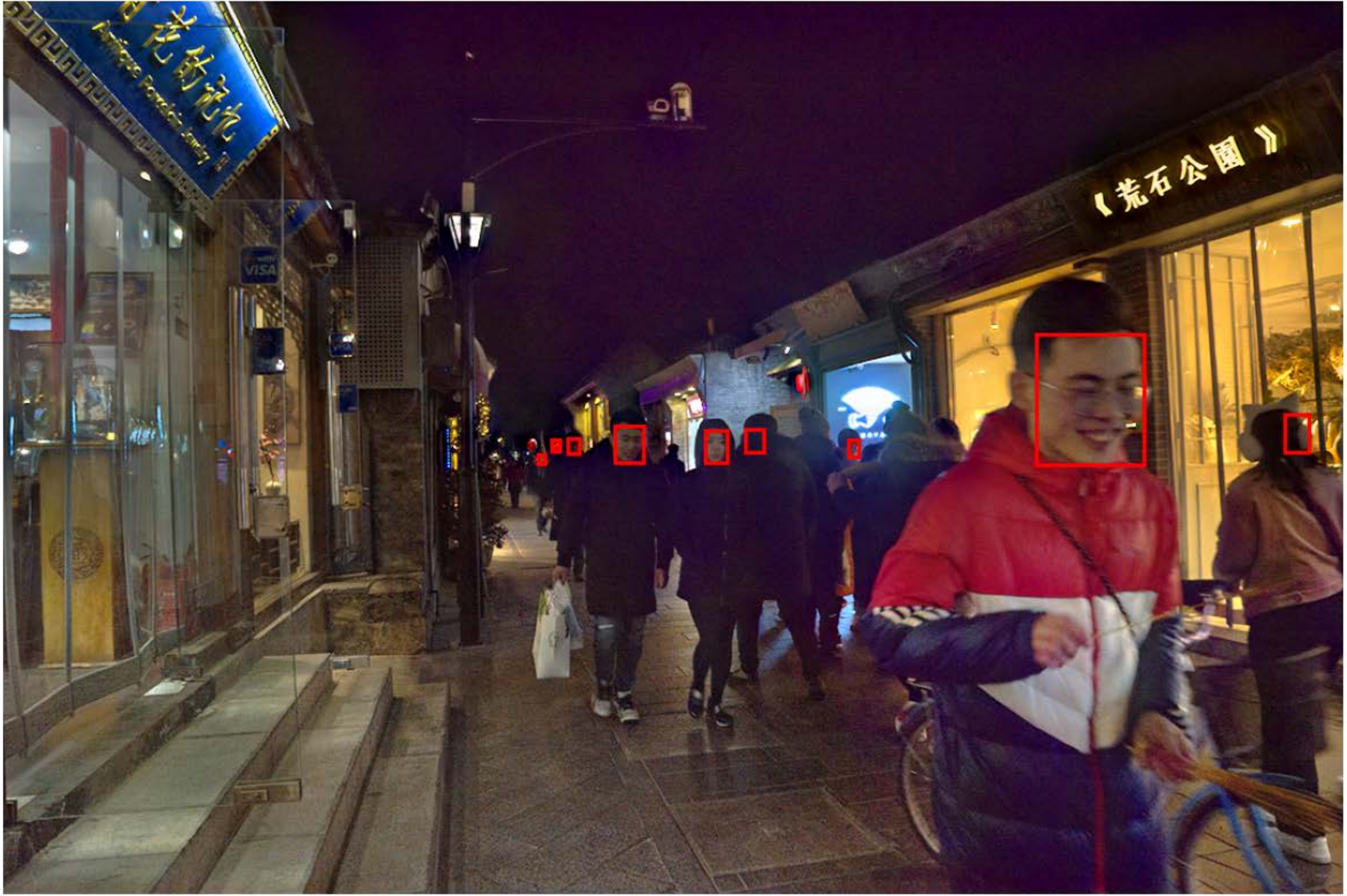}&
			\includegraphics[width=0.3\linewidth]{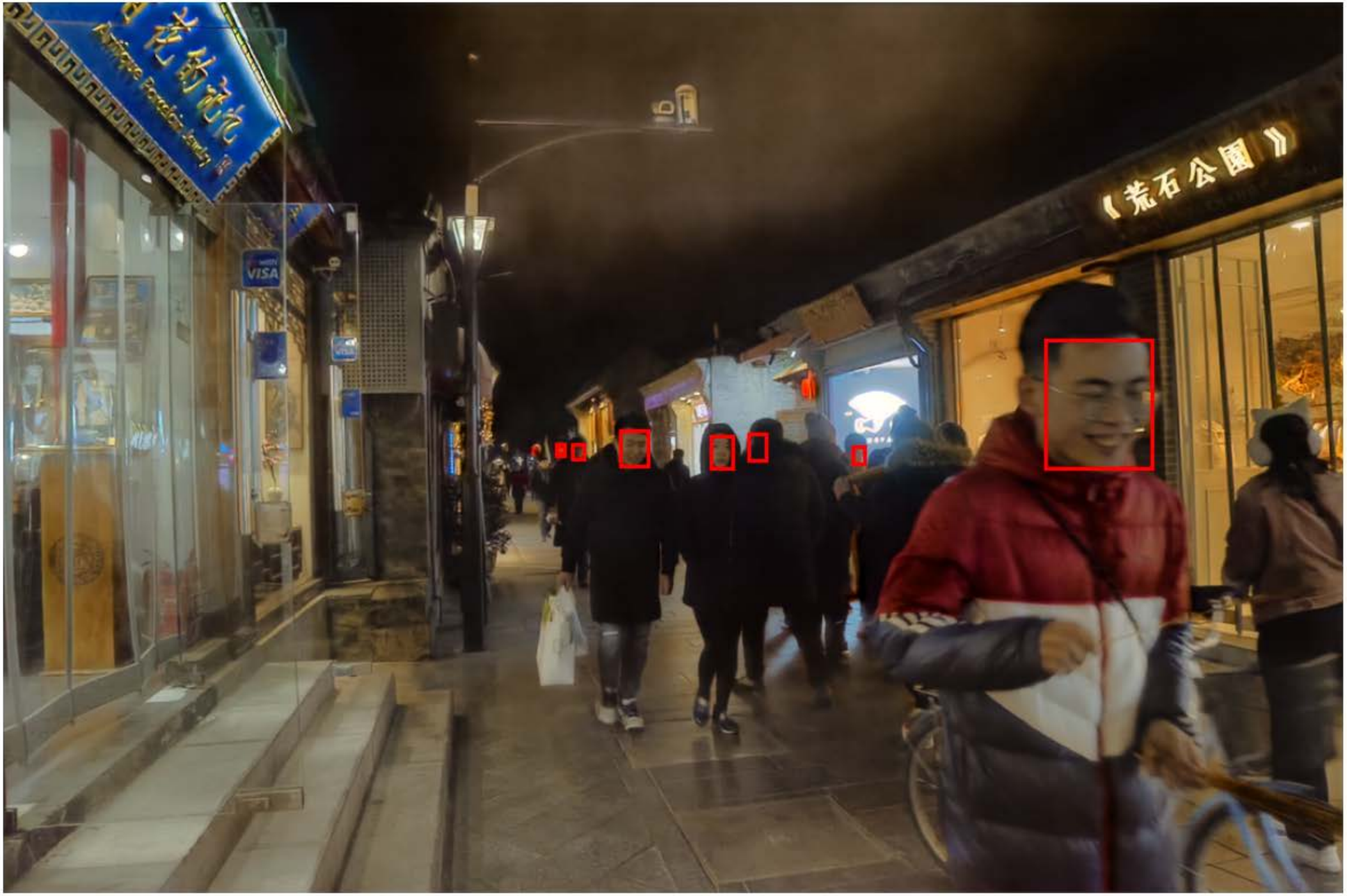}&
			\includegraphics[width=0.3\linewidth]{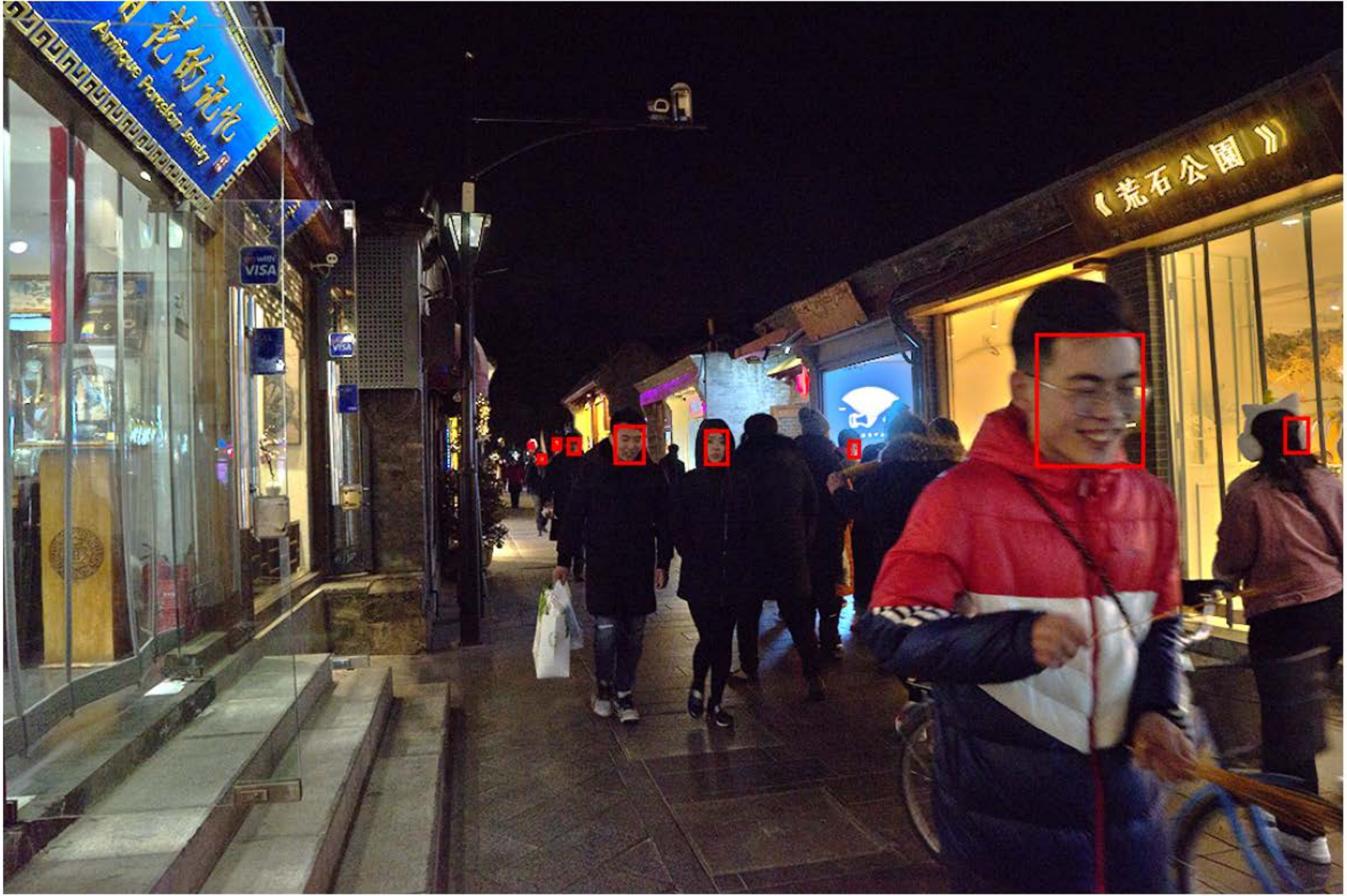}\\
			(j) EnlightenGAN \cite{EnlightenGAN} & (k) DRBN \cite{YangCVRP20} & (l) ExCNet \cite{ZhangACM191}\\
			\includegraphics[width=0.3\linewidth]{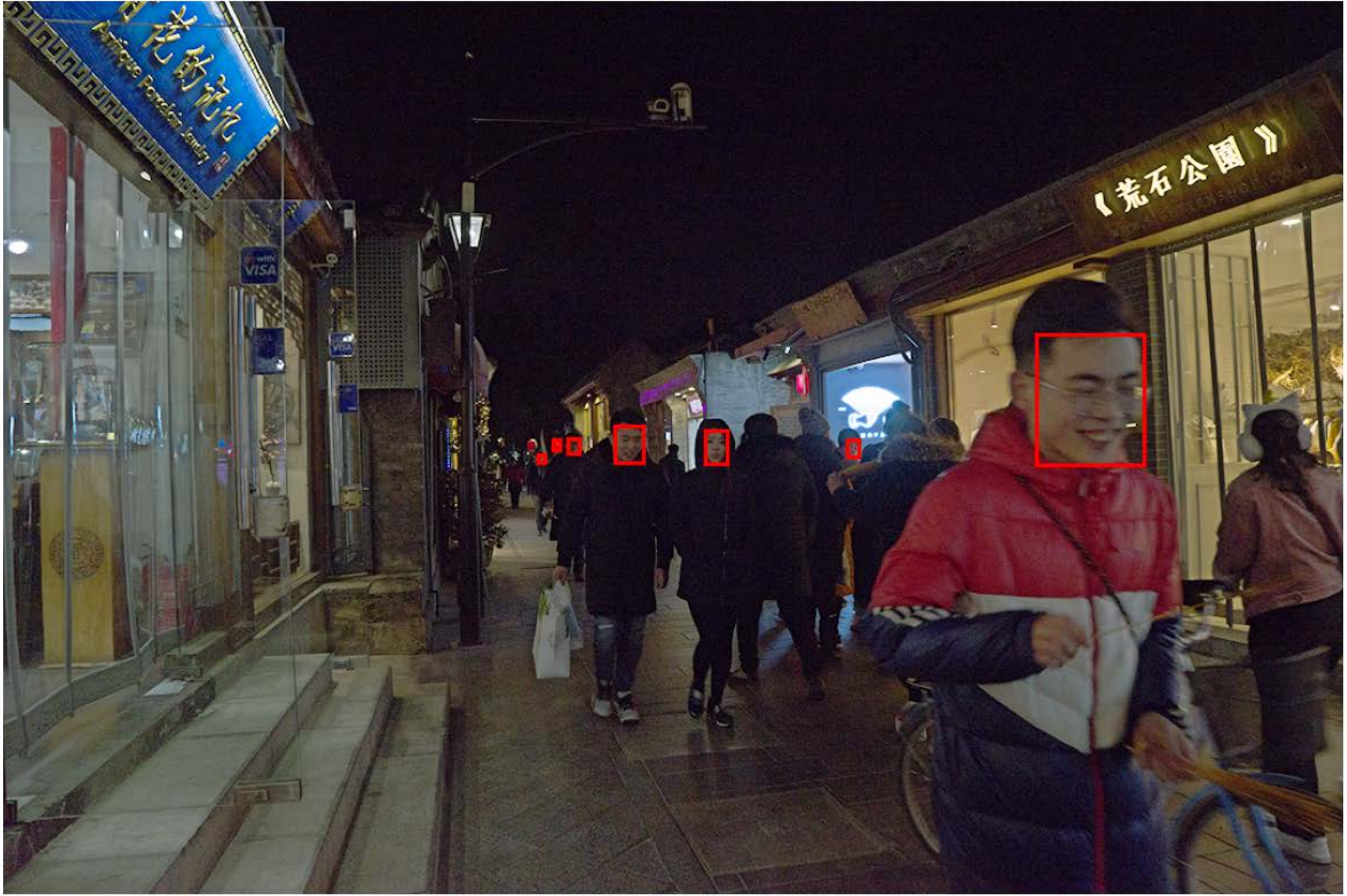}&
			\includegraphics[width=0.3\linewidth]{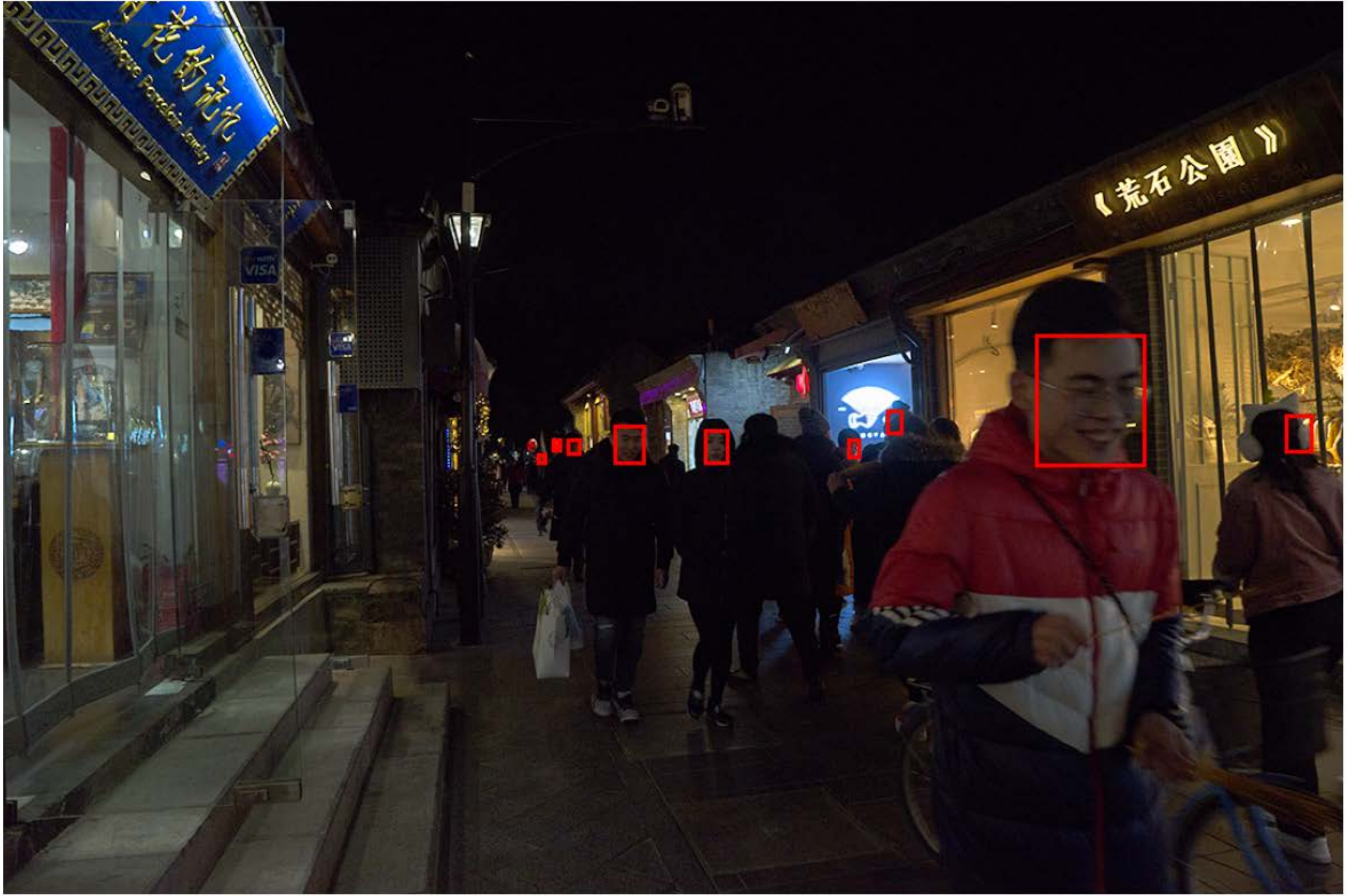}&
			~\\
			(m) Zero-DCE \cite{ZeroDCE} &  (n) 	RRDNet \cite{RRDNet}  &  \\
		\end{tabular}
	\end{center}
	\caption{Visual results of different methods on a low-light image sampled from  DARK FACE dataset  \cite{Yuan2019}. Better see with zoom in for the bounding boxes of faces.}
	\label{fig:face_visual1}
\end{figure*}

\begin{figure*} [h]
	\begin{center}
		\begin{tabular}{c@{ }c@{ }c@{ }}
			\includegraphics[width=0.3\linewidth]{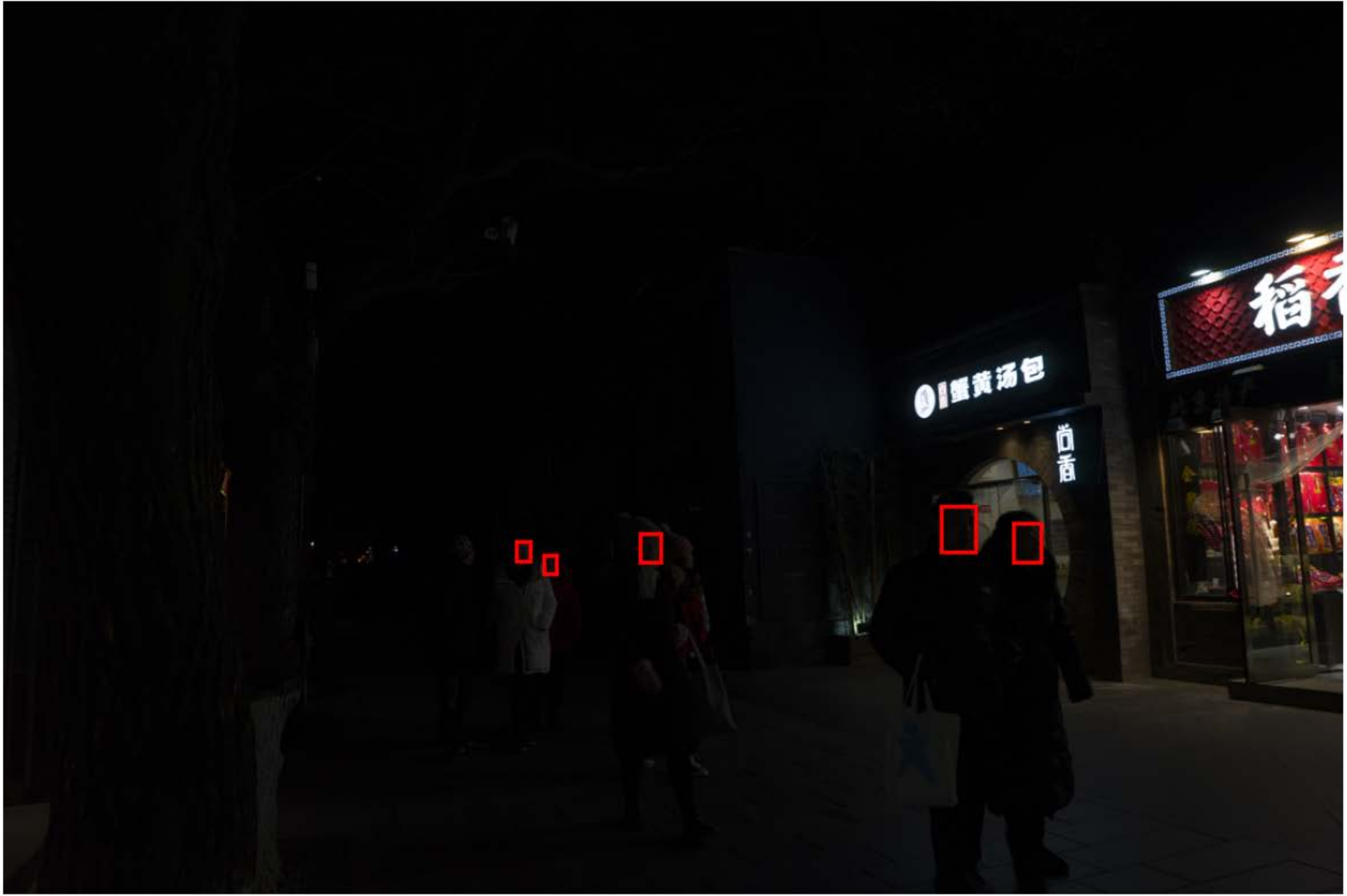}&
			\includegraphics[width=0.3\linewidth]{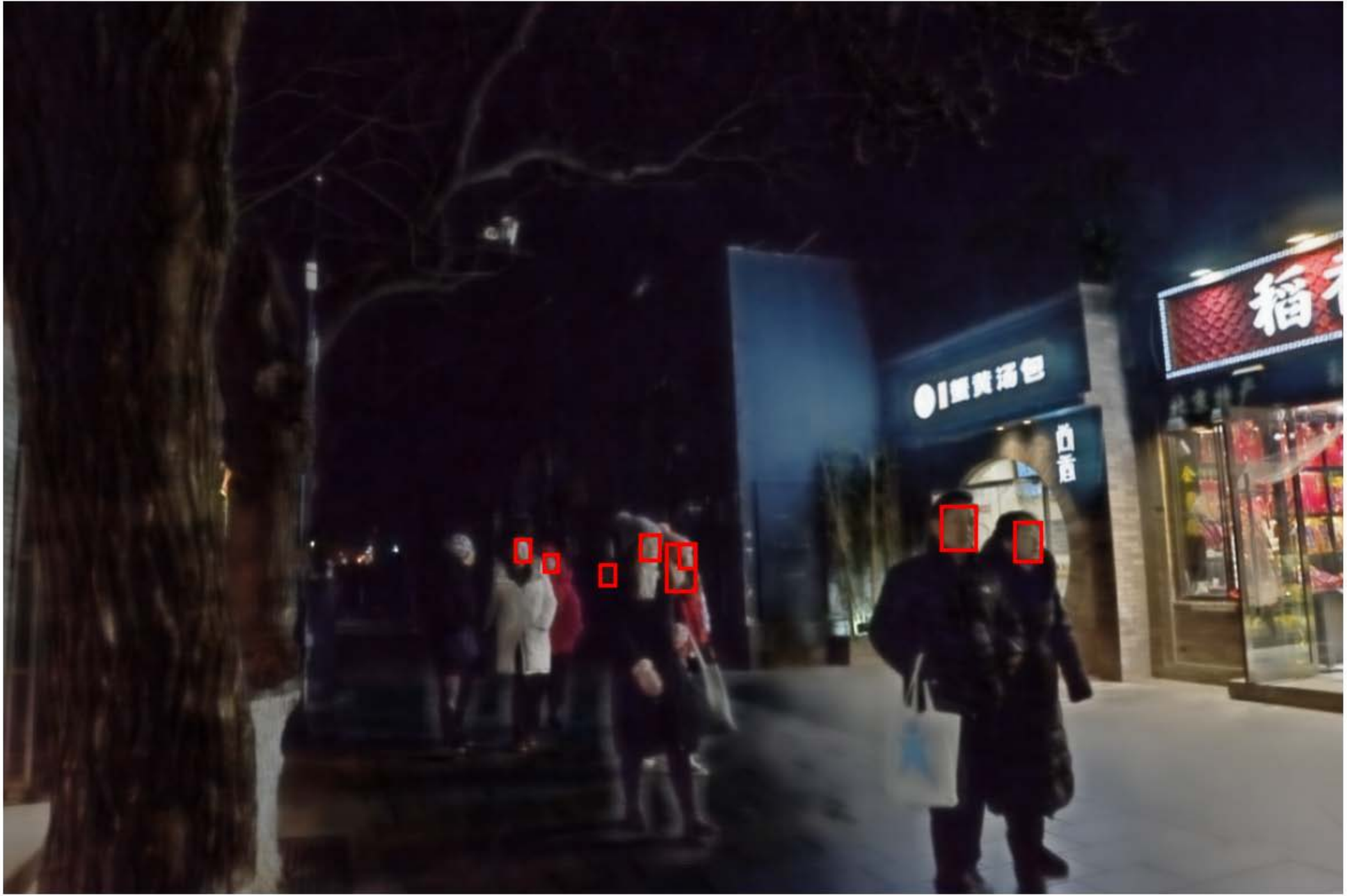}&
			\includegraphics[width=0.3\linewidth]{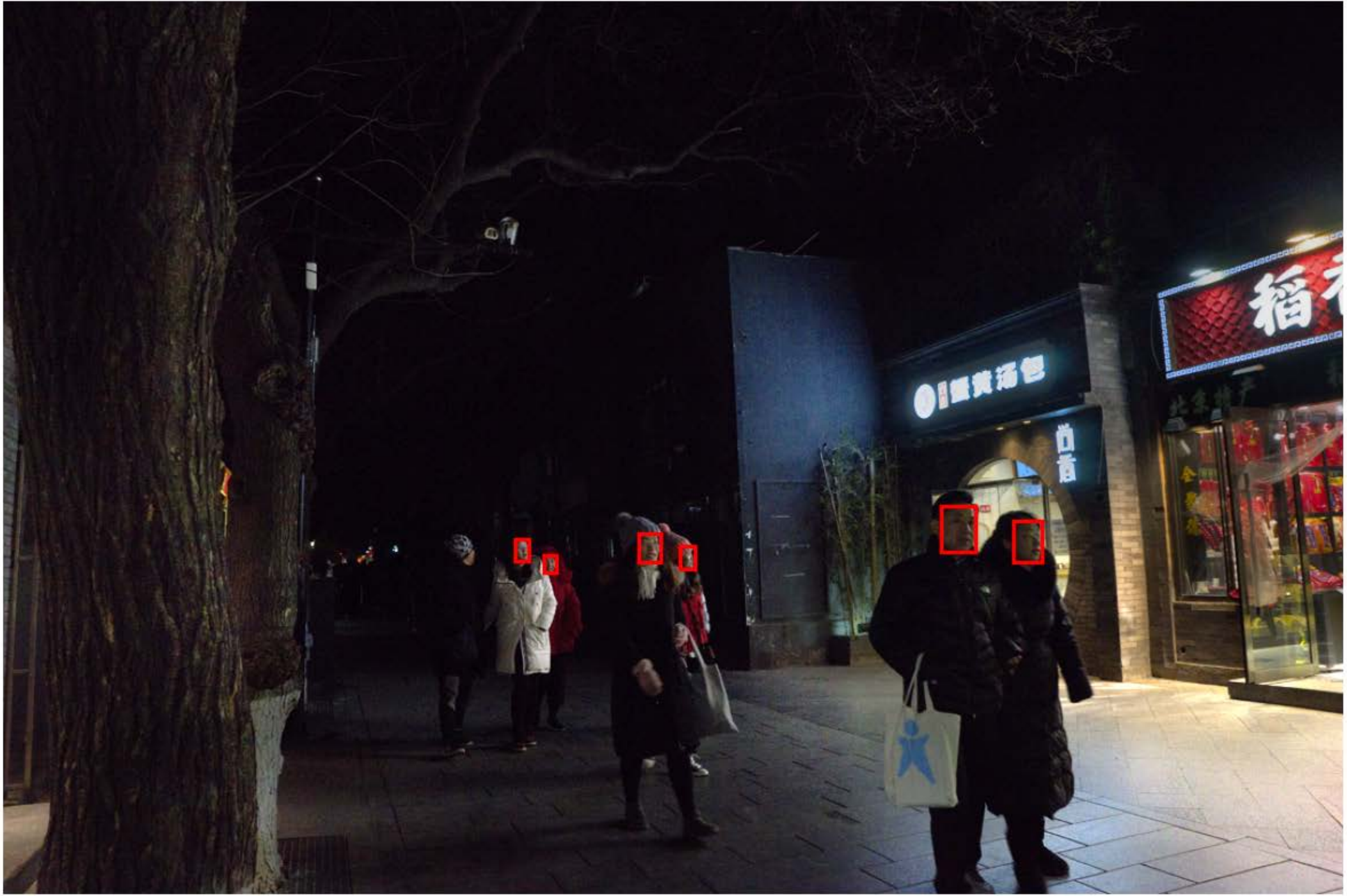}\\
			(a) input  & (b) LLNet \cite{LLNet}  &  (c) LightenNet \cite{LightenNet}\\
			\includegraphics[width=0.3\linewidth]{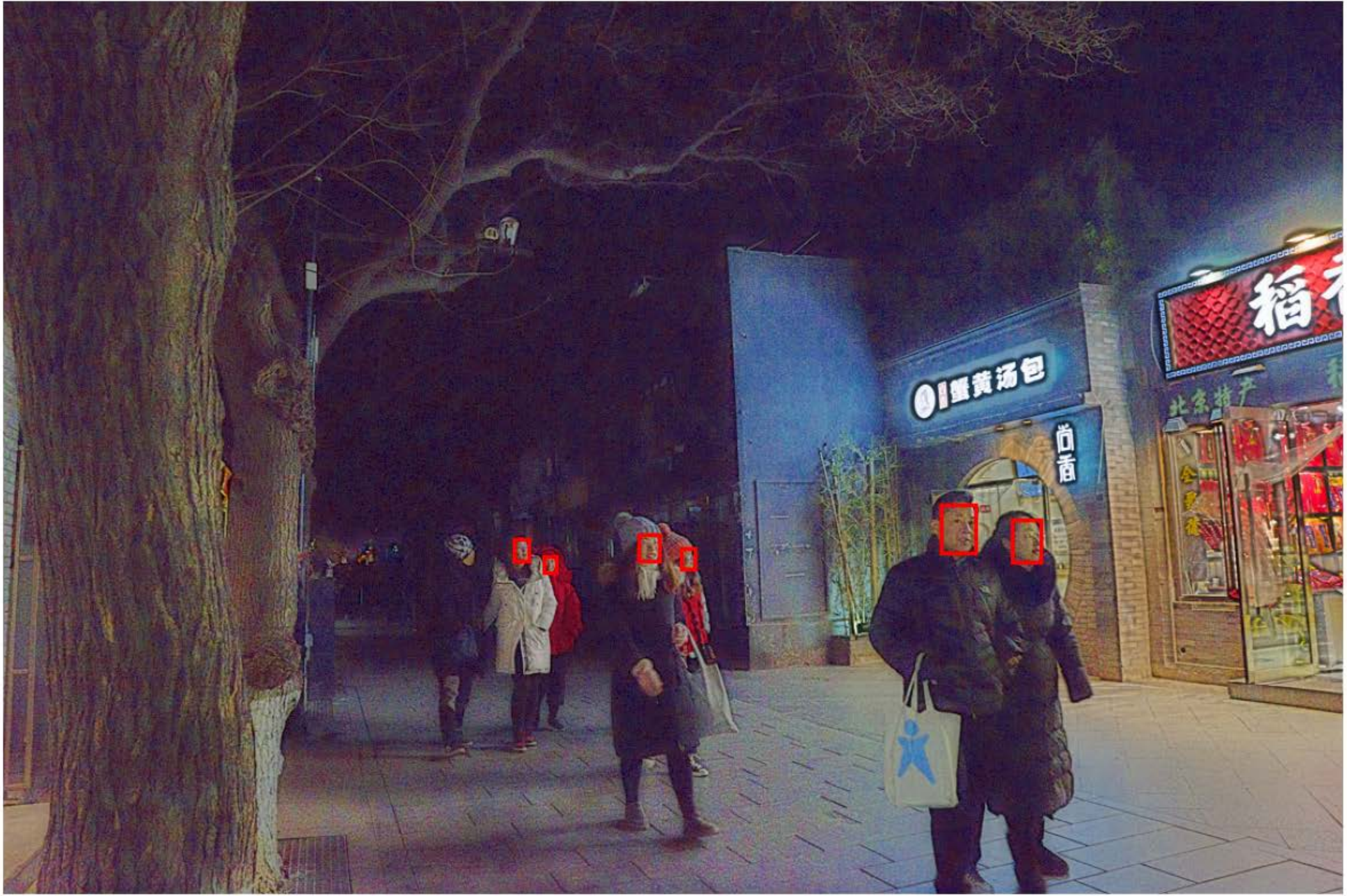}&
			\includegraphics[width=0.3\linewidth]{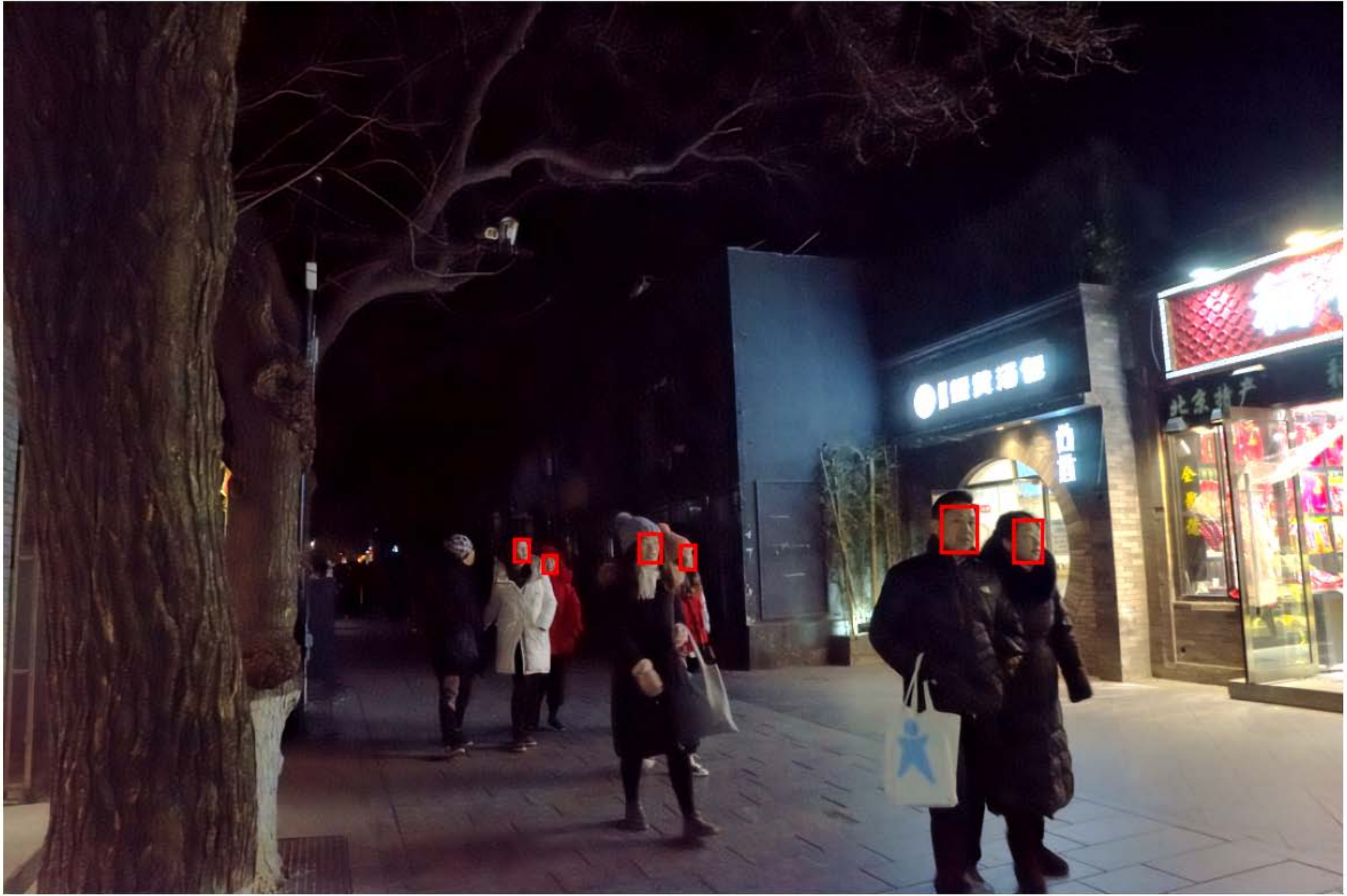}&
			\includegraphics[width=0.3\linewidth]{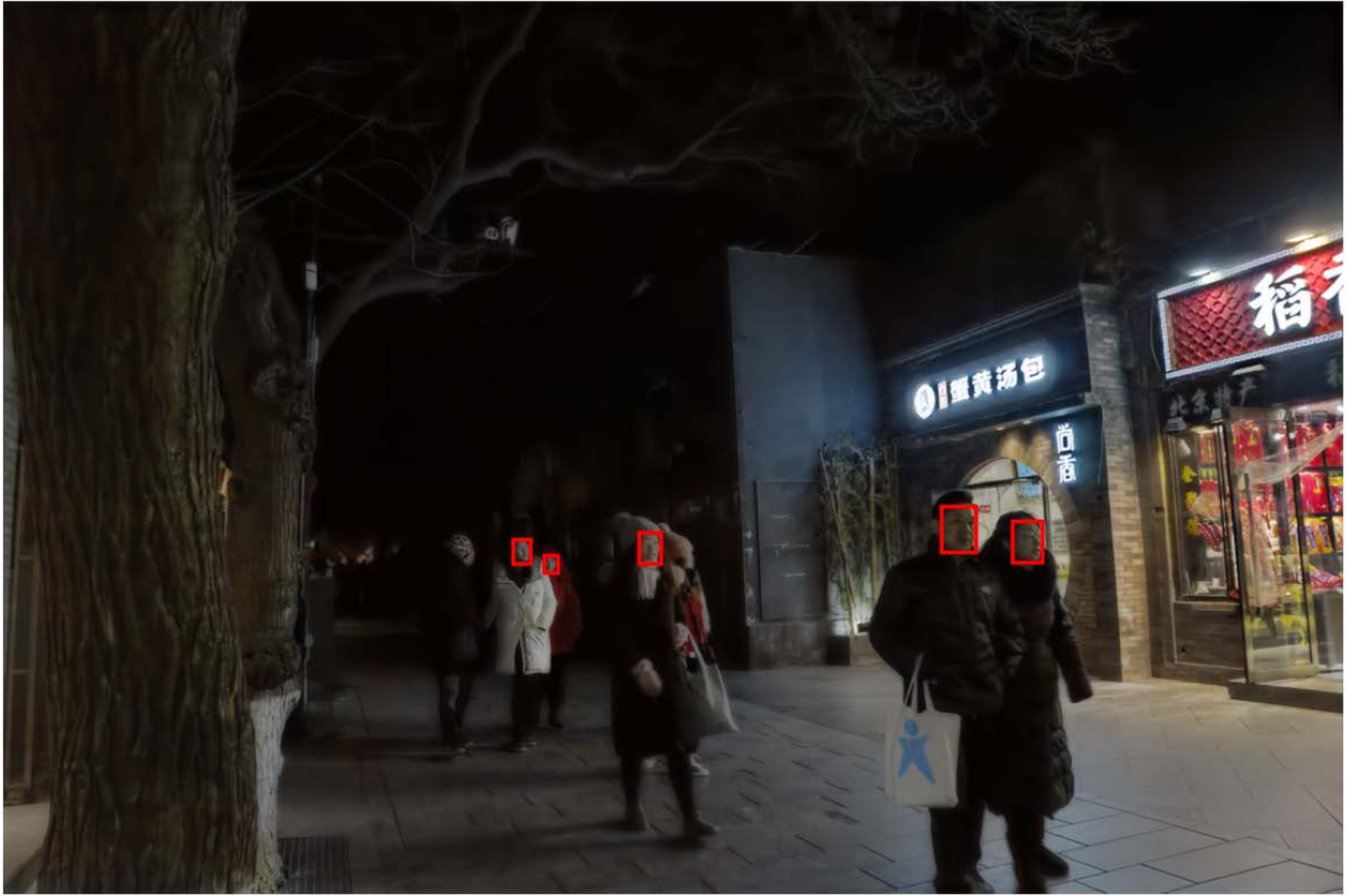}\\
			(d) Retinex-Net \cite{ChenBMVC18} & (e) MBLLEN \cite{LvBMVC2018} & (f) KinD \cite{ZhangACM19}\\
			\includegraphics[width=0.3\linewidth]{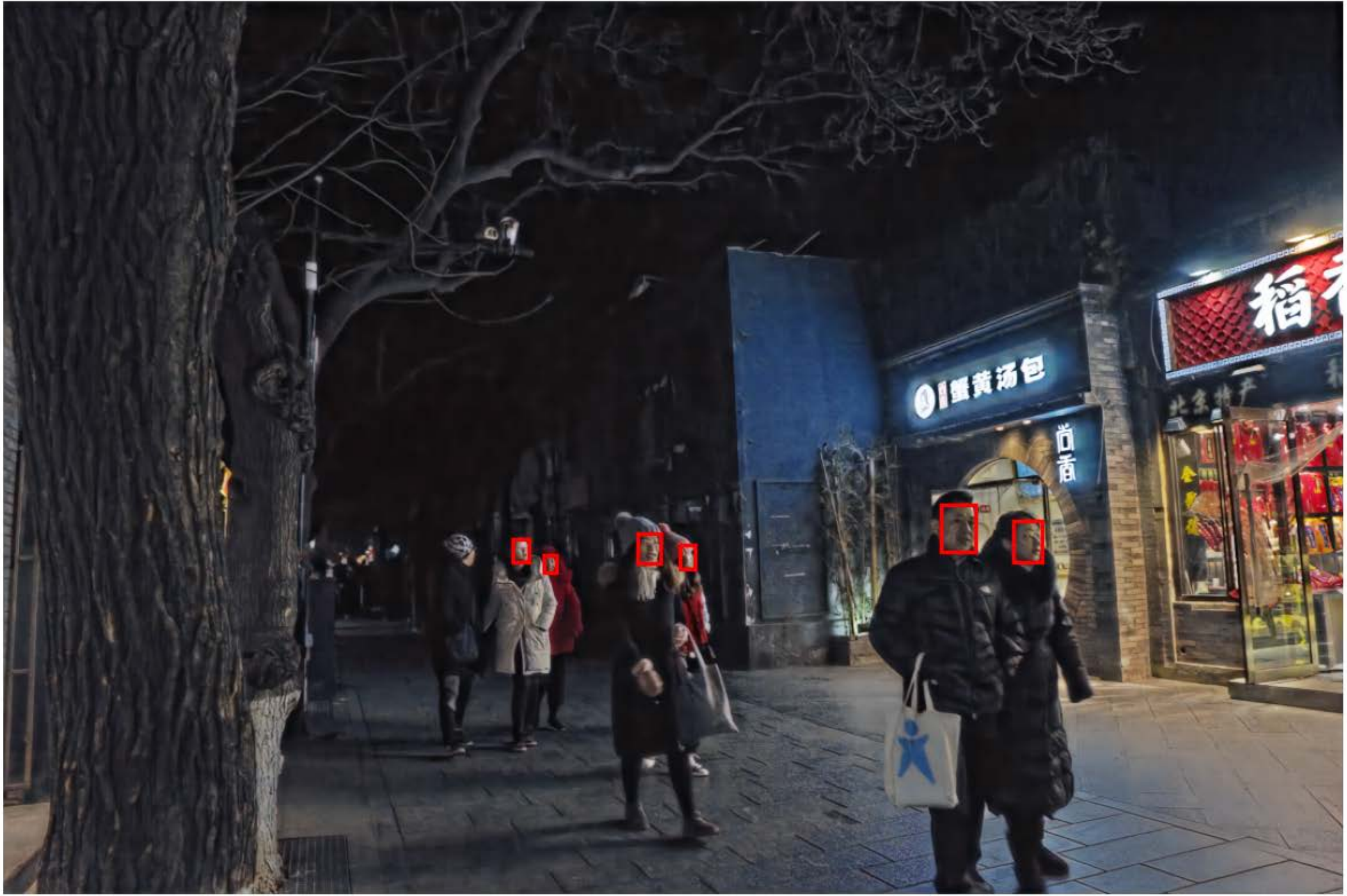}&
			\includegraphics[width=0.3\linewidth]{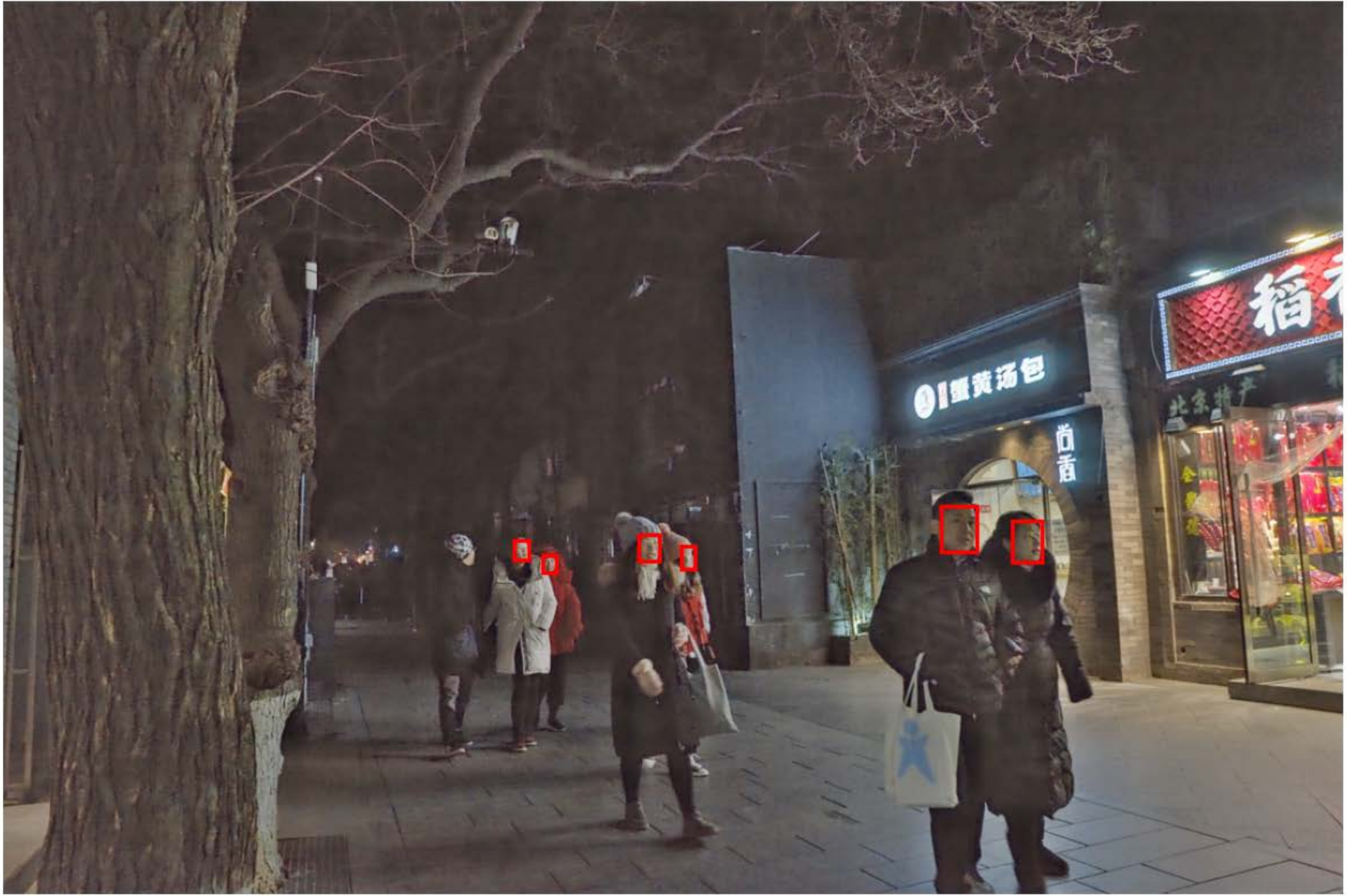}&
			\includegraphics[width=0.3\linewidth]{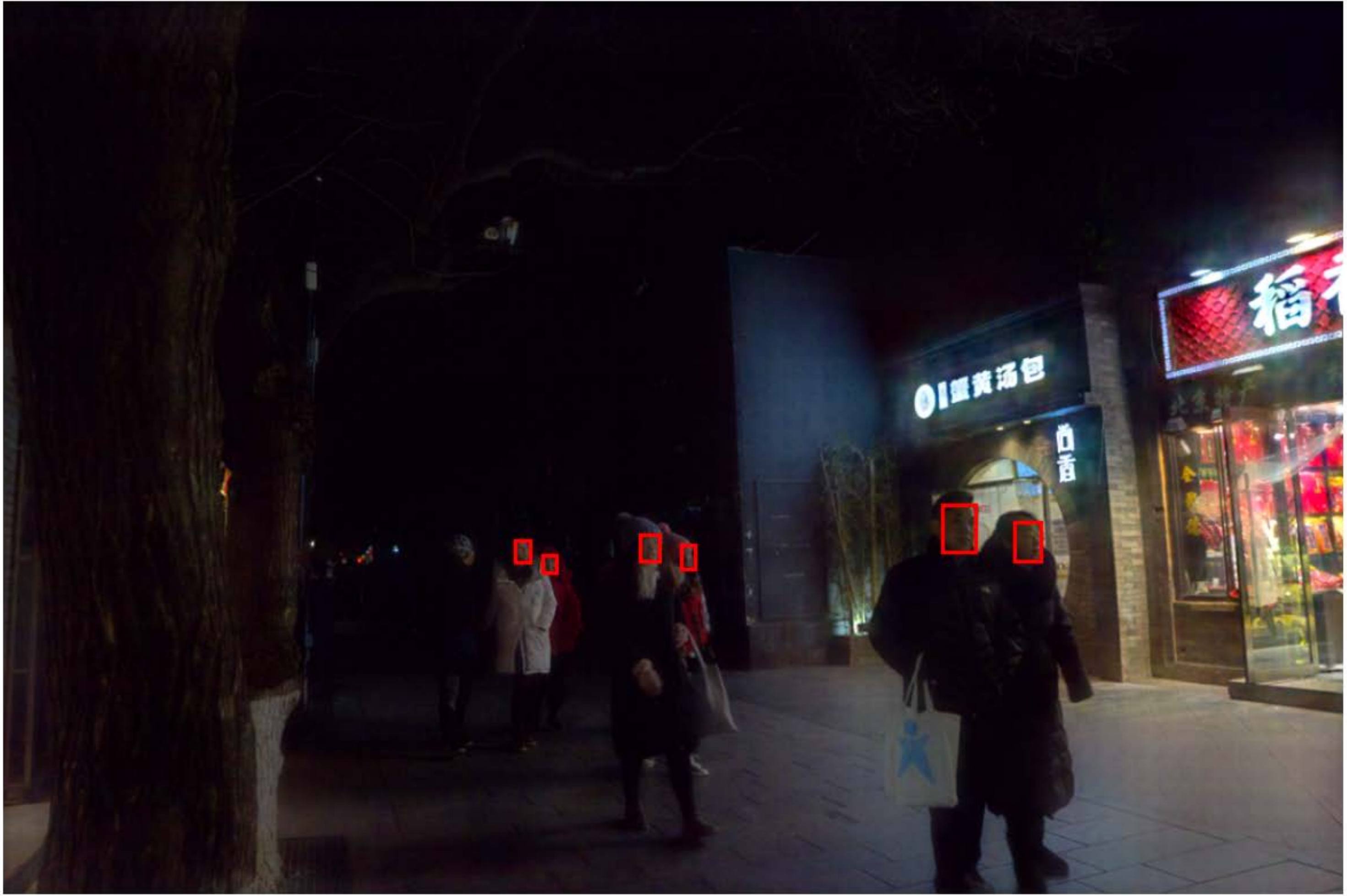}\\
			(g) KinD++ \cite{GuoIJCV2020} & (h) TBEFN \cite{TBEFN} &  (i)  	DSLR \cite{DSLR}\\
			\includegraphics[width=0.3\linewidth]{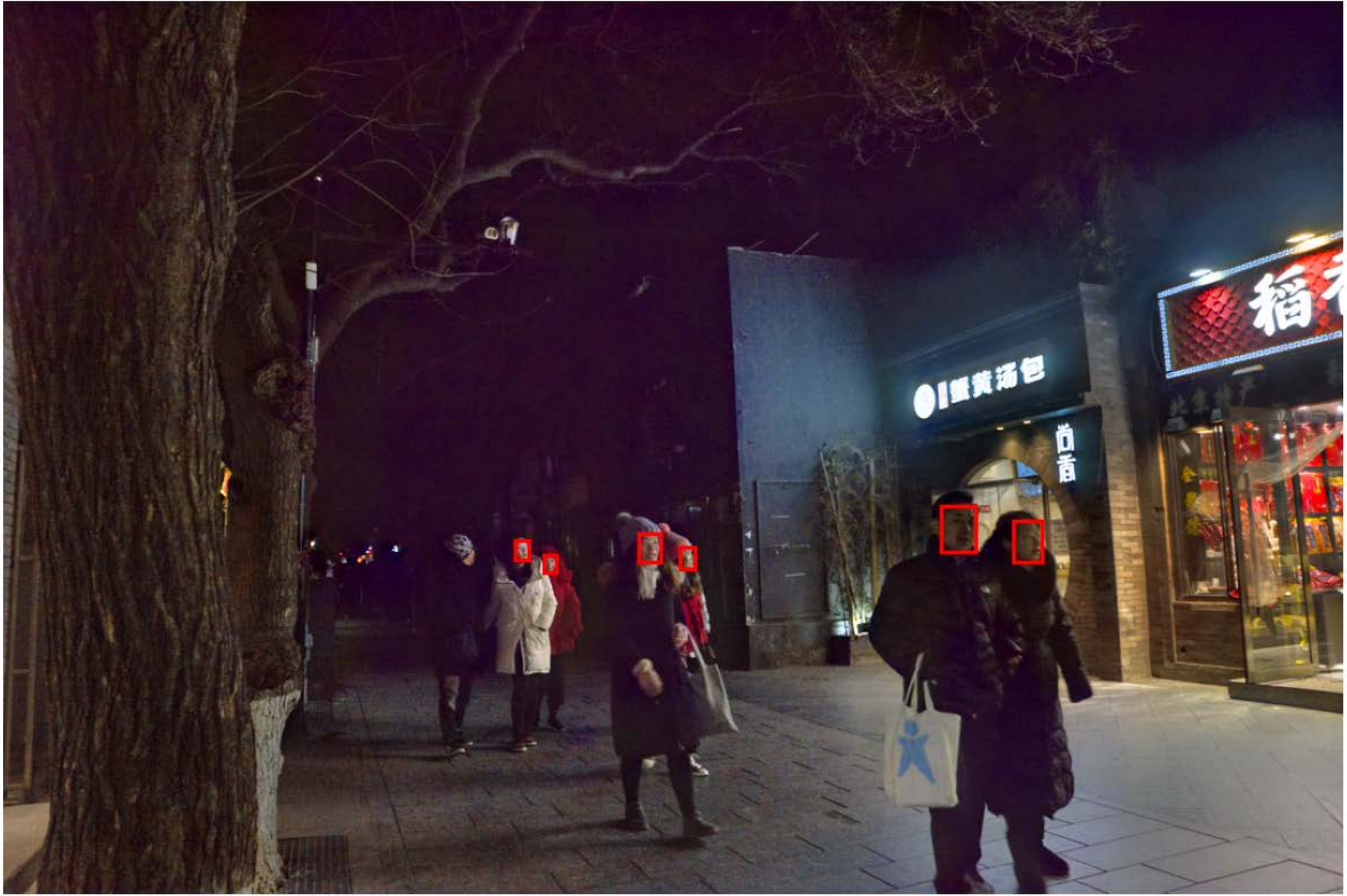}&
			\includegraphics[width=0.3\linewidth]{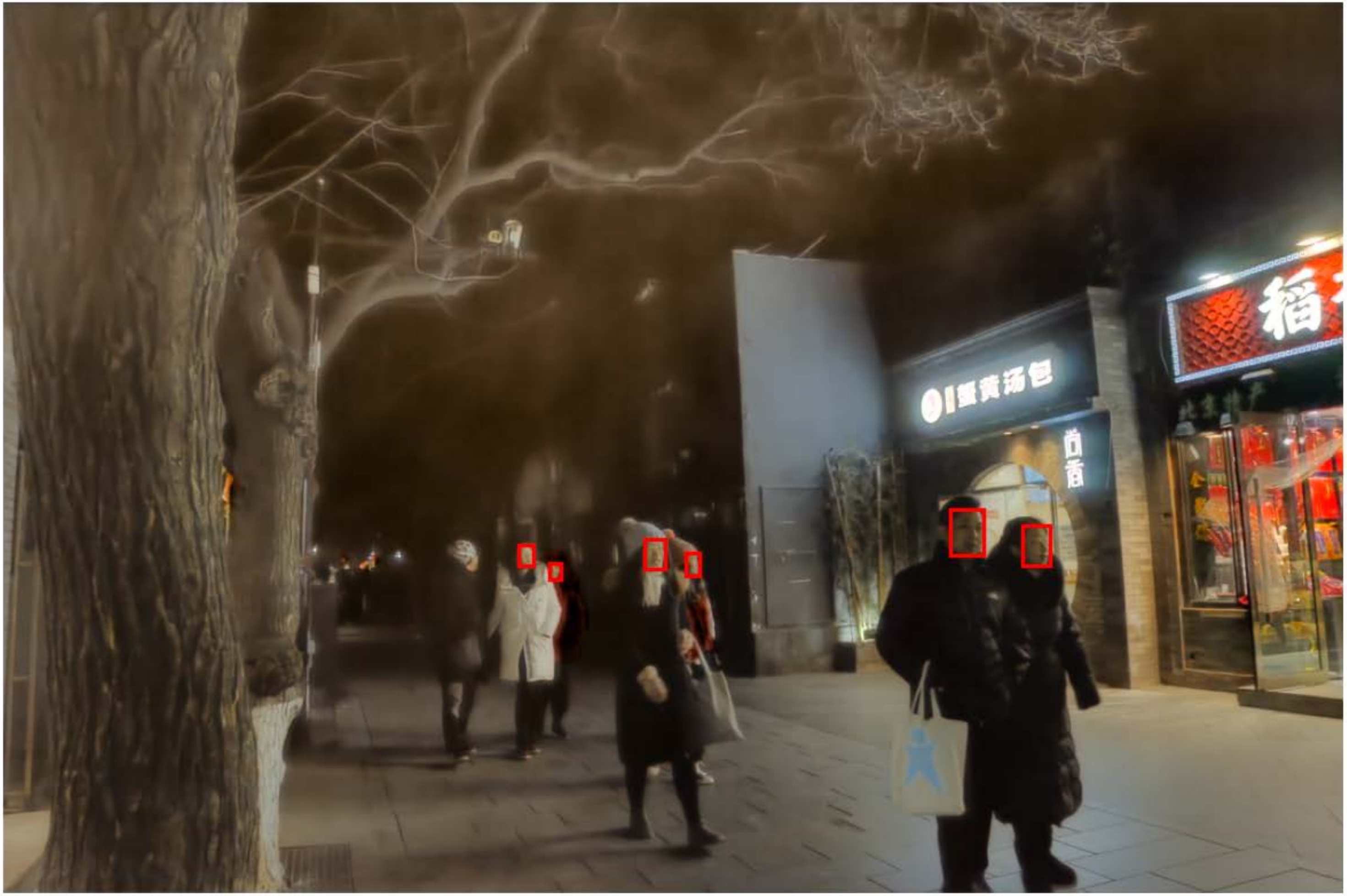}&
			\includegraphics[width=0.3\linewidth]{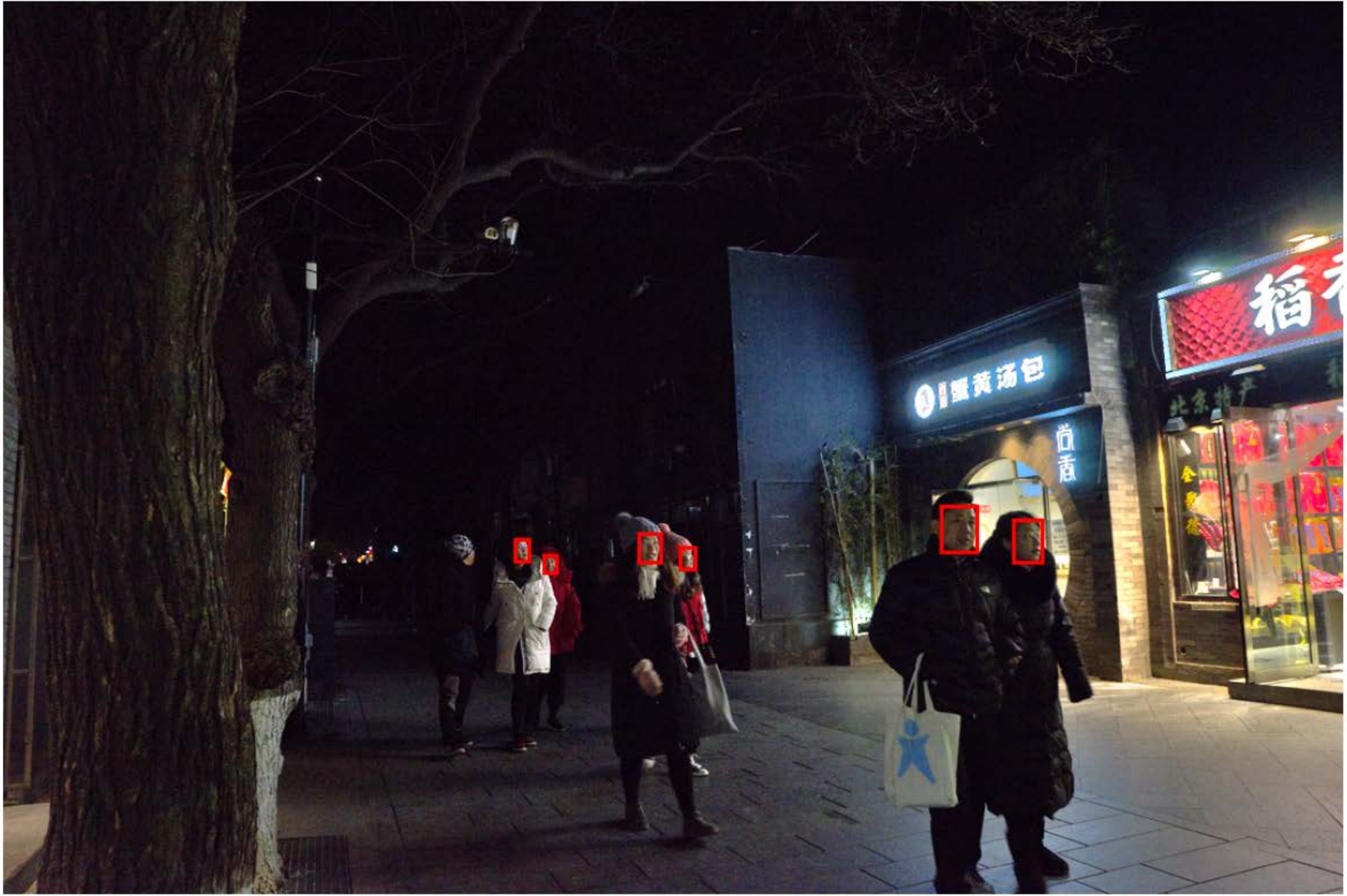}\\
			(j) EnlightenGAN \cite{EnlightenGAN} & (k) DRBN \cite{YangCVRP20} & (l) ExCNet \cite{ZhangACM191}\\
			\includegraphics[width=0.3\linewidth]{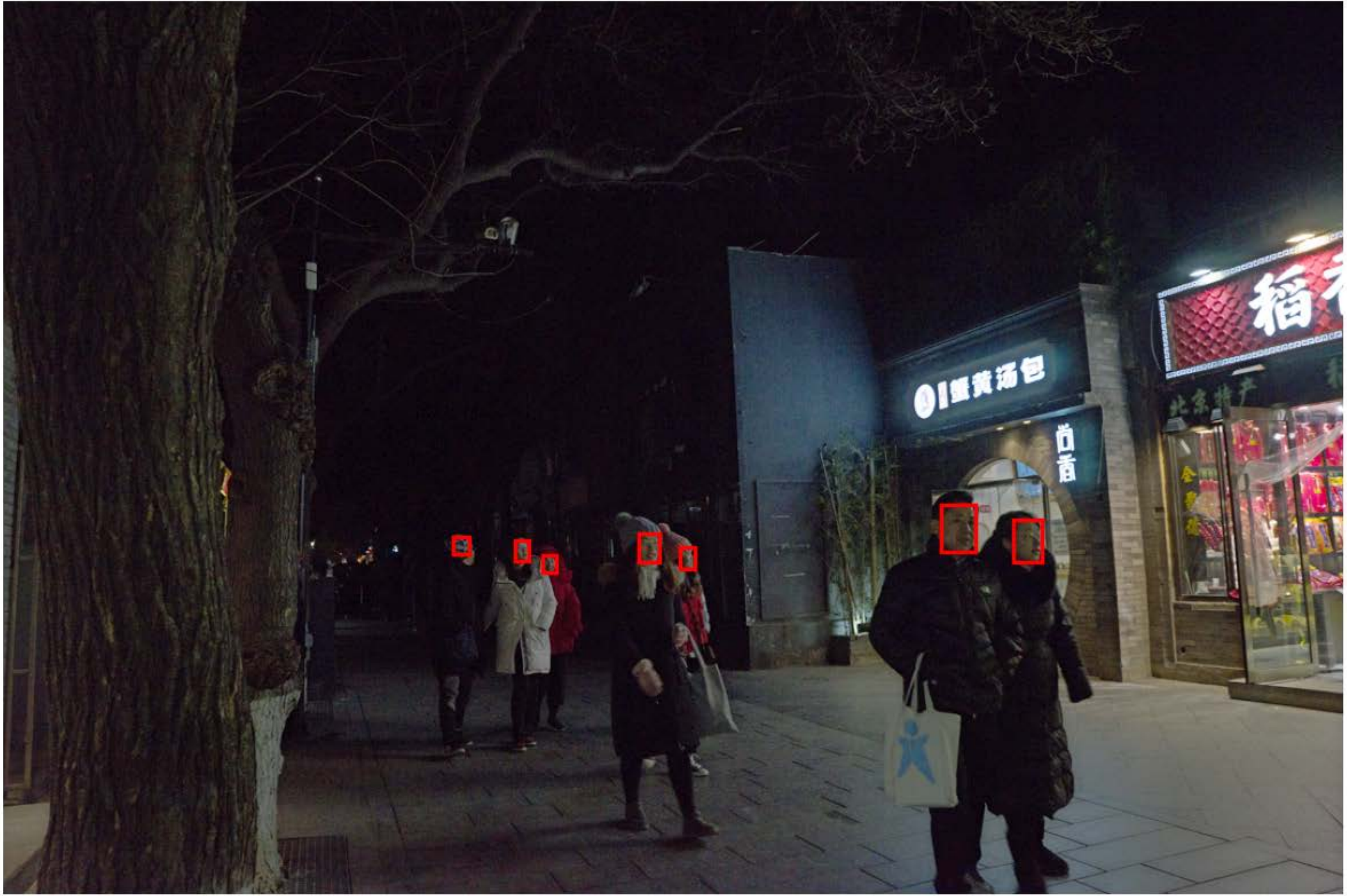}&
			\includegraphics[width=0.3\linewidth]{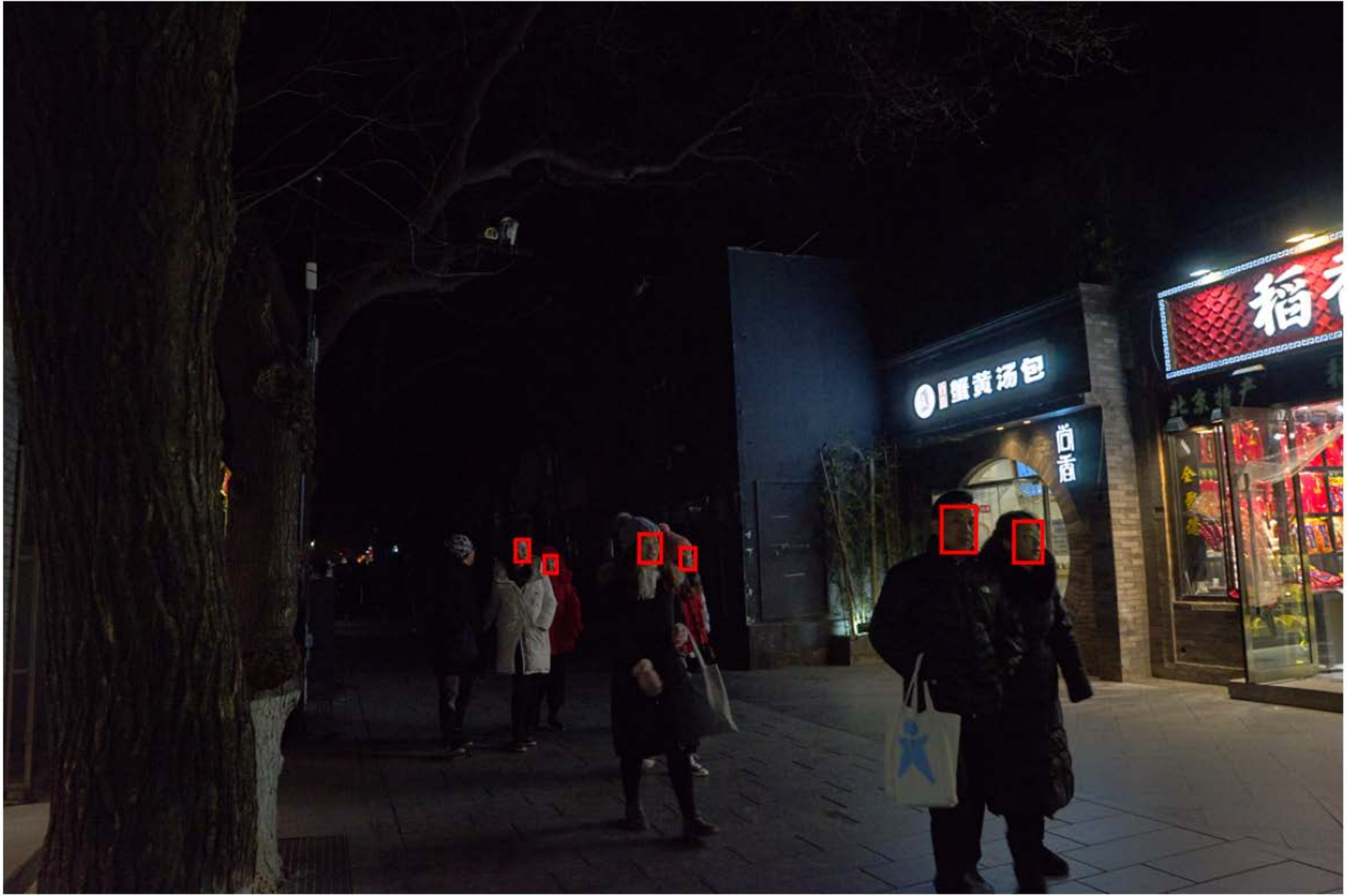}&
			~\\
			(m) Zero-DCE \cite{ZeroDCE} &  (n) 	RRDNet \cite{RRDNet}  &  \\
		\end{tabular}
	\end{center}
	\caption{Visual results of different methods on a low-light image sampled from  DARK FACE dataset  \cite{Yuan2019}. Better see with zoom in for the bounding boxes of faces.}
	\label{fig:face_visual2}
\end{figure*}

\begin{figure*} [h]
	\begin{center}
		\begin{tabular}{c@{ }c@{ }c@{ }}
			\includegraphics[width=0.3\linewidth]{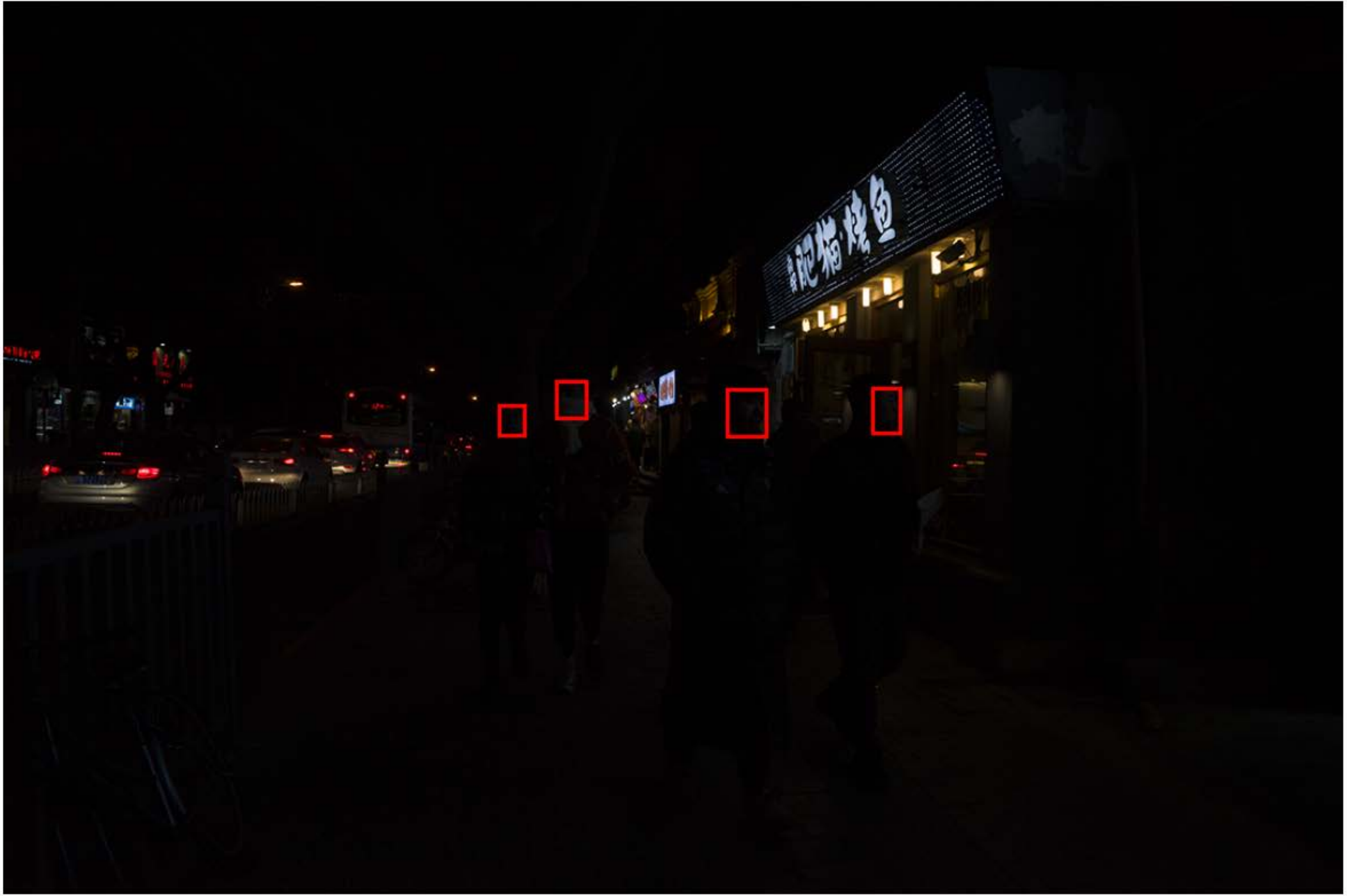}&
			\includegraphics[width=0.3\linewidth]{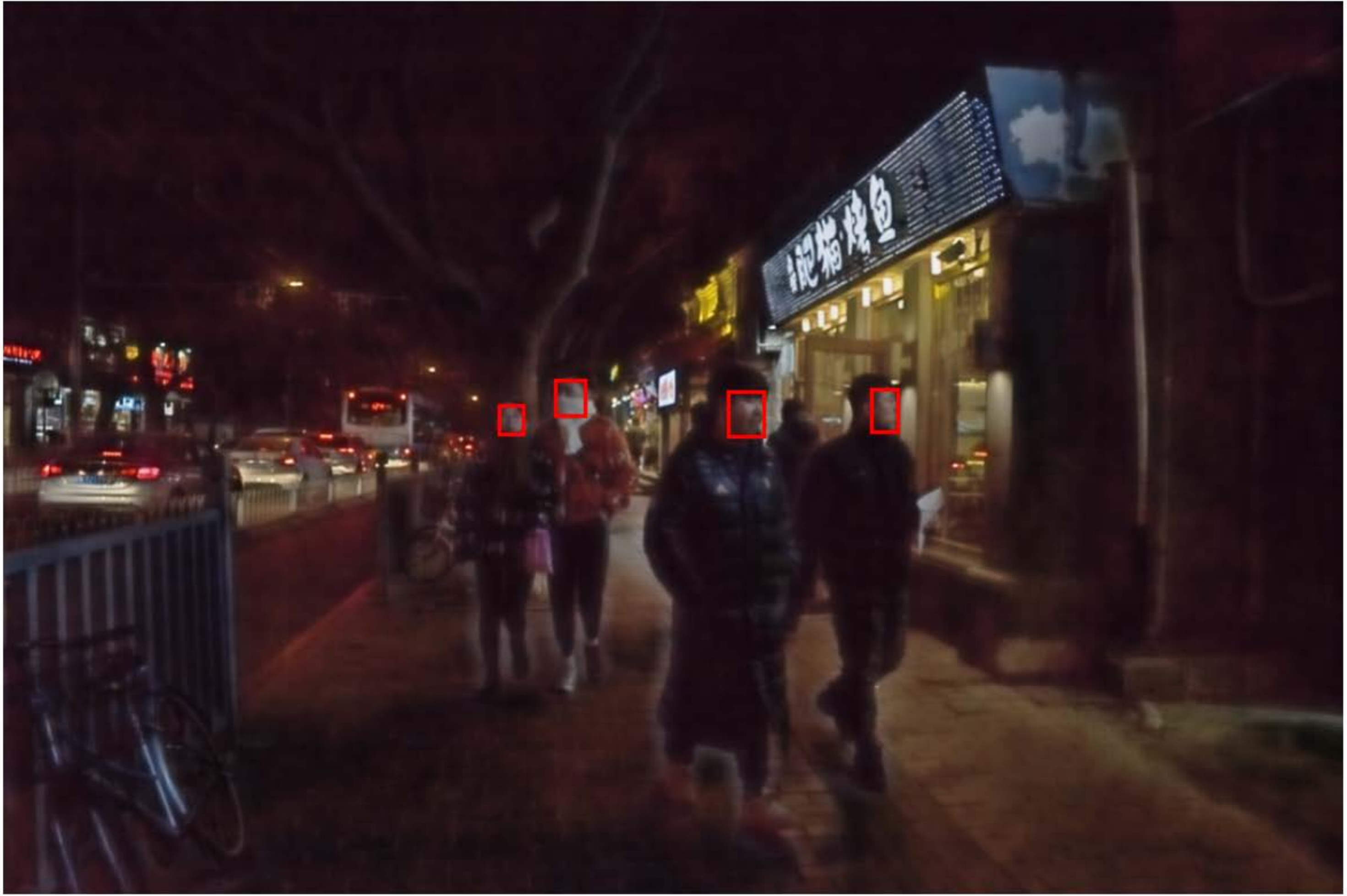}&
			\includegraphics[width=0.3\linewidth]{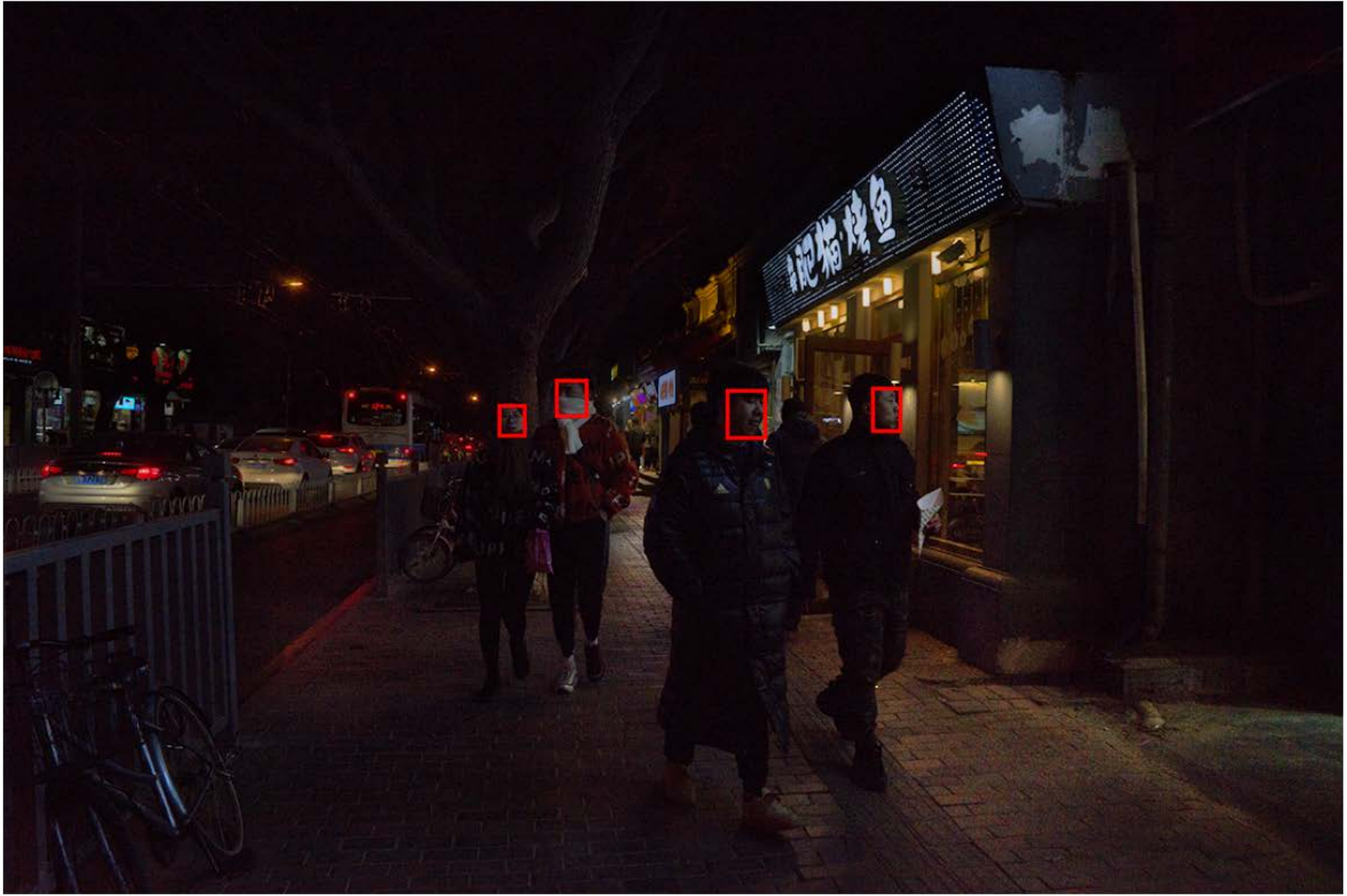}\\
			(a) input  & (b) LLNet \cite{LLNet}  &  (c) LightenNet \cite{LightenNet}\\
			\includegraphics[width=0.3\linewidth]{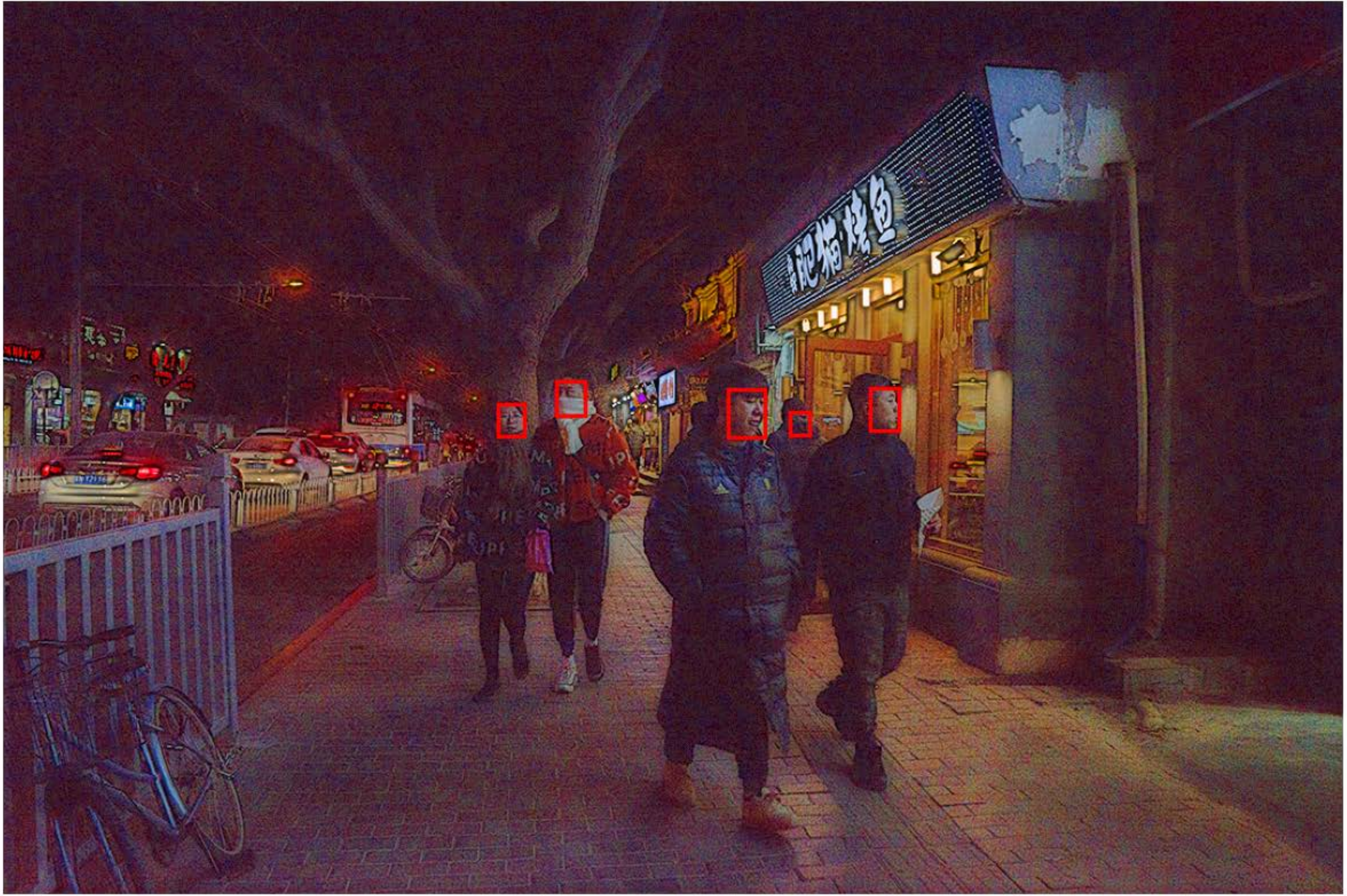}&
			\includegraphics[width=0.3\linewidth]{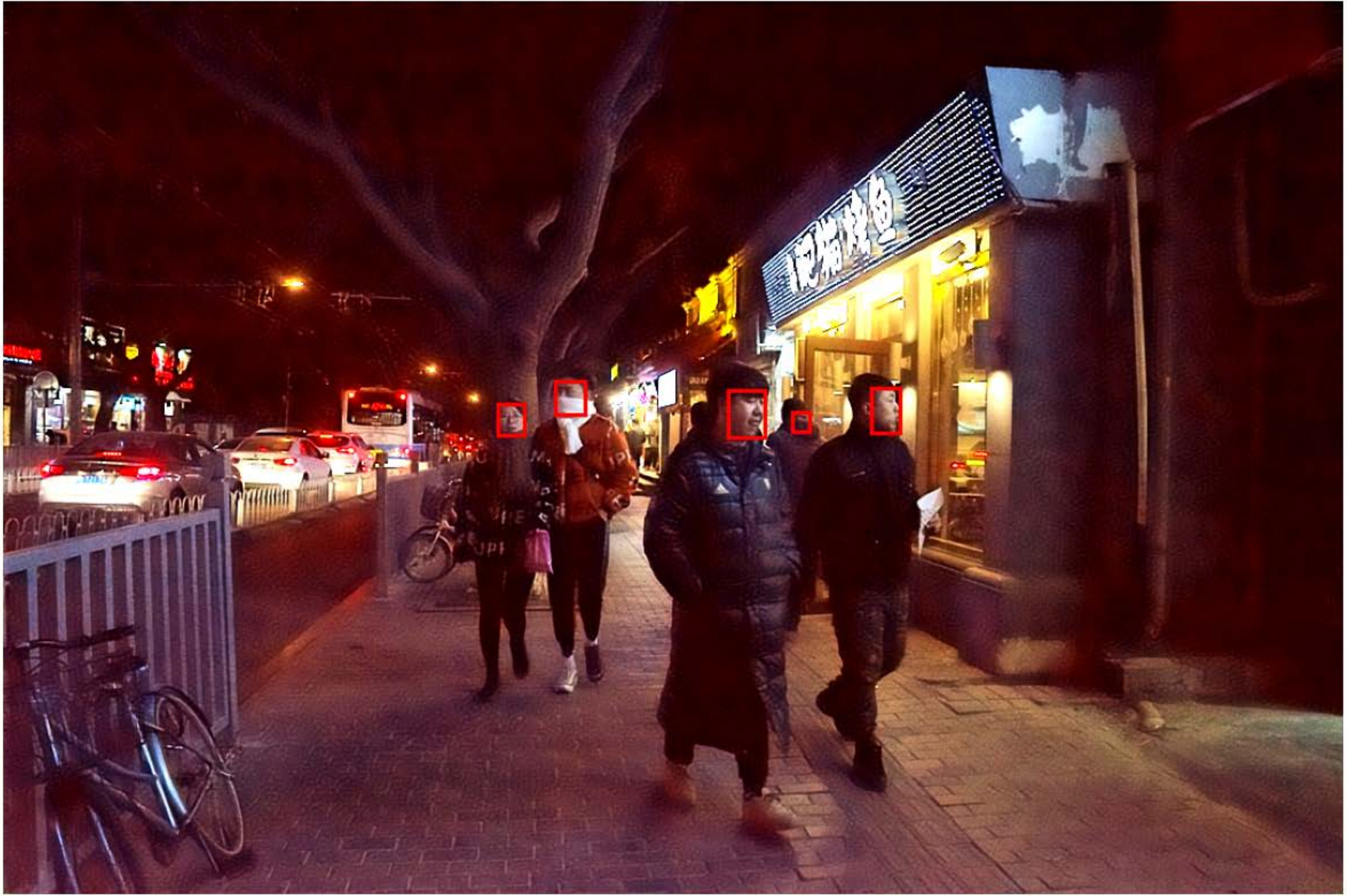}&
			\includegraphics[width=0.3\linewidth]{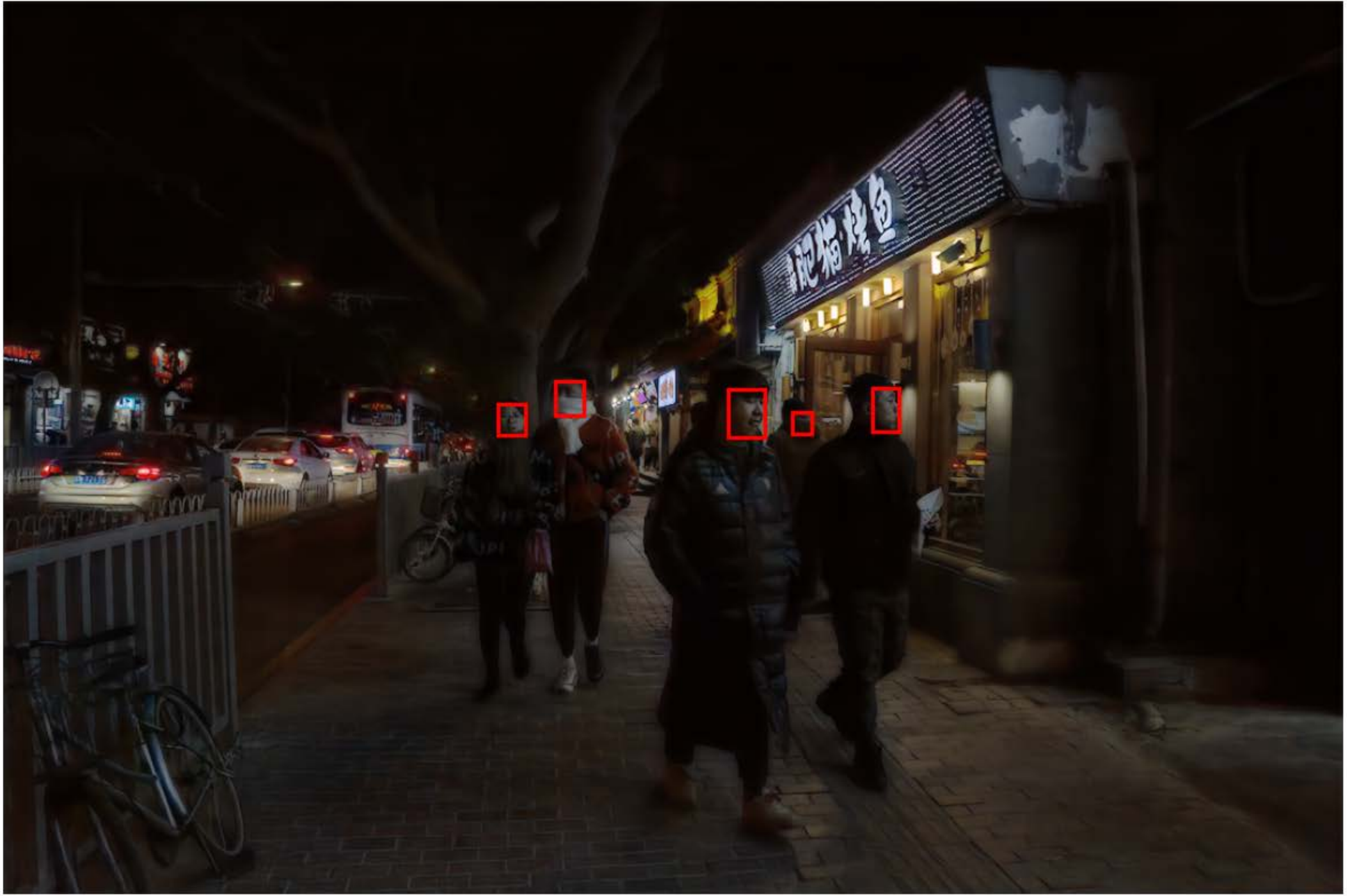}\\
			(d) Retinex-Net \cite{ChenBMVC18} & (e) MBLLEN \cite{LvBMVC2018} & (f) KinD \cite{ZhangACM19}\\
			\includegraphics[width=0.3\linewidth]{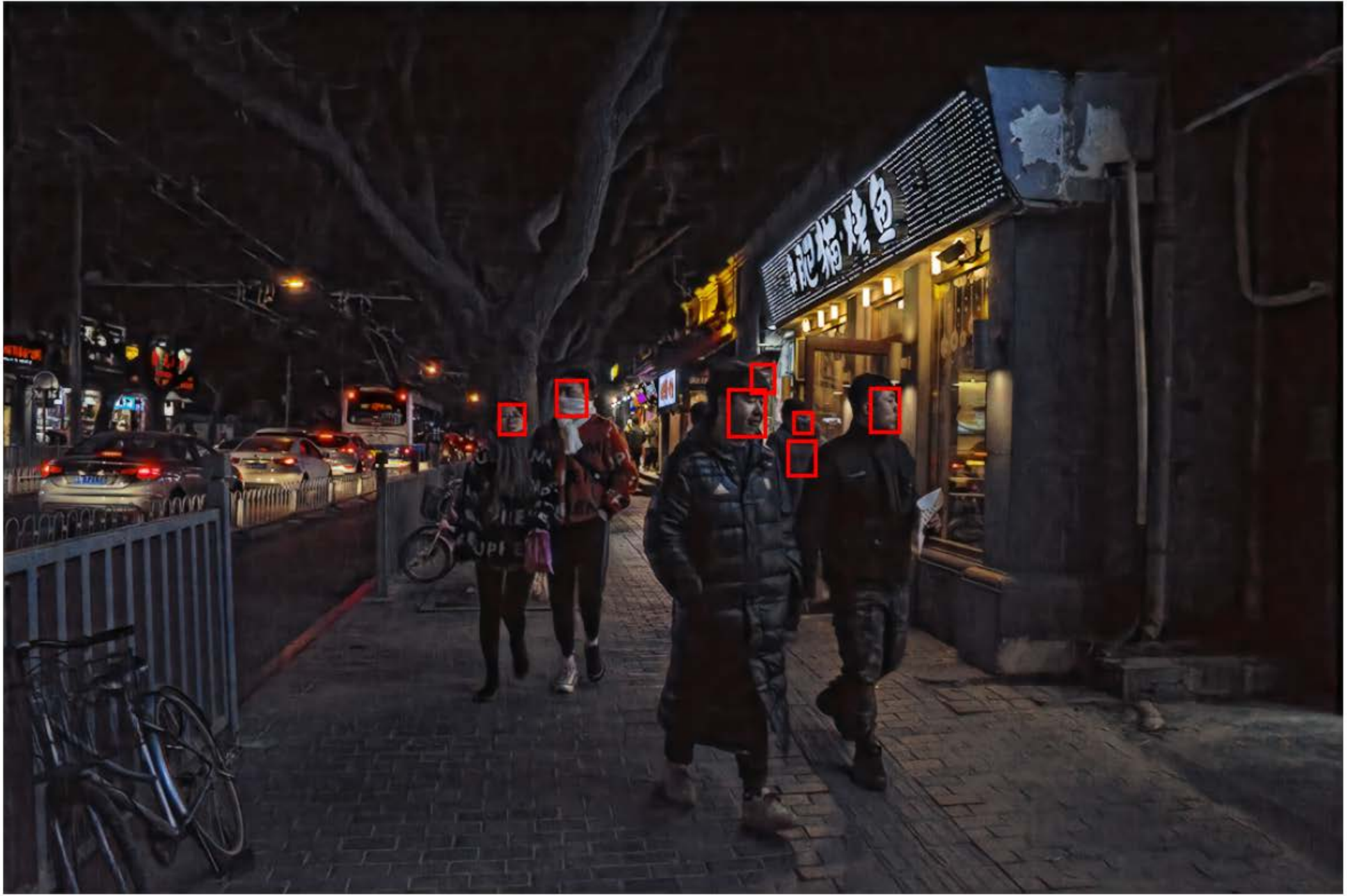}&
			\includegraphics[width=0.3\linewidth]{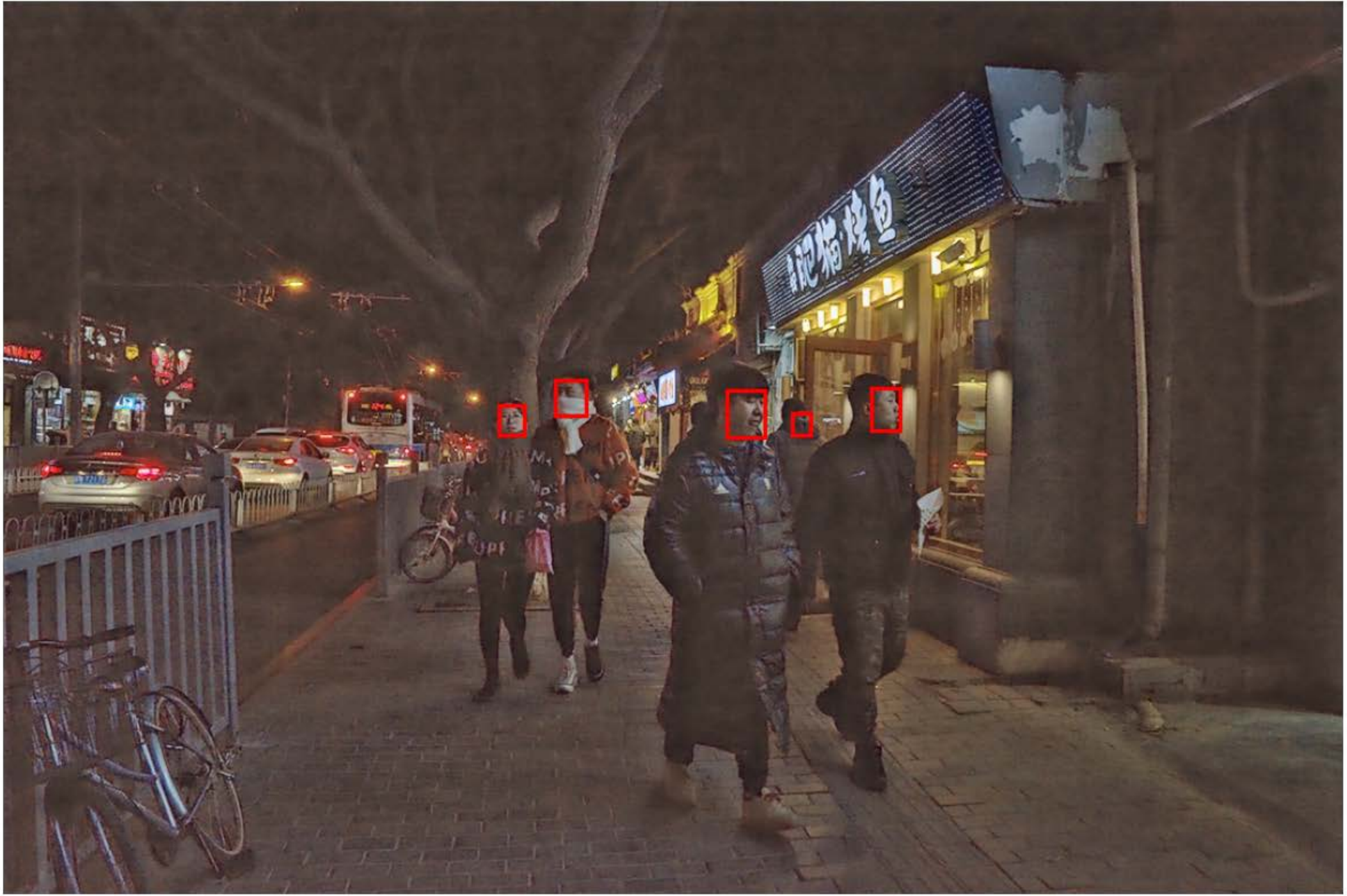}&
			\includegraphics[width=0.3\linewidth]{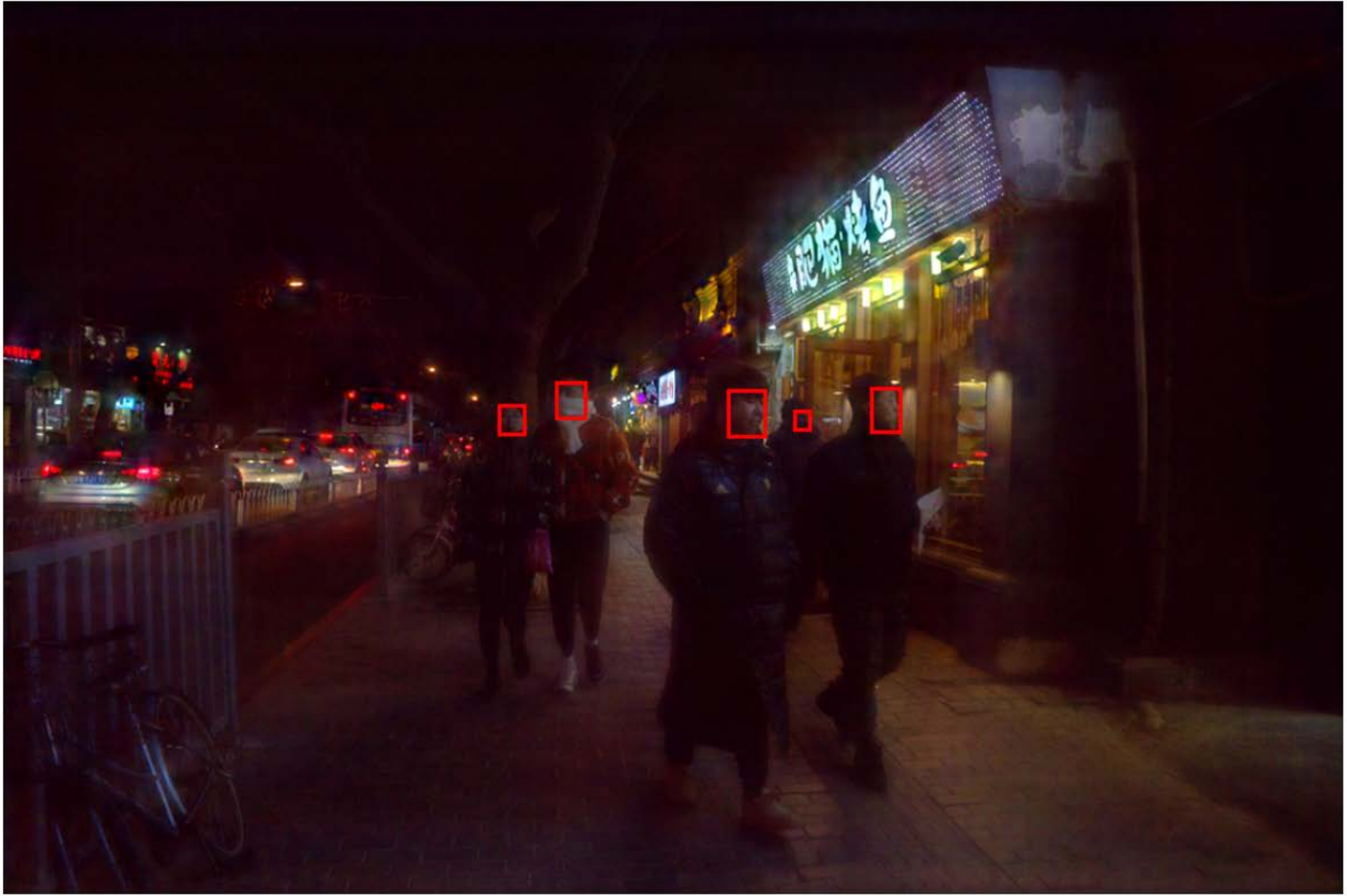}\\
			(g) KinD++ \cite{GuoIJCV2020} & (h) TBEFN \cite{TBEFN} &  (i)  	DSLR \cite{DSLR}\\
			\includegraphics[width=0.3\linewidth]{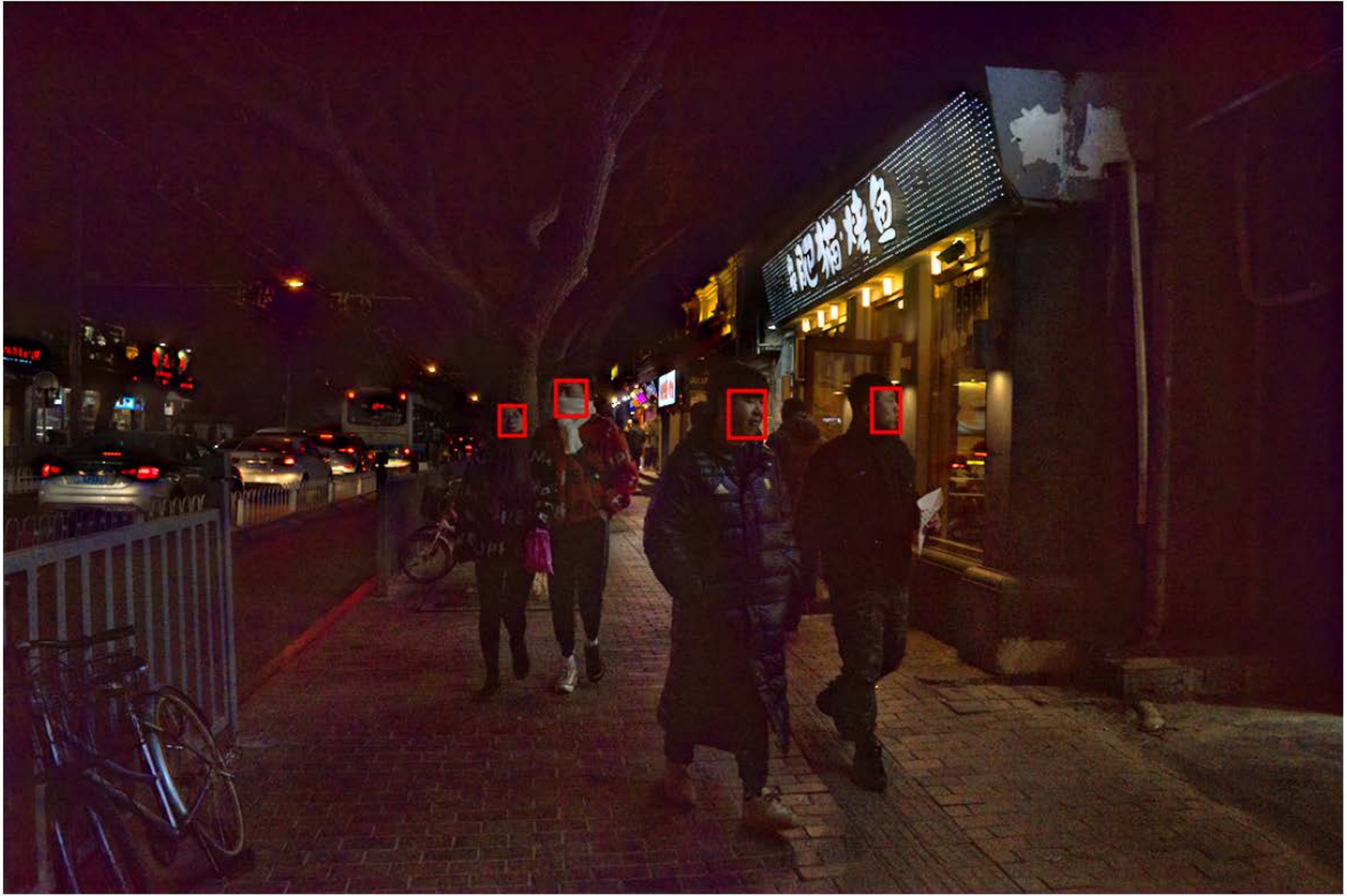}&
			\includegraphics[width=0.3\linewidth]{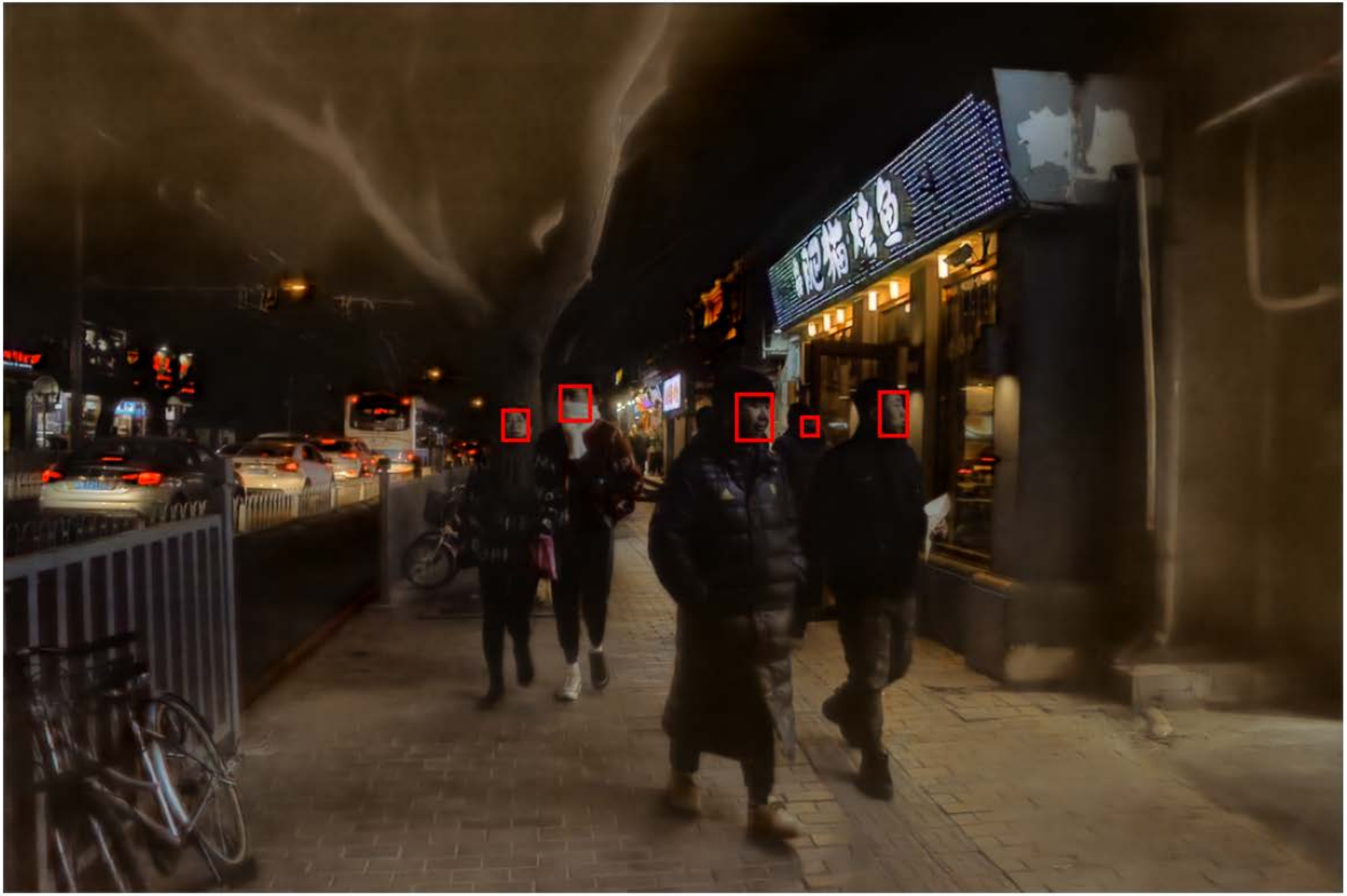}&
			\includegraphics[width=0.3\linewidth]{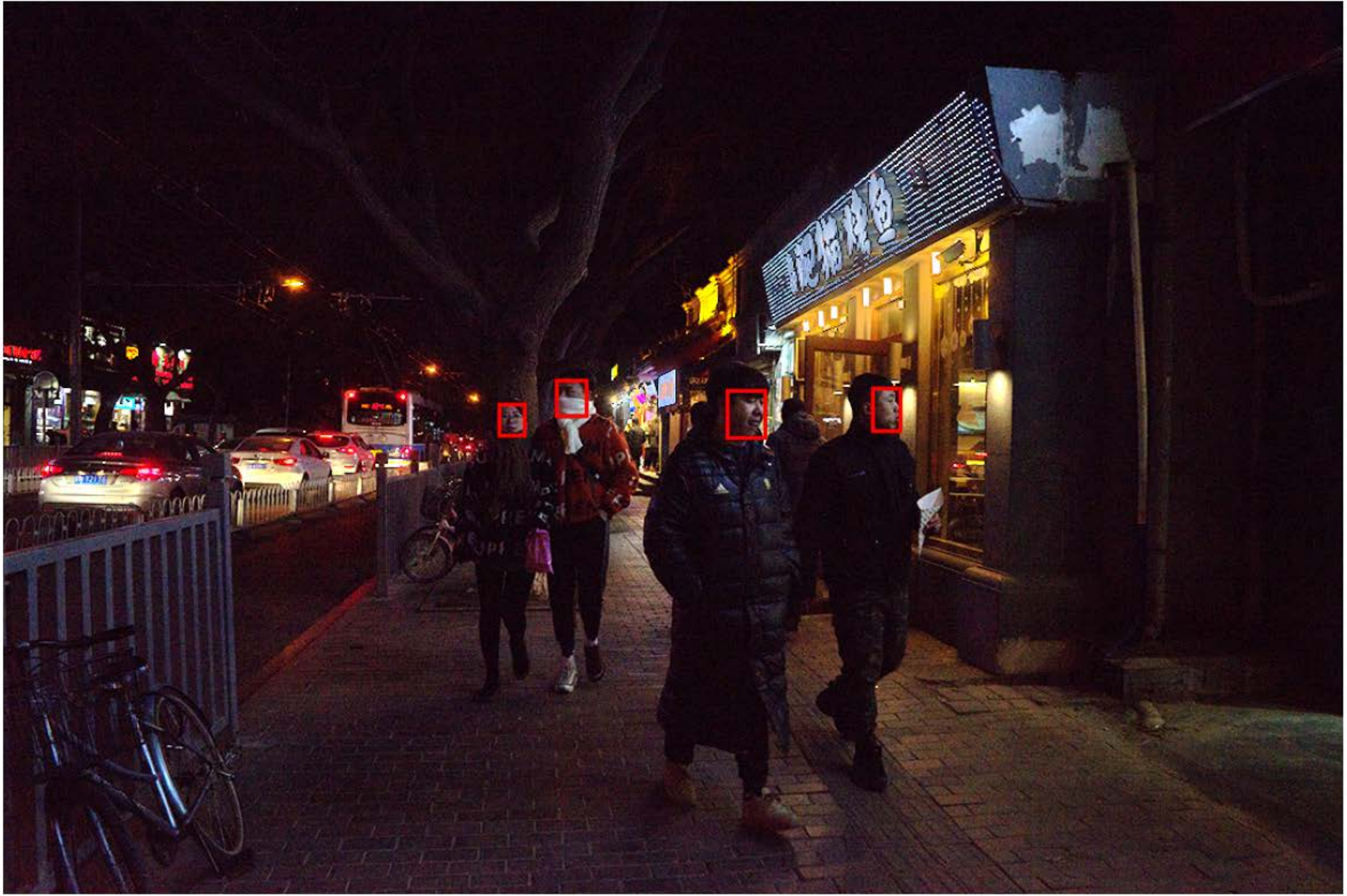}\\
			(j) EnlightenGAN \cite{EnlightenGAN} & (k) DRBN \cite{YangCVRP20} & (l) ExCNet \cite{ZhangACM191}\\
			\includegraphics[width=0.3\linewidth]{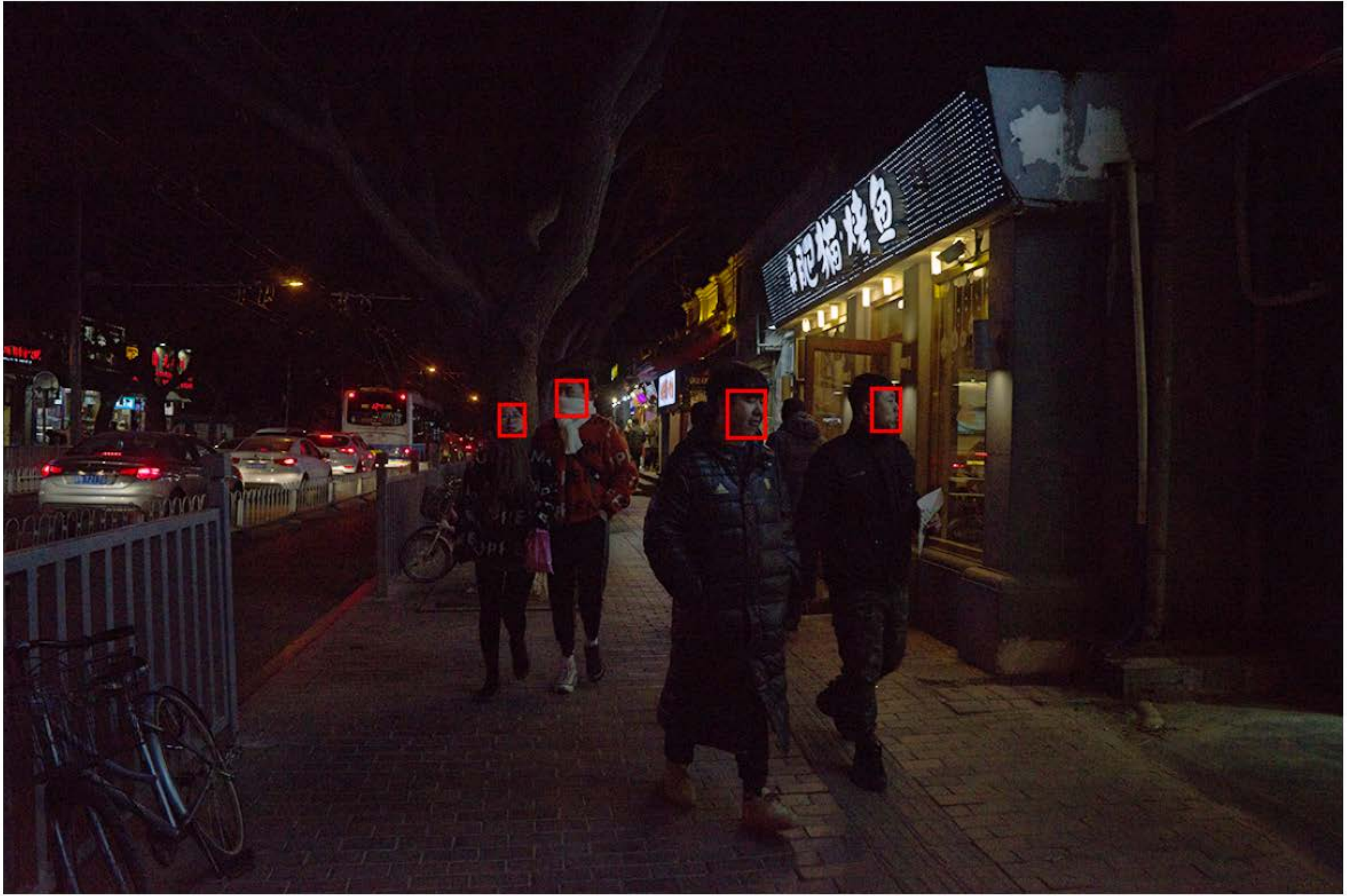}&
			\includegraphics[width=0.3\linewidth]{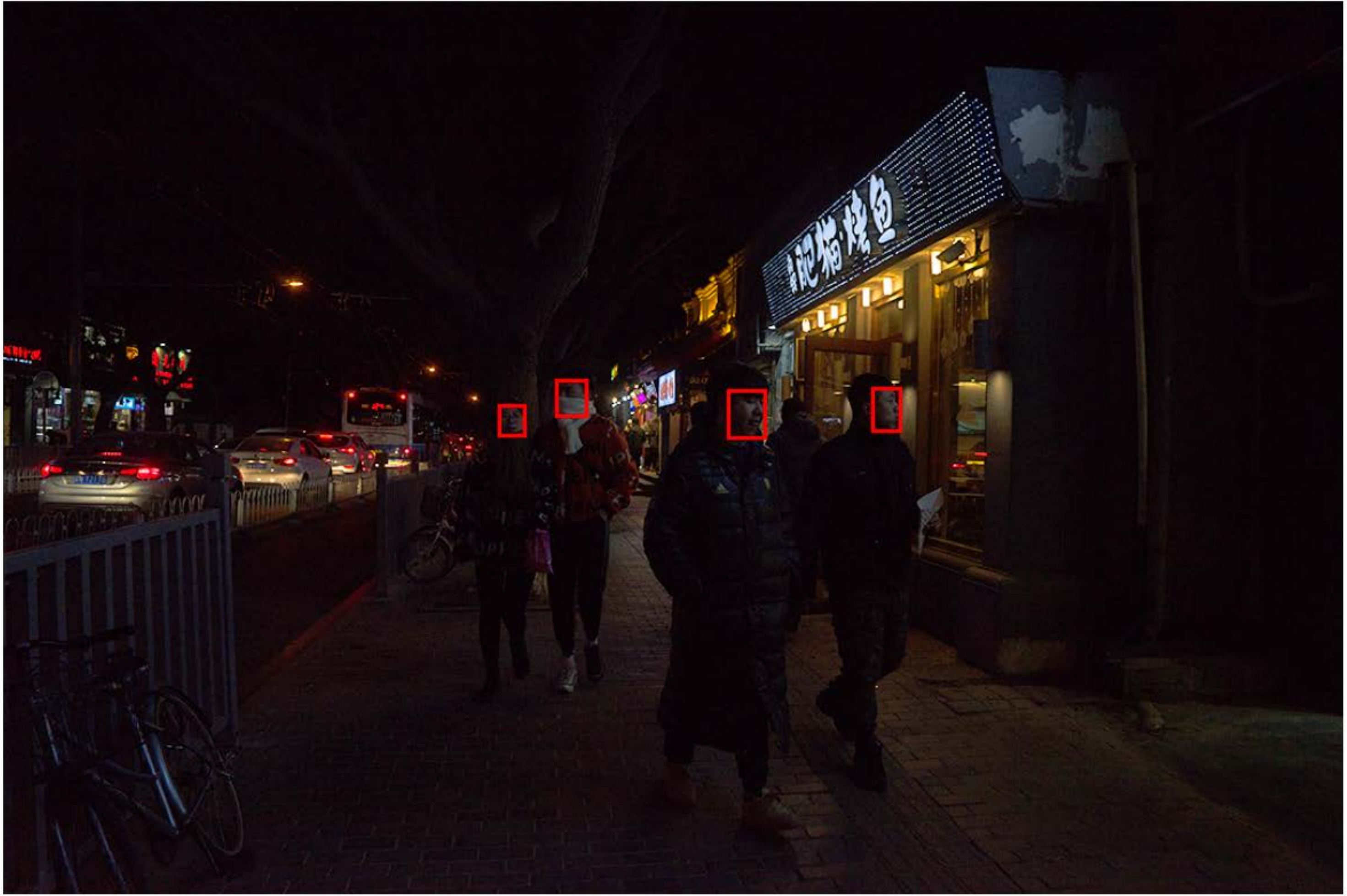}&
			~\\
			(m) Zero-DCE \cite{ZeroDCE} &  (n) 	RRDNet \cite{RRDNet}  &  \\
		\end{tabular}
	\end{center}
	\caption{Visual results of different methods on a low-light image sampled from  DARK FACE dataset  \cite{Yuan2019}. Better see with zoom in for the bounding boxes of faces.}
	\label{fig:face_visual3}
\end{figure*}

\begin{figure}[t]
	\centering  \centerline{\includegraphics[width=1\linewidth]{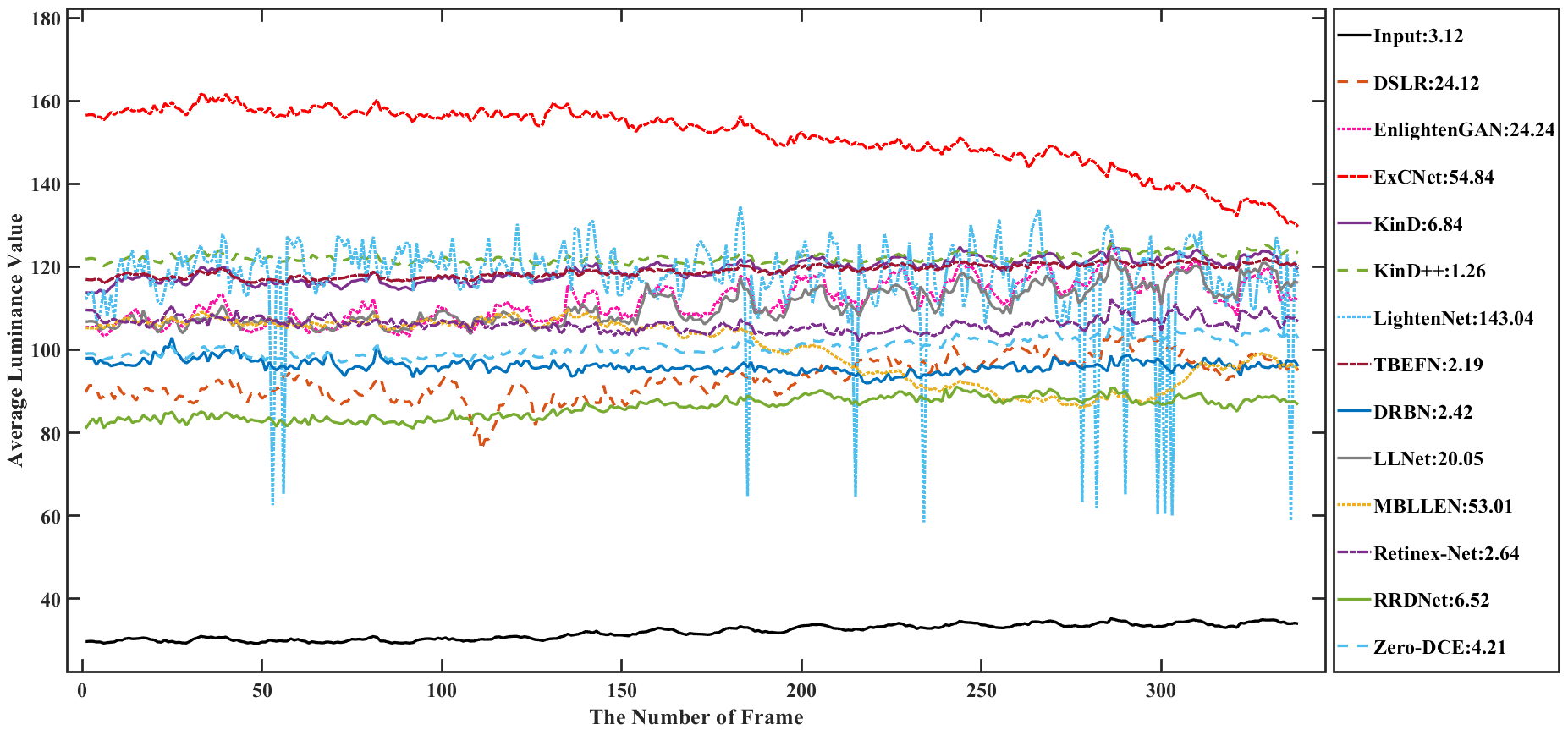}}
	\caption{Luminance curves of the video enhanced by different methods.  The curve is plotted by computing average luminance value of each bounding box in consecutive frames. The smoother the curve is, the better the method for video enhancement (i.e., temporal coherence) is. The numbers in the legend represent the average luminance variance values (the smaller, the better). The low-light video was taken by a Huawei Mate 20 Pro phone's camera.}
	\label{fig:curve1}
\end{figure}

\begin{figure}[t]
	\centering  \centerline{\includegraphics[width=1\linewidth]{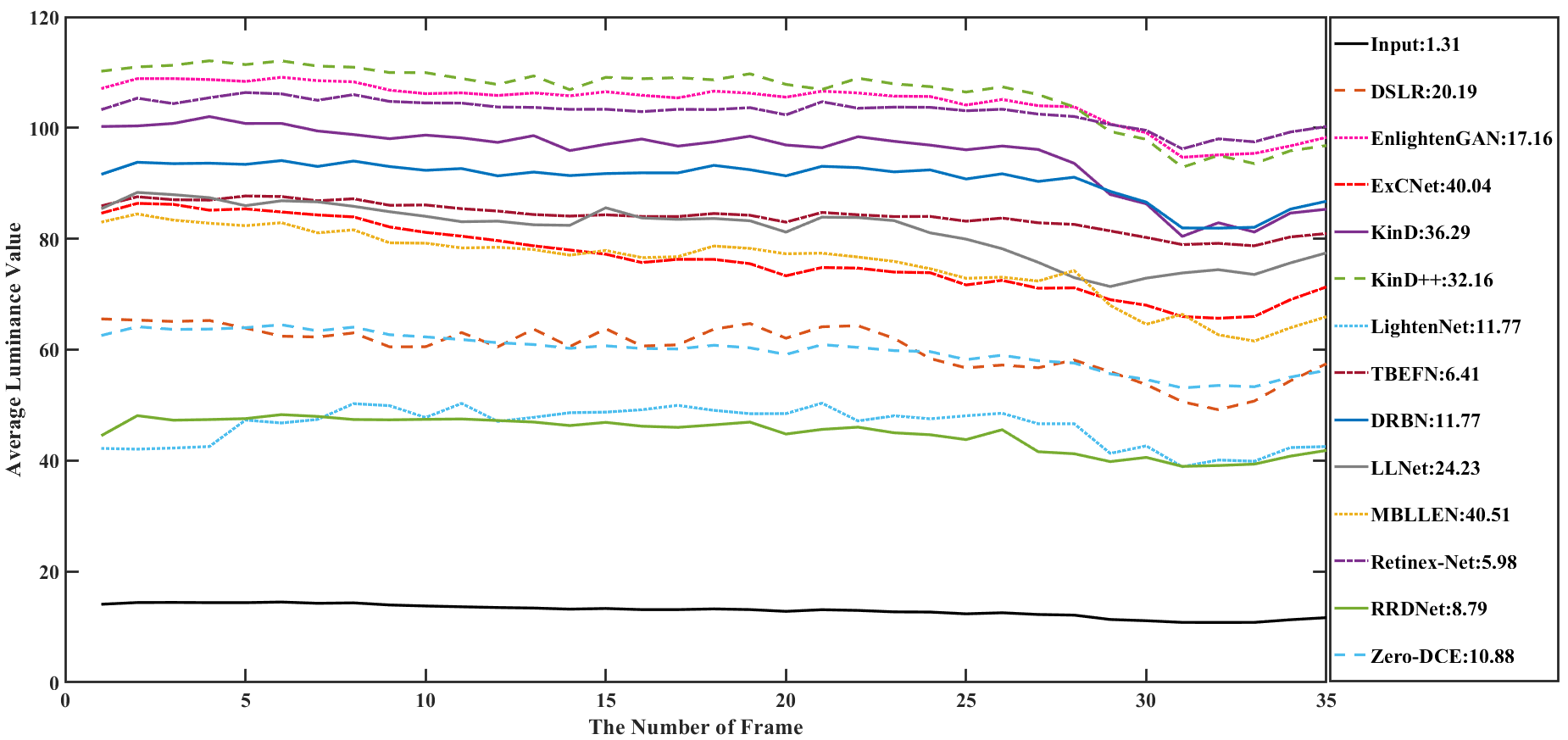}}
	\caption{Luminance curves of the video enhanced by different methods.  The curve is plotted by computing average luminance value of each bounding box in consecutive frames. The smoother the curve is, the better the method for video enhancement (i.e., temporal coherence) is. The numbers in the legend represent the average luminance variance values (the smaller, the better). The low-light video was taken by an iPhone 6s phone's camera.}
	\label{fig:curve2}
\end{figure}

\begin{figure}[t]
	\centering  \centerline{\includegraphics[width=1\linewidth]{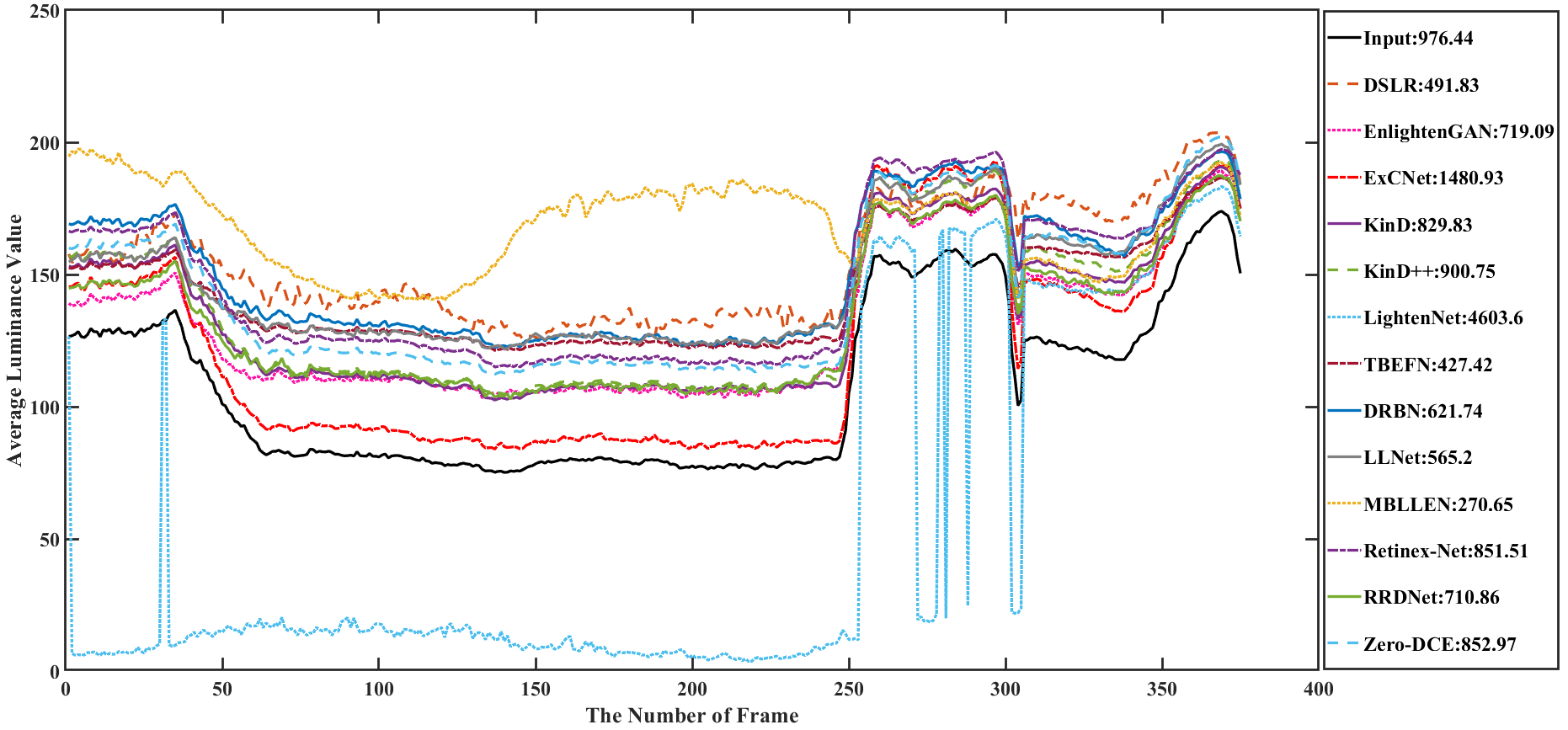}}
	\caption{Luminance curves of the video enhanced by different methods.  The curve is plotted by computing average luminance value of each bounding box in consecutive frames. The smoother the curve is, the better the method for video enhancement (i.e., temporal coherence) is. The numbers in the legend represent the average luminance variance values (the smaller, the better). The low-light video was taken by an iPhone 7 phone's camera.}
	\label{fig:curve3}
\end{figure}

\begin{figure}[t]
	\centering  \centerline{\includegraphics[width=1\linewidth]{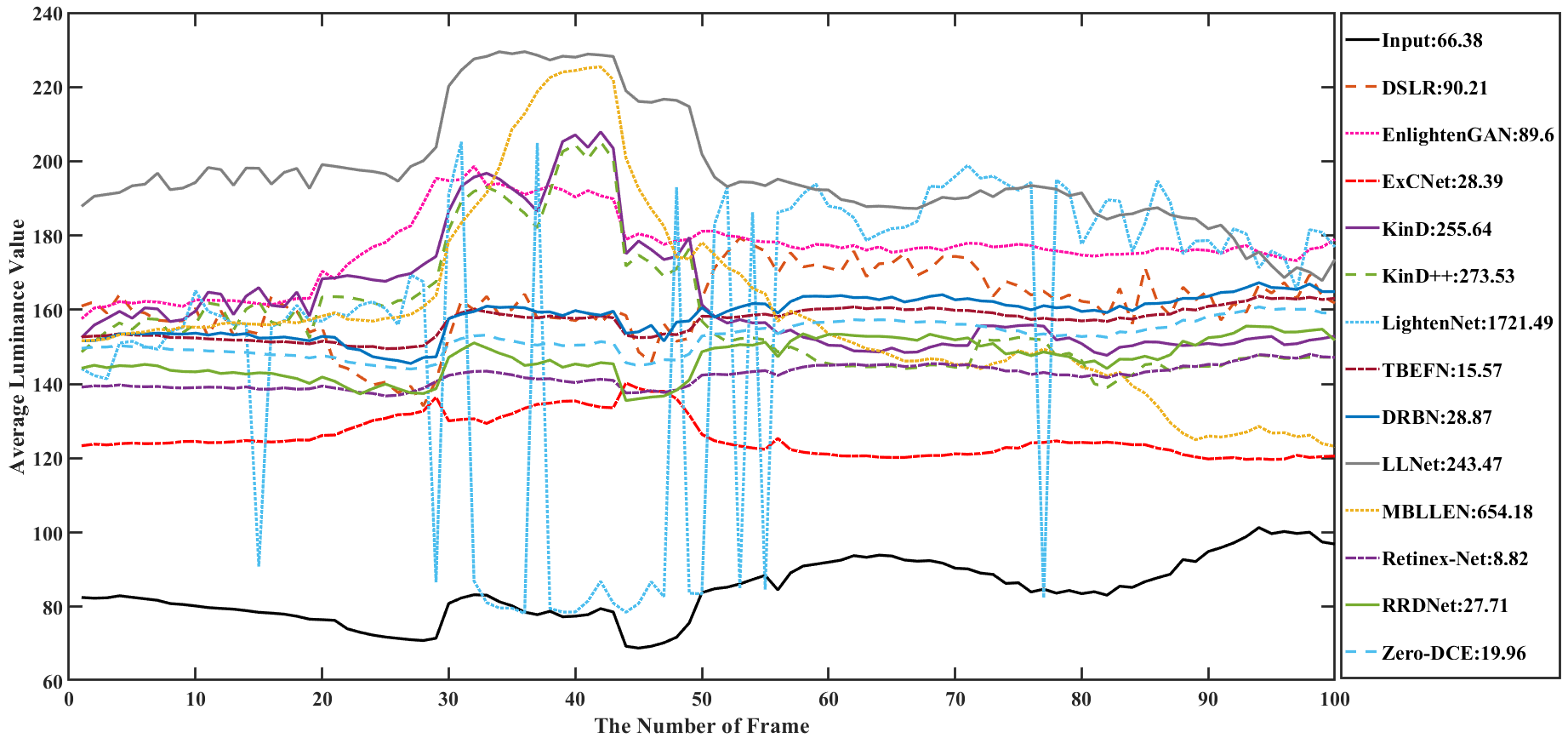}}
	\caption{Luminance curves of the video enhanced by different methods.  The curve is plotted by computing average luminance value of each bounding box in consecutive frames. The smoother the curve is, the better the method for video enhancement (i.e., temporal coherence) is. The numbers in the legend represent the average luminance variance values (the smaller, the better). The low-light video was taken by an iPhone7 Plus phone's camera.}
	\label{fig:curve4}
\end{figure}

\begin{figure}[t]
	\centering  \centerline{\includegraphics[width=1\linewidth]{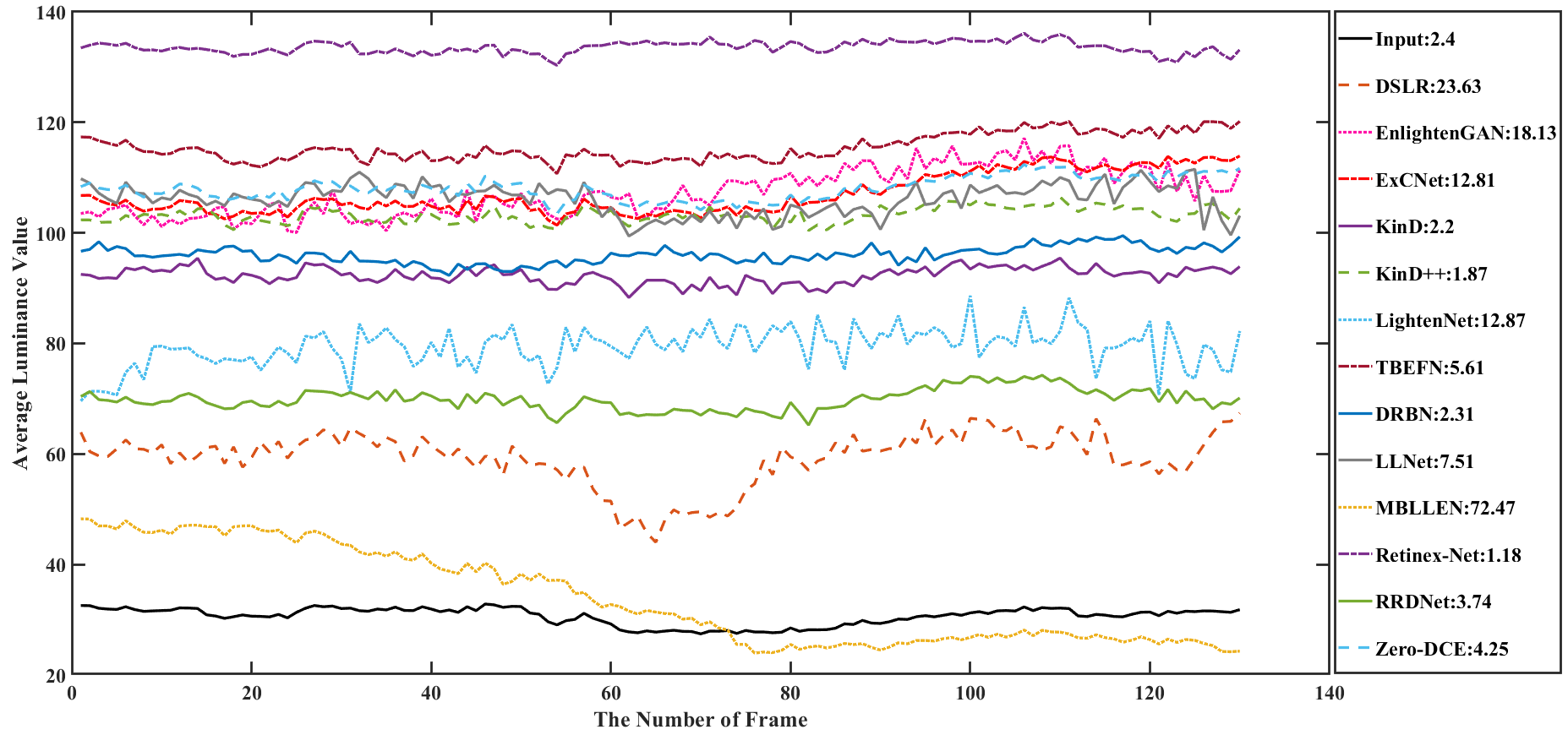}}
	\caption{Luminance curves of the video enhanced by different methods.  The curve is plotted by computing average luminance value of each bounding box in consecutive frames. The smoother the curve is, the better the method for video enhancement (i.e., temporal coherence) is. The numbers in the legend represent the average luminance variance values (the smaller, the better). The low-light video was taken by an iPhone8 Plus phone's camera.}
	\label{fig:curve5}
\end{figure}

\begin{figure}[t]
	\centering  \centerline{\includegraphics[width=1\linewidth]{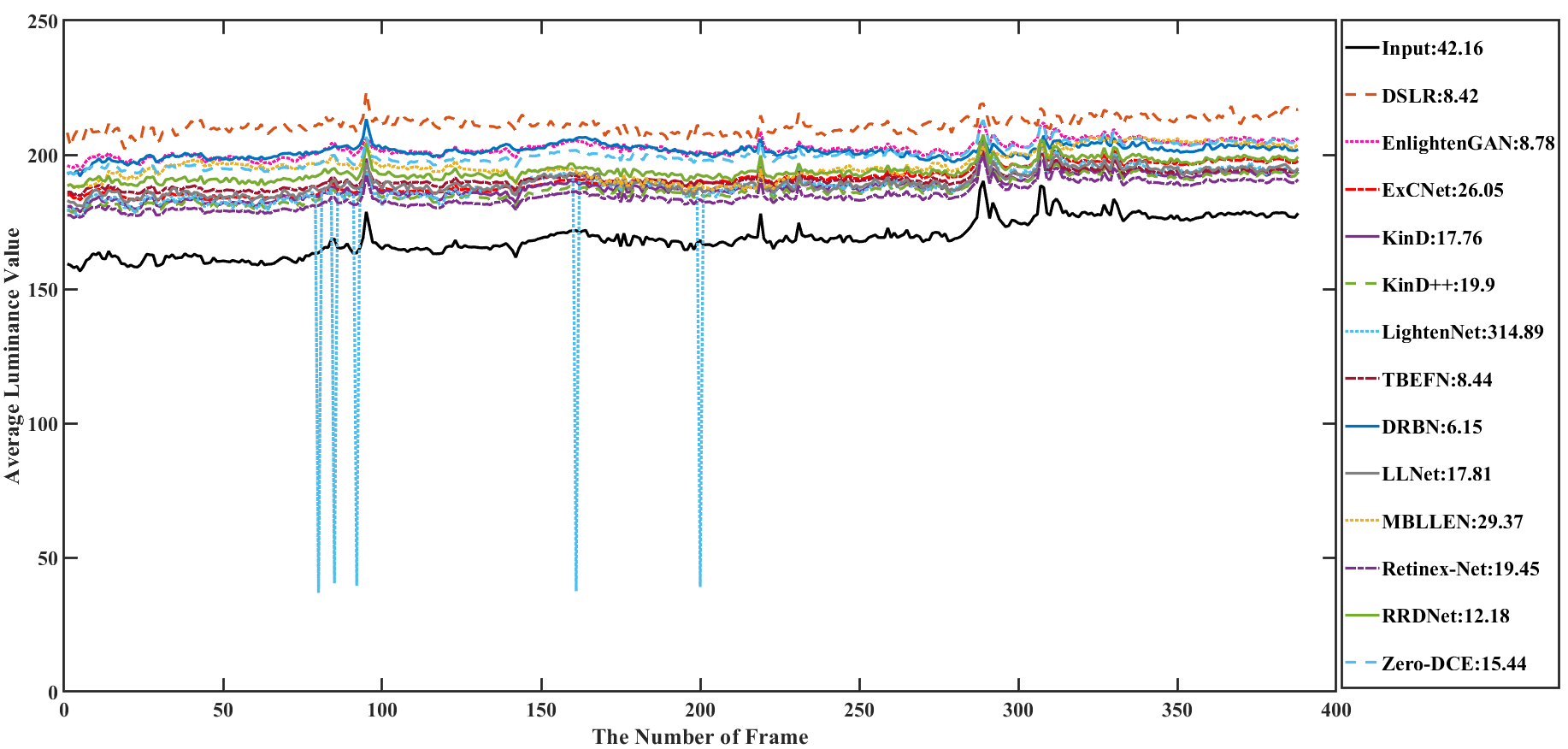}}
	\caption{Luminance curves of the video enhanced by different methods.  The curve is plotted by computing average luminance value of each bounding box in consecutive frames. The smoother the curve is, the better the method for video enhancement (i.e., temporal coherence) is. The numbers in the legend represent the average luminance variance values (the smaller, the better). The low-light video was taken by an iPhone 11 phone's camera.}
	\label{fig:curve6}
\end{figure}

\begin{figure}[t]
	\centering  \centerline{\includegraphics[width=1\linewidth]{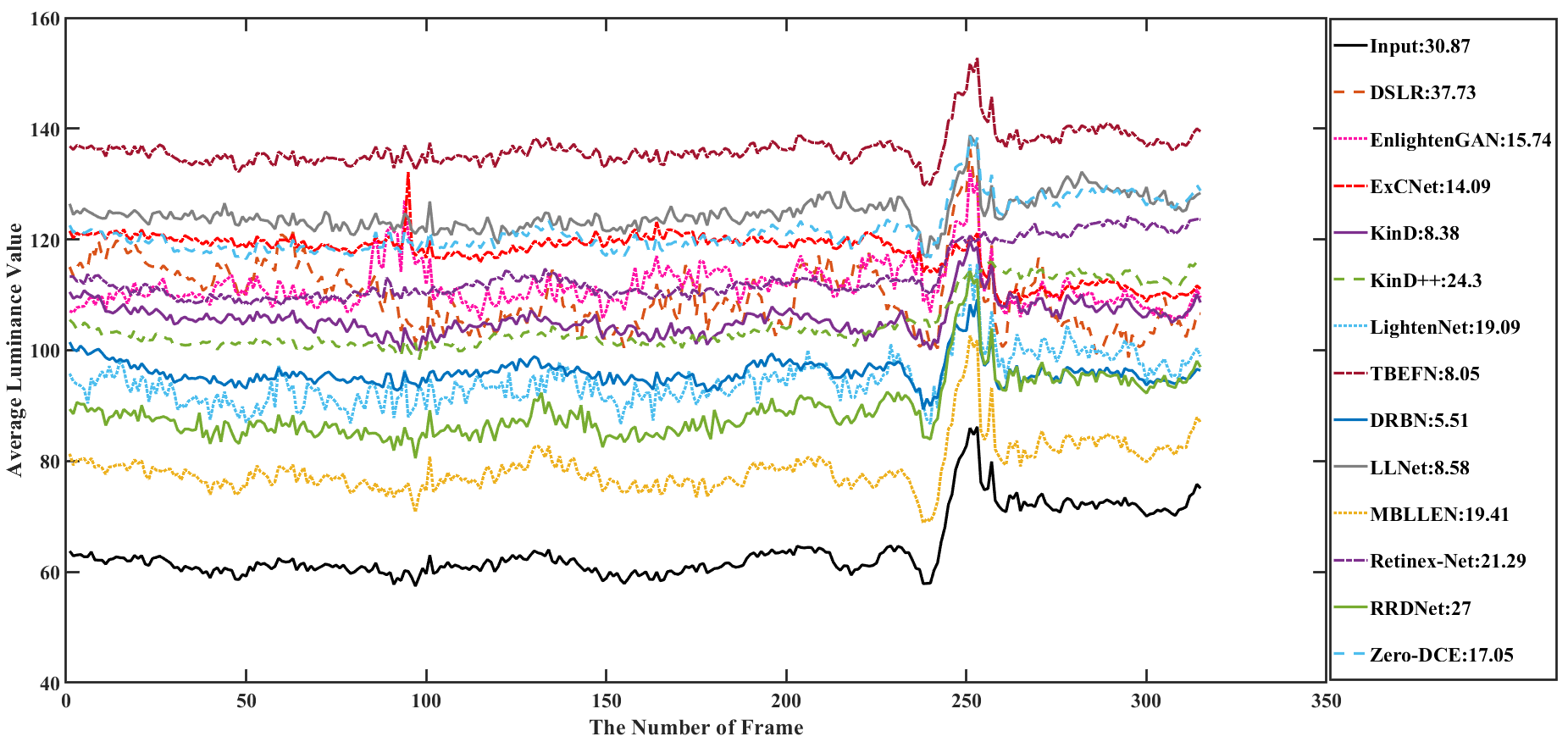}}
	\caption{Luminance curves of the video enhanced by different methods.  The curve is plotted by computing average luminance value of each bounding box in consecutive frames. The smoother the curve is, the better the method for video enhancement (i.e., temporal coherence) is. The numbers in the legend represent the average luminance variance values (the smaller, the better). The low-light video was taken by an iPhone11 Pro phone's camera.}
	\label{fig:curve7}
\end{figure}

\begin{figure}[t]
	\centering  \centerline{\includegraphics[width=1\linewidth]{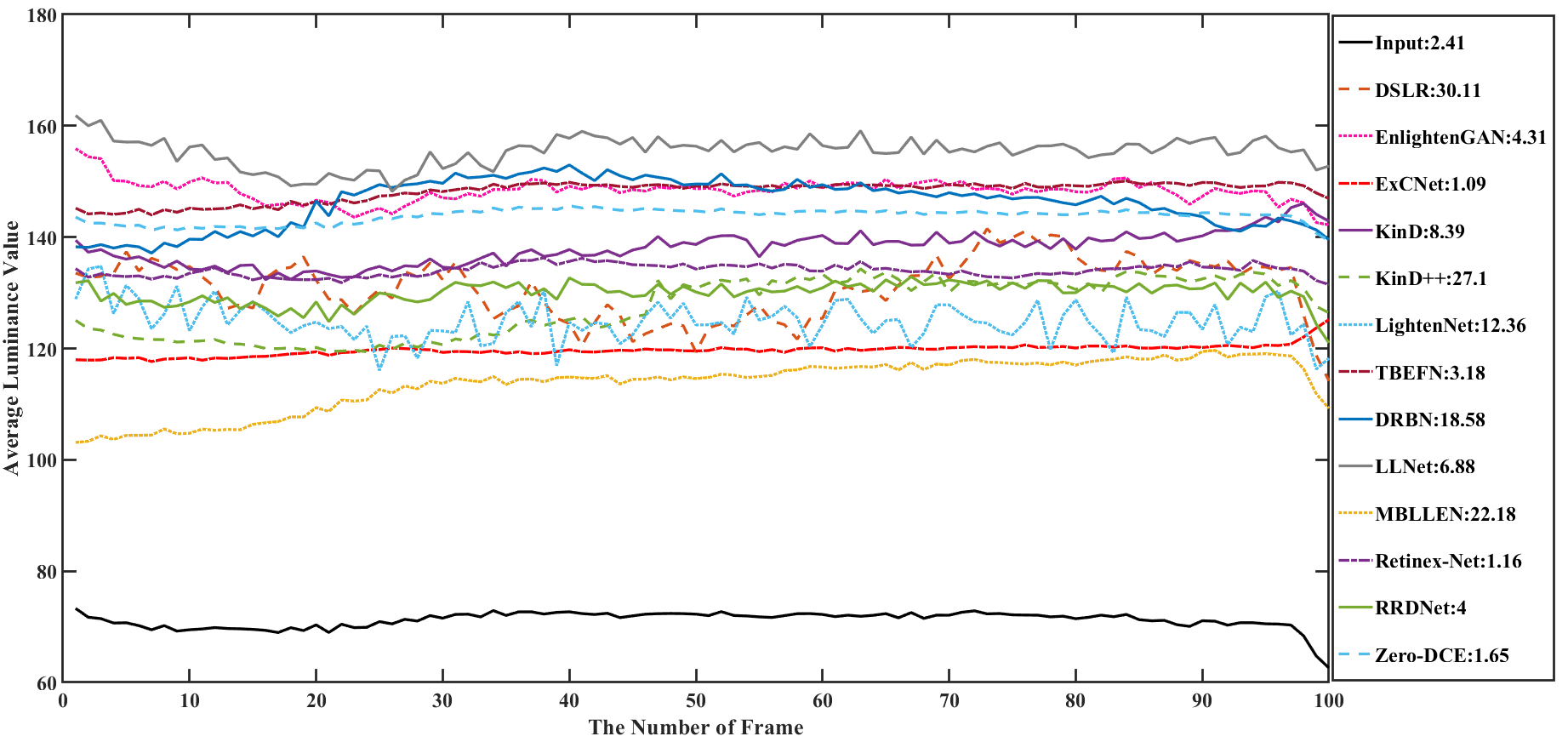}}
	\caption{Luminance curves of the video enhanced by different methods.  The curve is plotted by computing average luminance value of each bounding box in consecutive frames. The smoother the curve is, the better the method for video enhancement (i.e., temporal coherence) is. The numbers in the legend represent the average luminance variance values (the smaller, the better). The low-light video was taken by an iPhone SE phone's camera.}
	\label{fig:curve8}
\end{figure}

\begin{figure}[t]
	\centering  \centerline{\includegraphics[width=1\linewidth]{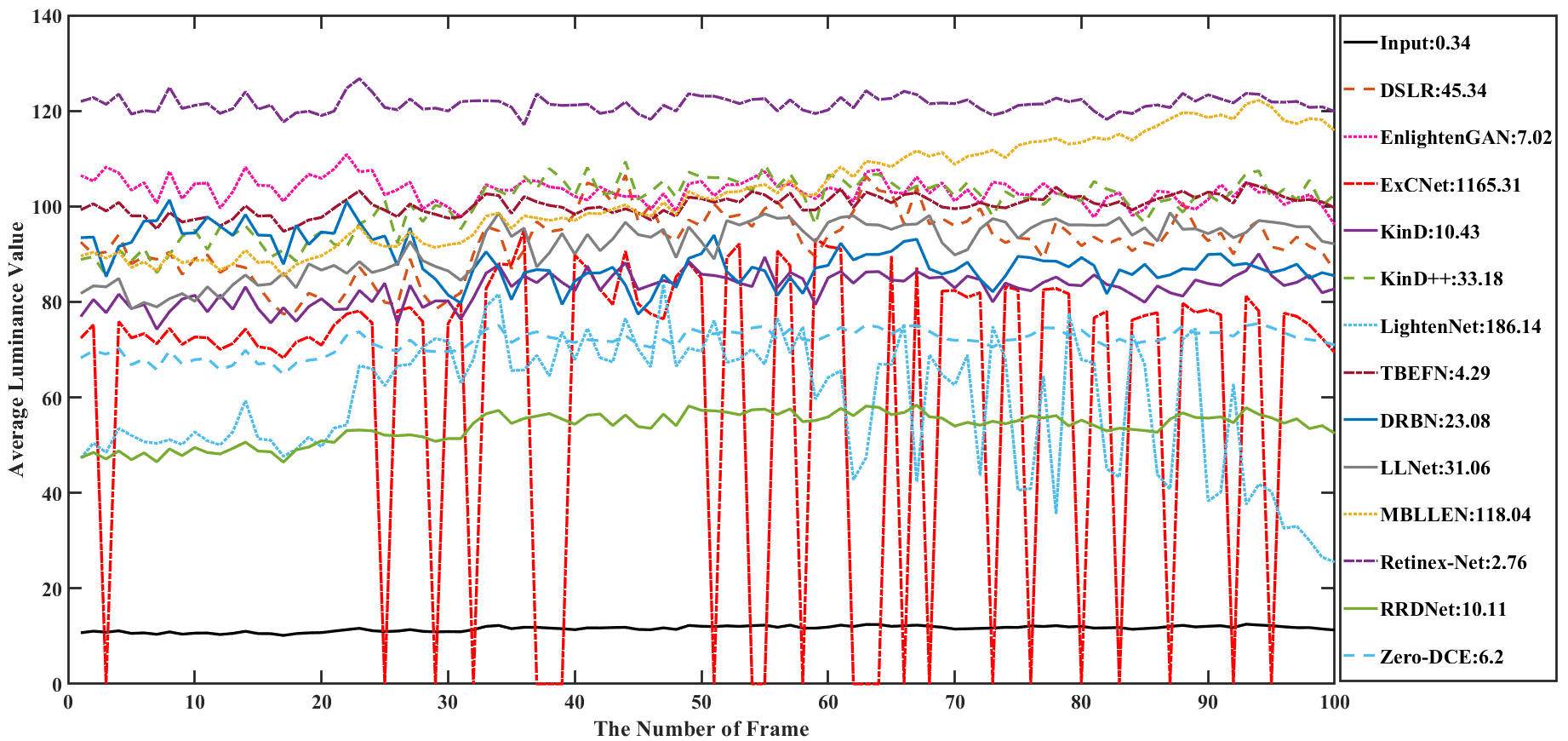}}
	\caption{Luminance curves of the video enhanced by different methods.  The curve is plotted by computing average luminance value of each bounding box in consecutive frames. The smoother the curve is, the better the method for video enhancement (i.e., temporal coherence) is. The numbers in the legend represent the average luminance variance values (the smaller, the better). The low-light video was taken by an iPhone XS phone's camera.}
	\label{fig:curve9}
\end{figure}

\begin{figure}[t]
	\centering  \centerline{\includegraphics[width=1\linewidth]{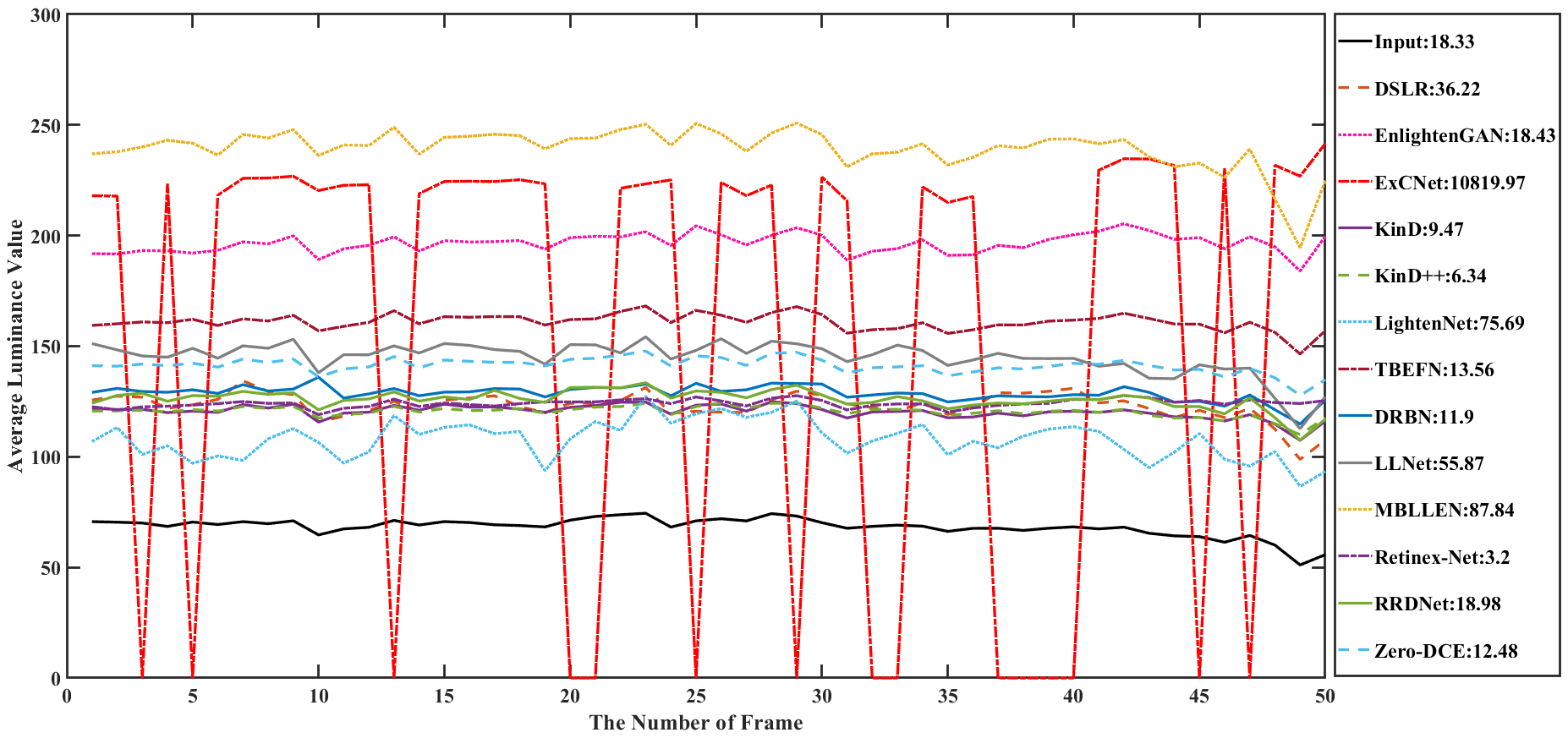}}
	\caption{Luminance curves of the video enhanced by different methods.  The curve is plotted by computing average luminance value of each bounding box in consecutive frames. The smoother the curve is, the better the method for video enhancement (i.e., temporal coherence) is. The numbers in the legend represent the average luminance variance values (the smaller, the better). The low-light video was taken by an OnePlus 5T phone's camera.}
	\label{fig:curve10}
\end{figure}

\clearpage

\ifCLASSOPTIONcaptionsoff
\newpage
\fi

{
	\bibliographystyle{IEEEtran}
	\bibliography{bibliography}
}
